\documentclass[11pt]{article} %
\usepackage{fullpage}
\usepackage{authblk}
\usepackage[sort&compress,numbers]{natbib}
\usepackage{comment}
\usepackage{color}
\usepackage{xcolor}
\usepackage{hyperref}
\usepackage{dsfont}
\usepackage{graphicx}
\hypersetup{
    colorlinks,
    linkcolor={red!50!black},
    citecolor={red!50!black},
    urlcolor={red!80!black}
}
\usepackage{enumitem}
\usepackage{mathrsfs}
\usepackage{mathtools}
\usepackage[font=small,labelfont=bf]{caption}
\usepackage{subcaption}
\usepackage{graphicx}

\usepackage{amsmath,amsthm,colonequals,etoolbox}
\usepackage{thmtools}

\usepackage{amssymb}

\usepackage{thmtools}
\usepackage{url}
\usepackage{cleveref}
\usepackage{microtype}
\usepackage{thm-restate}
\usepackage{array}
\usepackage{MnSymbol}
\usepackage{bm}
\usepackage{xcolor}

\allowdisplaybreaks

\renewcommand{\geq}{\geqslant}

\renewcommand{\leq}{\leqslant}
\renewcommand{\preceq}{\preccurlyeq}
\renewcommand{\succeq}{\succcurlyeq}

\newtheorem{thm}{Theorem}[section]
\newtheorem{mydef}[thm]{Definition}
\newtheorem{myprop}[thm]{Proposition}

\newtheorem{mycorollary}[thm]{Corollary}
\newtheorem{myfact}[thm]{Fact}

\newtheorem{mythm}[thm]{Theorem}

\newtheorem{ex}[thm]{Example}

\theoremstyle{definition}
\newtheorem{rmk}[thm]{Remark}

\DeclareMathOperator{\sech}{sech}

\DeclareMathOperator*{\argmin}{arg\,min}
\DeclareMathOperator*{\argmax}{arg\,max}

\DeclareMathOperator*{\diag}{\mathsf{diag}}
\DeclareMathOperator*{\blkdiag}{\mathsf{blk-diag}}

\DeclareMathOperator*{\polylog}{polylog}

\newcommand{\calL}{\mathcal{L}}
\newcommand{\calD}{\mathcal{D}}
\newcommand{\calP}{\mathcal{P}}

\newcommand{\R}{\ensuremath{\mathbb{R}}}

\newcommand{\N}{\ensuremath{\mathbb{N}}}

\newcommand{\norm}[1]{\lVert #1 \rVert}
\newcommand{\bignorm}[1]{\left\lVert #1 \right\rVert}

\newcommand{\ip}[2]{\ensuremath{\langle #1, #2 \rangle}}

\newcommand{\Var}{\mathrm{Var}}
\newcommand{\E}{\mathbb{E}}
\newcommand{\abs}[1]{\ensuremath{| #1 |}}
\newcommand{\bigabs}[1]{\ensuremath{\left| #1 \right|}}

\newcommand{\ceil}[1]{\lceil #1 \rceil}
\newcommand{\bigceil}[1]{\left\lceil #1 \right\rceil}

\newcommand{\ind}{\mathds{1}}
\renewcommand{\vec}{\mathrm{vec}}
\newcommand{\mat}{\mathrm{mat}}

\renewcommand{\Pr}{\mathbb{P}}
\newcommand{\T}{\mathsf{T}}

\newcommand{\calI}{\mathcal{I}}
\newcommand{\calG}{\mathcal{G}}
\newcommand{\calN}{\mathcal{N}}
\newcommand{\calE}{\mathcal{E}}

\newcommand{\calF}{\mathcal{F}}

\newcommand{\calR}{\mathcal{R}}
\newcommand{\calH}{\mathcal{H}}

\numberwithin{equation}{section}

\newcommand{\Symm}{\mathsf{Sym}}

\newcommand{\sfX}{\mathsf{X}}

\newcommand{\sfZ}{\mathsf{Z}}
\DeclarePairedDelimiterX{\infdivx}[2]{(}{)}{%
  #1\;\delimsize\|\;#2%
}
\newcommand{\KL}{\mathrm{KL}\infdivx}
\newcommand{\Hg}[2]{\mathrm{d}_H(#1, #2)}
\newcommand{\HgSq}[2]{\mathrm{d}_H^2(#1, #2)}

\newcommand{\FI}[2]{\mathrm{d}_{\mathrm{FI}}(#1, #2)}
\newcommand{\FISq}[2]{\mathrm{d}_{\mathrm{FI}}^2(#1, #2)}
\newcommand{\FIMax}[2]{\mathrm{d}_{\calI_{\max}}(#1, #2)}

\newcommand{\bbZ}{\mathbb{Z}}

\newcommand{\opnorm}[1]{\norm{#1}_{\mathrm{op}}}
\newcommand{\bigopnorm}[1]{\left\| #1 \right\|_{\mathrm{op}}}
\newcommand{\tvnorm}[1]{\norm{#1}_{\mathrm{TV}}}
\newcommand{\TV}[2]{\tvnorm{#1 - #2}}

\newcommand{\e}{\varepsilon}

\newcommand{\distconv}{\stackrel{\mathrm{d}}{\rightsquigarrow}}

\newcommand{\rmd}{\mathrm{d}}

\newcommand{\sfN}{\mathsf{N}}

\newcommand{\Lip}{\mathrm{Lip}}

\newcommand{\bbS}{\mathbb{S}}

\newcommand{\tsqrt}[1]{\scalebox{0.8}{$\sqrt{#1}$}}

\title{Nearly Instance-Optimal Parameter Recovery from Many Trajectories via Hellinger Localization}

\author[1]{Eliot Shekhtman\thanks{Equal contribution, order chosen randomly.}}
\author[2]{Yichen Zhou\protect\footnotemark[1]} %
\author[1]{Ingvar Ziemann}
\author[1]{Nikolai Matni}
\author[2]{Stephen Tu}

\affil[1]{Department of Electrical and Systems Engineering, University of Pennsylvania}
\affil[2]{Department of Electrical and Computer Engineering, University of Southern California}

\begin{document}

\maketitle

\begin{abstract}

Learning from sequential, temporally-correlated data is a core facet of modern machine learning and statistical modeling. Yet our fundamental understanding of sequential learning remains incomplete, particularly in the multi-trajectory setting where data consists of many independent realizations of a time-indexed stochastic process. This important regime both reflects modern training pipelines such as for large language and vision-language models, and offers the potential for learning
without the typical mixing assumptions (e.g., geometric/polynomial ergodicity) made in the classical single-trajectory case.
However, sharp instance-optimal bounds 
are known only for
least-squares regression problems with dependent covariates~\cite{tu2024learning,ziemann2024sharp};
for more general models or loss functions, the only broadly applicable guarantees
result from a simple %
reduction to either (i) i.i.d.\ learning, with effective sample size scaling only in the number of trajectories,
or (ii) an existing single-trajectory result when each individual trajectory mixes, with effective
sample size scaling as the full data budget deflated by a factor of the mixing-time.

In this work, we significantly broaden the scope of instance-optimal rates for parameter recovery in multi-trajectory settings via the \emph{Hellinger localization framework}, a general approach for maximum likelihood estimation.
Our method proceeds by first controlling the squared Hellinger distance at the \emph{path-measure} level via a reduction to i.i.d.\ learning, followed by localization as a quadratic form in parameter space weighted by the trajectory-level Fisher information matrix.
This yields instance-optimal parameter recovery bounds that scale with the full data budget, i.e., number of trajectories times trajectory length, under a broad set of conditions. 
We instantiate our framework across diverse case studies, including a simple mixture of Markov chains example, dependent linear regression under non-Gaussian noise (i.e., non-square losses), generalized linear models with non-monotonic activations, and linear-attention sequence models. %
In all cases, our parameter recovery bounds nearly match the instance-optimal rates implied by asymptotic normality, substantially improving over bounds from standard reductions.

\end{abstract}
\newpage

{\hypersetup{linkcolor=black}
\tableofcontents
}
\newpage

\section{Introduction}
\label{sec:intro}

Learning from sequential data is central to modern machine learning (ML) and statistical modeling, underpinning applications such as language modeling~\cite{brown2020gpt3}, speech recognition~\cite{baevski2020wav2vec}, time-series forecasting~\cite{salinas2020deepAR}, generalist robotics~\cite{zitkovich2023rt2}, neurological sequence analysis~\cite{schirrmeister2017deepEEG}, and many other examples. Yet, despite its importance and prevalence, our fundamental understanding of when learning from sequential streams is possible---and what the sharp, problem-specific sample complexities are---remains far less developed compared with the classical i.i.d.\ setting.

There are two predominant approaches to analyzing non-i.i.d.\ sequential learning setups. The first approach is to consider consuming data from a single stochastic process indexed by time $t$, and study what happens to the
estimator as time progresses forward.
In this paper, we will refer to a realization from a time-index stochastic process as a trajectory, and hence
we denote this approach as the \emph{single-trajectory} setting. 
This setting is challenging, as simple examples illustrate the necessity of imposing non-trivial assumptions about the long-running 
behavior of the underlying process. The strongest results hold under
assumptions about the rate of convergence (typically quantified via its mixing-time~\cite{bradley2005mixing}) of the stochastic process to its time-marginal distributions,
and provide bounds on e.g., the excess risk of the empirical risk minimizer (ERM) as time $t$ moves forward (see e.g.,~\cite{yu1994rates,karandikar2002,mohri2010stability,kuznetsov2017generalization}).
However, %
these type of results suffer from some key drawbacks:
(i) Many processes---e.g., human dialogue, periodic locomotion gaits, wearable sensor data---are interesting precisely because they do \emph{not} mix.
Even for processes that do mix, analytical bounds on the mixing-time can be quite conservative in high-dimension~\cite{qin2021limitations}, and
challenging to estimate numerically without assuming extra structure~\cite{qin2019spectralgap}.
(ii) Typical bounds suffer from \emph{sample deflation}, where the non-i.i.d.\ sequential rate is a factor
of the mixing-time larger than its corresponding i.i.d.\ rate (i.e., its effective sample size is deflated by the mixing-time);
furthermore, the non-i.i.d.\ rates are only valid after time $t$ exceeds some factor of the mixing-time, typically referred to as a burn-in time.

The second approach to studying sequential learning sits in-between the
classic i.i.d.\ setting, where every data point is independent,
and the single-trajectory setting, where every data point is correlated.
Instead, one assumes that many independent realizations of a stochastic process
are observed, a setup we will refer to as the \emph{multi-trajectory} setting (see e.g.,~\cite{hsu2012learningHMMs,thomas2015ope,chen2022mixtureLDS,tu2024learning}).
In the multi-trajectory setting, data within a trajectory is temporally correlated as usual, but importantly, data across different trajectories is independent.
The latter fact is crucial, as it enables reductions to i.i.d.\ learning 
only using minimal assumptions about the underlying process.
A further benefit of the multi-trajectory data model is that it is a much more
accurate description of the underlying training data for modern ML models which ingest sequential data---such as vision-language models (VLMs)~\cite{radford2021clip}, large language models (LLMs)~\cite{brown2020gpt3}, and generalist behavior policies for robotics~\cite{zitkovich2023rt2}---compared with the single-trajectory model.

Nevertheless, despite these advantages, many open questions still remain regarding the 
multi-trajectory model. 
A na{\"{i}}ve reduction to i.i.d.\ learning, where each trajectory is treated as a single data point, only yields rates where the effective sample size is the $m$, the number of trajectories. Here, the length $T$ of each trajectory\footnote{We assume for ease of exposition that each trajectory has the same length.} is absent from this sample size, which is in
general not the correct scaling.
On the other hand, embedding a multi-trajectory process into a single trajectory and appealing to a single-trajectory result yields either 
(i) similar bounds as the i.i.d.\ reduction if no assumption on the mixing-time of the individual trajectories is made, or 
(ii) bounds where the effective sample size 
scales as the $mT/\kappa$---i.e., \emph{total} number of data points available to the
learner divided by the mixing-time $\kappa$ of the individual trajectories.
While (ii) improves upon the i.i.d.\ reduction, the deflation factor is in general not optimal
for multi-trajectory settings. Indeed, a recent line of work~\cite{tu2024learning,ziemann2022learning,ziemann2024sharp}
shows that for square-loss regression from dependent covariates,
one can obtain---under a set of conditions
which do not generally require e.g., bounded mixing-times---finite-sample rates where 
the effective sample size not only improves to $mT$, 
but also the bounds nearly match those prescribed by asymptotic normality of 
maximum likelihood estimation (MLE). 
However, as their proof techniques are tailored specifically for square-loss, it is unclear how to generalize these results more broadly to e.g., MLE settings.

In this work, we significantly broaden our understanding of learning in the multi-trajectory setting by providing a general framework---which we call the \emph{Hellinger localization framework}---for deriving sharp instance-optimal parameter recovery rates for general maximum-likelihood estimation.
At a high-level, our framework proceeds in two main phases:
In the first phase, we utilize a reduction to i.i.d.\ learning which controls the
squared Hellinger distance between the \emph{path measures} of the MLE estimate and the 
underlying distribution, at a rate where the sample size is $m$, the number of trajectories.
In the second phase,
we utilize the fact that squared Hellinger distance is locally quadratic in the parameter space,
weighted by the Fisher information matrix of the underlying path measure. This has 
two key consequences: First, it allows us to extract out an additional scaling factor of $T$, the length of each trajectory, whenever the process contains sufficient excitation.
Second, the Fisher information matrix allows us
to derive instance-specific rates that match, up to logarithmic factors, those
prescribed by asymptotic normality of MLE.
Our framework thus yields 
instance-optimal bounds where the effective sample size contains all the observed data (i.e., scales as $mT$),
and applies broadly to maximum likelihood estimation problems.
Furthermore, as our framework relies only on 
bounded growth conditions of both the score function and the observed information matrix for localization,
it applies beyond the usual mixing processes and stable dynamics typically assumed in 
sequential learning.

To demonstrate the generality of our approach, we instantiate 
the Hellinger localization framework in four case studies:
(i) a simple mixture of Markov chains example,
(ii) a dependent linear regression setting under general (i.e., non-Gaussian) product-noise distributions,
(iii) a non-monotonic generalized linear model (GLM) example, and
(iv) a linear-attention~\cite{katharopoulos2020linearattention} sequence modeling problem.
For each of these case studies, our framework obtains near-optimal parameter recovery
error rates, yielding significant improvements over the rates obtained by standard reductions.

\paragraph{Paper organization.}
This manuscript is organized as follows.
In \Cref{sec:related_work}, we review related work.
\Cref{sec:general_framework} describes the multi-trajectory MLE problem setup,
reviews the standard i.i.d.\ and single-trajectory reductions in more detail,
and presents the Hellinger localization framework, including
a step-by-step guide describing how to instantiate the framework
for a general problem.
\Cref{sec:case_studies} contains the results of our four specific case studies;
for each case study, we also conduct a more thorough literature review on the specific problem
beyond what is described in \Cref{sec:related_work}.
\Cref{sec:conclusion} concludes the paper, with the appendices containing the deferred proofs. %

\section{Related Work}
\label{sec:related_work}

We first review the relevant literature for learning from non-i.i.d.\ sequential
data in the single-trajectory (single realization of a time-indexed stochastic process)
setting;
as already discussed briefly and will be reviewed in more detail in \Cref{sec:general_framework:MLE},
single-trajectory results can generally be used to analyze multi-trajectory settings.
The most common approach to analyzing single-trajectory learning
is to impose mixing-time assumptions on the underlying process (see e.g.,~\cite{yu1994rates,karandikar2002,mohri2010stability,kuznetsov2017generalization,meir2000nonparametric,mehryar2008noniid,duchi2012ergodic}); as a result, excess-risk or parameter recovery rates
are typically a factor of mixing-time
worse than their corresponding i.i.d.\ rates, as the standard blocking technique~\cite{yu1994rates} can only utilize one data point in every mixing-time size chunk. Recently, there have been a few improvements for various
problem settings. A line of work studying parameter recovery in linear dynamical systems
allows for the transition matrix $A$ to be marginally stable (i.e., $\rho \leq 1$)~\cite{simchowitz2018learningwithoutmixing} or even
unstable (i.e., $\rho > 1$)~\cite{sarkar2019linear,faradonbeh2018unstable}; both situations correspond to unbounded mixing-times,
with the former marginally stable setting also extended for a class of GLMs~\cite{kowshik2021nonlinear}.
For realizable non-linear regression problems with square loss, \cite{ziemann2022learning} shows that assuming a certain trajectory-level hyper-contractivity condition holds,
the sample size deflation in the excess risk bound can actually be removed, leaving the mixing-time dependence to only the burn-in time. The work \cite{ziemann2024sharp} further improves upon this result by obtaining 
variance-optimal rates under a (weakly) sub-Gaussian class (cf.~\cite{lecue2013learning}) assumption;
as this result is important context for our work, we review it in detail in \Cref{sec:general_framework:MLE}.
Despite improvements, these works either (a) only apply to a limited class of models, or (b) require the underlying process to mix, to satisfy additional technical assumptions (e.g., 
trajectory hyper-contractivity or sub-Gaussian class) that can be difficulty to verify, and only apply for the square-loss.

Next, we address literature directly studying the multi-trajectory setting (see e.g.,~\cite{hsu2012learningHMMs,thomas2015ope,chen2022mixtureLDS,tu2024learning,duchesne1996subspace,markovsky2015multiple,dean2020sample,zheng2021multitraj,xing2022multiplicative}). 
Most relevant to our work is \cite{tu2024learning},
which studies parameter recovery for linear least-squares regression from dependent covariates,
and derives instance-optimal rates that scale with the full dataset size $mT$ while
requiring no stability/mixing assumptions; we review these
results in more detail in \Cref{sec:general_framework:MLE}.
While this work also provides important context and motivation for our study,
their proof techniques---self-normalized martingales~\cite{abbasi2011online} and extensions of small-ball inequalities~\cite{mendelson2015learning}---are
tailored for the specific closed-form structure of the linear least-squares regression solution,
and do not admit obvious extensions
to more general setups. 
On the other hand, our approach is based more on information-theoretic concepts, building on a combination of 
techniques for analyzing density estimation~\cite{geer2000empirical,zhang2006densityestimation,foster2020nonlinear}, in conjunction with locally quadratic expansions of
$f$-divergences~\cite{polyanskiy2025information} (specifically, the squared Hellinger distance in our setting). 

We next briefly remark on other recent progress in learning from non-i.i.d.\ data sources. One line of work 
studies learning either regression functions or filters
from the perspective of online learning and regret minimization~\cite{hazan2017spectral,hazan2018spectral,block2022smoothedlearning,dogariu2025universal},
constructing a online predictor of future observations that is competitive over
a family of fixed predictors given perfect hindsight knowledge.
We view our results as complementary to this line of work---an interesting question
for future research is to study these regret minimization formulations
in settings where multi-trajectory data is revealed online to the player.
Another line of work considers non-temporal data correlations, specifically
learning Ising models from either a single sample (e.g.,~\cite{chatterjee2007spinglass,bhattacharya2018ising}),
or multiple independent samples (e.g.,~\cite{bresler2015ising,vuffray2016ising}). 
While these problem setups are not directly comparable, we believe it is interesting
future work to study whether our techniques can also be applied in such a setting.

Finally, the case studies we consider in \Cref{sec:case_studies} are special cases and/or
natural extensions of problem setups previously considered in the literature; we provide detailed
overview of problem-specific related work for each case study in its corresponding sub-section. %

\section{Problem Setup and General Framework}
\label{sec:general_framework}

In this section, we review background, outline our general problem formulation
and describe our new Hellinger localization framework. We first describe the notation used in our work.
\paragraph{Notation.}
For a vector $x \in \R^d$, we let $\norm{x}_p$ denote its $\ell_p$ norm; for $p=2$, we drop the subscript, i.e., $\norm{x}=\norm{x}_2$.
The notation $x^{\otimes 2} = xx^\T$ is shorthand for the outer product matrix.
Also, the notation $\mathrm{diag}(x) \in \R^{d \times d}$ is the diagonal matrix 
satisfying $\mathrm{diag}(x)_{ii} = x_i$ for $i \in [d]$.
Given a positive definite matrix $\Sigma \in \R^{d \times d}$, we let $\norm{x}_\Sigma = \sqrt{ x^\T \Sigma x}$ denote its weighed $\ell_2$ norm. 
For a matrix $M \in \R^{d \times k}$, we let $\opnorm{M}$, $\norm{M}_F$ denote its operator (maximum singular value) and Frobenius norm, respectively.
If $d=k$ and $M$ is positive semi-definite, we let $M^{1/2}$ denote its PSD square root.
If $M = M^\T$ is symmetric, we let the eigenvalues of $M$ be
denoted as $\lambda_i(M)$, $i \in [d]$, listed in non-increasing order.
The notation $\vec(M) \in \R^{dk}$ denotes vectorization of $M$; we follow the convention that vectorization is done in column-order, so that for size conforming matrices $A$, $M$, and $B$, the identity $\vec(AMB) = (B^\T \otimes A) \vec(M)$ holds, where $\otimes$ denotes the Kronecker product. 
We use $\mathrm{mat}(\cdot)$ to denote the inverse of $\vec(\cdot)$, i.e., 
$\mathrm{mat}(\vec(M)) = M$; the output dimension $d \times k$ of $\mathrm{mat}(\cdot)$ will be implicit from context.
Given a real-valued random variable $X$, we let $\norm{X}_{\calL^p(\rho)} = (\E_{\rho}[ \abs{X}^p ] )^{1/p}$ denote the $\calL^p(\rho)$ norm.
For a measure $\mu$, we let $\mu^{\otimes k}$ to denote its $k$-fold product measure.
Finally, the unit sphere in $\R^d$ is denoted
$\bbS^{d-1} := \{ x \in \R^d \mid \norm{x} = 1 \}$.

\subsection{Maximum Likelihood Estimation in Multi-Trajectory Settings}
\label{sec:general_framework:MLE}

We fix an index $T \in \N_+$ and 
consider a stochastic process $z_{1:T} := (z_t)_{t=1}^{T}$ taking values in $\sfZ$. 
Let $p_\star(z_{1:T})$ denote the joint distribution over $z_{1:T}$. We emphasize that the process $z_{1:T}$ is not
necessarily stationary nor ergodic, nor does it necessarily
have bounded mixing-times.
Our learner observes
$m \in \N_+$ independent trajectories $\calD_{m,T} := ((z^{(i)}_t)_{t=1}^{T})_{i=1}^{m}$
with each $z^{(i)}_{1:T} \sim p_\star(\cdot)$.
Fix a parametric class $\calP := \{ p_\theta(z_{1:T}) \mid \theta \in \Theta \}$ of distributions 
and consider the 
maximum-likelihood (MLE) estimator $\hat{p}_{m,T} \in \calP$ given the dataset $\calD_{m,T}$ as:
\begin{align}
    \hat{p}_{m,T} \in \argmax_{p_\theta \in \calP} \sum_{i=1}^{m} \log p_\theta(z^{(i)}_{1:T}). \label{eq:MLE}
\end{align}
In this work, we are interested in the finite-sample behavior of the MLE estimator $\hat{p}_{m,T}$
in the \emph{realizable}
setting, i.e., where $p _\star \in \calP$.
We impose some regularity conditions to make the analysis well-posed.
First, we endow $\sfZ$ with a base measure $\mu$ (e.g., counting measure for discrete $\sfZ$ or Lebesgue measure when $\sfZ$ is a subset of Euclidean space), and we overload $p_\theta$ to also denote the Radon-Nikodym density w.r.t.\ the corresponding base measure $\mu$ on $\sfZ$.
We also assume 
that (a) for $\mu^{\otimes T}$-a.e.\ $z_{1:T} \in \sfZ^T$,
the map $\theta \mapsto p_\theta(z_{1:T})$ is $C^2(\Theta_0)$ where $\Theta_0 \supseteq \Theta$ is an open set,
(b) that there exists a unique $\theta_\star \in \Theta$ such that $p_{\theta_\star}(\cdot) = p_\star(\cdot)$ (almost surely),
and that (c) the parameter set $\Theta$ is star-convex around $\theta_\star$.\footnote{That is for any $\theta \in \Theta$, $s \theta_\star + (1-s)\theta \in \Theta$ for all $s \in [0, 1]$.
As our analysis relies on local quadratic expansions, this technical assumption ensures that such expansions are indeed valid.
}
To set the stage for our results, 
let us first review what is known about the MLE estimator
$\hat{p}_{m,T}$. 

\paragraph{Asymptotic normality.}
First, we can use the lens of asymptotic normality to understand limiting behavior as $m \to \infty$ (but $T$ is fixed). To do this, we recall that
the Fisher information (FI) matrix for the \emph{trajectory} $z_{1:T}$ 
is defined as:
\begin{align}
    \calI(\theta) := -\E_{z_{1:T} \sim p_\theta}\left[ \nabla^2_\theta \log p_\theta(z_{1:T}) \right] = \E_{z_{1:T} \sim p_\theta}\left[ \nabla_\theta \log p_\theta(z_{1:T})^{\otimes 2} \right].
\end{align}
Under the assumption
that $\theta_\star \in \mathrm{int}(\Theta)$
and that the estimator $\hat{\theta}_{m,T}$ is consistent (i.e., $\hat{\theta}_{m,T} \to \theta_\star$ a.s.\ as $m \to \infty$),
standard asymptotic normality~\cite[see e.g.,][]{van2000asymptotic} for $M$-estimators yields:
\begin{align}
    \sqrt{m} \cdot \calI(\theta_\star)^{1/2} (\hat{\theta}_{m,T} - \theta_\star) \distconv \sfN(0, I_p), \label{eq:asymptotic_normality}
\end{align}
where $\distconv$ denotes convergence in distribution.
The condition \eqref{eq:asymptotic_normality} depends implicitly on the trajectory length $T$ through the FI matrix $\calI(\theta_\star)$. However, as the FI matrix $\calI(\theta)$
factorizes nicely across time:
\begin{align*}
    \calI(\theta) = -\E_{p_\theta}\left[ \nabla^2_\theta \log p_\theta(z_{1:T}) \right] = - \sum_{t=1}^{T} \E_{p_\theta}\left[ \nabla^2_\theta \log p_\theta(z_t \mid z_{1:t-1}) \right],
\end{align*}
we generically expect that $\calI(\theta)$ grows at least linearly with $T$, i.e., $\lambda_{\min}(\calI(\theta)) \geq \Omega(T)$. 
Hence, if we define the \emph{normalized} Fisher information matrix $\bar{\calI}(\theta) := T^{-1} \cdot \calI(\theta)$,
the result \eqref{eq:asymptotic_normality} 
implies that the limiting behavior as $m \to \infty$ scales with high probability as:
\begin{align}
    \norm{ \hat{\theta}_{m,T} - \theta_\star }^2_{\bar{\calI}(\theta_\star)} \lesssim \frac{p}{mT} \quad\text{and}\quad \norm{ \hat{\theta}_{m,T} - \theta_\star }^2 \lesssim \frac{p}{mT \cdot \lambda_{\min}(\bar{\calI}(\theta_\star))}. \label{eq:asymptotic_normality_parameter}
\end{align}
Therefore, as long as the normalized FI matrix 
provides sufficient excitation so that $\lambda_{\min}(\bar{\calI}(\theta_\star))$ does \emph{not} vanish to zero as the trajectory length $T$ increases,
then \eqref{eq:asymptotic_normality_parameter} implies that the squared parameter error decreases at a $1/(mT)$ rate, a rate which not involves \emph{all} the data points available in the training set, but as importantly is also \emph{instance-optimal}, containing instance-specific scaling through the FI matrix $\bar{\calI}(\theta_\star)$.
Hence, showing that \eqref{eq:asymptotic_normality_parameter}
holds in a non-asymptotic, finite number of trajectories 
regime under general conditions
serves as one of the main goals of this work. 

As we will discuss in detail in the remainder of this sub-section, 
finite-sample rates of the form \eqref{eq:asymptotic_normality_parameter}
are known to hold 
for the setting of least-squares regression over 
dependent covariates under various assumptions~\cite{tu2024learning,ziemann2024sharp};
unfortunately, these analysis techniques heavily
utilize the structure of the square-loss, and do not readily
extend to more general losses such as the log-loss for MLE.
Beyond the square-loss,
the most general rates comes from reductions to either (i) standard i.i.d.\ 
learning results or (ii) existing single-trajectory results;
the former yields rates exhibiting sub-optimal $1/m$ scaling,
whereas the latter inherits the single-trajectory stability assumptions
that are often unnecessary in the multi-trajectory case, and suffers from sample-deflation issues in the rates.
This motivates the need for developing a new approach for establishing
finite-sample instance-optimal rates for the multi-trajectory setting,
which we turn to in \Cref{sec:general_framework:main_results}.

\paragraph{Linear least-squares regression and linear system identification.}
One case where a non-asymptotic rate of the form \eqref{eq:asymptotic_normality_parameter}
is shown to hold in the literature is in the setting of 
linear least-squares regression over dependent covariates~\cite{tu2024learning}.
Specifically, consider the following linear dynamical system (LDS)
parameterized by $A \in \R^{d \times d}$:
\begin{align}
    z_{t+1} = A z_t + w_t, \quad w_t \sim \sfN(0, \sigma^2 I_d), \quad z_0 = 0. \label{eq:LDS}
\end{align}
For any $\theta = \vec(A) \in \R^{d^2}$, 
the FI matrix $\calI(\theta)$ takes on the form:
\begin{align*}
    \calI(\theta) = \frac{1}{\sigma^2} \left( \sum_{t=1}^{T-1} \Sigma_t(\mat(\theta)) \right) \otimes I_d , \quad \Sigma_t(A) = \E_{z_t \sim p_A}[z_tz_t^\T] = \sigma^2 \sum_{s=0}^{t-1} A^s (A^s)^\T.
\end{align*}
Hence letting $\hat{A}_{m,T} \in \argmin_{A \in \R^{d \times d}} \sum_{i=1}^{m} \sum_{t=1}^{T-1} \norm{ z_{t+1}^{(i)} - A z_t^{(i)}}^2$ denote the MLE \eqref{eq:MLE}
and $A_\star$ denote the true dynamics matrix generating the data via \eqref{eq:LDS}, 
plugging the FI matrix into \eqref{eq:asymptotic_normality_parameter} yields
\begin{align}
    \norm{ \hat{A}_{m,T} - A_\star }^2_{\Gamma_T(A_\star)} \lesssim \frac{\sigma^2 d^2}{mT} \quad\text{and}\quad  \norm{ \hat{A}_{m,T} - A_\star }^2_F \lesssim \frac{\sigma^2 d^2}{mT \cdot \lambda_{\min}(\Gamma_T(A_\star))}, \label{eq:LDS_asymptotic_normality}
\end{align}
where $\Gamma_t(A) = \frac{1}{t-1} \sum_{s=1}^{t-1} \Sigma_s(A)$.
In \cite{tu2024learning}, it is shown that
this rate \eqref{eq:LDS_asymptotic_normality} holds in expectation 
whenever $m \gtrsim d$,
and that this cut-off is sharp.
Similar results hold for the more general linear regression from LDS covariates.
We emphasize here that the rate from \eqref{eq:LDS_asymptotic_normality}
is truly a multi-trajectory phenomenon,
and is \emph{not} possible for arbitrary $A_\star$ from a single trajectory;
as shown in \cite{sarkar2019linear},
the MLE is not generally consistent in the single-trajectory setting when $A_\star$ is unstable.

\paragraph{Reductions to existing i.i.d./single-trajectory results.}
Beyond the linear least-squares regression setting, a simple generic approach
for deriving rates in the multi-trajectory setting is to invoke an existing i.i.d.\ and/or single-trajectory result. 
As an example of an i.i.d.\ reduction, using the standard
Rademacher complexity machinery for deriving risk bounds in independent settings~\cite{bartlett2002rademacher},\footnote{
For analyzing MLE, there are much sharper non-asymptotic analysis in the i.i.d.\ setting (e.g.,~\cite{spokoiny2012mle}), which do not require almost sure
bounds on log-likelihoods, contain the correct variance-optimal scaling, and also capture fast-rates in realizable settings.
We present the simplest result here to make our point clear.
}
we have that
the excess risk (in KL-divergence)
\begin{align}
    \mathsf{ER}(\theta) := \frac{1}{T}\KL{p_\star(z_{1:T})}{p_{\theta}(z_{1:T})} - \frac{1}{T} \inf_{\theta' \in \Theta}\KL{p_\star(z_{1:T})}{p_{\theta'}(z_{1:T})} \label{eq:KL_excess_risk}
\end{align}
of the MLE $\hat{\theta}_{m,T}$ satisfies with probability at least $1-\delta$,
\begin{align}
     \mathsf{ER}(\hat{\theta}_{m,T}) \leq \frac{2}{T}\sum_{t=1}^{T} \calR_m(\calG_t) + c_0 B \sqrt{\frac{\log(2/\delta)}{m}}, \label{eq:rademacher_complexity_scaling}
\end{align}
where $B$ bounds the log-likelihood $T^{-1} \cdot \log p_\theta(z_{1:T})$ a.s., $\calR_m(\calG_t)$ is the Rademacher complexity
of the function class
$\calG_t := \{ z_{1:T} \mapsto \log p_\theta(z_t \mid z_{1:t-1}) \mid \theta \in \Theta \}$,
and $c_0$ is a universal constant.
Assuming that each conditional $\abs{ \log p_\theta(z_t \mid z_{1:t-1}) } \leq O(1)$,
then we have $B \leq O(1)$ as well. We also generically expect that
$\calR_{m}(\calG_t) \leq O(\sqrt{p/m})$, which is the usual rate for parametric
function classes.
Furthermore, $\frac{1}{T}\KL{p_\star(z_{1:T})}{p_{\hat{\theta}_{m,T}}(z_{1:T})} \approx \frac{1}{2} \norm{ \theta_\star - \hat{\theta}_{m,T} }^2_{\bar{\calI}(\theta_\star)}$
asymptotically as $\hat{\theta}_{m,T} \to \theta_\star$, 
and hence the scaling w.r.t.\ $T$ in the excess risk \eqref{eq:KL_excess_risk} 
is the correct one for comparison to \eqref{eq:asymptotic_normality_parameter}.
Therefore, the general scaling for the RHS of \eqref{eq:rademacher_complexity_scaling} is of order
$\sqrt{p/m}$, i.e., the effective sample size is the number of trajectories $m$.
Note that in the realizable setting when 
$\inf_{\theta' \in \Theta}\KL{p_\star(z_{1:T})}{p_{\theta'}(z_{1:T})} =0$, 
the bound for \eqref{eq:rademacher_complexity_scaling}
can be improved to a fast-rate $p/m$ scaling with
local Rademacher complexities~\cite{bartlett2005localrademacher}.

Single-trajectory results can also be used for reductions,
by embedding the trajectories $\{z_{1:T}^{(i)}\}$ into one single trajectory
$\bar{z}_{1:mT} := ( z_{1:T}^{(1)}, \dots, z_{1:T}^{(m)})$.
For this discussion, we focus on results relying on $\beta$-mixing\footnote{
The $\beta$-mixing coefficients (cf.~\cite{bradley2005mixing,yu1994rates}) 
for $\{z_t\}_{t=1}^{\infty}$ are defined as
$\beta(k) := \sup_{j \in \N_+} \E_{z_{1:j}}[ \TV{\Pr_{z_{j+k:\infty}}(\cdot \mid z_{1:j})}{\Pr_{z_{j+k:\infty}}} ]$.
The process is called $\beta$-mixing if $\beta(k) \to 0$ as $k \to \infty$.} for concreteness,
noting that our discussion also applies to results that
rely on other definitions of mixing (e.g., $\phi$-mixing) 
in the literature.
We also assume the process $\{z_t\}$ is Markovian, as this makes the reduction
simpler to state.
Because $z_{1:T}^{(i)} \perp z_{1:T}^{(j)}$ whenever $i \neq j$, 
then we have that the $\beta$-mixing coefficients $\bar{\beta}(k)$ of $\{\bar{z}_t\}$ 
satisfy $\bar{\beta}(k) = \beta(k) \cdot \ind\{ k < T \}$, where $\beta(k)$ are the
$\beta$-mixing coefficients of $\{z_t\}$.
Hence, the embedded trajectory $\{\bar{z}_t\}$ is trivially $\beta$-mixing with mixing-time equal to $T$ without any assumption on $\beta(k)$.
Using this mixing-time and invoking a single-trajectory $\beta$-mixing result, such as from \cite{kuznetsov2017generalization},
without any further assumption on $\beta(k)$ yields a similar result to 
\eqref{eq:rademacher_complexity_scaling}.
However, if we further assume that $\beta(k) \leq C \exp( - \rho k )$ for some $\rho > 0$, and that $T / (2\kappa) \in \N_+$ for $\kappa := \ceil{\rho^{-1} \log(C mT/\delta)}$,
then we have the improved result: with probability at least $1-\delta$,
\begin{align}
    \mathsf{ER}(\hat{\theta}_{m,T}) \leq \frac{c_0'}{2\kappa}\sum_{j=1}^{2\kappa} \bar{\calR}_{mT/\kappa}^j + c_1' B \sqrt{\frac{\kappa \log(c_2'/\delta)}{mT}}, \label{eq:mixing_scaling}
\end{align}
where $\bar{\calR}_{mT/\kappa}^j$ denotes the \emph{de-coupled} Rademacher complexity:
\begin{align*}
    \bar{\calR}_{mT/\kappa}^j := \E \sup_{\theta \in \Theta} \frac{\kappa}{mT} \sum_{i=1}^{m} \sum_{\ell=1}^{\kappa} \e_{i,\ell} \log p_\theta( \tilde{z}^{(i)}_{(\ell-1)2\kappa + j} \mid \tilde{z}^{(i)}_{(\ell-1)2\kappa + j - 1}), \quad j \in [2\kappa],
\end{align*}
with the pair $(\tilde{z}^{(i)}_{(\ell-1)2\kappa + j - 1}, \tilde{z}^{(i)}_{(\ell-1)2\kappa + j})$ drawn from the same distribution as $(z_{(\ell-1)2\kappa + j - 1}, z_{(\ell-1)2\kappa + j})$, but \emph{independently} across $i \in [m]$ and $\ell \in [\kappa]$.
Similar to before, we generically expect that $\bar{\calR}^j_{mT/\kappa}$ scales as order $\sqrt{\kappa p/(mT)}$. Hence, the general scaling of the RHS of \eqref{eq:mixing_scaling}
is of order $\sqrt{\kappa p/(mT)}$.
Furthermore, as with the i.i.d.\ reduction, in the realizable setting local Rademacher arguments can also be used to improve the scaling of \eqref{eq:mixing_scaling} to the fast-rate $\kappa p / (mT)$.
This is an improvement over \eqref{eq:rademacher_complexity_scaling},
as the effective sample size increases from $m$ to $mT/\kappa$;
however, this sample size still remains deflated by $\kappa$, 
as a consequence of the standard blocking technique used for de-coupling.

\paragraph{Regression with square-loss.}
A recent line of work \cite{ziemann2022learning,ziemann2024sharp}
has shown that the sample size deflation described previously
can be removed in the special case of non-linear regression with the square-loss, in both parametric and
non-parametric regimes.
To make their results concrete, we consider the following 
parametric family of distribution $\calP$ over trajectories in $\R^d$:
\begin{align}
    z_{t+1} = f_{\theta}(z_t) + w_t, \quad w_t \sim \sfN(0, \sigma^2 I_d), \label{eq:parametric_problem}
\end{align}
coupled with the non-linear
least-squares estimator 
$\hat{\theta}_{m,T} \in \argmin_{\theta \in \Theta} \sum_{i=1}^{m} \sum_{t=1}^{T-1} \norm{z_{t+1}^{(i)} - f_\theta(z_t^{(i)})}^2$ 
with $\Theta \subseteq \R^p$,
which is precisely the MLE estimator for \eqref{eq:parametric_problem}.
We now specialize the main result of \cite[Theorem 3.1]{ziemann2024sharp} 
to the problem \eqref{eq:parametric_problem}.
Suppose that the following assumptions hold:\footnote{We state a clear set of assumptions, but not the most minimal, as \cite[Theorem 3.1]{ziemann2024sharp} is stated in broad generality.}
\begin{enumerate}[label=(\alph*)]
    \item \emph{(Realizable):} The process $\{z_t\}$ is generated by \eqref{eq:parametric_problem} 
    for some $\theta_\star \in \Theta$.
    
    \item \emph{(Stationary process):}
    The process $\{z_t\}$ has a stationary measure $\nu$, and $z_1 \sim \nu$.

    \item \emph{(Weakly sub-Gaussian):} The function class 
    $\calF' := \{ f_{\theta_1} - f_{\theta_2} \mid \theta_1, \theta_2 \in \Theta \}$ satisfies a \emph{weak-sub-Gaussian} condition~\cite[cf.][Def. 2.1]{ziemann2024sharp}:
    there exists $\eta \in (0, 1]$ and $L \geq 1$ such that $\norm{f}_{\Psi_p} \leq L \norm{f}_{\calL^2(\nu)}^\eta$ for all $f \in \calF'$, where $\norm{f}_{\Psi_p} := \sup_{k \in \N_+} k^{-1/p} \norm{f}_{\calL^p(\nu)}$.

    \item \emph{(Function class regularity):} 
    For every $x \in \R^d$, the map $\theta \mapsto f_\theta(x)$ is $L(x)$-Lipschitz, with $\norm{L(x)}_{\calL^2(\nu)} < \infty$.
    Furthermore, the set $\Theta \subseteq \R^p$ is a bounded set.

    \item \emph{(Burn-in):} Either (i) $mT \gtrsim \mathrm{poly}_\eta(T, p)$ or (ii)
    $\{z_t\}$ is $\beta$-mixing with $\beta(k) \leq C\exp(-\rho k)$
    and $T \gtrsim \kappa := \ceil{\rho^{-1} \log(C mT/\delta)}$, $m T \gtrsim \mathrm{poly}_\eta(\kappa, p)$, where 
    $\mathrm{poly}_\eta(\cdot)$ denotes that the polynomial dependence is a function of $\eta$.
    
\end{enumerate}
Then, the MLE estimator $\hat{\theta}_{m,T}$ satisfies with probability at least $1-\delta$,
\begin{align}
    \sigma^2 \cdot \mathsf{ER}(\hat{\theta}_{m,T}) \asymp \norm{ f_{\hat{\theta}_{m,T}} - f_{\theta_\star} }^2_{\calL^2(\nu)} \lesssim \frac{\sigma^2_{\mathrm{prox}}(p + \log(1/\delta))}{mT}, \label{eq:excess_risk_rate}
\end{align}
where $\sigma^2_{\mathrm{prox}} \geq \sigma^2$ is a variance proxy
which is determined from the specific choice of $(p, \eta)$ in Assumption (c). In the case where $p = \infty$, we have $\sigma^2_{\mathrm{prox}} = \sigma^2$.
Furthermore, when $p < \infty$ and $\eta = 1$, we
have that $\sigma^2_{\mathrm{prox}} = C_p \sigma^2 + o_{mT}(1)$
by the martingale Rosenthal inequality (cf.~\Cref{thm:rosenthal}), 
where $C_p$ is a constant only depending on $p$.
On the other hand, when $p < \infty$ and $\eta < 1$, the
precise relationship between $\sigma^2_{\mathrm{prox}}$ and
$\sigma^2$ is more complex.
We observe that the rate \eqref{eq:excess_risk_rate} is order-wise optimal from asymptotic normality (up to the variance proxy $\sigma^2_{\mathrm{prox}})$;
importantly, the rate \eqref{eq:excess_risk_rate}
has the correct dependence on the entire dataset size $mT$,
compared with the deflated rate $mT/\kappa$ from
the previous single-trajectory reduction.

However, Assumptions (a)-(e) can be restrictive and/or challenging to verify.
We first note that the stationary process Assumption (b) 
can be removed by using \cite[Theorem 4.1]{ziemann2022learning}
instead of \cite[Theorem\ 3.1]{ziemann2024sharp},
although the downside of this is that the
weakly sub-Gaussian Assumption (c) is then replaced
with a trajectory-level hyper-contractivity condition (cf.~\cite[Def.\ 4.1]{ziemann2022learning}) which is more
challenging to verify.
On the other hand, while the weakly sub-Gaussian Assumption (c) holds
broadly if $f_\theta$ is bounded and smooth in its input (cf.~\cite[Prop.\ 4.1]{ziemann2024sharp}), the constants $(L, \eta)$ provided %
depend poorly on the process dimension $d$, 
which yields burn-in times for $m, T$ that can depend exponentially in $d$; sharp control on the $(L, \eta)$ constants is
only currently available for simple function classes, e.g., 
linear function classes.

\paragraph{Summary.} The finite-sample behavior
of the MLE in multi-trajectory settings is currently most broadly
available through reduction to either an existing i.i.d.\ or single-trajectory result. In either case, there is a gap between the resulting bound
(cf.~\eqref{eq:rademacher_complexity_scaling} for the i.i.d.\ reduction and \eqref{eq:mixing_scaling} for the single-trajectory reduction)
compared with the optimal bound \eqref{eq:asymptotic_normality_parameter}
in terms of effective sample sizes.
In the case of least-squares regression (both for linear and more general parametric models), however, 
the finite-sample rate \eqref{eq:LDS_asymptotic_normality}
for linear regression and \eqref{eq:excess_risk_rate} 
for more general parametric regression matches the CLT-optimal
bound up to constant factors in the former, and up to
a variance-proxy factor in the latter.
This naturally raises the question whether optimal finite-sample
rates can be derived beyond the square-loss setting.
The proof techniques used for analyzing the square-loss (e.g., self-normalized martingales, small-ball inequalities) take advantage of either the closed-form nature of the 
linear regression solution, 
or specific properties of the square-loss such as the offset basic inequality~\cite[see e.g.,][]{liang2015offsetcomplexity},
and hence
do not readily generalize. 
This motivates the need for a different approach for establishing error bounds of the form \eqref{eq:asymptotic_normality_parameter} in more general settings.

\subsection{Analyzing MLE via Localization in Hellinger Distance}
\label{sec:general_framework:main_results}

We next develop a set of tools and a general five-step framework
for analyzing the MLE over a diverse set of problems.
The roadmap for the remainder of this section is as follows. 
We first develop tools in \Cref{sec:hellinger_distance_trajectory} 
to control the Hellinger distance
of the MLE solution to the true solution in terms of their length-$T$ trajectory (path) measures.
Next, we study in \Cref{sec:hellinger_fisher_equiv} 
how we can localize the Hellinger distance so that it approximately behaves like a
weighted Euclidean norm over the parameters, where the weight is determined by the Fisher information matrix at optimality.
Importantly, given sufficient trajectory-level excitation, the FI matrix scales with
the trajectory length $T$, providing the correct scaling with length of each trajectory.
Building on these mathematical tools, in \Cref{sec:twostatemc} we work through a simple illustrative example
combining these tools
to derive a sharp rate for parameter recovery in a two-state Markov chain.
Finally, we present our general Hellinger localization framework in \Cref{sec:hellinger_localization_framework}.

\paragraph{Divergence measures.}
For two measures $p, q$ over the same probability space, we define the Total-Variation (TV) distance, Hellinger distance, and Kullback-Leibler (KL) divergence as:
\begin{align*}
    \TV{p}{q} := \frac{1}{2} \int \abs{p - q}\, \rmd\mu, \quad \Hg{p}{q} := \sqrt{\int \left( \sqrt{p} - \sqrt{q} \right)^2 \rmd\mu}, \quad \KL{p}{q} := \E_{p}\left[ \log\frac{p}{q} \right].
\end{align*}
Note that the last definition of KL divergence require the absolute continuity condition $p \ll q$.
For what follows, we will often overload notation and write
e.g., $\Hg{\theta_1}{\theta_2} = \Hg{p_{\theta_1}}{p_{\theta_2}}$
for $\theta_1,\theta_2 \in \Theta$
(and similarly for TV distance and KL divergences).

\subsubsection{Control of Trajectory Measures in Hellinger Distance}
\label{sec:hellinger_distance_trajectory}

Our main approach is based on techniques used for studying
density estimation with maximum-likelihood. 
To set the stage for what follows, we first state a prototypical non-asymptotic
result from the study of maximum-likelihood estimators. 
The following instantiation is from \cite{foster2024behaviorcloning} and applied directly to
our problem setting \eqref{eq:MLE},
although it traces its roots back to the work of \cite{geer2000empirical,zhang2006densityestimation}.
Similar instantiations of the following result can also be found in more recent works~\cite{agarwal2020flambe,ge2024pretraining,ziemann2024short}.
\begin{mythm}[{cf.~\cite[Proposition B.1]{foster2024behaviorcloning}}]\label{thm:hellinger_bound_mle_og}
We have with probability at least $1-\delta$,
$$
    \HgSq{\hat{p}_{m,T}}{p_\star} \leq \inf_{\e > 0} \left\{ \frac{6 \log(2\mathcal{N}_\infty(\mathcal{P}, \e)/\delta)}{m} + 4\e \right\},
$$
where $\mathcal{N}_\infty(\mathcal{P},\e)$ is the $\e$-covering number of $\mathcal{P}$ in the max divergence.\footnote{
Specifically, a set $\calP' \subseteq \calP$ is an $\e$-covering in max divergence if for every $p \in \calP$ there exists a $p' \in \calP'$ such that 
for a.e.\ $z_{1:T} \in \sfZ^T$, $\log(p(z_{1:T})/p'(z_{1:T})) \leq \e$.
The quantity $\calN_\infty(\calP, \e)$ denotes the cardinality of the smallest such $\e$-covering.
}
\end{mythm}

\Cref{thm:hellinger_bound_mle_og} is a powerful result
in that it controls the Hellinger divergence of the \emph{trajectory (path)} distributions between the MLE estimator $\hat{p}_{m,T}(z_{1:T})$ and the true
data-generating trajectory distribution $p_\star(z_{1:T})$
at a $1/m$ rate, under fairly minimal assumptions
on $\calP$. In fact, the only assumption made on $\calP$ is that its max divergence covering number is bounded.
However, powerful as this result may be, extracting trajectory information out of the Hellinger divergence
in order to obtain $1/(mT)$ rates
is non-trivial, as the Hellinger distance does \emph{not} in general tensorize nicely
over non-product measures, unlike the KL-divergence.

Fortunately, some form of tensorization is indeed possible
when $\hat{p}_{m,T}$ is close enough to $p_\star$.
In particular, by an asymptotic argument~\cite[see e.g.][Theorem 7.23]{polyanskiy2025information}
for $\theta_0, \theta_1 \in \Theta$,
the following local expansion holds:
\begin{align}
    \HgSq{\theta_0}{\theta_1} = \frac{1}{4} \norm{\theta_0 -\theta_1}^2_{\calI(\theta_0)} + o(\norm{\theta_0-\theta_1}^2). \label{eq:hellinger_local_asymptotic_expansion}
\end{align}

Combining \eqref{eq:hellinger_local_asymptotic_expansion}
with \Cref{thm:hellinger_bound_mle_og} yields
a bound which resembles that of the CLT \eqref{eq:asymptotic_normality_parameter}.
Our key result is to quantify the region for which 
such a local expansion \eqref{eq:hellinger_local_asymptotic_expansion} holds, using a second-order Taylor expansion argument.
The argument proceeds in two steps.
Due to the nature of Taylor's theorem, we first need to uniformly control the performance of parameters within $\mathrm{conv}\{\hat{\theta}_{m,T}, \theta_\star\}$,\footnote{We define $\mathrm{conv}\{\theta_0, \theta_1\} := \{ (1-s) \theta_0 + s \theta_1 \mid s \in [0, 1]\}$.} which we do via a star-shaped variation of \Cref{thm:hellinger_bound_mle_og}.
We then Taylor expand the squared Hellinger distance
and characterize the necessary radius conditions for
\eqref{eq:hellinger_local_asymptotic_expansion} to hold.

To proceed, we first require the definition of an $\e$-cover in 
Hellinger distance.
\begin{mydef}[Hellinger cover]
\label{def:hellinger_cover}
A set $\calP' \subseteq \calP$ is an $\e$-covering of $\calP$ in 
Hellinger distance if for every $p \in \calP$, there exists a $p' \in \calP'$ such that
$\Hg{p}{p'} \leq \e$.
The $\e$-covering number of $\calP$ in Hellinger distance, denoted $\calN_H(\calP, \e)$, is defined as the cardinality of the smallest such $\e$-covering.
\end{mydef}
One issue which arises with either Hellinger or squared Hellinger distance is that it is not convex in its parameterization, i.e., in general we have neither $\theta \mapsto \Hg{\theta}{\theta_1}$ 
nor $\theta \mapsto \HgSq{\theta}{\theta_1}$
is convex for a fixed $\theta_1$;
$f$-divergences are jointly convex
in the space of \emph{probability measures}, but not necessarily the specific parameterization.
Thus, it will be necessary to consider another type of divergence.

For what follows, given $\theta_0, \theta_1 \in \Theta$, we define $\calI(\theta_0, \theta_1)$ as the matrix:
\begin{align*}
    \calI(\theta_0, \theta_1) := \int_0^1 \calI(\theta(s)) \rmd s, \quad \theta(s) := (1-s)\theta_0 + s \theta_1.
\end{align*}
Note that $\calI(\theta_0, \theta_1)$ is symmetric, i.e., $\calI(\theta_0, \theta_1) = \calI(\theta_1, \theta_0)$. 
We also assume there exists a positive definite matrix $\calI_{\max}$ such that
$\calI(\theta) \preccurlyeq \calI_{\max}$ for all $\theta \in \Theta$.
We use this to define both a symmetric averaged Fisher Information, and a max Fisher Information divergence measure:
\begin{align}
    \FI{p_{\theta_0}}{p_{\theta_1}} := \norm{ \theta_0 - \theta_1 }_{\calI(\theta_0, \theta_1)}, \quad \FIMax{p_{\theta_0}}{p_{\theta_1}} := \norm{\theta_0 - \theta_1}_{\calI_{\max}}.
\end{align}
The relationship $\FI{p_{\theta_0}}{p_{\theta_1}} \leq \FIMax{p_{\theta_0}}{p_{\theta_1}}$ is 
clear by definition.
Furthermore, the max FI measure exhibits the necessary
convexity of $\theta \mapsto \FIMax{\theta}{\theta_1}$ for all fixed $\theta_1$,
via the convexity of the weighted $\ell_2$ norm.
We now show the following connection that these two FI distances dominate the Hellinger distance, with proof deferred to \Cref{sec:appendix:framework}.
\begin{restatable}{myprop}{HellingerFIUpperBound}
\label{prop:hellinger_FI_upper_bound}
For any $\theta_0, \theta_1 \in \Theta$ such that $\mathrm{conv}(\theta_0, \theta_1) \subseteq \Theta$, we have:
\begin{align*}
    \Hg{p_{\theta_0}}{p_{\theta_1}} \leq \frac{1}{2} \FI{p_{\theta_0}}{p_{\theta_1}} \leq \frac{1}{2} \FIMax{p_{\theta_0}}{p_{\theta_1}}.
\end{align*}
\end{restatable}

Parallel to \Cref{def:hellinger_cover}, we also define a covering
in terms of the max FI divergence as follows.
\begin{mydef}[Max FI cover]
\label{def:max_FI_cover}
A set $\calP' \subseteq \calP$ is an $\e$-covering of $\calP$ in the max Fisher Information divergence if for every $p_\theta \in \calP$, there exists a $p_{\theta'} \in \calP'$ such that $\norm{\theta - \theta'}_{\calI_{\max}} \leq \e$.
The $\e$-covering number of $\calP$ in the max Fisher Information divergence, denoted $\calN_{\calI_{\max}}(\calP, \e)$, is
defined as the cardinality of the smallest such $\e$-covering.
\end{mydef}
Using this definition of $\e$-covering, we next introduce a \emph{discretized} version of the MLE estimator \eqref{eq:MLE}.
For $\e \geq 0$, we let $\calP_{\e} \subseteq \calP$ denote a minimal $\e$-covering of $\calP$ in either the Hellinger divergence (cf.~\Cref{def:hellinger_cover}) or max FI divergence (cf.~\Cref{def:max_FI_cover});\footnote{If there is not a unique minimal $\e$-covering, then we break ties in an arbitrary way so that $\calP_\e$ is not ambiguous.}
the specific divergence will be clear from context.
We then define the MLE over this set as:
\begin{align}
    \hat{p}^{\e}_{m,T} \in \argmax_{p \in \calP_{\e}} \sum_{i=1}^{m} \log p(z^{(i)}_{1:T}).\label{eq:discrete_MLE}
\end{align}
We also denote the parameters $\hat{\theta}^{\e}_{m,T} \in \Theta$ so that $\hat{p}^{\e}_{m,T} = p_{\hat{\theta}^{\e}_{m,T}}$. We further introduce the definition of the log-concavity of a parameterization of a density class.
\begin{mydef}
\label{def:log_concave}
We say that $\calP$ is \emph{log-concave} if $\Theta$ is convex, and furthermore for every
$\theta_0, \theta_1 \in \Theta$, $s \in [0, 1]$, and
$\mu^{\otimes T}$-a.e.\ $z \in \sfZ^T$, 
\begin{align*}
    \log p_{s \theta_0 + (1-s) \theta_1}(z) \geq s \log p_{\theta_0}(z) + (1-s) \log p_{\theta_1}(z).
\end{align*}
That is, for $\mu^{\otimes T}$-a.e.\ $z \in \sfZ^T$, the function $\theta \mapsto \log p_\theta(z)$ is concave over $\Theta$.
\end{mydef}
Finally, we define the max FI-diameter of $\Theta$ as:
\begin{align*}
    \mathrm{diam}(\Theta) := \sup_{\theta_0, \theta_1 \in \Theta} \norm{\theta_0 - \theta_1}_{\calI_{\max}}.
\end{align*}
We are now in a position to state the main result of the section. 
\begin{mythm}
\label{thm:hellinger_bound_MLE}
Fix $\delta \in (0, 1)$
and resolution $\e \in [0, \delta/(2\sqrt{2m})]$.
We have the following:
\begin{enumerate}[label=\textbf{(\alph*).}]
\item With probability at least $1-\delta$ over the data $\calD_{m,T}$,
the Hellinger divergence discretized MLE estimator $\hat{\theta}^{\e}_{m,T}$ satisfies:
\begin{align}
    \HgSq{\hat{\theta}^{\e}_{m,T}}{\theta_\star} \leq  \frac{4\log(2 \calN_{H}(\calP, \e)/\delta)}{m} + 2\e^2. \label{eq:hellinger_bound_MLE_discrete}
\end{align}
Furthermore, the same bound \eqref{eq:hellinger_bound_MLE_discrete}
holds for the max FI divergence discretized MLE estimator
with $\calN_{H}(\calP, \e)$ replaced with $\calN_{\calI_{\max}}(\calP, \e)$.
\item If we further assume that $\mathcal{P}$ is log-concave (cf.~\Cref{def:log_concave}), then
with probability at least $1-\delta$ over the data $\calD_{m,T}$,
the max FI divergence discretized MLE estimator $\hat{\theta}^{\e}_{m,T}$ satisfies:
\begin{align}
    \sup_{s \in [0, 1]} \HgSq{(1-s) \theta_\star + s \hat{\theta}^{\e}_{m,T}}{\theta_\star} \leq \inf_{\eta>0}\left\{\frac{6}{m} \log\left(\frac{2 \calN_{\calI_{\max}}(\calP, \e) }{\delta} \bigceil{\frac{1}{2\eta}} \right) + \frac{3\eta^2}{4} \mathrm{diam}^2(\Theta) + 3\e^2\right\}.\label{eq:fisher_bound_MLE_discrete}
\end{align}
\end{enumerate}
\end{mythm}
Before we turn to the proof of \Cref{thm:hellinger_bound_MLE}, several remarks are in order.
\begin{rmk}
\label{rmk:hellinger_new_vs_OG}
One key difference between 
\Cref{thm:hellinger_bound_mle_og}
and \Cref{thm:hellinger_bound_MLE} is that the former applies directly to the 
MLE estimator $\hat{\theta}_{m,T}$ \eqref{eq:MLE} over $\calP$,
whereas the latter applies to the discretized
MLE estimator $\hat{\theta}^\e_{m,T}$
\eqref{eq:discrete_MLE} over $\calP_\e$.
In practice there is no difference
between these two estimators at a sufficiently small $\e$ below floating point resolution.
However, from a theoretical perspective, the discrete estimator seems to exhibit more favorable properties than the exact MLE estimator. One of these properties is
allowing one to relax the covering requirement on
$\calP$ to either Hellinger \eqref{eq:hellinger_bound_MLE_discrete}
or max FI-divergence \eqref{eq:fisher_bound_MLE_discrete}, both which are less stringent than the max divergence covering in \Cref{thm:hellinger_bound_mle_og}, 
which requires an almost sure bound on the log-density ratio.
This is an important relaxation, as it allows us to handle trajectory distributions where the paths $z_{1:T}$ are not bounded almost surely;
in such situations the Hellinger/max-FI divergences can still be finite as we will see in the sequel.
We leave open the question of whether
rates of the form \eqref{eq:hellinger_bound_MLE_discrete} and 
\eqref{eq:fisher_bound_MLE_discrete} are possible for $\hat{p}_{m,T}$ without relying
on max divergence coverings, noting that 
some extra tail conditions on $\calP$ would be needed
to control the behavior of the empirical log likelihood
$\frac{1}{m} \sum_{i=1}^{m} \log p(z^{(i)}_{1:T})$.
\end{rmk}

\begin{rmk}
The key difference between 
\eqref{eq:hellinger_bound_MLE_discrete} and 
\eqref{eq:fisher_bound_MLE_discrete}
is that the former only controls
$\Hg{ \hat{\theta}^\e_{m,T} }{ \theta_\star }$,
whereas the latter controls 
$\Hg{ \theta }{ \theta_\star }$ along the entire
ray $\theta \in \mathrm{conv}\{ \hat{\theta}^\e_{m,T}, \theta_\star \}$.
Note that since in general neither Hellinger
nor squared Hellinger distance is convex in the 
\emph{parameter space},
the former in general does \emph{not} imply the latter.
Thus, \eqref{eq:fisher_bound_MLE_discrete} is a strictly stronger conclusion than \eqref{eq:hellinger_bound_MLE_discrete},
and therefore requires a stronger set of assumptions (e.g., log-concavity of $\calP$).
As we will see in the sequel, the conclusion
\eqref{eq:fisher_bound_MLE_discrete} will
play an important role in allowing $m \gtrsim \mathrm{polylog}(T)$ instead of $m \gtrsim T \cdot \mathrm{polylog}(T)$ minimum number of trajectories
for our CLT rates to hold. 
\end{rmk}

\begin{proof}[Proof of \Cref{thm:hellinger_bound_MLE}]
The proof follows the general structure of \cite[Proposition B.1]{foster2024behaviorcloning}, but includes a few crucial modifications.
Before we begin, we state the following upper bound on squared Hellinger distance which holds generally for two distributions $p, q$, which follows from the inequality $\log(1+x) \leq x$ for $x > -1$:
\begin{align}
    \frac{1}{2} \HgSq{p}{q} &\leq - \log\left( 1 - \frac{1}{2} \HgSq{p}{q} \right) \nonumber \\
    &= - \log\left( \E_{p} \left[\exp\left( -\frac{1}{2} \log\frac{p}{q} \right) \right] \right) = - \log\left( \E_{q} \left[\exp\left( -\frac{1}{2} \log\frac{q}{p} \right) \right] \right). \label{eq:squared_hellinger_upper_bound}
\end{align}

\paragraph{(a).}
Let $\calP_\e \subseteq \calP$ denote a minimal $\e$-covering of $\calP$ in the Hellinger distance (cf.~\Cref{def:hellinger_cover}).
Let us abbreviate $\hat{p}^\e := \hat{p}^\e_{m,T}$, and
for $p \in \calP$ let
$\varphi_\e[p] \in \calP_\e$ denote the closest element in the Hellinger cover, i.e., $\Hg{\varphi_\e[p]}{p} \leq \e$.
We first consider a hypothetical scenario where each $z^{(i)} := z^{(i)}_{1:T}$ in $\calD_{m,T}$ is drawn i.i.d.\ from $p_\star^\e := \varphi_\e[p_\star]$ instead of $p_\star$.
By combining \eqref{eq:squared_hellinger_upper_bound} and
\Cref{prop:logMGF_bound} with a union bound over $\calP_\e$,
we have with probability at least $1-\delta/2$ over $(p_\star^\e)^{\otimes m}$,
\begin{align*}
    \frac{1}{2} \HgSq{\hat{p}^\e}{p^\e_\star} \leq - \log\left( \E_{p_\star^\e} \left[\exp\left( -\frac{1}{2} \log\frac{p_\star^\e}{\hat{p}^\e} \right) \right] \right) &\leq \frac{1}{2m} \sum_{i=1}^{m} \log \frac{p_\star^\e(z^{(i)})}{\hat{p}^\e(z^{(i)})} + \frac{1}{m}\log\left(\frac{2\abs{\calP_\e}}{\delta}\right) 
    \leq \frac{1}{m}\log\left(\frac{2\abs{\calP_\e}}{\delta}\right),
\end{align*}
where the last inequality is since $\hat{p}^\e$ is the MLE over
$\calP_\e$ and $p^\e_\star \in \calP_\e$.
On the other hand, by triangle inequality for Hellinger distance
followed by the inequality $(a+b)^2 \leq 2(a^2+b^2)$ for $a,b \in \R$,
\begin{align*}
    \HgSq{\hat{p}^\e}{p_\star} \leq 2\HgSq{\hat{p}^\e}{p^\e_\star} + 2\HgSq{p^\e_\star}{p_\star} \leq \frac{4}{m}\log\left(\frac{2\abs{\calP_\e}}{\delta}\right) + 2 \e^2.
\end{align*}
Hence, we have shown that:
\begin{align*}
    \Pr_{\calD_{m,T} \sim (p^\e_\star)^{\otimes m}}\left\{ \HgSq{\hat{p}^\e[\calD_{m,T}]}{p_\star} > \frac{4}{m}\log\left(\frac{2\abs{\calP_\e}}{\delta}\right) + 2 \e^2 \right\} \leq \delta/2,
\end{align*}
where $\hat{p}^\e[\calD_{m,T}]$ is notation to emphasize that
$\hat{p}^\e$ is a function of the data $\calD_{m,T}$.
Recall that
$\TV{p}{q} \leq \Hg{p}{q}$ for two measures $p,q$ \cite[cf.][Section 7.3]{polyanskiy2025information}.
Hence we can change measure
between $\calD_{m,T} \sim (p^\e_\star)^{\otimes m}$ 
and $\calD_{m,T} \sim p_\star^{\otimes m}$ as follows:
\begin{align*}
    \Pr_{\calD_{m,T} \sim p_\star^{\otimes m}}\left\{ \HgSq{\hat{p}^\e[\calD_{m,T}]}{p_\star} > \frac{4}{m}\log\left(\frac{2\abs{\calP_\e}}{\delta}\right) + 2 \e^2 \right\} &\leq \TV{p_\star^{\otimes m}}{(p_\star^\e)^{\otimes m}} + \delta/2 \\
    &\leq \Hg{p_\star^{\otimes m}}{(p^\e_\star)^{\otimes m}} + \delta/2 \\
    &\leq \delta,
\end{align*}
where the last inequality follows from
\Cref{prop:hellinger_tensorization} 
since we have $\Hg{p_\star}{p_\star^\e} \leq \e \leq \delta/(2\sqrt{2m})$.
This establishes \eqref{eq:hellinger_bound_MLE_discrete}
for the Hellinger divergence discretized $\hat{\theta}^{\e}_{m,T}$.
The proof for the max FI divergence discretized estimator is nearly identical,
and hence omitted.

\paragraph{(b).}

For $p \in \calP$, let $\varphi_{\e}[p] \in \calP_{\e}$ denote the closest element in the
max FI-divergence $\e$-covering $\calP_{\e} \subseteq \calP$,
so that $\FIMax{\varphi_\e[p]}{p} \leq \e$ for all $p \in \calP$.
We define the following parameterization of $\mathrm{conv}\{ \theta, \hat{\theta}^\e_{m,T} \}$:
\begin{align*}
    \hat{\theta}^{\e}(s; \theta) &:= (1-s) \theta + s \hat{\theta}^{\e}_{m,T}.
\end{align*}
We also abbreviate $\hat{\theta}^{\e} = \hat{\theta}^{\e}_{m,T}$
and $\theta_\star^{\e} = \varphi_{\e}[\theta_\star]$.
Next, we let $s_1, \dots, s_N \in [0, 1]$ be a minimal $\eta$-covering of 
$[0, 1]$ in absolute value.
For any $s \in [0, 1]$,
letting $s_{\eta}$ denote its nearest element in the cover,
by the triangle inequality for Hellinger distance:
\begin{align}
    \Hg{\hat{\theta}^{\e}(s; \theta_\star)}{\theta_\star} &\leq \Hg{\hat{\theta}^{\e}(s; \theta_\star)}{\theta^{\e}_\star} + \Hg{\theta^{\e}_\star}{\theta_\star} \nonumber \\
    &\leq \Hg{\hat{\theta}^{\e}(s; \theta_\star^{\e})}{\theta_\star^{\e}} + \Hg{\hat{\theta}^{\e}(s; \theta_\star^{\e})}{\hat{\theta}^{\e}(s; \theta_\star)} +  \Hg{\theta^{\e}_\star}{\theta_\star} \nonumber \\
    &\leq \Hg{\hat{\theta}^{\e}(s_{\eta}; \theta_\star^{\e})}{\theta_\star^{\e}} + 
    \Hg{\hat{\theta}^{\e}(s; \theta_\star^{\e})}{\hat{\theta}^{\e}(s_{\eta}; \theta_\star^{\e})} \nonumber \\
    &\qquad +
    \Hg{\hat{\theta}^{\e}(s; \theta_\star^{\e})}{\hat{\theta}^{\e}(s; \theta_\star)} +  \Hg{\theta^{\e}_\star}{\theta_\star} \nonumber \\
    &\leq \Hg{\hat{\theta}^{\e}(s_{\eta}; \theta_\star^{\e})}{\theta_\star^{\e}} + 
    \frac{1}{2}\norm{\hat{\theta}^{\e}(s; \theta_\star^{\e})-\hat{\theta}^{\e}(s_{\eta}; \theta_\star^{\e})}_{\calI_{\max}} &&\text{[using \Cref{prop:hellinger_FI_upper_bound}]} \nonumber \\
    &\qquad +
    \frac{1}{2}\norm{\hat{\theta}^{\e}(s; \theta_\star^{\e})-\hat{\theta}^{\e}(s; \theta_\star)}_{\calI_{\max}} + \frac{1}{2}\norm{\theta^{\e}_\star - \theta_\star}_{\calI_{\max}} \nonumber \\
    &\leq \Hg{\hat{\theta}^{\e}(s_{\eta}; \theta_\star^{\e})}{\theta_\star^{\e}} + \frac{\abs{s-s_{\eta}}}{2} \norm{ \hat{\theta}^{\e} - \theta_\star^{\e} }_{\calI_{\max}} + \e \nonumber \\
    &\leq \Hg{\hat{\theta}^{\e}(s_{\eta}; \theta_\star^{\e})}{\theta_\star^{\e}} + \frac{\eta}{2} \mathrm{diam}(\Theta) + \e. \label{eq:hellinger_conv_triangle_eq}
\end{align}

As in the proof of (a), we first consider a hypothetical scenario where each $z^{(i)} := z^{(i)}_{1:T}$ in $\calD_{m,T}$ is drawn i.i.d.\ from $p_\star^{\e} := p_{\theta_\star^{\e}}$.
By combining \eqref{eq:squared_hellinger_upper_bound} and
\Cref{prop:logMGF_bound} with a union bound over both $\calP_\e$ and $\{s_k\}_{k=1}^{N}$,
we have with probability at least $1-\delta/2$ over $(p_\star^{\e})^{\otimes m}$,
abbreviating $\hat{\theta}^{\e}(s_{\eta}) := \hat{\theta}^{\e}(s_{\eta}; \theta_\star^{\e})$,
\begin{align}
    \frac{1}{2} \HgSq{\hat{\theta}^{\e}(s_{\eta})}{\theta_\star^{\e}} &\leq - \log\left( \E_{p_\star^{\e}} \left[\exp\left( -\frac{1}{2} \log\frac{p_\star^{\e}}{p_{\hat{\theta}^{\e}(s_{\eta})}} \right) \right] \right) \nonumber \\
    &\leq \frac{1}{2m} \sum_{i=1}^{m} \log \frac{p_\star^{\e}(z^{(i)})}{p_{\hat{\theta}^{\e}(s_{\eta})}(z^{(i)})} + \frac{1}{m}\log\left(\frac{2\abs{\calP_{\e}}}{\delta} \bigceil{\frac{1}{2\eta}} \right) 
    \leq \frac{1}{m}\log\left(\frac{2\abs{\calP_{\e}}}{\delta} \bigceil{\frac{1}{2\eta}} \right), \label{eq:hellinger_conv_ERM_bound}
\end{align}
where the last inequality holds from the following arguments. First note that log-concavity of $\calP$ means that for $\mu^{\otimes T}$ a.e.\ $z \in \sfZ^T$,
$- \log p_{\hat{\theta}^{\e}(s_{\eta})}(z) \leq -(1-s_{\eta}) \log p_{\theta_\star^{\e}}(z) - s_{\eta} \log p_{\hat{\theta}^{\e}}(z)$.
Hence,
\begin{align*}
    \log\frac{p_\star^\e(z)}{p_{\hat{\theta}^{\e}(s_{\eta})}(z)} &= \log p_\star^\e(z) - \log p_{\hat{\theta}^{\e}(s_{\eta})}(z) \\
    &\leq \log p_\star^\e(z) - (1-s_\eta) \log p_{\theta_\star^\e}(z) - s_\eta \log p_{\hat{\theta}^\e}(z) 
    = s_\eta \log\frac{p_{\theta_\star^\e}(z)}{p_{\hat{\theta}^\e}(z)},
\end{align*}
and therefore we have that the empirical log-likelihood ratio satisfies:
\begin{align*}
    \sum_{i=1}^{m} \log \frac{p_\star^{\e}(z^{(i)})}{p_{\hat{\theta}^{\e}(s_{\eta})}(z^{(i)})} \leq s_{\eta} \sum_{i=1}^{m} \log \frac{p_\star^{\e}(z^{(i)})}{ p_{\hat{\theta}^{\e}}(z^{(i)}) } \leq 0,
\end{align*}
where the last inequality holds since $\hat{\theta}^{\e}$ is a MLE over $\calP_{\e}$ and $p_\star^{\e} \in \calP_{\e}$.
Let us denote the event that \eqref{eq:hellinger_conv_ERM_bound} holds as
$\calE_1$.
On this event, we have by \eqref{eq:hellinger_conv_triangle_eq} and $(a+b+c)^2 \leq 3(a^2+b^2+c^2)$ for $a,b,c\in\R$,
on $\calE_1$, for every $s \in [0, 1]$,
\begin{align*}
    \HgSq{\hat{\theta}^{\e}(s;\theta_\star)}{\theta_\star} &\leq 3 \HgSq{\hat{\theta}^{\e}(s_{\eta}; \theta_\star^{\e})}{\theta_\star^{\e}} + \frac{3\eta^2}{4} \mathrm{diam}^2(\Theta) + 3\e^2 \\
    &\leq \frac{6}{m} \log\left(\frac{2\abs{\calP_{\e}}}{\delta} \bigceil{\frac{1}{2\eta}} \right) + \frac{3\eta^2}{4} \mathrm{diam}^2(\Theta) + 3\e^2.
\end{align*}
Hence, we have shown that:
\begin{align*}
    \Pr_{\calD_{m,T} \sim (p^{\e}_\star)^{\otimes m}}\left\{ \sup_{s \in [0, 1]} \HgSq{\hat{\theta}^{\e}[\calD_{m,T}](s; \theta_\star)}{\theta_\star} > \frac{6}{m} \log\left(\frac{2\abs{\calP_{\e}}}{\delta} \bigceil{\frac{1}{2\eta}} \right) + \frac{3\eta^2}{4} \mathrm{diam}^2(\Theta) + 3\e^2 \right\} \leq \delta/2.
\end{align*}
Above, as in part (a), we use the notation 
$\hat{\theta}^{\e}[\calD_{m,T}]$ to emphasize
the dependence of the estimator on the data $\calD_{m,T}$.
We now take similar steps as in part (a)
to change measure 
between $\calD_{m,T} \sim (p^\e_\star)^{\otimes m}$ 
and $\calD_{m,T} \sim p_\star^{\otimes m}$:
\begin{align*}
    &\Pr_{\calD_{m,T} \sim (p_\star)^{\otimes m}}\left\{ \sup_{s \in [0, 1]} \HgSq{\hat{\theta}^{\e}[\calD_{m,T}](s; \theta_\star)}{\theta_\star} > \frac{6}{m} \log\left(\frac{2\abs{\calP_{\e}}}{\delta} \bigceil{\frac{1}{2\eta}} \right) + \frac{3\eta^2}{4} \mathrm{diam}^2(\Theta) + 3\e^2 \right\} \\
    &\leq \TV{p_\star^{\otimes m}}{(p_\star^{\e})^{\otimes m}} + \delta/2 
    \leq \Hg{p_\star^{\otimes m}}{(p^{\e}_\star)^{\otimes m}} + \delta/2 
    \leq \delta,
\end{align*}
where the last inequality follows from
\Cref{prop:hellinger_tensorization} 
since we have $\Hg{p_\star}{p_\star^{\e}} \leq \e \leq \delta/(2\sqrt{2m})$.
This establishes the result.
\end{proof}

\subsubsection{Equivalence of Hellinger Distance and Fisher-weighted Metric}
\label{sec:hellinger_fisher_equiv}

\Cref{thm:hellinger_bound_MLE} is, at its core, a result about i.i.d.\ learning. 
It however contains a rich amount of information about 
trajectories \emph{within} the divergence term $\HgSq{\theta^\e_{m,T}}{\theta_\star}$.
In this section, we study how to extract this information out of the Hellinger divergence.
As previously discussed, this is challenging
as neither the Hellinger nor squared Hellinger distance tensorizes across $z_{1:T}$ in non-i.i.d.\ settings.
However, when $\hat{\theta}^\e_{m,T}$ is close to $\theta_\star$, such a tensorization is indeed possible, as observed via the asymptotic expansion \eqref{eq:hellinger_local_asymptotic_expansion}.
Our next result quantifies the radius of validity for this expansion, through a second-order Taylor expansion analysis.

\begin{myprop}
\label{prop:hellinger_perturbation}
Fix any $\theta_0, \theta_1 \in \Theta$ where for all $\theta \in \mathrm{conv}\{\theta_0, \theta_1\}$, we have
(a) $\theta \in \Theta$ and (b) 
$\calI(\theta) \succ 0$.
Define the quantities:
\begin{align}
    B_1(\theta_0, \theta_1) &:= \sup_{\theta \in \mathrm{conv}\{\theta_0,\theta_1\}} \sup_{v \in \bbS^{p-1}} \norm{ \ip{v}{\calI(\theta)^{-1/2} \nabla_\theta \log p_\theta(z_{1:T})}}_{\calL^4(p_\theta)}, \label{eq:hellinger_MDS} \\
    B_2(\theta_0, \theta_1) &:= \sup_{\theta \in \mathrm{conv}\{\theta_0,\theta_1\}} \sup_{v \in \bbS^{p-1}} \norm{ \ip{v}{\calI(\theta)^{-1/2} \nabla^2_\theta \log p_\theta(z_{1:T}) \calI(\theta)^{-1/2} v}}_{\calL^2(p_\theta)}. \label{eq:hellinger_hessian}
\end{align}
\begin{enumerate}[label=\textbf{(\alph*).}]
\item Suppose that the following condition holds:
\begin{align}
    \sup_{\theta \in \mathrm{conv}\{ \theta_0, \theta_1 \}} \Hg{\theta_0}{\theta} \leq \frac{1}{16\sqrt{2}} \min\left\{ \frac{1}{B_1^2(\theta_0, \theta_1)}, \frac{1}{B_2(\theta_0, \theta_1)} \right\}. \label{eq:hellinger_perturbation_radius}
\end{align}
Then the following inequalities hold:
\begin{align}
    \frac{3}{16} \norm{\theta_0-\theta_1}^2_{\calI_2(\theta_0, \theta_1)} \leq \HgSq{\theta_0}{\theta_1} \leq \frac{5}{16} \norm{\theta_0-\theta_1}^2_{\calI_2(\theta_0, \theta_1)}, \quad \calI_2(\theta_0, \theta_1) := 2 \int_0^1 (1-s) \calI(\theta(s)) \rmd s, \label{eq:hellinger_perturbation_avg_FI}
\end{align}
where $\theta(s) := (1-s) \theta_0 + s \theta_1$.

\item If in addition to \eqref{eq:hellinger_perturbation_radius} holding,
we furthermore have that:
\begin{align}
    \sup_{\theta \in \mathrm{conv}\{\theta_0, \theta_1\}} \opnorm{ \calI(\theta_0)^{-1/2} \calI(\theta) \calI(\theta_0)^{-1/2} - I_p } \leq \frac{1}{2},
    \label{eq:hellinger_perturbation_FI_radius}
\end{align}
then we also have the following inequalities:
\begin{align}
    \frac{3}{32} \norm{\theta_0-\theta_1}^2_{\calI(\theta_0)} \leq \HgSq{\theta_0}{\theta_1} \leq \frac{15}{32} \norm{\theta_0-\theta_1}^2_{\calI(\theta_0)}. \label{eq:hellinger_perturbation}
\end{align}
\end{enumerate}
\end{myprop}

\begin{rmk}
The constants in
\eqref{eq:hellinger_perturbation_avg_FI} and
\eqref{eq:hellinger_perturbation} can be made arbitrarily close to $1/4$ (cf.~\eqref{eq:hellinger_local_asymptotic_expansion}) at the expense of decreasing the constants in the local radius conditions \eqref{eq:hellinger_perturbation_radius}
and \eqref{eq:hellinger_perturbation_FI_radius}.
\end{rmk}

\begin{proof}[Proof of \Cref{prop:hellinger_perturbation}]
For $\theta \in \Theta$ and $z \in \sfZ^T$, define the function $h(\theta; z) := \sqrt{p_\theta(z)}$.
Abbreviating $\mu = \mu^{\otimes T}$,
we take the first and second derivatives of both
$\theta \mapsto h(\theta; z)$ and $\theta \mapsto \HgSq{\theta_0}{\theta}$:
\begin{align*}
    \nabla_\theta h(\theta; z) &= \frac{1}{2} \sqrt{p_\theta(z)} \nabla_\theta \log p_\theta(z), \\
    \nabla_\theta^2 h(\theta; z) &= \sqrt{p_\theta(z)} \left[ \frac{1}{2} \nabla^2_\theta \log p_\theta(z) + \frac{1}{4} \nabla_\theta \log p_\theta(z) \nabla_\theta \log p_\theta(z)^\T \right], \\
    \nabla_\theta \HgSq{\theta_0}{\theta} &= 2 \int ( h(\theta; z) - h(\theta_0; z)) \nabla_\theta h(\theta; z) \rmd\mu, \\
    \nabla^2_\theta \HgSq{\theta_0}{\theta} &= 2 \int ( h(\theta; z) - h(\theta_0; z)) \nabla^2_\theta h(\theta; z) \rmd \mu + 2 \int \nabla_\theta h(\theta; z)\nabla_\theta h(\theta; z)^\T \rmd \mu =: \calH(\theta; \theta_0).
\end{align*}
We therefore have the identity for the Hessian $\calH(\theta; \theta_0)$:
\begin{align*}
    \calH(\theta; \theta_0) &= 2 \int ( h(\theta; z) - h(\theta_0; z)) \nabla^2_\theta h(\theta; z) \rmd \mu + \frac{1}{2} \int \nabla_\theta \log p_{\theta}(z) \nabla_\theta \log p_{\theta}(z)^\T  p_\theta(z) \rmd \mu \\
    &= 2 \int ( h(\theta; z) - h(\theta_0; z)) \nabla^2_\theta h(\theta; z) \rmd \mu + \frac{1}{2} \calI(\theta).
\end{align*}
Using the second-order integral version of Taylor's theorem and expanding $\theta \mapsto \HgSq{\theta_0}{\theta}$ around $\theta=\theta_0$,
with shorthand notation $\theta_s := \theta(s)$ and
$\calI_s := \calI(\theta(s))$ for $s \in [0, 1]$, we have:
\begin{align}
    \HgSq{\theta_0}{\theta_1} &= \int_0^1 (1-s) \Delta^\T \calH(\theta_s; \theta_0) \Delta \rmd s \nonumber \\
    &= \int_0^1 (1-s) \Delta^\T \left[  2 \int ( h(\theta_s; z) - h(\theta_0; z)) \nabla^2_\theta h(\theta_s; z) \rmd\mu + \frac{1}{2} \calI_s \right] \Delta \rmd s \nonumber \\
    &= \frac{1}{2} \int_0^1 (1-s) \Delta^\T \calI_s \Delta \rmd s + 2 \int_0^1 (1-s) \int ( h(\theta_s; z) - h(\theta_0; z)) \Delta^\T \nabla^2_\theta h(\theta_s; z) \Delta \rmd\mu \rmd s. \label{eq:hellinger_sq_second_order_expansion}
\end{align}

\paragraph{(a).}
Fix a vector $q \in \R^p$ and $s \in [0, 1]$.
We first bound:
\begin{align*}
    &\bigabs{ \int (h(\theta_s;z)-h(\theta_0;z)) q^\T \calI_s^{-1/2} \nabla^2_\theta h(\theta_s; z) \calI_s^{-1/2} q \rmd \mu } \\
    &\stackrel{(a)}{\leq} \sqrt{ \int (h(\theta_s;z) - h(\theta_0;z))^2 \rmd \mu} \sqrt{  \int (q^\T \calI_s^{-1/2} \nabla^2_\theta h(\theta_s; z) \calI_s^{-1/2} q   )^2 \rmd\mu  } \\
    &= \Hg{\theta_0}{\theta_s} \sqrt{  \int (q^\T \calI_s^{-1/2} \nabla^2_\theta h(\theta_s; z) \calI_s^{-1/2} q   )^2 \rmd\mu } \\
    &\stackrel{(b)}{\leq} \norm{q}^2 \Hg{\theta_0}{\theta_s} \sqrt{ \frac{1}{2} B_2^2 + \frac{1}{8} B_1^4 }
    \stackrel{(c)}{\leq} \sqrt{2} \norm{q}^2 \Hg{\theta_0}{\theta_s} \max\{B_1^2, B_2\}
    \stackrel{(d)}{\leq} \frac{1}{16}\norm{q}^2,
\end{align*}
where (a) is Cauchy-Schwarz,
(b) follows by
the following bounds with $q_s := \calI_s^{-1/2} q$
and using the inequality $(a+b)^2 \leq 2(a^2+b^2)$ for $a,b \in \R$:
\begin{align*}
    \int (q_s^\T \nabla^2_\theta h(\theta_s; z) q_s)^2 \rmd \mu &= \int p_{\theta_s}(z) \left[ \frac{1}{2} q_s^\T \nabla^2_\theta \log p_{\theta_s}(z) q_s + \frac{1}{4} \ip{q_s}{\nabla_\theta \log p_{\theta_s}(z)}^2 \right]^2 \rmd\mu \\
    &\leq \int\left[ \frac{1}{2} (q_s^\T \nabla^2_\theta \log p_{\theta_s}(z) q_s)^2+ \frac{1}{8} \ip{q_s}{\nabla_\theta \log p_{\theta_s}(z)}^4\right] p_{\theta_s}(z) \rmd \mu \\
    &= \frac{1}{2} \norm{ \ip{q}{\calI_s^{-1/2} \nabla_\theta^2 \log p_{\theta_s}(z) \calI_s^{-1/2} q} }_{\calL^2(p_{\theta_s})}^2 + \frac{1}{8} \norm{ \ip{q}{\calI_s^{-1/2} \nabla_\theta \log p_{\theta_s}(z)}}^4_{\calL^4(p_{\theta_s})} \\
    &\leq \norm{q}^4 \left[\frac{1}{2} B_2^2 + \frac{1}{8} B_1^4 \right],
\end{align*}
(c) follows from the basic inequalities:
$$\sqrt{ \frac{1}{2} B_2^2 + \frac{1}{8} B_1^4 } \leq B_1^2/\sqrt{8} + B_2/\sqrt{2} \leq (B_1^2 + B_2)/\sqrt{2} \leq \sqrt{2} \max\{ B_1^2, B_2 \},$$
and (d) follows by our stated assumption \eqref{eq:hellinger_sq_second_order_expansion}.
Hence, setting $q = \calI_s^{1/2} \Delta$, we have:
\begin{align*}
    &\bigabs{ \int (h(\theta_s;z)-h(\theta_0;z)) \Delta^\T \nabla^2_\theta h(\theta_s; z) \Delta \rmd \mu } \\
    &= \bigabs{ \int (h(\theta_s;z)-h(\theta_0;z)) (\calI_s^{1/2} \Delta)^\T \calI_s^{-1/2} \nabla^2_\theta h(\theta_s; z) \calI_s^{-1/2} (\calI_s^{1/2} \Delta) \rmd \mu } 
    \leq \frac{1}{16} \norm{\Delta}^2_{\calI_s} .
\end{align*}
Now, utilizing the second order expansion from \eqref{eq:hellinger_sq_second_order_expansion},
\begin{align*}
    \HgSq{\theta_0}{\theta_1} &\geq \frac{1}{2} \int_0^1 (1-s) \Delta^\T \calI_s \Delta \rmd s - 2 \int_0^1 (1-s) \bigabs{ \int ( h(\theta_s; z) - h(\theta_0; z)) \Delta^\T \nabla^2_\theta h(\theta_s; z) \Delta \rmd\mu  } \rmd s \\
    &\geq \frac{1}{2} \int_0^1 (1-s) \norm{\Delta}_{\calI_s}^2 \rmd s - \frac{1}{8} \int_0^1 (1-s) \norm{\Delta}_{\calI_s}^2  \rmd s 
    = \frac{3}{8} \int_0^1 (1-s) \norm{\Delta}_{\calI_s}^2 \rmd s = \frac{3}{16} \norm{\Delta}_{\calI_2(\theta_0, \theta_1)}^2.
\end{align*}
The upper bound is established in a nearly identical way, which yields
\eqref{eq:hellinger_perturbation_avg_FI}.

\paragraph{(b).}
Using \eqref{eq:hellinger_perturbation_FI_radius}, we conclude that
for all $s \in [0, 1]$,
\begin{align*}
    \frac{1}{2} I_p \preccurlyeq \calI_0^{-1/2} \calI_s \calI_0^{-1/2} \preccurlyeq \frac{3}{2} I_p.
\end{align*}
From this, we conclude that:
\begin{align*}
    \frac{1}{2} \calI(\theta_0) \preccurlyeq \calI_2(\theta_0, \theta_1) \preccurlyeq \frac{3}{2} \calI(\theta_0),
\end{align*}
from which \eqref{eq:hellinger_perturbation} 
follows from plugging the above semidefinite inequalities into
\eqref{eq:hellinger_perturbation_avg_FI}.
\end{proof}

\Cref{prop:hellinger_perturbation} shows that the
region for which the asymptotic expansion \eqref{eq:hellinger_local_asymptotic_expansion} holds
is governed by two key conditions:
(i) the value of $\sup_{\theta \in \mathrm{conv}\{\theta_0, \theta_1\}} \Hg{\theta_0}{\theta}$ being small enough relative to the inverse of the moment bounds
$B_1^2(\theta_0, \theta_1)$ and $B_2(\theta_0, \theta_1)$
(cf.~\eqref{eq:hellinger_perturbation_radius}),
and (ii) the parameters $\theta_0, \theta_1$ being close 
enough as measured through the corresponding FI matrices (cf.~\eqref{eq:hellinger_perturbation_FI_radius}).
In \Cref{sec:hellinger_localization_framework}, 
we describe the Hellinger localization framework, which gives a general recipe
for verifying these conditions so that \Cref{thm:hellinger_bound_MLE}
can be used in conjunction with \Cref{prop:hellinger_perturbation}
to establish non-asymptotic rates for the MLE which exhibit the CLT
scaling \eqref{eq:asymptotic_normality_parameter}.
Before describing our general framework, we first work through
a specific example next in \Cref{sec:twostatemc},
which will set the stage for the general recipe. 

\subsection{A Two-State Markov Chain Example}
\label{sec:twostatemc}

We now demonstrate how the combination of \Cref{thm:hellinger_bound_MLE} and \Cref{prop:hellinger_perturbation} gives us a nearly optimal parameter recovery bound via a simple example. We consider a two-state discrete-time Markov chain, where $\sfZ = \{0, 1\}$, $\Theta = [\mu, 1-\mu]$ for some $\mu \in (0, 1/2)$,
and $\calP$ is the set of all two-state Markov chains with $z_1 \sim \rho_1$ (independent of $\theta$) and one-step transition probability:
\begin{align*}
    p_\theta(z_{t+1} \mid z_t) = \theta \ind\{z_{t+1} = z_t\} + (1-\theta) \ind\{z_{t+1} \neq z_t\}, \quad \theta \in \Theta.
\end{align*}
For what follows, we will assume that $T \geq 2$ 
(otherwise no information about the Markov chain transition probabilities is revealed).
We assume $p_\star \in \calP$ with parameter $\theta_\star \in \Theta$. 
While this specific problem is simple enough that its MLE estimator can be studied via only elementary concentration inequalities 
(which we discuss at the end), 
we utilize our framework to analyze this problem in order to illustrate both the mechanics and relative sharpness of our arguments. 

\paragraph{Roadmap.}
We first compute the FI matrix $\calI(\theta)$ in addition to its uniform bound
$\calI_{\max}$.
From these quantities, we bound the covering number $\calN_{\calI_{\max}}(\calP, \e)$
and invoke \Cref{thm:hellinger_bound_MLE} (b) for log-concave $\calP$;
this yields control of the Hellinger distance 
between $\theta_\star$ and every element in $\mathrm{conv}\{ \theta_\star, \hat{\theta}^\e_{m,T} \}$, where $\hat{\theta}^\e_{m,T}$ denotes the discretized MLE estimator.\footnote{Specially, $\hat{\theta}^\e_{m,T}$ denotes the max FI divergence discretized MLE estimator at resolution $\e=1/(2\sqrt{2m})$.}
With this bound in hand, we then estimate the quantities $B_1, B_2$ (cf.\ \eqref{eq:hellinger_MDS}, \eqref{eq:hellinger_hessian}); a key intermediate step is to 
show that control of the Hellinger distance from \Cref{thm:hellinger_bound_MLE}
implies a direct $O(1/m)$ bound on the squared parameter error, which we then use to localize the $B_1, B_2$
computation in a small neighborhood around $\theta_\star$.
At this point, we are now able to invoke \Cref{prop:hellinger_perturbation} (a) 
to boost our bound on the squared parameter error to $O(1/(mT))$;
the only missing piece is that this bound is not variance-optimal.
However, by using this bound to establish that the condition \eqref{eq:hellinger_perturbation_FI_radius} holds, we finally conclude by invoking \Cref{prop:hellinger_perturbation} (b), which yields
the instance-optimal rate.

Towards carrying out this plan, we first introduce some notation.
Let us denote $\sigma^2_\star := \theta_\star (1-\theta_\star)$,
which is the variance of a $\mathrm{Bern}(\theta_\star)$ distribution,
and governs the curvature of the FI matrix $\calI(\theta_\star)$.
We also define the set $\Theta_{\sigma_\star} := \{ \theta \in \Theta \mid \abs{\theta - \theta_\star} \leq \sigma_\star^2/2 \}$ for localization purposes:
observe that for $\theta \in \Theta_{\sigma_\star}$, 
we have 
$\frac{1}{\theta(1-\theta)} \leq \frac{2}{\sigma^2_\star}$,
a key inequality we will use in our computations.

\paragraph{FI matrix, covering number, and Hellinger bound.}
We first gather the results of some straightforward computations
in \Cref{sec:appendix:two_state}:
\begin{align}
    \calI(\theta) = \frac{T-1}{\theta(1-\theta)}, \quad \sup_{\theta \in \Theta} \calI(\theta) \leq \calI_{\max} := \frac{T-1}{\mu(1-\mu)}, \quad \mathrm{diam}(\Theta) \leq \frac{T-1}{\mu(1-\mu)}.
\end{align}
From this, we bound covering number of $\calP$ in the max FI-divergence (cf.~\Cref{def:max_FI_cover}) as:
\begin{align*}
    \calN_{\calI_{\max}}(\calP, \e) \leq \calN_{\abs{\cdot}}\left([\mu, 1-\mu], \frac{\mu(1-\mu) \e}{T-1}\right) \leq \bigceil{\frac{T-1}{2\mu(1-\mu)\e}}.
\end{align*}
Now we are in a position to apply \Cref{thm:hellinger_bound_MLE}.
Setting $\e = \delta/(2\sqrt{2m})$ and $\eta = 1/\mathrm{diam}(\Theta)$,
we obtain with probability at least $1-\delta$,
the max FI divergence MLE estimator satisfies:
\begin{align}
    \sup_{s \in [0, 1]} \HgSq{(1-s)\theta_\star + s\hat{\theta}^{\e}_{m,T}}{\theta_\star} &\lesssim \frac{\log(mT/(\mu\delta))}{m}. \label{eq:two_state_hellinger_bound}
\end{align}
Let us denote the event in \eqref{eq:two_state_hellinger_bound} by $\calE_1$.

\paragraph{Estimate $B_1$ and $B_2$.}
We will estimate $B_1(\theta_0, \theta_1)$ and
$B_2(\theta_0, \theta_1)$ over $\theta_0,\theta_1 \in \Theta_{\sigma_\star}$.
Before we proceed, we first utilize 
\eqref{eq:two_state_hellinger_bound} to construct
a condition on $m$ such that $\hat{\theta}_{m,T}^\e \in \Theta_{\sigma_\star}$ on $\calE_1$.
Specifically:
\begin{align*}
    \Hg{\hat{p}^\e_{m,T}}{p_\star} &\stackrel{(a)}{\geq} \Hg{\hat{p}^\e_{m,T}(z_1, z_2)}{p_\star(z_1, z_2)} 
    \stackrel{(b)}{\geq} \TV{\hat{p}^\e_{m,T}(z_1, z_2)}{p_\star(z_1, z_2)} \\
    &\stackrel{(c)}{=} \E_{z_1 \sim \rho_1}[ \TV{\mathrm{Bern}(\hat{\theta}^\e_{m,T})}{\mathrm{Bern}(\theta_\star)} ] 
    = \abs{ \hat{\theta}^\e_{m,T} - \theta_\star },
\end{align*}
where (a) uses the 
data processing inequality for $f$-divergences,
(b) uses the inequality $\TV{p}{q} \leq \Hg{p}{q}$ for two measures $p,q$,
and (c) uses the fact that $z_1 \sim \rho_1$ irregardless of $\theta$.
Hence, if $m \gtrsim \sigma^{-4}_\star\log(mT/(\mu\delta))$, then we have that 
$\hat{\theta}^\e_{m,T} \in \Theta_{\sigma_\star}$ on $\calE_1$.
By convexity of $\Theta_{\sigma_\star}$, this implies that
$\mathrm{conv}\{\hat{\theta}^\e_{m,T}, \theta_\star\} \subset \Theta_{\sigma_\star}$.
By \Cref{prop:log_dominance},
it suffices to take $m \gtrsim \sigma_\star^{-4} \log(T/ (\mu \delta))$.
In summary:
\begin{align}
    m \gtrsim \sigma_\star^{-4} \log( T/ (\mu \delta)) \Longrightarrow \mathrm{conv}\{\hat{\theta}^\e_{m,T}, \theta_\star\} \subset \Theta_{\sigma_\star} \textrm{ on } \calE_1. \label{eq:two_state_m_req_E1}
\end{align}
Next, we show in \Cref{sec:appendix:two_state} that:
\begin{align*}
    \E_{p_\theta}[ (\partial_\theta \log p_\theta(z_{1:T}))^4 ] &= (T-1)\left(\frac{1}{\theta^3} + \frac{1}{(1-\theta)^3}\right) + 3(T-1)(T-2)\frac{1}{\theta^2(1-\theta)^2}, \\
    \E_{p_\theta}[ (\partial_\theta^2 \log p_\theta(z_{1:T}))^2 ] &= (T-1) \left( \frac{1}{\theta^3} + \frac{1}{(1-\theta)^3}\right) + (T-1)(T-2) \frac{1}{\theta^2(1-\theta)^2}.
\end{align*}
Hence, for any $\theta \in \Theta_{\sigma_\star}$, we have:
\begin{equation}\label{eq:two_state_B1_B2_prelim}
\begin{aligned}
    \calI(\theta)^{-2} \max\{ \E_{p_\theta}[ (\partial_\theta \log p_\theta(z_{1:T}))^4 ], \E_{p_\theta}[ (\partial_\theta^2 \log p_\theta(z_{1:T}))^2 ] \} 
    \lesssim \max\left\{\frac{1}{T \sigma^2_\star}, 1\right\}.
\end{aligned}
\end{equation}
Therefore, we have established
\begin{align}
    \sup_{\theta_1,\theta_2 \in \Theta_{\sigma_\star}} \max\{ B_1^2(\theta_1, \theta_2), B_2(\theta_1, \theta_2) \} \lesssim \max\left\{\frac{1}{\sigma_\star\sqrt{T}}, 1 \right\}. \label{eq:two_state_B1_B2_bounds}
\end{align}

\paragraph{Parameter error bound.}

We first verify the condition in 
\eqref{eq:hellinger_perturbation_radius}
for $\theta_0 = \theta_\star$ and $\theta_1 = \hat{\theta}^\e_{m,T}$.
By combining \eqref{eq:two_state_hellinger_bound}, 
\eqref{eq:two_state_B1_B2_bounds}, and \Cref{prop:log_dominance},
it suffices to choose an $m$ satisfying:
\begin{align}
    mT \gtrsim \frac{1}{\sigma^2_\star} \log(1/(\mu\delta)), \quad m \gtrsim \log(T/(\mu\delta)). \label{eq:two_state_m_req_B1_B2}
\end{align}
Thus combining all requirements on $m,T$ from \eqref{eq:two_state_hellinger_bound},
\eqref{eq:two_state_m_req_E1}, and \eqref{eq:two_state_m_req_B1_B2}:
\begin{align*}
    m \gtrsim \sigma^{-4}_\star \log(T/(\mu\delta)) \Longrightarrow \text{\eqref{eq:hellinger_perturbation_radius} holds on $\calE_1$ for $\theta_0 = \theta_\star$ and $\theta_1 = \hat{\theta}^\e_{m,T}$}. 
\end{align*}
Therefore by \eqref{eq:hellinger_perturbation_avg_FI} from \Cref{prop:hellinger_perturbation}, we have on $\calE_1$:
\begin{align*}
    \calI(\theta_\star, \hat{\theta}^\e_{m,T}) \abs{\hat{\theta}_{m,T} - \theta_\star}^2 \lesssim \HgSq{\hat{\theta}^\e_{m,T}}{\theta_\star} \lesssim \frac{\log(mT/(\mu\delta))}{m}.
\end{align*}
To lower bound the LHS, observe that
for any $\theta \in \Theta$, $\calI(\theta) \geq 4(T-1)$.
Hence for any $\theta_0, \theta_1 \in \Theta$,
$\calI(\theta_0, \theta_1) \gtrsim T$,
which implies that on $\calE_1$,
\begin{align}
    \abs{\hat{\theta}_{m,T}^\e - \theta_\star}^2 \lesssim \frac{ \log(mT/(\mu\delta))}{mT}. \label{eq:two_state_param_error}
\end{align}

\paragraph{Verify FI radius.}

We first observe for any $\theta_0, \theta_1 \in \Theta_{\sigma_\star}$,
\begin{align}
    \abs{\calI(\theta_0)^{-1} \calI(\theta_1) - 1} &= \bigabs{ \frac{\theta_0 (1-\theta_0)}{\theta_1 (1-\theta_1)} - 1} 
    \leq \frac{2\abs{\theta_0-\theta_1}}{\sigma^2_\star}. 
    \label{eq:two_state_FI_metric}
\end{align}
From \eqref{eq:two_state_param_error} and \eqref{eq:two_state_FI_metric}, we have on $\calE_1$:
\begin{align*}
    \sup_{\theta \in \mathrm{conv}\{ \hat{\theta}^\e_{m,T}, \theta_\star \}} \abs{ \calI(\theta_\star)^{-1} \calI(\theta) - 1 } \lesssim \sup_{\theta \in \mathrm{conv}\{ \hat{\theta}^\e_{m,T}, \theta_\star \}} \frac{\abs{\theta - \theta_\star}}{\sigma^2_\star} = \frac{\abs{\hat{\theta}^\e_{m,T} - \theta_\star}}{\sigma^2_\star} \lesssim \frac{1}{\sigma^2_\star} \sqrt{ \frac{ \log(mT/(\mu\delta))}{mT} }.
\end{align*}
By another application of \Cref{prop:log_dominance},
we can ensure the FI radius condition \eqref{eq:hellinger_perturbation_FI_radius} holds on $\calE_1$ by setting
$mT \gtrsim \sigma^{-4}_\star \log(1/(\mu\delta))$,
which is already implied by $m \gtrsim \sigma^{-4}_\star \log(T/(\mu\delta))$.
\Cref{prop:hellinger_perturbation} now yields via \eqref{eq:hellinger_perturbation} that on $\calE_1$,
\begin{align*}
    \calI(\theta_\star) \abs{\hat{\theta}^\e_{m,T} - \theta_\star}^2 \lesssim \HgSq{\hat{\theta}^\e_{m,T}}{\theta_\star} \lesssim \frac{\log(mT/(\mu\delta))}{m}.
\end{align*}

\paragraph{Final result.}
Combining the previous arguments, the final result is that
as long as $m$ satisfies:
\begin{align}
    m \gtrsim \frac{1}{\sigma^4_\star} \log\left(\frac{T}{\mu\delta}\right), \label{eq:two_state_m_req}
\end{align}
then with probability at least $1-\delta$ (over $\calD_{m,T}$),
\begin{align}
    \abs{\hat{\theta}^\e_{m,T} - \theta_\star}^2 \lesssim \frac{\sigma^2_\star \log(mT/(\mu\delta))}{mT}. \label{eq:two_state_rate}
\end{align}

\paragraph{Sharpness of the result.}
We now evaluate the sharpness of this result by
providing an elementary solution based on sub-Exponential tail inequalities for Binomial distributions. 
We show in \Cref{sec:appendix:two_state} that
there exists an event $\calE_2$ with probability at least $1-\delta$ that satisfies
\begin{align}
    mT \gtrsim \sigma^{-2}_\star \log(1/\delta) \Longrightarrow \abs{\hat{\theta}_{m,T} - \theta_\star}^2 \lesssim \frac{\sigma_\star^2 \log(1/\delta)}{mT} \textrm{ on $\calE_2$}. \label{eq:two_state_optimal_rate}
\end{align}
Comparing both the requirement on $mT$ in \eqref{eq:two_state_optimal_rate} to \eqref{eq:two_state_m_req},
in addition to the final error rate to \eqref{eq:two_state_rate},
we see that the result utilizing our framework is 
sharp up to log factors in the final error rate, but 
misses a few factors in the requirement on $m$.
In particular, \eqref{eq:two_state_optimal_rate}
shows that $mT \gtrsim \tilde{O}( \sigma_\star^{-2} )$ suffices 
to enter the CLT rate regime, but \eqref{eq:two_state_m_req} requires
the more conservative bound $m \gtrsim \tilde{O}(\sigma_\star^{-4})$.
We note that the source of this conservatism is due to 
(a) the use of the data processing inequality to lower bound
$\Hg{\hat{p}^\e_{m,T}}{p_\star} \geq \Hg{\hat{p}^\e_{m,T}(z_1, z_2)}{p_\star(z_1, z_2)}$
and (b) further lower bounding
$\Hg{\hat{p}^\e_{m,T}(z_1, z_2)}{p_\star(z_1, z_2)} \geq \TV{\hat{p}^\e_{m,T}(z_1, z_2)}{p_\star(z_1, z_2)}$.
The DPI inequality (a) is lossy in this case, since it is possible to prove (at least when $\rho_1 = \mathrm{Unif}(\{1, 2\})$) that the tensorization property 
$\HgSq{\theta_0}{\theta_1} = 1 - (1 - \HgSq{\mathrm{Bern}(\theta_0)}{\mathrm{Bern}(\theta_1)})^{T-1}$ actually holds~\cite[see e.g.][Lemma 5]{daskalakis2018testingmarkov}.
Going from Hellinger to TV distance in (b) is lossy as well since
the TV distance between two Bernoulli distributions loses all local curvature information. 
Nevertheless, we see that our general framework is able to capture the qualitative aspects of this problem which arise from a problem-specific analysis.

\subsection{Hellinger Localization Framework}
\label{sec:hellinger_localization_framework}

Previously in \Cref{sec:twostatemc}, we saw a specific example of how \Cref{prop:hellinger_perturbation} was combined with \Cref{thm:hellinger_bound_MLE} to establish non-asymptotic rates for MLE which exhibit nearly optimal CLT scaling from \eqref{eq:asymptotic_normality}.
In this section, we will utilize these two results to provide a general recipe, which we call the \emph{Hellinger localization framework}, for establishing rates.
Importantly, our framework in addition to being general purpose, does not inherently rely on any mixing, ergodicity, or stationarity properties of the process $z_{1:T}$, but instead relies on the presence of multiple independent trajectories 
 to allow us to learn from possibly non-stationary and/or non-mixing processes.
Before we present the main framework, we introduce a key identifiability condition which plays an important role.
\begin{mydef}[Hellinger Identifiability]
\label{def:identifiability}
We say that the parametric family $\calP$ satisfies $(\gamma_1, \gamma_2)$-\emph{Hellinger identifiability} (or $(\gamma_1,\gamma_2)$-\emph{identifiability}) about the point $\theta_\star \in \Theta$ if for every $p_\theta \in \calP$:
\begin{align*}
    \Hg{p_{\theta}}{p_{\theta_\star}} \leq \gamma_1 \Longrightarrow \norm{\theta-\theta_\star} \leq \gamma_2 \cdot \Hg{p_{\theta}}{p_{\theta_\star}}.
\end{align*}
\end{mydef}
\Cref{def:identifiability} is in essence the minimal set of assumptions needed for parameter recovery;
fortunately it is not hard to see that under our stated assumptions
\Cref{def:identifiability} holds for some $(\gamma_1, \gamma_2)$ under fairly generic conditions (see \Cref{prop:hellinger_identifiability_generic} for a precise statement). 
On the other, obtaining problem specific constants---especially \emph{sharp} constants i.e., $\gamma_2^{-1} \asymp \sqrt{\lambda_{\min}(\calI(\theta_\star))}$ as we expect from asymptotic normality (cf.~\eqref{eq:asymptotic_normality})---is non-trivial, and one of our main contributions.
Indeed, our work can be contextualized as starting from a fairly
sub-optimal pair $(\gamma_1, \gamma_2)$, and bootstrapping such
a pair into a nearly optimal one; see \Cref{rmk:single_step_identifiability}
for one particular method for obtaining a starting pair $(\gamma_1, \gamma_2)$.
The specific steps of this recipe, which mirror the steps
taken for the example in \Cref{sec:twostatemc},
are as follows:
\begin{enumerate}[label=Step~\arabic*., labelsep=1em, align=left, leftmargin=*, itemsep=0.5em]
 \item \textbf{Hellinger bound.} If the density class is log-concave (cf.~\Cref{def:log_concave}), or one can prove that 
 $\theta \mapsto \HgSq{\theta}{\theta_\star}$ is convex
 in the \emph{parameter space}, 
 we estimate the covering number
 $\calN_{\calI_{\max}} \equiv \calN_{\calI_{\max}}(\calP, \e)$ at resolution $\e \asymp \delta/\sqrt{m}$
 for $\calP$, and apply \Cref{thm:hellinger_bound_MLE}, specifically \eqref{eq:fisher_bound_MLE_discrete}, with $\eta \asymp (\mathrm{diam}(\Theta) \sqrt{m})^{-1}$ to
 obtain the following event $\calE_1$ that holds with probability at least $1-\delta$ for a universal $c_0$:
 \begin{equation}
 \sup_{\theta \in \mathrm{conv}\{ \hat{\theta}^\e_{m,T}, \theta_\star \}}\HgSq{\theta}{\theta_\star} \lesssim m^{-1}\log(c_0 m \cdot \mathrm{diam}(\Theta) \calN_{\calI_{\max}}/\delta). \label{eq:metricentropyhellinger}
 \end{equation}
 Otherwise without log-concavity of $\calP$, 
 we rely on \Cref{prop:hellinger_FI_upper_bound} to derive the bound:
\begin{align}
     \sup_{\theta \in \mathrm{conv}\{ \hat{\theta}^\e_{m,T}, \theta_\star \}} \HgSq{\theta}{\theta_\star} \leq \sup_{\theta \in \mathrm{conv}\{ \hat{\theta}^\e_{m,T}, \theta_\star \}} \frac{1}{4} \FISq{\theta}{\theta_\star} \leq \frac{\lambda_{\max}(\calI_{\max})}{4} \norm{\hat{\theta}^\e_{m,T}-\theta_\star}^2, \label{eq:framework_non_log_concave_bound}
\end{align}
Next, under the assumption of $(\gamma_1, \gamma_2)$-identifiability (cf.~\Cref{def:identifiability}),
the previous inequality implies:
\begin{align}
    \HgSq{\hat{\theta}^\e_{m,T}}{\theta_\star} \leq \gamma_1^2 \Longrightarrow \sup_{\theta \in \mathrm{conv}\{ \hat{\theta}^\e_{m,T}, \theta_\star \}} \HgSq{\theta}{\theta_\star} \leq \frac{\gamma_2^2 \lambda_{\max}(\calI_{\max})}{4} \cdot \HgSq{\hat{\theta}^\e_{m,T}}{\theta_\star}. \label{eq:hellinger_error_via_identifiability}
\end{align}
Applying \Cref{thm:hellinger_bound_MLE}, specifically \eqref{eq:hellinger_bound_MLE_discrete}, 
we obtain the event $\calE_1$ with probability $1-\delta$:
\begin{align*}
    \HgSq{\hat{\theta}^\e_{m,T}}{\theta_\star} \lesssim m^{-1} \log(c_0 \calN_{\calI_{\max}}/\delta).
\end{align*}
Hence as long as 
\begin{align}
    m \gtrsim \gamma_1^{-2} \cdot \log(c_0 \calN_{\calI_{\max}}/\delta), \label{eq:framework_gamma_1_bound}
\end{align}
then on $\calE_1$:
\begin{align}
    \sup_{\theta \in \mathrm{conv}\{ \hat{\theta}^\e_{m,T}, \theta_\star \}} \HgSq{\theta}{\theta_\star} \lesssim m^{-1} \gamma_2^2 \lambda_{\max}(\calI_{\max}) \log(c_0 \calN_{\calI_{\max}}/\delta). \label{eq:framework_hellinger_conv_hull_bound_general}
\end{align}
 \item \textbf{Estimate $B_1$ and $B_2$.} We next compute upper bounds for $B_1 \equiv B_{1}(\hat{\theta}^\e_{m,T}, \theta_\star)$ and $B_2 \equiv B_{2}(\hat{\theta}^\e_{m,T}, \theta_\star)$ defined in \eqref{eq:hellinger_MDS} and \eqref{eq:hellinger_hessian}. 
 As these quantities are random variables due to the presence of $\hat{\theta}^\e_{m,T}$, we often
 approximate $B_1, B_2$ by taking a supremum over a larger set $\Theta' \subseteq \Theta$ for which we can ensure that $\mathrm{conv}\{\hat{\theta}^\e_{m,T}, \theta_\star\} \subseteq \Theta'$ on $\calE_1$.
 For some problems, it suffices to take $\Theta'=\Theta$. However, for other problems, sharper estimates can be derived by more refined $\Theta'$. We will provide examples of both in the sequel.
 
 Control of $B_{1}$ relies on the fact that $\nabla_{\theta}\log p_{\theta}(z_{1:T})$ forms a martingale in the realizable setting, and utilizes estimates from e.g., Burkholder's martingale Rosenthal inequality (see \Cref{thm:rosenthal} for a precise statement). Control of $B_2$ is often more straightforward, and a simple triangle inequality often suffices. 
 Regarding scaling of $B_1, B_2$, in the examples we work through in the sequel both scale at most poly-logarithmically in $T$.
 
 \item \textbf{Parameter error bound.} Once $B_1, B_2$ are controlled, then
the upper bound on $\sup_{\theta \in \mathrm{conv}\{ \hat{\theta}^\e_{m,T}, \theta_\star\} } \Hg{\theta}{\theta_\star}$ (which holds on $\calE_1$) derived in Step 1 can be used to establish condition  \eqref{eq:hellinger_perturbation_radius}.
Concretely, this is done via a requirement on the minimum number
of trajectories $m$.
For the case when $\calP$ is log-concave, this requirement
scales as 
\begin{align}
    m \gtrsim \max\{ B_1^4, B_2^2 \} \log(c_0 m \cdot \mathrm{diam}(\Theta) \calN_{\calI_{\max}}/\delta), \label{eq:framework_B1_B2_log_concave}
\end{align}
whereas in the general case the trajectory requirement scales as
\begin{align}
    m \gtrsim \max\{ B_1^4, B_2^2 \} \gamma_2^2 \lambda_{\max}(\calI_{\max}) \log(c_0 \calN_{\calI_{\max}}/\delta). \label{eq:framework_B1_B2_general}
\end{align}
Given condition \eqref{eq:hellinger_perturbation_radius}, \Cref{prop:hellinger_perturbation} yields the following bound on the parameter error:
\begin{align}
    \norm{\hat{\theta}^\e_{m,T} - \theta_\star}^2_{\calI_2(\theta_\star, \hat{\theta}^\e_{m,T})} \lesssim m^{-1} \log(c_0 \calN_{\calI_{\max}}/\delta). \label{eq:framework_param_error}
\end{align}
While this rate is still not quite the CLT rate \eqref{eq:asymptotic_normality_parameter} as the dependence on
$\calI_2(\theta_\star, \hat{\theta}^\e_{m,T})$ is not necessarily variance optimal, for many practical applications this rate may be sufficient, especially if it is possible to show that $\calI_2(\theta_\star, \hat{\theta}^\e_{m,T}) \succeq \Omega(T) \cdot I_p$ on $\calE_1$. 

 \item \textbf{Verify FI radius.} 
 In order to apply the second part \Cref{prop:hellinger_perturbation} (i.e., obtain the bound \eqref{eq:hellinger_perturbation}), the FI radius condition \eqref{eq:hellinger_perturbation_FI_radius}
 remains to be verified
 for $\theta_0 = \theta_\star$ and $\theta_1 = \hat{\theta}^\e_{m,T}$.
 To do this, we often rely on the following upper bound:
 \begin{align*}
    \sup_{\theta \in \mathrm{conv}\{\theta_\star, \hat{\theta}^\e_{m,T}\}} \opnorm{ \calI(\theta_\star)^{-1/2} \calI(\theta) \calI(\theta_\star)^{-1/2} - I_p } &\leq  \sup_{\theta \in \mathrm{conv}\{\theta_\star, \hat{\theta}^\e_{m,T}\}} \frac{\opnorm{ \calI(\theta)  - \calI(\theta_\star) }}{\lambda_{\min}\left(\calI(\theta_\star)\right)}.
 \end{align*}
We then proceed to show a bound on $\opnorm{ \calI(\theta_0) - \calI(\theta_1) }$ for $\theta_0, \theta_1 \in \Theta$ of the following form (see \Cref{prop:FI_lipschitz} for a precise statement):
\begin{align*}
    \opnorm{ \calI(\theta_0) - \calI(\theta_1) } \lesssim T\left[ L \norm{\theta_0-\theta_1} + B_{\calI} \Hg{\theta_0}{\theta_1} \right],
\end{align*}
for suitable Lipschitz-like constants $L, B_{\calI}$.
This implies that
\begin{align*}
    \sup_{\theta \in \mathrm{conv}\{\theta_\star, \hat{\theta}^\e_{m,T}\}} \frac{\opnorm{ \calI(\theta)  - \calI(\theta_\star) }}{\lambda_{\min}\left(\calI(\theta_\star)\right)} \lesssim \frac{1}{\lambda_{\min}(\bar{\calI}(\theta_\star))} \big[ L \norm{\hat{\theta}^\e_{m,T} - \theta_\star} + B_{\calI} \sup_{\theta \in \mathrm{conv}\{ \theta_\star, \hat{\theta}^\e_{m,T}\}} \Hg{\theta}{\theta_\star}  \big].
\end{align*}
The RHS of this expression can be bounded by combining
either \eqref{eq:metricentropyhellinger} or \eqref{eq:framework_hellinger_conv_hull_bound_general} (depending on which one holds)
with \eqref{eq:framework_param_error}.
 Altogether, we have that condition \eqref{eq:hellinger_perturbation_FI_radius} holds given a minimum amount of trajectories $m$ when $\calP$ is log-concave:
 \begin{align}
     mT \gtrsim \frac{L^2\log(c_0 \calN_{\calI_{\max}}/\delta)}{\lambda_{\min}^2(\bar{\calI}(\theta_\star)) \underline{\mu}} \quad \textrm{and} \quad m \gtrsim \frac{B_{\calI}^2\log(c_0m \cdot \mathrm{diam}(\Theta) \calN_{\calI_{\max}}/\delta)}{\lambda_{\min}^2(\bar{\calI}(\theta_\star))}, \label{eq:framework_log_concave_FI_lipschitz}
 \end{align}
 where $\underline{\mu} := \lambda_{\min}( \calI_2(\theta_\star, \hat{\theta}^\e_{m,T}) )/T$.
 On the other hand, if $\calP$ is not log-concave, we have
 that the sufficient condition for \eqref{eq:hellinger_perturbation_FI_radius} to hold
 is 
 \begin{align}
    mT \gtrsim \frac{L^2\log(c_0 \calN_{\calI_{\max}}/\delta)}{\lambda_{\min}^2(\bar{\calI}(\theta_\star)) \underline{\mu}} \quad \textrm{and} \quad m \gtrsim \frac{B_{\calI}^2 \gamma_2^2 \lambda_{\max}(\calI_{\max}) \log(c_0 \calN_{\calI_{\max}}/\delta)}{\lambda^2_{\min}(\bar{\calI}(\theta_\star))}. \label{eq:framework_general_FI_lipschitz}
 \end{align}
 
\item \textbf{Final result.} Combining all previous steps, we have that as long as:
\begin{enumerate}[label=(\roman*)]
    \item \textit{For log-concave $\calP$:} the conditions \eqref{eq:framework_B1_B2_log_concave} and \eqref{eq:framework_log_concave_FI_lipschitz} on $m$ hold,
    \item \textit{For non-log-concave $\calP$:} the conditions 
    \eqref{eq:framework_gamma_1_bound}, \eqref{eq:framework_B1_B2_general}, 
    and \eqref{eq:framework_general_FI_lipschitz} on $m$ hold,
\end{enumerate}
then we obtain the following rate with probability at least $1-\delta$:
\begin{align*}
    \norm{\hat{\theta}^\e_{m,T}-\theta_\star}^2_{\bar{\calI}(\theta_\star)} \lesssim \frac{\log(c_0 \calN_{\calI_{\max}}/\delta)}{mT}.
\end{align*}
For the parametric function classes
considered in this work, we will generally have
$\log(\calN_{\calI_{\max}}/\delta) \lesssim p \log(mT/\delta)$
due to the standard volumetric estimate, which yields 
the CLT rate \eqref{eq:asymptotic_normality_parameter}
up to logarithmic factors.
However, depending
on the properties of the stochastic process generated by $p_\theta$, 
in particular the growth rate of the typical realization of $z_{1:T}$, the dependence on $T$
may be worse; an example of this will be given in \Cref{sec:case_studies:l1_regression}.

We conclude by making a brief remark
regarding the required scaling on $m$.
For the log-concave $\calP$ case, the required conditions generically
yield the form $m \gtrsim \mathrm{polylog}(T)$ (ignoring all other problem parameters)
whereas for the non-log-concave $\calP$ case, the scaling 
requirement increases to $m \gtrsim T \cdot \mathrm{polylog}(T)$.
For the latter case, we believe the linear scaling in $T$ 
to be an artifact
of our analysis strategy, in particular the step
taken in \eqref{eq:framework_non_log_concave_bound}.
We leave improving this step to future work.

\end{enumerate}

\begin{rmk}[Single-step Hellinger Identifiability]
\label{rmk:single_step_identifiability}
One simple method we utilize to obtain sub-optimal---but \emph{problem specific}---Hellinger identifiability (cf.~\Cref{def:identifiability}) constants is
through the use of the \emph{data processing inequality} (DPI)
for $f$-divergences. 
Suppose that $z_1 \sim \rho_1$ for all $\theta \in \Theta$. Then we have:
\begin{align*}
    \E_{z_1 \sim \rho_1}[ \HgSq{p_{\theta}(z_2 \mid z_1)}{p_{\theta_\star}(z_2 \mid z_1)} ] = \HgSq{ p_{\theta}(z_{1:2}) }{ p_{\theta_\star}(z_{1:2}) } \leq \HgSq{p_\theta}{p_{\theta_\star}}, 
\end{align*}
where the equality holds from
\cite[Prop.~7.2]{polyanskiy2025information}
and the inequality is the DPI for $f$-divergences
\cite[cf.][Thm.~7.4]{polyanskiy2025information}.
While the inequality above is often lossy, it is in practice
often much easier to prove identifiability using the single-step distributions, i.e.,
\begin{align*}
     \E_{z_1 \sim \rho_1}[ \HgSq{p_{\theta}(z_2 \mid z_1)}{p_{\theta_\star}(z_2 \mid z_1)} ] \leq \gamma_1^2 \Longrightarrow \norm{\theta-\theta_\star}^2 \leq \gamma_2^2 \cdot \E_{z_1 \sim \rho_1}[ \HgSq{p_{\theta}(z_2 \mid z_1)}{p_{\theta_\star}(z_2 \mid z_1)} ].
\end{align*}
We will show several examples of this in the sequel.
\end{rmk}

The remainder of this paper is dedicated to realizing the Hellinger localization framework on a diverse set of estimation problems, which we turn to in \Cref{sec:case_studies}.

\section{Case Studies}
\label{sec:case_studies}

This section contains the four
multi-trajectory
parameter recovery with ERM 
case studies that we
consider in this work:
(i) a mixture of two-state Markov chains (\Cref{sec:case_studies:nonmixing}),
(ii) a linear regression from dependent covariates
setup with general (i.e., non-Gaussian) product-noises (\Cref{sec:case_studies:l1_regression}),
(iii) a GLM setup with a non-expansive, non-monotonic
activation function (\Cref{sec:case_studies:glm}),
and (iv) a simple linear-attention sequence model
(\Cref{sec:case_studies:sequence_modeling}). The problem setup and analysis
for each case study is fairly self-contained, and
can be read in any order.

    \subsection{Mixture of Two-State Markov Chains}
\label{sec:case_studies:nonmixing}

We build on the example from \Cref{sec:twostatemc} by considering the following mixture formulation.
Suppose we have two Markov chains $M^{(0)}$, $M^{(1)}$,
and a Bernoulli distribution $P$ on $\{0, 1\}$.
The generative process we consider proceeds by first
sampling $B \sim P$ and $z_1 \sim \rho_1$ independently,
and then generating $z_{t+1} \mid z_t, B$ from
$M^{(B)}_{z_t, z_{t+1}}$.
The goal is to recover the 
parameters for the two Markov chains $M^{(0)}$, $M^{(1)}$ 
given $m$ trajectories of length $T$ from this process ($\calD_{m,T}$), where $B^{(i)}$ is unobserved for each trajectory $i$.
Such a problem is a special case of learning from mixtures of Markov chains~\cite{gupta2016mixturesofMCs,pmlr-v202-kausik23a}.
Our motivation for studying this problem is two-fold: (a) the parameters of both Markov chains clearly cannot be learned in a single-trajectory setting, necessitating a multi-trajectory approach, and
(b) the trajectory process $\{z_t\}$ is \emph{not} $\alpha$-mixing, but we can still apply the Hellinger localization framework to derive sharp rates
directly for the MLE. 

Let us define $\calP$ as instances of this Markov chain mixture with transition matrices:
\begin{align}
M^{(0)}=\begin{pmatrix}\theta_{0} & 1-\theta_{0} \\ 1-\theta_{0} & \theta_{0}\end{pmatrix},\quad M^{(1)}=\begin{pmatrix}\theta_{1} & 1-\theta_{1} \\ 1-\theta_{1} & \theta_{1}\end{pmatrix}, \label{eq:markov_two_state_mixture}
\end{align}
where $\theta = (\theta_0, \theta_1) \in \Theta := [\mu, 1-\mu]^2$ with $0 < \mu < 1/2$,
$P = \mathrm{Bern}(1/2)$, and $\rho_1 = \mathrm{Unif}(\{1, 2\})$.
In the remainder of the section, we will use $\Pr_{\theta}$ to denote a trajectory measure on the space of $z_{1:\infty}$ realized by a fixed parameter $\theta\in\Theta$, and for $i=0,1$, $\Pr_{\theta}^{(i)}(\cdot):=\Pr_{\theta}\left(\cdot\;|\;B=i\right)$. We further use $\E_{\theta}$ and $\E_{\theta}^{(i)}$ to denote expectation under $\Pr_{\theta}$ and $\Pr_{\theta}^{(i)}$ respectively,
so that $\E_\theta[X] = \frac{1}{2} \left(\E_\theta^{(0)}[X] + \E_\theta^{(1)}[X]\right)$
for any random variable $X$.

\paragraph{Mixtures are not $\alpha$-mixing.}
We give a short argument illustrating the lack of $\alpha$-mixing
for the process $\{z_t\}$. We fix a $p_\theta \in \calP$ with $\theta_0 \neq \theta_1$.
First, we recall the definition of $\alpha$-mixing and
introduce some notation.
For $a \leq b$, we let $z_{a:b} = (z_a, \dots, z_b)$, and
we let $\sigma(z_{a:b})$ denote the $\sigma$-algebra generated
by the subsequence $z_{a:b}$. 
The $\alpha$-mixing coefficients are defined as
(cf.~\cite[Eq. 2.2]{bradley2005mixing}, \cite[Def. 2.2]{yu1994rates}):
\begin{align}
    \alpha(k) := \sup_{j \in \N_+} \sup\left\{ \abs{\Pr_\theta( A_j \cap B_{j,k} ) - \Pr_\theta(A_j) \Pr_\theta(B_{j,k})} \mid A_j \in \sigma(z_{1:j}), B_{j,k} \in \sigma(z_{j+k:\infty}) \right\}, \quad k \in \N_+.
\end{align}
The process $\{z_t\}$ is denoted $\alpha$-mixing if $\alpha(k) \to 0$ as $k \to \infty$.
We consider $\alpha$-mixing in this section, as it is the \emph{weakest} notion of
dependency used in the literature; in particular
it is known that $\psi$-mixing $\Rightarrow$
$\phi$-mixing $\Rightarrow$ $\beta$-mixing $\Rightarrow$ $\alpha$-mixing, and furthermore $\rho$-mixing $\Rightarrow$ $\alpha$-mixing as well~\cite{bradley2005mixing}.

We now proceed as follows. A simple computation shows that for $A_j \in \sigma(z_{1:j}), B_{j,k} \in \sigma(z_{j+k:\infty})$:
\begin{align*}
    \Pr_\theta(B_{j,k} \mid A_j) - \Pr_\theta(B_{j,k}) &= \left( \Pr_\theta(B=0\mid A_j) - \frac{1}{2} \right) \left( \Pr_\theta^{(0)}(B_{j,k}) - \Pr_\theta^{(1)}(B_{j,k}) \right) + \Delta(A_j, B_{j,k}),
\end{align*}
where
\begin{align*}
    \Delta(A_j, B_{j,k}) &:= \Pr_\theta(B=0\mid A_j)\left[ \Pr_\theta^{(0)}(B_{j,k} \mid A_j) - \Pr_\theta^{(0)}(B_{j,k}) \right] \\
    &\qquad + \Pr_\theta(B=1\mid A_j)\left[ \Pr_\theta^{(1)}(B_{j,k} \mid A_j) - \Pr_\theta^{(1)}(B_{j,k}) \right].
\end{align*}
Hence, by triangle inequality,
\begin{align*}
    &\abs{\Pr_\theta( A_j \cap B_{j,k} ) - \Pr_\theta(A_j) \Pr_\theta(B_{j,k})} \\
    &\geq \Pr_\theta(A_j) \bigabs{ \Pr_\theta(B=0\mid A_j) - \frac{1}{2} } \bigabs{ \Pr_\theta^{(0)}(B_{j,k}) - \Pr_\theta^{(1)}(B_{j,k})  } - \abs{\Delta(A_j, B_{j,k})}.
\end{align*}
We now select $j=2$, and let $k \in \N_+$. We define the events $A_2$ and $B_{2,k}$ to be:
\begin{align*}
    A_2 := \{ z_{1:2} = (1, 1) \}, \quad B_{2,k} := \{ z_{2+k:2+k+1} = (1, 1) \}.
\end{align*}
We next observe that for $i \in \{0, 1\}$,
\begin{align*}
    \Pr_\theta^{(i)}(B_{2,k}) = \Pr_\theta^{(i)}( z_{2+k} = 1, z_{2+k+1} = 1) = \frac{\theta_i}{2},
\end{align*}
and hence $\abs{\Pr_\theta^{(0)}(B_{j,k}) - \Pr_\theta^{(1)}(B_{j,k})} = \abs{\theta_0-\theta_1}/2$.
We also note that
$\lim_{k \to \infty} \abs{\Delta(A_2, B_{2,k})} = 0$, 
since we have $\lim_{k \to \infty} \abs{ \Pr_\theta^{(i)}( z_{j+k} = 1\mid z_j = 1 ) - 1/2 } = 0$ by the ergodicity of the 
individual Markov chains $M^{(0)}, M^{(1)}$~\cite[see e.g.,][]{levin2017markov}.
On the other hand,
\begin{align*}
    \Pr_\theta(B=0 \mid A_2) = \frac{\Pr_\theta^{(0)}(A_2)}{\Pr_\theta^{(0)}(A_2) + \Pr_\theta^{(1)}(A_2)} = \frac{\theta_0}{\theta_0 + \theta_1},
\end{align*}
and hence $\abs{ \Pr_\theta(B=0 \mid A_2) - 1/2 } > 0$ as
$\theta_0 \neq \theta_1$.
Therefore, we have
\begin{align*}
    \liminf_{k \to \infty} \alpha(k) &\geq \liminf_{k \to \infty} \left( \Pr_\theta(A_2) \bigabs{ \frac{\theta_0}{\theta_0 + \theta_1} - \frac{1}{2} } \frac{\abs{\theta_0 - \theta_1}}{2} - \abs{\Delta(A_2, B_{2,k})} \right) \\
    &= \Pr_\theta(A_2) \bigabs{ \frac{\theta_0}{\theta_0 + \theta_1} - \frac{1}{2} } \frac{\abs{\theta_0 - \theta_1}}{2} - \limsup_{k \to \infty} \abs{\Delta(A_2, B_{2,k})} \\
    &= \Pr_\theta(A_2) \bigabs{ \frac{\theta_0}{\theta_0 + \theta_1} - \frac{1}{2} } \frac{\abs{\theta_0 - \theta_1}}{2} > 0,
\end{align*}
and hence we have that $\{z_t\}$ is not $\alpha$-mixing.

\begin{rmk}
Although $\{z_{t}\}_{t=1}^{\infty}$ is not $\alpha$-mixing, it is ergodic as $M^{(0)}$ and $M^{(1)}$ admit the same stationary distribution, which implies any time average of single trajectory will converge to the same marginal expectation. In general, the mixture of Markov chain process will be non-ergodic and non-mixing as long as the candidate transition matrices have different stationary distributions. 
\end{rmk}

\begin{rmk}
We remark that mixing coefficients are not
fully standardized in the literature, and depending on what
specific definition is adopted, the trajectory $\{ z_t \}$ could
be considered mixing.
As a specific example, in
\cite{kuznetsov2017generalization}
a weaker definition of $\beta$-mixing is considered,
defined as
$\beta(k) = \sup_{t \geq 1} \E_{z_{1:t}}[ \TV{ \Pr_{z_{t+k}}(\cdot \mid z_{1:t})}{\Pr_{z_{t+k}}(\cdot) } ]$.
Under this definition
$\{z_t\}$ is actually $\beta$-mixing, since 
the $M^{(i)}$'s admit the same stationary distribution.
However, there are many ways to modify the 
mixture model \eqref{eq:markov_two_state_mixture}
so that it is not $\beta$-mixing under this more relaxed definition. A simple modification is
\begin{align*}
M^{(0)}=\begin{pmatrix}\theta_{0} & 1-\theta_{0} \\ 1-\theta_{0}' & \theta_{0}'\end{pmatrix},\quad M^{(1)}=\begin{pmatrix}\theta_{1} & 1-\theta_{1} \\ 1-\theta_{1}' & \theta_{1}'\end{pmatrix},
\end{align*}
where $\theta_i' = \theta_i + \tau$ (modulating by one if necessary), where $\tau$ is a fixed offset.
Another option is to consider general two-state Markov chains (so that $\theta$ contains four total parameters).
For both modifications, the structure of the proof to be presented remains the same, although the detailed calculations may be different, especially for the general two-state parameterization. 
\end{rmk}

Towards stating our main result, we define the following quantities for $\theta \in (0, 1)^2$:
\begin{align*}
\mathrm{Gap}(\theta):=\left|\theta_{0}-\theta_{1}\right|, \quad \bar{\sigma}^2(\theta) := \max_{i=0,1} \sigma_i^2(\theta), \quad \underline{\sigma}^2(\theta) := \min_{i=0,1} \sigma_i^2(\theta).
\end{align*}
The following is our main parameter recovery bound for the mixture of Markov chains problem.
\begin{thm}\label{thm:mixturemc}
Fix $\delta \in (0, 1)$,
and define the constants $\rho_{\star} := \mathrm{Gap}(\theta_{\star})$
and $\sigma_{\min}^{2}:=\mu(1-\mu)$.
Suppose that the following conditions hold:
\begin{enumerate}[label=(\alph*)]
    \item $\theta_{\star,0} > \theta_{\star,1}$, %
    \item $T \gtrsim \max\left\{ \frac{\bar{\sigma}^2(\theta_\star)}{\rho_{\star}^4}, \frac{1}{\rho_{\star}^2}  \right\} \log^2(1/\mu) \log\left( \frac{ \bar{\sigma}^2(\theta_\star) }{\underline{\sigma}^4(\theta_\star)} \cdot \frac{\log(1/\mu)}{\rho_{\star} } \right)$,
    \item $m \gtrsim \max\left\{ \frac{1}{\rho_\star^4}, \frac{1}{\rho_\star^2 \underline{\sigma}^4(\theta_\star)}, \frac{T}{\rho_\star^2 \underline{\sigma}^{2}(\theta_{\star})} \right\} \log\left( \max\left\{ \frac{1}{\rho_\star^4}, \frac{1}{\rho_\star^2 \underline{\sigma}^4(\theta_\star)}, \frac{T}{\rho_\star^2 \underline{\sigma}^{2}(\theta_{\star})} \right\}  \frac{T}{\sigma^2_{\min} \delta} \right)$,
\end{enumerate}
Let $\hat{\theta}^\e_{m,T}$ denote the max FI discretized MLE estimator \eqref{eq:discrete_MLE} at resolution $\e=\delta/(2\sqrt{2m})$,
and suppose the MLE estimator
satisfies $\left( \hat{\theta}^\e_{m,T} \right)_0 \geq \left( \hat{\theta}^\e_{m,T} \right)_1$.
With probability at least $1-\delta$,
\begin{align}
\norm{\hat{\theta}^\e_{m,T} - \theta_\star}_{\bar{\calI}(\theta_{\star})}^2 \lesssim \frac{1}{mT}\log \left(\frac{mT}{\sigma_{\min}^{2}\delta } \right) \quad \textrm{and} \quad
\norm{\hat{\theta}^\e_{m,T} - \theta_\star}^2 \lesssim \frac{\bar{\sigma}^{2}(\theta_{\star})}{mT}\log \left(\frac{mT}{\sigma_{\min}^{2}\delta } \right). \label{eq:mixturemc_rate}
\end{align}

\end{thm}

Some remarks are in order for \Cref{thm:mixturemc}.
First, we note that Assumption (a) in \Cref{thm:mixturemc} does not change the generality of the result, as the distribution $p_\theta(z_{1:T})$ is invariant under parameter permutation (since $B$ is sampled uniformly over the two choices), and hence we can assume wlog that $\theta_{\star,0} > \theta_{\star,1}$. Furthermore, given an MLE $\hat{\theta}^{\e}_{m,T}$, we can always assume wlog $\left(\hat{\theta}^{\e}_{m,T}\right)_{0}>\left(\hat{\theta}^{\e}_{m,T}\right)_{1}$, otherwise we just permute the estimator. 
Second, we remark that \Cref{thm:mixturemc} is nearly fully \emph{instance dependent}, i.e., both the requirements $m, T$ and the final parameter error bound on do not involve the global bound $\mu$ outside of poly-logarithmic factors, but instead depend on 
the problem-specific parameters $\rho_\star, \underline{\sigma}(\theta_\star), \bar{\sigma}(\theta_\star)$ of the true, data-generating distribution.
Third, the rate prescribed by \eqref{eq:mixturemc_rate}
is in general not improvable, as it matches the two-state Markov chain
argument in \Cref{sec:twostatemc}, specifically the optimal rate in \eqref{eq:two_state_optimal_rate}.

We now discuss the requirements on $T$ via Assumption (b), and $m$ via Assumption (c).
Starting with the requirement on $T$ in Assumption (b),
the role of this condition is to ensure that 
there is sufficient information within a trajectory
to distinguish which chain most likely generated 
the data \emph{if the true parameters were known}; hence the scaling of $\mathrm{poly}(1/\mathrm{Gap}(\theta_\star))$ is quite intuitive.
On the other hand, the requirement on $m$ in Assumption (c) parallels that of \eqref{eq:two_state_m_req}
in the two-state Markov chain case (cf.~\Cref{sec:twostatemc}). The biggest difference is that in Assumption (c), we have the scaling of $m \gtrsim \tilde{\Omega}(T)$ instead of 
$m \gtrsim \tilde{\Omega}(1)$ in \eqref{eq:two_state_m_req}.
This comes the non-concavity of the MLE for the mixture problem, which required us to use \eqref{eq:framework_B1_B2_general}, compared with the concave MLE for the two-state mixture; as noted in \Cref{sec:hellinger_localization_framework}, 
the requirements on $m$ are worse for non-concave problems.
\paragraph{Comparison to existing results.} Learning mixture of Markov chains has been recently studied by a few authors~\cite{pmlr-v202-kausik23a,spaeh2023mixture, gupta2016mixturesofMCs}. From this set of works, most related to ours is \cite{pmlr-v202-kausik23a}, where the authors develop efficient algorithms for clustering and estimating the family of transition matrices. 
Adapted to our specific setting, \cite[Theorem 4]{pmlr-v202-kausik23a} reads that when:
$$
m\gtrsim \frac{\tau_{\mathrm{mix}}\log(1/\delta)}{\rho_{\star}^{3}},\quad T\gtrsim \tau_{\mathrm{mix}}\log(1/\delta), \quad 
\tau_{\mathrm{mix}} := \max_{i=0,1} \frac{1}{\min\{\theta_{\star,i}, 1 - \theta_{\star,i}\}},
$$
then their algorithm recovers an estimate $\hat{\theta}_{m,T}$ that satisfies with probability at least $1-\delta$:
\begin{align}
\norm{\hat{\theta}_{m,T}-\theta_{\star}}^{2}\lesssim \frac{\tau_{\mathrm{mix}}^{2/3}}{T^{2/3}}\frac{\log(1/\delta)}{m
}. \label{eq:mixturemc_rate_kausik}
\end{align}
We note that although the result from \cite{pmlr-v202-kausik23a} has less stringent requirements on both
the minimum number of trajectories $m$ and trajectory length $T$ compared with \Cref{thm:mixturemc},
the final rate \eqref{eq:mixturemc_rate_kausik}
has both (i) a $1/ (m T^{2/3})$ scaling in comparison
to a $1/(mT)$ scaling in \eqref{eq:mixturemc_rate},
and (ii) also scales proportion to $\tau_{\mathrm{mix}}^{2/3}$ as opposed to
$\bar{\sigma}^2(\theta_\star)$ in \eqref{eq:mixturemc_rate};
note that in general $\tau_{\mathrm{mix}}$ can grow arbitrarily large as $\theta_\star$ approaches the boundary of the positive orthant, whereas $\bar{\sigma}^2(\theta_\star) \leq 1/4$ always.
On the other hand, as mentioned previously, the
work \cite{pmlr-v202-kausik23a} provides an efficient algorithm which can also learn the distribution of the latent variable $B$, whereas our result \Cref{thm:mixturemc} uses the MLE estimate which,
in this case, requires maximizing a non-concave objective and does not handle the case where the
distribution of $B$ must be jointly learned.
Extensions of our analysis to more general mixture setups, in addition to practical algorithms such as
expectation maximization~\cite{McLachlan2000}, is 
left as interesting future work. In \Cref{subsec:mixture_general}, we comment in more detail on how
our proof techniques may be generalized to other mixture recovery problems.

\subsubsection{Preliminary Results for \Cref{thm:mixturemc}}

Our analysis resembles that of the two-state Markov chain case (cf.~\Cref{sec:twostatemc}), given that each trajectory can be associated with a particular chain if the trajectory length $T$ is sufficiently long. In the following, we state a few auxiliary results that will be crucial towards enabling our
analysis. The following subset $\Theta' \subset \Theta$ 
plays an important role in localizing the problem-specific parameters:
\begin{align}
    \Theta' := \left\{ \theta \in \Theta \mid \norm{\theta-\theta_\star} \leq \min\{\mathrm{Gap}(\theta_\star)/(2\sqrt{2}), \underline{\sigma}^2(\theta_\star)/2\} \right\}. \label{eq:mixture_Theta_prime}
\end{align}

\begin{myprop}\label{lemma:ergodictheorem}
Fix $\theta = (\theta_0, \theta_1) \in (0, 1)^2$ and suppose that $\theta_0 \neq \theta_1$.
Define for $i \in \{0, 1\}$:
\begin{align*}
    \Delta_i(\theta) := \KL{\mathrm{Bern}(\theta_i)}{\mathrm{Bern}(\theta_{1-i})}, \quad
    \sigma_i^2(\theta) := \theta_i(1-\theta_i).
\end{align*}
Fix $\e, \delta \in (0, 1)$, and suppose that $T$ satisfies:
\begin{align*}
    T \gtrsim \max_{i=0,1} \max\left\{ \frac{\ell^2(\theta) \sigma_i^2(\theta) \log(2/\delta)}{\Delta_i^2(\theta)}, \frac{1}{\Delta_i(\theta)}\left[ \log\left(\frac{1}{\e}\right) + \ell(\theta) \log\left(\frac{2}{\delta}\right) \right] \right\}, \quad  \ell(\theta) := \bigabs{ \log\left( \frac{\theta_0(1-\theta_{1})}{\theta_{1} (1-\theta_0)} \right) }.
\end{align*}
Denote the posterior density of $B$ evaluated at $0$:
$$
w_{\theta}(z_{1:T}) := p(B=0\;|\;z_{1:T})=\frac{p_{\theta_0}(z_{1:T})}{p_{\theta_0}(z_{1:T}) + p_{\theta_1}(z_{1:T})}.
$$
We have:
\begin{align*}
    \Pr^{(0)}_\theta\left( w_{\theta}(z_{1:T}) \geq 1 - \e \right) \geq 1 - \delta, \quad \Pr^{(1)}_\theta\left( w_{\theta}(z_{1:T}) \leq \e \right) \geq 1 - \delta.
\end{align*}
\end{myprop}
\begin{proof}
To ease notation, we let $T' := T-1$ for what follows.
Given $z_{1:T}$, we have that
\begin{align*}
    \log r_{\theta}(z_{1:T}) &:= \log\frac{p_{\theta_0}(z_{1:T})}{p_{\theta_1}(z_{1:T})} = \log\left(\frac{\theta_0}{\theta_1}\right) \cdot \sum_{t=1}^{T'} \ind\{ z_{t+1} = z_t \} + \log\left(\frac{1-\theta_0}{1-\theta_1}\right) \cdot \sum_{t=1}^{T'} \ind\{ z_{t+1} \neq z_t \} \\
    &=: \log\left(\frac{\theta_0}{\theta_1}\right) \cdot N_{\mathrm{stay}}(z_{1:T}) + \log\left(\frac{1-\theta_0}{1-\theta_1}\right) \cdot (T'- N_{\mathrm{stay}}(z_{1:T})).
\end{align*}
Next, since conditioned on $B=i$, the random variable
$N_{\mathrm{stay}}(z_{1:T}) \sim \mathrm{Bin}(T', \theta_i)$,
and therefore the conditional MGF of $N_{\mathrm{stay}}(z_{1:T})$
is: 
\begin{align*}
    \E^{(i)}_\theta[ \exp( \lambda N_{\mathrm{stay}}(z_{1:T}) ) ] 
    = (e^\lambda \theta_i + (1-\theta_i))^{T'}.
\end{align*}
Hence,
we have 
\begin{align*}
    \log \E^{(i)}_\theta[ \exp( \lambda (N_{\mathrm{stay}}(z_{1:T}) - T' \theta_i ) ] = T' \log( e^\lambda \theta_i + (1-\theta_i)) - T'\theta_i \lambda.
\end{align*}
By \Cref{prop:log_mgf_sub_exp}, we have that with probability at least $1-\delta$ over $z_{1:T} \sim p_{\theta_i}$,
\begin{align*}
    \bigabs{ N_{\mathrm{stay}}(z_{1:T}) - T' \theta_i } \leq 2\sqrt{2e T' \sigma_i^2(\theta) \log(2/\delta)} + 2 \log(2/\delta).
\end{align*}
Let us temporarily call this event $\calE$. On event $\calE$ under $p_{\theta_0}$,
\begin{align*}
    &\bigabs{\log r_{\theta}(z_{1:T}) - T'\KL{\mathrm{Bern}(\theta_0)}{\mathrm{Bern}(\theta_1)}} \\
    &= \bigabs{\log\left(\frac{\theta_0}{\theta_1}\right)\left( N_{\mathrm{stay}}(z_{1:T}) - T'\theta_0 \right) + \log\left(\frac{1-\theta_0}{1-\theta_1}\right)\left( T'\theta_0 - N_{\mathrm{stay}}(z_{1:T}) \right)} \\
    &= \bigabs{ \log\left( \frac{\theta_0(1-\theta_1)}{\theta_1 (1-\theta_0)} \right) } \abs{  N_{\mathrm{stay}}(z_{1:T}) - T' \theta_0 } \\
    &\lesssim \bigabs{ \log\left(  \frac{\theta_0(1-\theta_1)}{\theta_1 (1-\theta_0)} \right) } ( \sqrt{T \sigma_0^2(\theta) \log(2/\delta)} +  \log(2/\delta)  ) \\
    &=: \Psi_1^{(0)} \sqrt{T} + \Psi_2^{(0)},
\end{align*}
where
\begin{align*}
    \Psi_1^{(0)} := \ell(\theta) \sqrt{\sigma_0^2(\theta) \log(2/\delta)}, \quad \Psi_2^{(0)} := \ell(\theta) \log(2/\delta).
\end{align*}
Therefore we have on event $\calE$, there exists a universal $c > 0$ such that:
\begin{align}
    \log r_{\theta}(z_{1:T}) \geq T' \Delta_0(\theta) - c(\Psi_1^{(0)} \sqrt{T} + \Psi_2^{(0)}).   \label{eq:ergodictheorem_ratiobound}
\end{align}
Now, we observe that for any $\e \in (0, 1)$,
\begin{align*}
    w_{\theta}(z_{1:T}) \geq 1-\e \Longleftrightarrow \log r_{\theta}(z_{1:T}) \geq \log\left(\frac{1-\e}{\e}\right).
\end{align*}
To achieve the claimed result, it remains to derive sufficient conditions on $T$ so that the RHS of \eqref{eq:ergodictheorem_ratiobound} is lower bounded by $\log((1-\e)/\e)$.
First, we require that 
\begin{align*}
    T'\Delta_0(\theta)/2 \geq c\Psi_1^{(0)}\sqrt{T} \Longleftrightarrow T \gtrsim (\Psi_1^{(0)})^2/\Delta_0^2(\theta).
\end{align*}
We also require that
\begin{align*}
    T'\Delta_0(\theta)/2 \geq \log\left(\frac{1-\e}{\e}\right) + c\Psi_2^{(0)} \Longleftarrow T \gtrsim \frac{1}{\Delta_0(\theta)}\left[  \log\left(\frac{1}{\e}\right) + \Psi_2^{(0)} \right].
\end{align*}
Hence, we have that as long as:
\begin{align*}
    T \gtrsim \max\left\{ \frac{(\Psi_1^{(0)})^2}{\Delta_0^2(\theta)},  \frac{1}{\Delta_0(\theta)}\left[  \log\left(\frac{1}{\e}\right) + \Psi_2^{(0)}\right]  \right\},
\end{align*}
then we have
\begin{align*}
    \Pr^{(0)}_\theta\{ w_{\theta}(z_{1:T}) \geq 1 - \e \} \geq 1 - \delta.
\end{align*}
The proof for $\Pr^{(1)}_\theta$ proceeds exactly the same as above with the roles of 
$\theta_0, \theta_1$ swapped.
\end{proof}

\begin{rmk}
We note that one could get a similar result by applying \cite[Theorem 3.9]{Paulin2015}. However, our requirement on $T$ from the above lemma does not depend linearly on the inverse spectral gap $\asymp [\min\{ \theta_i, 1-\theta_i \}]^{-1}$ of the chain defined by individual $\theta_{i}$'s.
\end{rmk}

\begin{mycorollary}\label{corollary:uniformergodic} Fix $\theta_{\star}=(\theta_{\star,0},\theta_{\star,1}) \in \Theta$ and $\e,\delta\in(0,1)$. Suppose $T$ satisfies
\begin{align*}
    T \gtrsim \max\left\{ \frac{\bar{\sigma}^2(\theta_\star)}{\mathrm{Gap}^4(\theta_\star)}\cdot \log^2(1/\mu)\log\left(\frac{2}{\delta}\right), \frac{1}{\mathrm{Gap}^2(\theta_\star)} \left[ \log\left(\frac{1}{\e}\right) + \log(1/\mu)\log\left(\frac{2}{\delta}\right)  \right]  \right\}.
\end{align*}
Then for any $\theta \in \Theta'$, we have:
\begin{align*}
    \Pr^{(0)}_\theta\left( w_{\theta}(z_{1:T}) \geq 1 - \e \right) \geq 1 - \delta, \quad \Pr^{(1)}_\theta\left( w_{\theta}(z_{1:T}) \leq \e \right) \geq 1 - \delta.
\end{align*}
\end{mycorollary}
\begin{proof}
First by Pinsker's inequality, we have:
\begin{align*}
\Delta_i(\theta) = \KL{\mathrm{Bern}(\theta_i)}{\mathrm{Bern}(\theta_{1-i})} \geq 2 \TV{\mathrm{Bern}(\theta_i)}{\mathrm{Bern}(\theta_{1-i})}^2 = 2 \mathrm{Gap}^2(\theta), \quad i \in \{0, 1\}.
\end{align*}
Next, we have that for $\theta \in \Theta'$:
\begin{align*}
    \mathrm{Gap}(\theta) &= \abs{\theta_0 - \theta_1} 
    = \abs{(\theta_{\star,0} - \theta_{\star,1}) + (\theta_0 - \theta_{\star,0}) - (\theta_1 - \theta_{\star,1}) } \\
    &\geq \mathrm{Gap}(\theta_\star) - \norm{\theta - \theta_\star}_1 
    \geq \mathrm{Gap}(\theta_\star) - \sqrt{2}\norm{\theta-\theta_\star} \geq \mathrm{Gap}(\theta_\star)/2.
\end{align*}
Consequently for $\theta \in \Theta'$ and $i \in \{0,1\}$,
$\Delta_i(\theta) \geq \mathrm{Gap}^2(\theta_\star)/2$.
Next, for any $\theta \in \Theta$, we have:
\begin{align*}
    \ell(\theta) = \bigabs{ \log\left( \frac{\theta_0(1-\theta_1)}{\theta_1(1-\theta_0)} \right)} \leq \bigabs{\log\left( \frac{1-\theta_0}{\theta_0} \right)} + \bigabs{\log\left( \frac{1-\theta_1}{\theta_1} \right)} \leq 2 \max_{\theta \in [\mu, 1-\mu]} \bigabs{\log\left(\frac{1-\theta}{\theta}\right)} \leq 2\log(1/\mu),
\end{align*}
since we assumed $\mu < 1/2$. 
Finally, for any $\theta \in \Theta'$ and $i \in \{0, 1\}$,
\begin{align*}
    \abs{ \sigma_i^2(\theta) - \sigma_i^2(\theta_\star) } \leq \abs{\theta_i - \theta_{\star,i}} \leq \norm{\theta - \theta_\star} \leq \sigma_i^2(\theta_\star)/2 \quad&\Longrightarrow\quad \sigma_i^2(\theta) \leq 3\sigma_i^2(\theta_\star)/2 \\
    &\Longrightarrow\quad \bar{\sigma}^2(\theta) \leq 3\bar{\sigma}^2(\theta_\star)/2.
\end{align*}
The claim now follows from \Cref{lemma:ergodictheorem}.
\end{proof}

\begin{mycorollary}
\label{corollary:lengthrequirement}
Fix $\theta_{\star}=(\theta_{\star,0},\theta_{\star,1}) \in \Theta$
and $\e \in (0, 1)$, and suppose $T$ satisfies:
\begin{align*}
    T \gtrsim \max\left\{ \frac{\bar{\sigma}^2(\theta_\star)}{\mathrm{Gap}^4(\theta_\star)}, \frac{1}{\mathrm{Gap}^2(\theta_\star)}  \right\} \log^2(1/\mu) \log\left(\frac{2}{\e}\right).
\end{align*}
Then for any $\theta \in \Theta'$, we have:
\begin{align*}
    \E^{(0)}_\theta[ \abs{w_{\theta}(z_{1:T}) - 1} ] \leq \e, \quad \E^{(1)}_{\theta}[ w_{\theta}(z_{1:T}) ] \leq \e.
\end{align*}
Furthermore, fix any $k \in \N_+$ and $\eta \in (0, 1)$, and suppose that $T$ satisfies:
\begin{align*}
    T \gtrsim \max\left\{ \frac{\bar{\sigma}^2(\theta_\star)}{\mathrm{Gap}^4(\theta_\star)}, \frac{1}{\mathrm{Gap}^2(\theta_\star)}  \right\} \log^2(1/\mu) \log\left( \frac{2}{\eta} \left( \frac{k \log(1/\mu)}{\mathrm{Gap}(\theta_\star) } \right)^k  \right).
\end{align*}
Then for any $\theta \in \Theta'$, we have:
\begin{align*}
    \E^{(0)}_\theta[ \abs{w_{\theta}(z_{1:T}) - 1} ] \leq \frac{\eta}{T^k}, \quad \E^{(1)}_{\theta}[ w_{\theta}(z_{1:T}) ] \leq \frac{\eta}{T^k}.
\end{align*}
\end{mycorollary}
\begin{proof}
For the first part of the statement,
we temporarily denote $\calE := \{ \abs{w_{\theta}(z_{1:T}) - 1} > \e/2 \}$.
Then we have, using $w_{\theta}(z_{1:T}) \in [0, 1]$:
\begin{align*}
    \E^{(1)}_\theta[ \abs{w_{\theta}(z_{1:T}) - 1} ] &= \E^{(1)}_\theta[ \abs{w_{\theta}(z_{1:T}) - 1}\cdot \ind\{\calE\} ] + \E^{(1)}_\theta[ \abs{w_{\theta}(z_{1:T}) - 1}\cdot \ind\{\calE^c\} ] \\
    &\leq \Pr^{(1)}_\theta( \abs{w_{\theta}(z_{1:T}) - 1} > \e/2 ) + \e/2 \leq \e/2 + \e/2 = \e,
\end{align*}
where the last inequality holds from 
\Cref{corollary:uniformergodic}, given our requirement on $T$.
The result for $\E^{(1)}_\theta[ w_{\theta}(z_{1:T}) ] \leq \e$ follows
using same proof.

For the second part of the statement, we invoke the first part
with $\e = \eta/T^k$, and use \Cref{prop:log_dominance}
to recover the dependence on $T$.
Specifically, we require
\begin{align*}
    T \geq A \log(2 T^k / \eta), \quad A := c_0 \max\left\{ \frac{\bar{\sigma}^2(\theta_\star)}{\mathrm{Gap}^4(\theta_\star)}, \frac{1}{\mathrm{Gap}^2(\theta_\star)}  \right\} \log^2(1/\mu).
\end{align*}
It suffices to require that:
\begin{align*}
    T/2 \geq A \log(2/\eta), \quad T/2 \geq kA \cdot \log{T}.
\end{align*}
Applying \Cref{prop:log_dominance} to the second inequality it suffices for
$T \gtrsim kA \log(kA)$.
Hence, simplifying further by bounding $\bar{\sigma}^2(\theta_\star) \leq 1/4$ yields the claim.
\end{proof}

\begin{myprop}\label{corollary:eigenvaluebound}
Fix $\theta_{\star}=(\theta_{\star,0},\theta_{\star,1}) \in \Theta$.
Suppose $T$ satisfies:
\begin{align*}
    T \gtrsim \max\left\{ \frac{\bar{\sigma}^2(\theta_\star)}{\mathrm{Gap}^4(\theta_\star)}, \frac{1}{\mathrm{Gap}^2(\theta_\star)}  \right\} \log^2(1/\mu) \log\left( \frac{ \bar{\sigma}^2(\theta_\star) }{\underline{\sigma}^4(\theta_\star)} \cdot \frac{\log(1/\mu)}{\mathrm{Gap}(\theta_\star) }   \right).
\end{align*}
We have for any $\theta \in \Theta'$:
\begin{align*}
    \frac{1}{4} \begin{pmatrix} \mathcal{I}(\theta_{0}) & 0 \\0 & \mathcal{I}(\theta_{1})\end{pmatrix} \preccurlyeq \calI(\theta) \preccurlyeq \frac{3}{4} \begin{pmatrix} \mathcal{I}(\theta_{0}) & 0 \\0 & \mathcal{I}(\theta_{1})\end{pmatrix},
\end{align*}
where $\mathcal{I}(\theta_{0})$ and $\mathcal{I}(\theta_{1})$ are 
the Fisher information of the individual two-state Markov chains under $\theta_{0}$ and $\theta_{1}$ respectively (cf.~\Cref{sec:twostatemc}),
i.e., $\mathcal{I}(\theta_{i})=\frac{T-1}{\theta_{i}(1-\theta_{i})}$
for $i \in \{0, 1\}$.
\end{myprop}
\begin{proof}
We start with calculating relevant derivatives for our mixture model $\calP$.
We recall that the parameter space is $\Theta = [\mu, 1-\mu]^2$.
Given $\theta \in \Theta$,
the log likelihood ratio is
$$
\log p_{\theta}(z_{1:T}) = \log \left(\frac{1}{2}p_{\theta_0}(z_{1:T})+\frac{1}{2}p_{\theta_1}(z_{1:T})\right)=\log \left(p_{\theta_0}(z_{1:T})+p_{\theta_1}(z_{1:T})\right)-\log 2.
$$
Therefore the first order information is
\begin{equation}\label{eq:mixture_twostate_gradient}
\nabla_{\theta}\log p_{\theta}(z_{1:T}) =\begin{pmatrix}
w_{\theta}(z_{1:T})\partial_{\theta_{0}}\log p_{\theta_0}(z_{1:T})  & (1-w_{\theta}(z_{1:T}))\partial_{\theta_{1}}\log p_{\theta_1}(z_{1:T})
\end{pmatrix}^{\T}.
\end{equation}
We now compute the second order information:
\begin{equation}\label{eq:mixture_twostate_hessian}
\begin{aligned}
\partial^{2}_{\theta_{0}}\log p_{\theta}(z_{1:T})&=
w_{\theta}(z_{1:T})(1-w_{\theta}(z_{1:T}))\left(\partial_{\theta_{0}}\log p_{\theta_{0}}(z_{1:T})\right)^{2}+w_{\theta}(z_{1:T})\partial^{2}_{\theta_{0}}\log p_{\theta_{0}}(z_{1:T}),\\
\partial^{2}_{\theta_{1}}\log p_{\theta}(z_{1:T})&=
w_{\theta}(z_{1:T})(1-w_{\theta}(z_{1:T}))\left(\partial_{\theta_{1}}\log p_{\theta_{1}}(z_{1:T})\right)^{2}+(1-w_{\theta}(z_{1:T}))\partial^{2}_{\theta_{1}}\log p_{\theta_{1}}(z_{1:T}),\\
\partial_{\theta_{1}}\partial_{\theta_{0}}\log p_{\theta}(z_{1:T})&=\partial_{\theta_{0}}\partial_{\theta_{1}}\log p_{\theta}(z_{1:T})=w_{\theta}(z_{1:T})(w_{\theta}(z_{1:T})-1)\partial_{\theta_{0}}\log p_{\theta_{0}}(z_{1:T})\partial_{\theta_{1}}\log p_{\theta_{1}}(z_{1:T}).
\end{aligned}
\end{equation}
Denoting
\begin{align*}
H_{\theta}(z_{1:T})&:=\begin{pmatrix} \partial^{2}_{\theta_{0}}\log p_{\theta}(z_{1:T}) & \partial_{\theta_{1}}\partial_{\theta_{0}}\log p_{\theta}(z_{1:T}) \\
\partial_{\theta_{0}}\partial_{\theta_{1}}\log p_{\theta}(z_{1:T}) & \partial^{2}_{\theta_{1}}\log p_{\theta}(z_{1:T})\end{pmatrix} = \nabla^2_\theta \log p_\theta(z_{1:T}),
\end{align*}
we compute the Fisher information of $\theta$ as:
\begin{align*}
\mathcal{I}(\theta)&=-\E_\theta\left[H_{\theta}(z_{1:T})\right]=\frac{1}{2}\E^{(0)}_\theta\left[-H_{\theta}(z_{1:T})\right]+\frac{1}{2}\E^{(1)}_\theta\left[-H_{\theta}(z_{1:T})\right],
\end{align*}
We now further define
\begin{align*}
\E^{(0)}_\theta\left[-H_{\theta}(z_{1:T})\right]&=\begin{pmatrix} \E_\theta^{(0)}\left[-\partial^{2}_{\theta_{0}}\log p_{\theta_{0}}(z_{1:T})\right] & 0 \\0 & 0\end{pmatrix}+\E^{(0)}_\theta\left[-\left(H_{\theta}(z_{1:T})-\begin{pmatrix} \partial^{2}_{\theta_{0}}\log p_{\theta_{0}}(z_{1:T}) & 0 \\0 & 0\end{pmatrix}\right)\right],\\
&:=\begin{pmatrix} \mathcal{I}(\theta_{0}) & 0 \\0 & 0\end{pmatrix}+\E^{(0)}_\theta\left[E_{0}(\theta)\right],\\
\E^{(1)}_\theta\left[-H_{\theta}(z_{1:T})\right]&=\begin{pmatrix} 0 & 0 \\0 & \E^{(1)}_\theta\left[-\partial^{2}_{\theta_{1}}\log p_{\theta_{1}}(z_{1:T})\right]\end{pmatrix}+\E^{(1)}_\theta\left[-\left(H_{\theta}(z_{1:T})-\begin{pmatrix} 0 & 0 \\0 & \partial^{2}_{\theta_{1}}\log p_{\theta_{1}}(z_{1:T})\end{pmatrix}\right)\right],\\
&:=\begin{pmatrix} 0 & 0 \\0 & \mathcal{I}(\theta_{1})\end{pmatrix}+\E^{(1)}_\theta\left[E_{1}(\theta)\right].
\end{align*}
We highlight the following expressions for $E_0, E_1$:
\begin{align*}
& \left(E_0(\theta)\right)_{11}=
(1-w_{\theta}(z_{1:T}))\left(\partial^{2}_{\theta_{0}}\log p_{\theta_{0}}(z_{1:T})-w_{\theta}(z_{1:T})\left(\partial_{\theta_{0}}\log p_{\theta_{0}}(z_{1:T})\right)^{2}\right),\\
& \left(E_1(\theta)\right)_{22}=w_{\theta}(z_{1:T})\left(\partial^{2}_{\theta_{1}}\log p_{\theta_{1}}(z_{1:T})-(1-w_{\theta}(z_{1:T}))\left(\partial_{\theta_{1}}\log p_{\theta_{1}}(z_{1:T})\right)^{2}\right).
\end{align*}
With this error decomposition, we have that $\calI(\theta)$ can be written as:
\begin{align}
\mathcal{I}(\theta)=\frac{1}{2}\begin{pmatrix} \mathcal{I}(\theta_{0}) & 0 \\0 & \mathcal{I}(\theta_{1})\end{pmatrix}+\frac{1}{2}\left(\E^{(0)}_\theta\left[E_{0}(\theta)\right]+\E_\theta^{(1)}\left[E_{1}(\theta)\right]\right).\label{eq:mixture_fisherdecomp}
\end{align}
We next bound:
\begin{align}
    \frac{1}{2}\opnorm{\E^{(0)}_\theta\left[E_{0}(\theta)\right]+\E^{(1)}_\theta\left[E_{1}(\theta)\right]}
&\leq \frac{1}{2}\left(\opnorm{ \E^{(0)}_\theta\left[E_{0}(\theta)\right]}+\opnorm{\E^{(1)}_\theta\left[E_{1}(\theta)\right]}\right)\nonumber\\
&\leq \frac{1}{2}\left( \E^{(0)}_\theta\left[\opnorm{ E_{0}(\theta)}\right]+\E^{(1)}_\theta\left[\opnorm{ E_{1}(\theta)}\right]\right)\nonumber\\
&\leq  \E^{(0)}_\theta\left[\sup_{1\leq i,j\leq 2}\left|\left(E_{0}(\theta)\right)_{ij}\right|\right]+\E^{(1)}_\theta\left[\sup_{1\leq i,j\leq 2}\left|\left(E_{1}(\theta)\right)_{ij}\right|\right],\label{eq:mixture_stability_inequality}
\end{align}
where in the penultimate inequality we applied
Jensen's inequality, and in the last step we apply the simple inequality
$\opnorm{A} \leq \norm{A}_F \leq \sqrt{nm} \cdot \max_{i \in [m], j \in [n]} \abs{A_{ij}}$ for $A \in \R^{m \times n}$.
We now recall from the two-state Markov chain example (\Cref{sec:twostatemc}) the following computation for $i=0,1$,
\begin{align*}
    \partial_{\theta_{i}} \log p_{\theta_{i}}(z_{1:T}) &= \sum_{t=1}^{T-1}\left(\frac{1}{\theta_{i}} \ind\{ z_{t+1} = z_{t} \} - \frac{1}{1-\theta_{i}} \ind\{ z_{t+1} \neq z_{t} \}\right), \\
    \partial^2_{\theta_{i}} \log p_{\theta_{i}}(z_{1:T}) &= - \sum_{t=1}^{T-1}\left( \frac{1}{\theta_{i}^2} \ind\{z_{t+1} = z_{t}\} + \frac{1}{(1-\theta_{i})^2} \ind\{z_{t+1} \neq z_{t}\} \right).
\end{align*}
It is clear that we have
\begin{align*}
\abs{\partial_{\theta_{i}} \log p_{\theta_{i}}(z_{1:T})} \leq (T-1)/\underline{\sigma}^2(\theta), \quad 
 -(T-1)/\underline{\sigma}^4(\theta) \leq \partial^2_{\theta_{i}} \log p_{\theta_{i}}(z_{1:T}) \leq 0, 
\end{align*}
Hence, it is straightforward to show the following almost surely bounds:
\begin{equation}
\begin{aligned}
    \abs{ \partial^2_{\theta_0} \log p_\theta(z_{1:T}) } &\leq w_{\theta}(z_{1:T}) (T-1)^2 / \underline{\sigma}^4(\theta), \\
    \abs{ \partial^2_{\theta_1} \log p_\theta(z_{1:T}) } &\leq (1-w_{\theta}(z_{1:T}) ) (T-1)^2 / \underline{\sigma}^4(\theta), \\
    \abs{ \partial_{\theta_1} \partial_{\theta_0} \log p_\theta(z_{1:T}) } &\leq w_{\theta}(z_{1:T})(1-w_{\theta}(z_{1:T})) (T-1)^2 / \underline{\sigma}^4(\theta) .
\end{aligned}\label{eq:markov_mixture_almost_sure_bounds}
\end{equation}
From these bounds we can conclude that:
\begin{align*}
    \sup_{1\leq i,j\leq 2}\left|\left(E_{0}(\theta)\right)_{ij}\right|\leq 2(1 - w_{\theta}(z_{1:T})) (T-1)^{2} / \underline{\sigma}^4(\theta), \quad
    \sup_{1\leq i,j\leq 2}\left|\left(E_{1}(\theta)\right)_{ij}\right|\leq 2A_{\theta}(z_{1:T})(T-1)^{2} / \underline{\sigma}^4(\theta).
\end{align*}
Next, we define
\begin{align*}
    \bar{\calI}_s(\theta) := \frac{1}{2}\begin{pmatrix} \bar{\calI}(\theta_{0}) & 0 \\0 & \bar{\calI}(\theta_{1})\end{pmatrix}, \quad E_{\bar{\calI}(\theta)} := \bar{\calI}(\theta) - \bar{\calI}_s(\theta).
\end{align*}
From \eqref{eq:mixture_stability_inequality}, we have the following bound:
\begin{align}
     \opnorm{E_{\bar{\calI}(\theta)}} &\leq 2 (T-1) / \underline{\sigma}^4(\theta)  \cdot \underbrace{\left(\E^{(0)}_\theta\left[1 - w_{\theta}(z_{1:T})\right]+\E^{(1)}_\theta\left[w_{\theta}(z_{1:T})\right]\right)}_{:= \zeta}. \label{eq:mixture_error_bound_zeta}
\end{align}
Since $E_{\bar{\calI}(\theta)}$ is symmetric, this implies:
\begin{align*}
    - 2(T-1)\zeta/\underline{\sigma}^4(\theta) \cdot I_2 \preccurlyeq E_{\bar{\calI}(\theta)} \preccurlyeq 2(T-1)\zeta/\underline{\sigma}^4(\theta) \cdot I_2.
\end{align*}
Therefore,
\begin{align*}
    \bar{\calI}(\theta) &= \bar{\calI}_s(\theta) + E_{\bar{\calI}(\theta)} \\
    &\succcurlyeq \bar{\calI}_s(\theta) - 2(T-1)\zeta/\underline{\sigma}^4(\theta) \cdot I_2 \\
    &= \bar{\calI}_s(\theta) - 2(T-1)\zeta/\underline{\sigma}^4(\theta)  \cdot \bar{\calI}_s^{1/2}(\theta) \bar{\calI}_s^{-1}(\theta)  \bar{\calI}_s^{1/2}(\theta)  \\
    &\succcurlyeq \bar{\calI}_s(\theta) - \frac{2(T-1) \zeta \lambda_{\max}(\bar{\calI}_s^{-1}(\theta) )}{ \underline{\sigma}^4(\theta) } \cdot \bar{\calI}_s(\theta) \\
    &= \left( 1 - \frac{4(T-1) \zeta \bar{\sigma}^2(\theta) }{ \underline{\sigma}^4(\theta)  } \right) \bar{\calI}_s(\theta).
\end{align*}
A nearly identical argument shows that
$\bar{\calI}(\theta) \preccurlyeq \left( 1 + \frac{4(T-1) \zeta \bar{\sigma}^2(\theta) }{ \underline{\sigma}^4(\theta)  } \right) \bar{\calI}_s(\theta)$.
Hence, if we choose
$\zeta \leq \frac{\underline{\sigma}^4(\theta)}{8\bar{\sigma}^2(\theta)} \cdot \frac{1}{T-1}$, we will have the desired
inequality:
\begin{align*}
    \frac{1}{2} \bar{\calI}_s(\theta) \preccurlyeq \bar{\calI}(\theta) \preccurlyeq \frac{3}{2} \bar{\calI}_s(\theta).
\end{align*}
To conclude, we see that for any $\theta \in \Theta'$ and $i \in \{0, 1\}$,
\begin{align*}
    \abs{ \sigma_i^2(\theta) - \sigma_i^2(\theta_\star) } \leq \abs{\theta_i - \theta_{\star,i}} \leq \norm{\theta - \theta_\star} \leq \sigma_i^2(\theta_\star)/2 \quad&\Longrightarrow\quad   \sigma_i^2(\theta_\star)/2 \leq \sigma_i^2(\theta) \leq 3 \sigma_i^2(\theta)/2.
\end{align*}
The RHS above implies that:
\begin{align}
    \underline{\sigma}^4(\theta) \geq \underline{\sigma}^4(\theta_\star)/4, \quad \bar{\sigma}^2(\theta) \leq 3\bar{\sigma}^2(\theta_\star).\label{eq:mixture_unifvar}
\end{align}
Consequently, we have for $\theta \in \Theta'$:
\begin{align*}
    \zeta \leq \frac{\underline{\sigma}^4(\theta_\star)}{48 \bar{\sigma}^2(\theta_\star)} \cdot \frac{1}{T-1} \Longrightarrow \zeta \leq \frac{\underline{\sigma}^4(\theta)}{8\bar{\sigma}^2(\theta)} \cdot \frac{1}{T-1}
\end{align*}
We now apply \Cref{corollary:lengthrequirement}
with $\eta=\frac{\underline{\sigma}^4(\theta_\star)}{48\bar{\sigma}^2(\theta_\star)}$ and $k=1$, from which the result follows.

\end{proof}

\begin{myprop}
\label{prop:mixture_identifiability}
Suppose that $T \geq 3$.
The family of densities defined by \eqref{eq:markov_two_state_mixture} over $\Theta_{+} := \{\theta \in \Theta \mid \theta_0 \geq \theta_1 \}$
is $\left(\mathrm{Gap}^2(\theta_\star)/44, 13/\mathrm{Gap}(\theta_\star)\right)$-Hellinger identifiable (cf.~\Cref{def:identifiability}) 
around $\theta_{\star}\in\Theta_{+}$. 
\end{myprop}
\begin{proof}
We first generate a table of transition probabilities for any $\theta \in \Theta$
over the first three elements $(z_1,z_2,z_3)$, leading to eight possibilities:
\begin{table}[h!]
\centering
\begin{tabular}{|c|c|}
\hline
$(z_1, z_2, z_3)$ & $p_\theta(z_1, z_2, z_3)$ \\
\hline
$(1,1,1)$, $(2,2,2)$ & $\frac{1}{4}\left( \theta_0^2 + \theta_1^2 \right)$ \\
$(1,1,2)$, $(1,2,2)$, $(2,2,1)$, $(2,1,1)$ & $\frac{1}{4}\left( \theta_0(1-\theta_0) + \theta_1(1-\theta_1) \right)$ \\
$(1,2,1)$, $(2,1,2)$ & $\frac{1}{4}\left( (1-\theta_0)^2 + (1-\theta_1)^2 \right)$ \\
\hline
\end{tabular}
\caption{A enumeration of the probabilities 
$p_\theta(z_{1:3})$
over the first three elements $(z_1, z_2, z_3)$ in $z_{1:T}$.}
\label{table:transition_probabilities}
\end{table}

We now use \Cref{table:transition_probabilities} to 
establish a lower bound for
the TV distance $\TV{p_{\theta}(z_1, z_2, z_{3})}{p_{\theta_{\star}}(z_1, z_2, z_{3})}$
in terms of the parameters $\theta$ and $\theta_\star$:
\begin{align*}
    \TV{p_{\theta}(z_{1:3})}{p_{\theta_{\star}}(z_{1:3})}
    &= \frac{1}{2} \sum_{(z_{1:3}) \in \{1,2\}^3} \abs{ p_\theta(z_{1:3}) - p_{\theta_\star}(z_{1:3}) } \\
    &= \frac{1}{4} \abs{ \theta_0^2 + \theta_1^2 - \theta_{\star,0}^2 - \theta_{\star,1}^2 } + \frac{1}{4} \abs{ (1-\theta_0)^2 + (1-\theta_1)^2 - (1-\theta_{\star,0})^2 - (1-\theta_{\star,1})^2 } \\
    &\qquad + \frac{1}{2} \abs{ \theta_0(1-\theta_0) + \theta_1(1-\theta_1) - \theta_{\star,0}(1-\theta_{\star,0}) - \theta_{\star,1}(1-\theta_{\star,1})  } \\
    &= \frac{1}{4} \abs{ \norm{\theta}^2 - \norm{\theta_\star}^2 } + \frac{1}{4} \abs{ \norm{\ind - \theta}^2 - \norm{\ind - \theta_\star}^2 } + \frac{1}{2} \abs{ \ip{\theta}{\ind - \theta} - \ip{\theta_\star}{\ind - \theta_\star} } \\
    &\geq \frac{1}{4} \abs{ \norm{\theta}^2 - \norm{\theta_\star}^2 } + \frac{1}{4} \abs{ \norm{\ind - \theta}^2 - \norm{\ind - \theta_\star}^2 } \\
    &= \frac{1}{4} \abs{ \norm{\theta}^2 - \norm{\theta_\star}^2 } + \frac{1}{4} \abs{  \norm{\theta}^2 - \norm{\theta_\star}^2 + 2(\norm{\theta_\star}_1 - \norm{\theta}_1)  } \\
    &\geq \max\left\{ \frac{1}{4} \abs{ \norm{\theta}^2 - \norm{\theta_\star}^2 }, \frac{1}{2} \abs{ \norm{\theta}_1 - \norm{\theta_\star}_1 } \right\}.
\end{align*}
Hence by the data processing inequality, we have
\begin{align*}
    \Hg{p_{\theta}}{p_{\theta_\star}} \geq \max\left\{ \frac{1}{4} \abs{ \norm{\theta}^2 - \norm{\theta_\star}^2 }, \frac{1}{2} \abs{ \norm{\theta}_1 - \norm{\theta_\star}_1 } \right\}.
\end{align*}
Suppose we can control $\Hg{p_{\theta}}{p_{\theta_{\star}}} \leq \e$, this would imply that:
$$
\left|\left\Vert \theta\right\Vert^{2}-\left\Vert \theta_{\star}\right\Vert^{2}\right|\leq 4\e, \quad \left|\left\Vert \theta\right\Vert_{1}-\left\Vert \theta_{\star}\right\Vert_{1}\right|\leq 2\e,
$$
i.e., denoting $a:=\theta_0$, and $b:=\theta_1$, we have:
\begin{subequations}\label{eq:S_and_Q}
\begin{align}
S &:= a^{2} + b^{2} \in 
    \left[ \left\Vert \theta_{\star}\right\Vert^{2} - 4\e, \,
           \left\Vert \theta_{\star}\right\Vert^{2} + 4\e \right], \label{eq:S_equation}\\
Q &:= a + b \in 
    \left[ \left\Vert \theta_{\star}\right\Vert_{1} - 2\e, \,
           \left\Vert \theta_{\star}\right\Vert_{1} + 2\e \right]. \label{eq:Q_equation}
\end{align}
\end{subequations}
Our next step will be to solve for $(a, b)$ in terms of $(S, Q)$.
To do this, we first will argue that the
discriminant $2S-Q^2 > 0$.
We define the following quantities:
\begin{align*}
    \Delta_S := S - \norm{\theta_\star}^2, \quad \Delta_Q := Q - \norm{\theta_\star}_1, \quad \rho_\star := \mathrm{Gap}(\theta_\star).
\end{align*}
With this notation,
\begin{align*}
    2 S - Q^2 &= 2(\norm{\theta_\star}^2 + \Delta_S) - (\norm{\theta_\star}_1^2 + \Delta_Q)^2 \\
    &= 2 \norm{\theta_\star}^2 - \norm{\theta_\star}^2_1 + 2\Delta_S - 2 \Delta_Q \norm{\theta_\star}_1 - \Delta_Q^2 \\
    &= \rho_\star^2 + 2\Delta_S - 2 \Delta_Q \norm{\theta_\star}_1 - \Delta_Q^2 =: \rho_\star^2 + \Delta.
\end{align*}
Now observe that since $\e \leq \sqrt{2}$,
we have $\abs{\Delta} \leq 8\e + 8\e + 4\e^2 \leq 22\e$,
and hence:
\begin{align*}
    \e \leq \rho_\star^2/44 \Longrightarrow 2S-Q^2 \in [ \rho_\star^2/2, 3 \rho_\star^2/2].
\end{align*}
This shows that $2S-Q^2 > 0$, and therefore the following is the unique solution with $a > b$ to \eqref{eq:S_and_Q}:
$$
a=\frac{Q+\sqrt{2S-Q^{2}}}{2}, \quad b=\frac{Q-\sqrt{2S-Q^{2}}}{2}.
$$
We next consider how $\sqrt{2S-Q^2}$ scales as a function of the perturbation $\Delta$.
Let us define $f(\Delta) := \sqrt{\rho_\star^2 + \Delta}$,
which has derivative $f'(\Delta) = \frac{1}{2\sqrt{\rho_\star^2 + \Delta}}$.
By concavity of square-root, we have:
\begin{align*}
    f(\Delta) \leq f(0) + f'(0) \Delta \leq \rho_\star + \frac{1}{2\rho_\star} \abs{\Delta}.
\end{align*}
On the other hand, by the mean value theorem for some $c$ with $\abs{c} \leq \abs{\Delta}$:
\begin{align*}
    f(\Delta) = f(0) + f'(c) \Delta \geq \rho_\star - \frac{1}{\sqrt{2}\rho_\star} \abs{\Delta}.
\end{align*}
We can make this interval symmetric:
\begin{align*}
    f(\Delta) \in \left[ \rho_\star - \frac{1}{\sqrt{2}\rho_\star} \abs{\Delta}, \rho_\star + \frac{1}{\sqrt{2}\rho_\star} \abs{\Delta} \right].
\end{align*}
Now we can conclude our analysis.
We first observe that:
\begin{align*}
    \norm{\theta_\star}_1 + \rho_\star = 2 \theta_{\star,0}, \quad \norm{\theta_\star}_1 - \rho_\star = 2 \theta_{\star,1}.
\end{align*}
Next, we start with $a$. We have:
\begin{align*}
    a &= \frac{Q + \sqrt{2S-Q^2}}{2} = \frac{\norm{\theta_\star}_1 + \Delta_Q + f(\Delta) }{2} \leq \frac{\norm{\theta_\star}_1 + \rho_\star + \Delta_Q + \abs{\Delta}/(\sqrt{2}\rho_\star)}{2} \\
    &\leq \frac{ 2 \theta_{\star,0} + 2\e + 22\e/(\sqrt{2} \rho_\star)}{2} \leq \theta_{\star,0} + (1+11/\sqrt{2}) \e/\rho_\star.
\end{align*}
A similar argument shows that $a \geq \theta_{\star,0} - (1+5\sqrt{2})\e/\rho_\star$, and hence
we have:
\begin{align*}
    \abs{a - \theta_{\star,0}} \leq (1+11/\sqrt{2})\e/\rho_\star.
\end{align*}
Furthermore, a similar argument also shows that:
\begin{align*}
    \abs{b - \theta_{\star,1}} \leq (1+11/\sqrt{2})\e/\rho_\star.
\end{align*}
Hence, we have $\norm{\theta-\theta_\star} \leq \sqrt{2} \norm{\theta-\theta_\star}_\infty \leq 13 \e / \rho_\star$.
Therefore, we have shown that for all $\theta \in \Theta'$:
\begin{align*}
     \norm{ \theta - \theta_\star } \leq \frac{13}{\rho_\star} \Hg{p_\theta}{p_{\theta_\star}} .
\end{align*}
\end{proof}
\begin{rmk}[On identifiability up to permutation]
We note that picking a uniform distribution for $B$ illustrates one key issue in mixture models, where we can only identify parameters up to a permutation. Therefore we have to assume some additional distinguishability between parameters (e.g., the restricted subset $\Theta_{+}$) to guarantee unique identifiability. 
We discuss this issue in more detail in \Cref{subsec:mixture_general}.
\end{rmk}

\subsubsection{Proof of \Cref{thm:mixturemc}}\label{sec:mixture_proof}
For compactness of notation, in the proof we will use
$$
w_{\theta}^{(0)}(z_{1:T}):=w_{\theta}(z_{1:T}),\quad w_{\theta}^{(1)}(z_{1:T}):=1-w_{\theta}^{(0)}(z_{1:T}).
$$
\paragraph{Covering Number Bound.}
We first compute an upper bound $\calI_{\max}$ such that
$\calI(\theta) \preccurlyeq \calI_{\max}$ for all $\theta \in \Theta$. Applying \Cref{prop:diagmajorization} and recall the related computation in \Cref{corollary:eigenvaluebound}, specifically \eqref{eq:markov_mixture_almost_sure_bounds}, we have:
\begin{align*}
\calI(\theta)&\preccurlyeq \begin{pmatrix} \E_{\theta}\left[\left|\partial^{2}_{\theta_{0}}\log p_{\theta}(z_{1:T})\right|+\left|\partial_{\theta_{1}}\partial_{\theta_{0}}\log p_{\theta}(z_{1:T})\right|\right] & 0\\
0 & \E_{\theta}\left[\left|\partial^{2}_{\theta_{1}}\log p_{\theta}(z_{1:T})\right|+\left|\partial_{\theta_{1}}\partial_{\theta_{0}}\log p_{\theta}(z_{1:T})\right|\right]\end{pmatrix}\\
&\preccurlyeq \frac{5}{4}\frac{(T-1)^{2}}{\underline{\sigma}^{4}(\theta)}I_2.
\end{align*}
Hence, we have for all $\theta \in \Theta$:
\begin{align*}
    \calI(\theta) \preccurlyeq \calI_{\max} := \frac{5(T-1)^2}{4 \sigma_{\min}^4} I_2. %
\end{align*}
Consequently, for any $\theta,\theta' \in \Theta$,
$$
\mathrm{d}_{\calI_{\max}}\left(p_{\theta},p_{\theta'}\right) = \left\Vert \theta-\theta'\right\Vert_{\mathcal{I}_{\max}}=\sqrt{\frac{5}{4}}\frac{T-1}{\sigma_{\min}^{2}}\left\Vert \theta-\theta'\right\Vert.
$$
Therefore we can upper bound the metric entropy
$$
\log\mathcal{N}_{\calI_{\max}}\left(\mathcal{P},\e\right)\leq \log \mathcal{N}_{\Vert\cdot\Vert}\left([\mu,1-\mu]^2,\sqrt{\frac{4}{5}}\frac{\sigma_{\min}^{2}\e}{T-1}\right)\leq 2 \log \left(\sqrt{\frac{5}{4}}\frac{T-1}{\sigma_{\min}^{2} \e} \right)+\log 2.
$$
We now first apply \Cref{thm:hellinger_bound_MLE} (a) with resolution $\e= \delta/(2\sqrt{2m})$,
which yields an event $\calE_{1}$ with probability at least $1-\delta$, where on $\calE_{1}$:
\begin{align}
     \HgSq{\hat{\theta}^{\e}_{m,T}}{\theta_\star} \leq   \frac{ 8 \log \left(\sqrt{10}\frac{\sqrt{m}(T-1)}{\sigma_{\min}^{2} \delta} \right)+8\log 2+4\log(1/\delta)+\delta^{2}/4}{m}\leq \frac{21 \log \left(\frac{4mT}{\sigma_{\min}^{2} \delta} \right)}{m}, \label{eq:mixture_hellinger_sq_inequality}
\end{align}
where the last inequality follows from by assumption, $m\geq 4$, and the observation
\begin{align*}
\log \left(\sqrt{10}\frac{\sqrt{m}(T-1)}{\sigma_{\min}^{2} \delta} \right)\geq \log \left(8\sqrt{10}\frac{1}{\delta}\right)\geq \max\left\{\log 2,\log(1/\delta),\delta^{2}\right\},\quad\forall\;\delta\in (0,1).
\end{align*}
For the remainder of the proof, we will assume we are on 
the event $\calE_{1}$.
Invoking the Hellinger identifiability (\Cref{prop:mixture_identifiability}) and \Cref{prop:log_dominance}:
\begin{align*}
m\geq \frac{42}{\gamma_{1}^{2}}\log \left(\frac{168T}{\gamma_{1}^{2}\sigma_{\min}^{2}\delta } \right)&\implies m\geq \frac{21}{\gamma_{1}^{2}}\log \left(\frac{4mT}{\sigma_{\min}^{2}\delta } \right)\\
&\implies \Hg{\hat{\theta}^{\e}_{m,T}}{\theta_{\star}}\leq \gamma_{1}\\
&\implies \norm{\hat{\theta}^{\e}_{m,T}-\theta_{\star}}\leq \gamma_{2}\Hg{\hat{\theta}^{\e}_{m,T}}{\theta_{\star}}\leq 5\gamma_{2}\sqrt{\frac{  \log \left(\frac{4mT}{\sigma_{\min}^{2} \delta} \right)}{m}}.
\end{align*}
We can now additionally impose (cf.~\Cref{prop:log_dominance})
$$
m\geq \frac{50\gamma_{2}^{2}}{c^{2}}\log\left(\frac{200\gamma_{2}^{2}T}{c^{2}\sigma_{\min}^{2}\delta}\right)\implies m\geq \frac{25\gamma_{2}^{2}\log\left(\frac{4mT}{\sigma_{\min}^{2}\delta}\right)}{c^{2}}, %
$$
where $c$ is some constant radius. 
This results in the following key identity, which will be used in the sequel:
\begin{align}
m\geq \max\left\{\frac{42}{\gamma_{1}^{2}}\log \left(\frac{168T}{\gamma_{1}^{2}\sigma_{\min}^{2}\delta } \right),\frac{50\gamma_{2}^{2}}{c^{2}}\log\left(\frac{200\gamma_{2}^{2}T}{c^{2}\sigma_{\min}^{2}\delta}\right)\right\}\implies \norm{\hat{\theta}^{\e}_{m,T}-\theta}\leq c.\label{eq:mixture_prelim}
\end{align}

\paragraph{Estimate $B_1$ and $B_2$.}
We now estimate $B_1(\hat{\theta}^{\e}_{m,T}, \theta_{\star})$ and
$B_2(\hat{\theta}^{\e}_{m,T}, \theta_{\star})$. 
Our first step is to guarantee
that $\hat{\theta}^{\e}_{m,T} \in \Theta'$ (cf. \eqref{eq:mixture_Theta_prime}), i.e.,
$$
\norm{\hat{\theta}^{\e}_{m,T}-\theta_{\star}}\leq \min\left\{\frac{\rho_{\star}}{2\sqrt{2}}, \frac{\underline{\sigma}^2(\theta_\star)}{2}\right\}.
$$
In view of \eqref{eq:mixture_prelim}, it suffices to require 
\begin{align}
m\geq \max\left\{\frac{42}{\gamma_{1}^{2}}\log \left(\frac{168T}{\gamma_{1}^{2}\sigma_{\min}^{2}\delta } \right),\frac{50\gamma_{2}^{2}}{\min\left\{\frac{\rho_{\star}}{2\sqrt{2}}, \frac{\underline{\sigma}^2(\theta_\star)}{2}\right\}^{2}}\log\left(\frac{200\gamma_{2}^{2}T}{\min\left\{\frac{\rho_{\star}}{2\sqrt{2}}, \frac{\underline{\sigma}^2(\theta_\star)}{2}\right\}^{2}\sigma_{\min}^{2}\delta}\right)\right\}.
\end{align} 
We abbreviate the above to be written as:
\begin{align}
    m \gtrsim \max\left\{ \frac{1}{\rho_\star^4}, \frac{1}{\rho_\star^2 \underline{\sigma}^4(\theta_\star)} \right\} \log\left( \max\left\{ \frac{1}{\rho_\star^4}, \frac{1}{\rho_\star^2 \underline{\sigma}^4(\theta_\star)} \right\}  \frac{T}{\sigma^2_{\min} \delta} \right). \label{eq:mixtureradius}
\end{align}
In addition to \eqref{eq:mixtureradius}, further requiring
\begin{align}
T \gtrsim \max\left\{ \frac{\bar{\sigma}^2(\theta_\star)}{\rho_{\star}^4}, \frac{1}{\rho_{\star}^2}  \right\} \log^2(1/\mu) \log\left( \frac{ \bar{\sigma}^2(\theta_\star) }{\underline{\sigma}^4(\theta_\star)} \cdot \frac{\log(1/\mu)}{\rho_{\star} } \right),\label{eq:mixture_Treq}
\end{align}
implies by \Cref{corollary:eigenvaluebound} that
for any $\theta\in\text{conv}\left\{\hat{\theta}^{\e}_{m,T},\theta_{\star}\right\}$:
\begin{align}
    \frac{1}{4} \calI_{\diag}(\theta) \preccurlyeq \calI(\theta) \preccurlyeq \frac{3}{4} \calI_{\diag}(\theta), \quad \calI_{\diag}(\theta) := \diag\left\{ \calI(\theta_0), \calI(\theta_1) \right\} = \diag\left\{ \frac{T-1}{\sigma_0^2(\theta)}, \frac{T-1}{\sigma_1^2(\theta)} \right\}. \label{eq:mixture_FI_upper_lower_diag_bounds}
\end{align}
And therefore applying \Cref{prop:loewnerprecond}, we have for any $\theta\in\text{conv}\left\{\hat{\theta}^{\e}_{m,T},\theta_{\star}\right\}$:

\begin{align*}
 &\quad\sup_{\Vert v\Vert=1}\left\Vert \left \langle v \calI(\theta)^{-1/2}\nabla_\theta \log p_\theta(z_{1:T})\right\rangle\right\Vert_{\calL^{4}(p_{\theta})}\\
 &\leq \sup_{\Vert v\Vert=1}\left\Vert\left\langle v \left(\tfrac{1}{4}\right)^{-1/2}\diag\left\{\calI(\theta_{0})^{-1/2},\calI(\theta_{1})^{-1/2}\right\}\nabla_\theta \log p_\theta(z_{1:T})\right\rangle \right\Vert_{\calL^{4}(p_{\theta})}\\
 &=2\sup_{\Vert v\Vert=1}\left\Vert \sum_{i=0,1}\left(v_{i}\sqrt{\tfrac{\theta_{i}(1-\theta_{i})}{T-1}}w_{\theta}^{(i)}(z_{1:T})\partial_{\theta_{i}}\log p_{\theta_{i}}(z_{1:T})\right)\right\Vert_{\calL^{4}(p_{\theta})}\\
 &\lesssim \underbrace{\left\Vert \sqrt{\tfrac{\theta_{0}(1-\theta_{0})}{T-1}}w_{\theta}^{(0)}(z_{1:T})\partial_{\theta_{0}}\log p_{\theta_{0}}(z_{1:T})\right\Vert_{\calL^{4}(p_{\theta})}}_{:=(I)}+\underbrace{\left\Vert \sqrt{\tfrac{\theta_{1}(1-\theta_{1})}{T-1}}w_{\theta}^{(1)}(z_{1:T})\partial_{\theta_{1}}\log p_{\theta_{1}}(z_{1:T})\right\Vert_{\calL^{4}(p_{\theta})}}_{:=(II)}.
\end{align*}
Notice each term \emph{without} the density ratio $w_{\theta}^{(i)}(z_{1:T})$ resembles that what we have seen in \Cref{sec:twostatemc}. Hence we upper bound $(I)$ as:
\begin{equation}\label{eq:mixture_B1_bound_I}
\begin{aligned}
(I)&=\left(\frac{\theta_{0}^{2}(1-\theta_{0})^{2}}{(T-1)^{2}}\E_{\theta}\left[\left(w_{\theta}^{(0)}(z_{1:T})\partial_{\theta_{0}}\log p_{\theta_{0}}(z_{1:T})\right)^{4}\right]\right)^{1/4}\\
&\leq\left(\frac{\theta_{0}^{2}(1-\theta_{0})^{2}}{(T-1)^{2}}\E^{(0)}_{\theta}\left[\left(\partial_{\theta_{0}}\log p_{\theta_{0}}(z_{1:T})\right)^{4}\right]+\frac{\theta_{0}^{2}(1-\theta_{0})^{2}}{(T-1)^{2}}\E^{(1)}_{\theta}\left[\left(w_{\theta}^{(0)}(z_{1:T})\partial_{\theta_{0}}\log p_{\theta_{0}}(z_{1:T})\right)^{4}\right]\right)^{1/4}\\
&\lesssim \left(\frac{1}{T\underline{\sigma}^{2}(\theta)}+O(1)+\frac{(T-1)^{2}}{\underline{\sigma}^{4}(\theta)}\E^{(1)}_{\theta}\left[\left|w_{\theta}^{(0)}(z_{1:T})\right|\right]\right)^{1/4}\\
&\lesssim \left(\frac{1}{T\underline{\sigma}^{2}(\theta)}+O(1)\right)^{1/4}\lesssim \left(\frac{1}{T\underline{\sigma}^{2}(\theta_{\star})}+O(1)\right)^{1/4},
\end{aligned}
\end{equation}
where the third to last step follows from \eqref{eq:two_state_B1_B2_prelim}, and the last two steps we first set a large enough $T$ such that the last term is also $O(1)$, for which our assumption \eqref{eq:mixture_Treq} would suffice together with \eqref{eq:mixture_unifvar}. A similar argument shows that
$$
(II)\lesssim \left(\frac{1}{T\underline{\sigma}^{2}(\theta_{\star})}+O(1)\right)^{1/4},
$$
and therefore we can conclude
$$
B_{1}^{4}\lesssim \max\left\{\frac{1}{T\underline{\sigma}^{2}(\theta_{\star})},1\right\}.
$$
Now for $B_2$, we proceed as we did for $B_1$.
For any $\theta\in\text{conv}\left\{\hat{\theta}^{\e}_{m,T},\theta_{\star}\right\}$:

\begin{align*}
 &\quad\sup_{\Vert v\Vert=1}\left\Vert\left\langle v \calI(\theta)^{-1/2}\nabla^{2}_\theta \log p_\theta(z_{1:T})\calI(\theta)^{-1/2}v\right\rangle\right\Vert_{\calL^{2}(p_{\theta})}\\
 &\leq\left(\tfrac{1}{4}\right)^{-1}\left(\sup_{\Vert v\Vert=1}\left\Vert v_{0}^{2}\tfrac{\theta_{0}(1-\theta_{0})}{(T-1)}w_{\theta}^{(0)}(z_{1:T})w_{\theta}^{(1)}(z_{1:T})\left(\partial_{\theta_{0}}\log p_{\theta_{0}}(z_{1:T})\right)^{2}\right\Vert_{\calL^{2}(p_{\theta})}\right.\\
 &\quad +\sup_{\Vert v\Vert=1}\left\Vert v_{0}^{2}\tfrac{\theta_{0}(1-\theta_{0})}{(T-1)}w_{\theta}^{(0)}(z_{1:T})\partial^{2}_{\theta_{0}}\log p_{\theta_{0}}(z_{1:T})\right\Vert_{\calL^{2}(p_{\theta})}\\
 &\quad+ \sup_{\Vert v\Vert=1}\left\Vert v_{1}^{2}\tfrac{\theta_{1}(1-\theta_{1})}{(T-1)}w_{\theta}^{(0)}(z_{1:T})w_{\theta}^{(1)}(z_{1:T})\left(\partial_{\theta_{1}}\log p_{\theta_{1}}(z_{1:T})\right)^{2}\right\Vert_{\calL^{2}(p_{\theta})}\\
 &\quad +\sup_{\Vert v\Vert=1}\left\Vert v_{1}^{2}\tfrac{\theta_{1}(1-\theta_{1})}{(T-1)}w_{\theta}^{(1)}(z_{1:T})\partial^{2}_{\theta_{1}}\log p_{\theta_{1}}(z_{1:T})\right\Vert_{\calL^{2}(p_{\theta})}\\
 &\left.\quad+ \sup_{\Vert v\Vert=1}\left\Vert v_{0}v_{1}\left(\sum_{i=0,1}\tfrac{\theta_{i}(1-\theta_{i})}{(T-1)}\right)w_{\theta}^{(0)}(z_{1:T})w_{\theta}^{(1)}(z_{1:T})\partial_{\theta_{0}}\log p_{\theta_{0}}(z_{1:T})\partial_{\theta_{1}}\log p_{\theta_{1}}(z_{1:T})\right\Vert_{\calL^{2}(p_{\theta})}\right)\\
 &\lesssim \underbrace{\left\Vert w_{\theta}^{(0)}(z_{1:T})\left(\sqrt{\tfrac{\theta_{0}(1-\theta_{0})}{(T-1)}}\partial_{\theta_{0}}\log p_{\theta_{0}}(z_{1:T})\right)^{2}\right\Vert_{\calL^{2}(p_{\theta})}}_{:=(III)} +\underbrace{\left\Vert \tfrac{\theta_{0}(1-\theta_{0})}{(T-1)}w_{\theta}^{(0)}(z_{1:T})\partial^{2}_{\theta_{0}}\log p_{\theta_{0}}(z_{1:T})\right\Vert_{\calL^{2}(p_{\theta})}}_{:=(IV)}\\
 &\;\;+ \underbrace{\left\Vert w_{\theta}^{(1)}(z_{1:T})\left(\sqrt{\tfrac{\theta_{1}(1-\theta_{1})}{(T-1)}}\partial_{\theta_{1}}\log p_{\theta_{1}}(z_{1:T})\right)^{2}\right\Vert_{\calL^{2}(p_{\theta})}}_{:=(V)} +\underbrace{\left\Vert \tfrac{\theta_{1}(1-\theta_{1})}{(T-1)}w_{\theta}^{(1)}(z_{1:T})\partial^{2}_{\theta_{1}}\log p_{\theta_{1}}(z_{1:T})\right\Vert_{\calL^{2}(p_{\theta})}}_{:=(VI)}\\
 &\;\;+ \underbrace{\tfrac{\bar{\sigma}^{2}(\theta)}{T-1}\left\Vert w_{\theta}^{(0)}(z_{1:T})w_{\theta}^{(1)}(z_{1:T})\partial_{\theta_{0}}\log p_{\theta_{0}}(z_{1:T})\partial_{\theta_{1}}\log p_{\theta_{1}}(z_{1:T})\right\Vert_{\calL^{2}(p_{\theta})}}_{:=(VII)}.
\end{align*}
Now using similar arguments for the upper bound of $(I)$ in \eqref{eq:mixture_B1_bound_I}, we can upper bound $(III)$:
\begin{align*}
(III)&=\left(\frac{\theta_{0}^{2}(1-\theta_{0})^{2}}{(T-1)^{2}}\E_{\theta}\left[\left(\tsqrt{w_{\theta}^{(0)}(z_{1:T})}\partial_{\theta_{0}}\log p_{\theta_{0}}(z_{1:T})\right)^{4}\right]\right)^{1/2}\\
&\leq\left(\frac{\theta_{0}^{2}(1-\theta_{0})^{2}}{(T-1)^{2}}\E^{(0)}_{\theta}\left[\left(\partial_{\theta_{0}}\log p_{\theta_{0}}(z_{1:T})\right)^{4}\right]+\frac{\theta_{0}^{2}(1-\theta_{0})^{2}}{(T-1)^{2}}\E^{(1)}_{\theta}\left[\left(\tsqrt{w_{\theta}^{(0)}(z_{1:T})}\partial_{\theta_{0}}\log p_{\theta_{0}}(z_{1:T})\right)^{4}\right]\right)^{1/2}\\
&\lesssim \left(\frac{1}{T\underline{\sigma}^{2}(\theta)}+O(1)+\frac{(T-1)^{2}}{\underline{\sigma}^{4}(\theta)}\E^{(1)}_{\theta}\left[\left|w_{\theta}^{(0)}(z_{1:T})\right|\right]\right)^{1/2}\\
&\lesssim \left(\frac{1}{T\underline{\sigma}^{2}(\theta_{\star})}+O(1)\right)^{1/2},
\end{align*}
and $(IV)$:
\begin{align*}
(IV)&=\left(\frac{\theta_{0}^{2}(1-\theta_{0})^{2}}{(T-1)^{2}}\E_{\theta}\left[\left(w_{\theta}^{(0)}(z_{1:T})\partial_{\theta_{0}}^{2}\log p_{\theta_{0}}(z_{1:T})\right)^{2}\right]\right)^{1/2}\\
&\leq\left(\frac{\theta_{0}^{2}(1-\theta_{0})^{2}}{(T-1)^{2}}\E^{(0)}_{\theta}\left[\left(\partial_{\theta_{0}}^{2}\log p_{\theta_{0}}(z_{1:T})\right)^{2}\right]+\frac{\theta_{0}^{2}(1-\theta_{0})^{2}}{(T-1)^{2}}\E^{(1)}_{\theta}\left[\left(w_{\theta}^{(0)}(z_{1:T})\partial_{\theta_{0}}^{2}\log p_{\theta_{0}}(z_{1:T})\right)^{2}\right]\right)^{1/2}\\
&\lesssim \left(\frac{1}{T\underline{\sigma}^{2}(\theta)}+O(1)+\frac{1}{\underline{\sigma}^{4}(\theta)}\E^{(1)}_{\theta}\left[\left|w_{\theta}^{(0)}(z_{1:T})\right|\right]\right)^{1/2}\\
&\lesssim \left(\frac{1}{T\underline{\sigma}^{2}(\theta_{\star})}+O(1)\right)^{1/2}.
\end{align*}
By symmetry between $\theta_0$ and $\theta_1$, we have
$$
(V)\lesssim \left(\frac{1}{T\underline{\sigma}^{2}(\theta_{\star})}+O(1)\right)^{1/2}, \quad(VI)\lesssim \left(\frac{1}{T\underline{\sigma}^{2}(\theta_{\star})}+O(1)\right)^{1/2}.
$$
Finally, we can upperbound $(VII)$:
\begin{align*}
(VII)
&\leq\frac{\bar{\sigma}^{2}(\theta)(T-1)}{\underline{\sigma}^{4}(\theta)}\left(\E^{(0)}_{\theta}\left[\left(w_{\theta}^{(1)}(z_{1:T})\right)^{2}\right]+\E^{(1)}_{\theta}\left[\left(w_{\theta}^{(0)}(z_{1:T})\right)^{2}\right]\right)^{1/2}\\
&\lesssim \frac{\bar{\sigma}^{2}(\theta_{\star})(T-1)}{\underline{\sigma}^{4}(\theta_{\star})}\left(\E^{(0)}_{\theta}\left[\left|w_{\theta}^{(1)}(z_{1:T})\right|\right]+\E^{(1)}_{\theta}\left[\left|w_{\theta}^{(0)}(z_{1:T})\right|\right]\right)^{1/2}\\
&\lesssim O(1),
\end{align*}
where at the second step we again applied \eqref{eq:mixture_unifvar}. Therefore
$B_{2}^{2}\lesssim \max\left\{\frac{1}{T\underline{\sigma}^{2}(\theta_{\star})},1\right\}$, and hence:
\begin{align}
\max\{B_{1}^{4},B_{2}^{2}\}\lesssim \max\left\{\frac{1}{T\underline{\sigma}^{2}(\theta_{\star})},1\right\}. \label{eq:mixture_B1_B2_bound}
\end{align}

\paragraph{Parameter error bound.}
Our first step is to define a more refined 
$\calI_{\max}'$ for $\theta \in \Theta'$.
From \Cref{corollary:eigenvaluebound},
for any $\theta \in \Theta'$:
$$
\lambda_{\max}(\calI(\theta))\leq \frac{T}{\underline{\sigma}^{2}(\theta)}\leq \frac{4T}{\underline{\sigma}^{2}(\theta_{\star})}.
$$
Therefore, we have:
\begin{align*}
    \theta \in \Theta' \Longrightarrow \calI(\theta) \preccurlyeq \frac{4T}{\underline{\sigma}^2(\theta_\star)} I_2 =: \calI_{\max}'.
\end{align*}
Hence, following \eqref{eq:hellinger_error_via_identifiability}
using $\calI_{\max}'$ instead of $\calI_{\max}$, we obtain that:
\begin{align*}
    \Hg{\hat{\theta}^{\e}_{m,T}}{\theta_\star} \leq \gamma_1 \Longrightarrow \sup_{\theta \in \mathrm{conv}\{ \hat{\theta}^{\e}_{m,T}, \theta_\star \}} \HgSq{\theta}{\theta_\star} \leq \frac{\gamma_2^2 T}{\underline{\sigma}^2(\theta_\star)} \HgSq{\hat{\theta}^{\e}_{m,T}}{\theta_\star} \lesssim \frac{\gamma_2^2 T \log(mT/(\sigma^2_{\min} \delta))}{ \underline{\sigma}^2(\theta_\star) m},
\end{align*}
where recall that the last inequality is from \eqref{eq:mixture_hellinger_sq_inequality}.
Therefore, combining the above inequality with
\eqref{eq:mixture_B1_B2_bound} and \Cref{prop:log_dominance},
condition \eqref{eq:hellinger_perturbation_radius}
holds if in addition to \eqref{eq:mixtureradius} and \eqref{eq:mixture_Treq} we also have:
\begin{align}
m\gtrsim \max\left\{\frac{1}{\rho_\star^2 \underline{\sigma}^{4}(\theta_{\star})},\frac{T}{\rho_\star^2 \underline{\sigma}^{2}(\theta_{\star})}\right\} \log \left(  \max\left\{\frac{1}{\rho_\star^2 \underline{\sigma}^{4}(\theta_{\star})},\frac{T}{\rho_\star^2 \underline{\sigma}^{2}(\theta_{\star})}\right\} \frac{T}{\sigma_{\min}^2 \delta} \right).
\label{eq:mixture_m_identifiability}
\end{align}
Applying \Cref{prop:hellinger_perturbation} (a), 
we obtain from \eqref{eq:hellinger_perturbation_avg_FI}:
$$
\norm{\hat{\theta}^{\e}_{m,T} - \theta_\star}^2_{{\calI_2}(\theta_\star, \hat{\theta}^{\e}_{m,T})} \lesssim m^{-1}\log \left(\frac{mT}{\sigma_{\min}^{2}\delta } \right).
$$
Furthermore, from \eqref{eq:mixture_FI_upper_lower_diag_bounds}, we have that $\calI_2(\theta_\star, \hat{\theta}^{\e}_{m,T}) \succcurlyeq c_0 T \cdot I_2$ for some $c_0 > 0$, 
and hence the following parameter error bound holds as well:
\begin{equation}
\norm{\hat{\theta}^{\e}_{m,T} - \theta_\star}^2 \lesssim \frac{1}{mT}\log \left(\frac{mT}{\sigma_{\min}^{2}\delta } \right).\label{eq:mixturemc_prelim2}
\end{equation}
\paragraph{Verify FI radius.}

For the variance-weighted CLT rate, we want to show the FI radius condition \eqref{eq:hellinger_perturbation_FI_radius}. 
By combining \Cref{prop:loewnerprecond2} and
\eqref{eq:mixture_FI_upper_lower_diag_bounds}, we have
for any $\theta \in \Theta'$,
\begin{align*}
    \opnorm{\calI(\theta_{\star})^{-1/2}\calI(\theta)\calI(\theta_{\star})^{-1/2}-I} &=\opnorm{\calI(\theta_{\star})^{-1/2}\left(\calI(\theta)-\calI(\theta_{\star})\right)\calI(\theta_{\star})^{-1/2}} \\
    &\leq 4\opnorm{\calI_{\diag}(\theta_{\star})^{-1/2}\left(\calI(\theta)-\calI(\theta_{\star})\right)\calI_{\diag}(\theta_{\star})^{-1/2}}.
\end{align*}
Hence condition \eqref{eq:hellinger_perturbation_FI_radius}
is implied by:
\begin{align}
    \opnorm{\calI_{\diag}(\theta_{\star})^{-1/2}\left(\calI(\theta)-\calI(\theta_{\star})\right)\calI_{\diag}(\theta_{\star})^{-1/2}} \leq 1/8. \label{eq:mixture_firadius1}
\end{align}
Now recall from the error decomposition in  \eqref{eq:mixture_fisherdecomp}, we have
\begin{align*}
\mathcal{I}(\theta) &=
 \frac{1}{2}\calI_{\diag}(\theta) +\frac{1}{2}\left(\E^{(0)}_\theta\left[E_{0}(\theta)\right]+\E^{(1)}_\theta\left[E_{1}(\theta)\right]\right),\\
\mathcal{I}(\theta_{\star}) &= \frac{1}{2} \calI_{\diag}(\theta_\star) +\frac{1}{2}\left(\E^{(0)}_{\theta_{\star}}\left[E_{0}(\theta_{\star})\right]+\E^{(1)}_{\theta_{\star}}\left[E_{1}(\theta_{\star})\right]\right).
\end{align*}
Hence denoting 
\begin{align*}
E(\theta,\theta_{\star}) &:= \left( \calI(\theta) - \calI(\theta_{\star}) \right) - \left( \frac{1}{2}\calI_{\diag}(\theta) - \frac{1}{2}\calI_{\diag}(\theta_\star) \right) \\
&=\frac{1}{2}\left(\E^{(0)}_\theta\left[E_{0}(\theta)\right]+\E^{(1)}_\theta\left[E_{1}(\theta)\right]\right) -\frac{1}{2}\left(\E^{(0)}_{\theta_{\star}}\left[E_{0}(\theta_{\star})\right]+\E^{(1)}_{\theta_{\star}}\left[E_{1}(\theta_{\star})\right]\right),
\end{align*}
we have \eqref{eq:mixture_firadius1} is equivalent to
\begin{align*}
    \bigopnorm{ \calI_{\diag}(\theta_\star)^{-1/2}\left(  \frac{1}{2} \calI_{\diag}(\theta) - \frac{1}{2} \calI_{\diag}(\theta_\star) + E(\theta, \theta_\star) \right) \calI_{\diag}(\theta_\star)^{-1/2} } \leq 1/8.
\end{align*}
By triangle inequality, a further sufficient condition is:
\begin{align}
&\left\Vert\diag\left\{\frac{\calI(\theta_{0})}{\calI(\theta_{\star,0})}-1,\frac{\calI(\theta_{1})}{\calI(\theta_{\star,1})}-1\right\}\right\Vert_{\mathsf{op}}=\max\left\{\left|\frac{\calI(\theta_{0})}{\calI(\theta_{\star,0})}-1\right|,\left|\frac{\calI(\theta_{1})}{\calI(\theta_{\star,1})}-1\right|\right\}\leq \frac{1}{8},\label{eq:mixture_firadius2}\\
&\bigopnorm{ \calI_{\diag}^{-1/2}(\theta_\star) E(\theta, \theta_\star) \calI_{\diag}^{-1/2}(\theta_\star) } \leq \frac{1}{16}.\label{eq:mixture_firadius3}
\end{align}
From \Cref{sec:twostatemc}, specifically, \eqref{eq:two_state_FI_metric}, we have that \eqref{eq:mixture_firadius2} can be satisfied by requiring
$$
\frac{\sqrt{2}}{\underline{\sigma}^{2}(\theta_{\star})}\norm{\theta-\theta_{\star}}\leq\frac{2}{\underline{\sigma}^{2}(\theta_{\star})}\max\left\{\left|\theta_{0}-\theta_{0,\star}\right|,\left|\theta_{1}-\theta_{1,\star}\right|\right\}\leq \frac{1}{8}.
$$
So it suffices to require
$$
\frac{\norm{\hat{\theta}^{\e}_{m,T}-\theta_{\star}}^{2}}{\underline{\sigma}^{4}(\theta_{\star})}\leq \frac{1}{128},
$$
and in view of \eqref{eq:mixturemc_prelim2} and \Cref{prop:log_dominance}, this holds if:
\begin{align*}
mT\gtrsim \frac{1}{\underline{\sigma}^{4}(\theta_{\star})}\log\left(\frac{1}{\underline{\sigma}^{4}(\theta_{\star})\sigma_{\min}^{2}\delta}\right).%
\end{align*}
This condition is however already implied by \eqref{eq:mixture_m_identifiability} (up to adjusting constant factors).
We now address condition \eqref{eq:mixture_firadius3}.
Recalling $\calI_{\diag}(\theta_\star) \succcurlyeq (T-1) \cdot I_2$,
we have:
\begin{align*}
    &\bigopnorm{ \calI_{\diag}^{-1/2}(\theta_\star) E(\theta, \theta_\star) \calI_{\diag}^{-1/2}(\theta_\star) } \stackrel{(a)}{\leq} \frac{1}{T-1} \opnorm{E(\theta, \theta_\star)} \\
 &\stackrel{(b)}{\lesssim}  \frac{T-1}{\underline{\sigma}^4(\theta)}  \left(\E^{(0)}_\theta\left[w_{\theta}^{(1)}(z_{1:T})\right]+\E^{(1)}_\theta\left[w_{\theta}^{(0)}(z_{1:T})\right]\right)+\frac{T-1}{\underline{\sigma}^4(\theta_{\star})}  \left(\E^{(0)}_{\theta_{\star}}\left[ w_{\theta_{\star}}^{(1)}(z_{1:T})\right]+\E^{(1)}_{\theta_{\star}}\left[w_{\theta_{\star}}^{(0)}(z_{1:T})\right]\right)\\
&\stackrel{(c)}{\lesssim}  \frac{T-1}{\underline{\sigma}^4(\theta_{\star})}  \left(\E^{(0)}_\theta\left[w_{\theta}^{(1)}(z_{1:T})\right]+\E^{(1)}_\theta\left[w_{\theta}^{(0)}(z_{1:T})\right]+\E^{(0)}_{\theta_{\star}}\left[w_{\theta_{\star}}^{(1)}(z_{1:T})\right]+\E^{(1)}_{\theta_{\star}}\left[w_{\theta_{\star}}^{(0)}(z_{1:T})\right]\right) \stackrel{(d)}{\lesssim} O(1),
\end{align*}
where (a) uses \Cref{prop:loewnerprecond2},
(b) uses \eqref{eq:mixture_error_bound_zeta}
from the proof of \Cref{corollary:eigenvaluebound}, and
(c) uses \eqref{eq:mixture_unifvar}, and
(d) uses the requirement on $T$ from \eqref{eq:mixture_Treq} (possibly adjusting constant factors as necessary), and follows the arguments bounding $\zeta$ in 
\eqref{eq:mixture_error_bound_zeta} from \Cref{corollary:eigenvaluebound}.

\paragraph{Final result.}
We now have the necessary requirements to invoke
\Cref{prop:hellinger_perturbation} (b) to conclude:
$$
\norm{\hat{\theta}^\e_{m,T} - \theta_\star}_{\calI(\theta_{\star})}^2 \lesssim \frac{1}{m}\log \left(\frac{mT}{\sigma_{\min}^{2}\delta } \right).
$$
In particular, combined with \Cref{corollary:eigenvaluebound}, this
also implies
$$
\norm{\hat{\theta}^\e_{m,T} - \theta_\star}^2 \lesssim \frac{\bar{\sigma}^{2}(\theta_{\star})}{mT}\log \left(\frac{mT}{\sigma_{\min}^{2}\delta } \right).
$$
We conclude by summarizing the requirements on $m, T$. In total, 
we require conditions \eqref{eq:mixtureradius}, \eqref{eq:mixture_Treq}, and \eqref{eq:mixture_m_identifiability} to hold.
These are readily simplified into
assumptions (b) and (c) in the theorem statement,
from which the result follows.

\subsubsection{Extensions of Proof Techniques to More General Mixture Problems}
\label{subsec:mixture_general}

We now discuss the extent to which
the proof strategy for \Cref{sec:case_studies:nonmixing} extends
to general mixture distributions.
We consider
a density class $\calP:=\left\{p_{\theta} \mid \theta\in\Theta^{k}\right\}$ over $k$ parameters $\{\theta_{1},\ldots,\theta_{k}\}\subset\Theta$ where for all $i\neq j$, $\theta_{i}\neq \theta_{j}$, and a multinomial distribution on the index set $[k$] parameterized by $f:=(f_{1},\ldots,f_{k-1}) \in \Delta^{k-1}_\mu$, where $f_{k} := 1-\sum_{i=1}^{k-1}f_{i}$ and
$$
\Delta^{k-1}_\mu := \left\{f\in \R^{k-1} \,\Big|\, \sum_{i=1}^{k-1}f_{i} \leq 1 - \mu, \,\, f_{i}\geq \mu \,\,\forall i \in [k-1] \right\}
$$
denotes the $\mu$-strict interior of $k$-dimensional probability simplex, for $0 < \mu \leq 1/k$. 
We require the weights $f_k$ to be in the strict interior for regularity (e.g., differentiability and exchanging derivatives with integrals) reasons. 
The data-generating process we consider 
proceeds similarly to what we considered in \Cref{sec:case_studies:nonmixing}.
Specifically, a latent variable $B\in [k]$ is first drawn according to $\{f_{i}\}_{i=1}^{k}$, which is then used condition the data $z_{1:T}\sim p_{\theta_{B}}$.

\paragraph{Mixture with known weights.} 
We first suppose $f$ is known. 
Hence the densities $p_{\theta}\in\calP$ are:
$$
p_{\theta}(z_{1:T}) =\sum_{i=1}^{k}f_{i}p_{\theta_i}(z_{1:T}),
$$
where 
$$
\theta:=\begin{pmatrix}\theta_{1}^{\T} & \ldots & \theta_{k}^{\T}\end{pmatrix}^{\T}=\begin{pmatrix}\theta_{1,1} & \ldots & \theta_{1,d} & \ldots & \theta_{k,1} & \ldots & \theta_{k,d}\end{pmatrix}^{\T}\in \Theta^{k}
$$
is the joint parameter to be learned.
A simple computation of the first order information yields:
\begin{align}
    \nabla_{\theta}\log p_{\theta}(z_{1:T}) = \begin{pmatrix}
    w_{\theta}^{(1)}(z_{1:T})\nabla_{\theta_{1}}\log p_{\theta_1}(z_{1:T})^{\T}  & \ldots &w_{\theta}^{(k)}(z_{1:T}) \nabla_{\theta_{k}}\log p_{\theta_k}(z_{1:T})^{\T}
    \end{pmatrix}^{\T}, \label{eq:mixture_general_gradient}
\end{align}
where $w_{\theta}^{(B)}(z_{1:T})$ is the posterior density of $B$ given $z_{1:T}$:
$$
w_{\theta}^{(i)}(z_{1:T}) := p_\theta(B=i\;|\;z_{1:T})=\frac{p_\theta(B=i)p_\theta(z_{1:T}\;|\;B=i)}{\sum_{i=1}^{k} p_\theta(B=i,z_{1:T})}=\frac{f_{i}p_{\theta_{i}}(z_{1:T})}{\sum_{i=1}^{k}f_{i}p_{\theta_{i}}(z_{1:T})}.
$$
We now compute the second order information in blocks: for $1\leq i\neq j\leq k$,
\begin{equation}\label{eq:mixture_general_hessian_components}
\begin{aligned}
H_{\theta}^{(ii)}(z_{1:T}) &:= \nabla^{2}_{\theta_{i}}\log p_{\theta}(z_{1:T}) \\
&=
w_{\theta}^{(i)}(z_{1:T})\left(1-w_{\theta}^{(i)}(z_{1:T})\right)\left(\nabla_{\theta_{i}}\log p_{\theta_{i}}(z_{1:T})\right)^{\otimes2}+w_{\theta}^{(i)}(z_{1:T})\nabla^{2}_{\theta_{i}}\log p_{\theta_{i}}(z_{1:T}),\\
H_{\theta}^{(ij)}(z_{1:T}) &:= \nabla_{\theta_{j}}\nabla_{\theta_{i}}\log p_{\theta}(z_{1:T})=w_{\theta}^{(i)}(z_{1:T})w_{\theta}^{(j)}(z_{1:T})\nabla_{\theta_{i}}\log p_{\theta_{i}}(z_{1:T})\nabla_{\theta_{j}}\log p_{\theta_{j}}(z_{1:T})^{\T},
\end{aligned}
\end{equation}
and the Hessian matrix 
$\nabla^2_\theta \log p_\theta(z_{1:T})$
is
\begin{equation}\label{eq:mixture_general_hessian}
H_{\theta}(z_{1:T}) :=
\begin{pmatrix}
H_{\theta}^{(11)}(z_{1:T}) & H_{\theta}^{(12)}(z_{1:T}) & \cdots & H_{\theta}^{(1k)}(z_{1:T}) \\
H_{\theta}^{(21)}(z_{1:T}) & H_{\theta}^{(22)}(z_{1:T}) & \cdots & H_{\theta}^{(2k)}(z_{1:T}) \\
\vdots & \vdots & \ddots & \vdots \\
H_{\theta}^{(k1)}(z_{1:T}) & H_{\theta}^{(k2)}(z_{1:T}) & \cdots & H_{\theta}^{(kk)}(z_{1:T})
\end{pmatrix}.
\end{equation}
We note the similarity between (\ref{eq:mixture_general_gradient}--\ref{eq:mixture_general_hessian}) and (\ref{eq:mixture_twostate_gradient}--\ref{eq:mixture_twostate_hessian}). We first consider the covering number bound. We assume for each mixture component $1\leq i\leq k$, we have 
the almost sure bounds:
$$
\norm{\nabla_{\theta_{i}}\log p_{\theta_{i}}(z_{1:T})}\lesssim c_{i}T,\quad \opnorm{\nabla^{2}_{\theta_{i}}\log p_{\theta_{i}}(z_{1:T})}\lesssim c'_{i}T,
$$
where $c_{i},\;c'_{i}$ are some constants depending on system parameters such as state dimension, etc. 
As we will see in later sections, this is possible to show for many systems.
We then have for $1\leq i,j\leq k$,
$$
\opnorm{H_{\theta}^{(ij)}(z_{1:T})}\lesssim \left(\max_{1\leq i\leq k}c_{i}\right)^2T^{2}.
$$
Therefore we can apply \Cref{prop:blockmajorization} to get
\begin{align*}
H_{\theta}(z_{1:T})&\preceq \blkdiag\left\{\left(\sum_{j=1}^{k}\opnorm{H_{\theta}^{(1j)}(z_{1:T})}\right)I_{d},\;\ldots,\;\left(\sum_{j=1}^{k}\opnorm{H_{\theta}^{(kj)}(z_{1:T})}\right)I_{d}\right\} \preceq k\left(\max_{1\leq i\leq k}c_{i}\right)^2T^{2}I_{kd},
\end{align*}
which implies $\calI(\theta)=-\mathbb{E}_{\theta}\left[H_{\theta}(z_{1:T})\right]\preceq k\left(\max_{1\leq i\leq k}c_{i}\right)^2T^{2}I_{kd}$. That is, we have
$$
\calI_{\max}:=k\left(\max_{1\leq i\leq k}c_{i}\right)^2T^{2}I_{kd},
$$
which gives us a bound on the metric entropy under $\text{FI}$-norm:
$$
\log\mathcal{N}_{\mathrm{FI}}\left(\mathcal{P},\e\right)\lesssim kd \log \left(\left.k\left(\max_{1\leq i\leq k}c_{i}\right)T\right/\e\right).
$$
This allows us to carry out the analysis done in Step 1. 

For Step 2 onward, we inspect $w_{\theta}^{(i)}(z_{1:T})$:
\begin{align*}
w_{\theta}^{(i)}(z_{1:T})&=\frac{1}{1+\frac{\sum_{j\neq i}f_{j}p_{\theta_{j}}(z_{1:T})}{f_{i}p_{\theta_{i}}(z_{1:T})}}=\frac{1}{1+\sum_{j\neq i}\frac{f_{j}}{f_{i}}\frac{p_{\theta_{j}}(z_{1:T})}{p_{\theta_{i}}(z_{1:T})}}.
\end{align*}
Ideally, each component of the mixture should be identifiable from the others given long enough trajectories. Hence, a natural assumption is that when $B=i$ (i.e., $z_{1:T}\sim p_{\theta_{i}}$), for any $j\neq i$ the following ergodic condition holds: there exists some constant $\Delta_{ij}>0$ such that
\begin{align}
    \frac{1}{T} \log\left( \frac{p_{\theta_j}(z_{1:T})}{p_{\theta_i}(z_{1:T})} \right) \overset{T \to \infty}{\longrightarrow} - \Delta_{ij}\quad \text{a.s.} \label{eq:mixture_general_assump}
\end{align}
The immediate implication of \eqref{eq:mixture_general_assump} is that when $B=i$:
$$
w_{\theta}^{(i)}(z_{1:T})=\frac{1}{1+\sum_{j\neq i}\frac{f_{j}}{f_{i}}\exp\left\{\log\left( \frac{p_{\theta_j}(z_{1:T})}{p_{\theta_i}(z_{1:T})} \right)\right\}}\overset{T \to \infty}{\longrightarrow} \frac{1}{1+\sum_{j\neq i}\frac{f_{j}}{f_{i}}\exp\left\{-\infty\cdot \Delta_{ij}\right\}}=1\quad \text{a.s.}
$$
On a flip side, when $B=k\neq i$, we have:
\begin{align*}
w_{\theta}^{(i)}(z_{1:T})=\frac{1}{1+\sum_{j\neq i}\frac{f_{j}}{f_{i}}\exp\left\{\log\left( \frac{p_{\theta_j}(z_{1:T})}{p_{\theta_i}(z_{1:T})} \right)\right\}}&\leq \frac{1}{1+\frac{f_{k}}{f_{i}}\exp\left\{\log\left( \frac{p_{\theta_k}(z_{1:T})}{p_{\theta_i}(z_{1:T})} \right)\right\}}\\
&\overset{T \to \infty}{\longrightarrow} \frac{1}{1+\frac{f_{k}}{f_{i}}\exp\left\{\infty\cdot \Delta_{ki}\right\}}=0\quad \text{a.s.}
\end{align*}
Since $w_{\theta}^{(i)}(z_{1:T})\geq 0$, we have $w_{\theta}^{(i)}(z_{1:T})\overset{T \to \infty}{\longrightarrow}0$ a.s.\ when $B=k\neq i$. Therefore we conclude
\begin{equation}\label{eq:mixture_general_posterior_collapse}
w_{\theta}^{(i)}(z_{1:T})\overset{T \to \infty}{\longrightarrow} \ind\{B = i \} \quad \text{a.s.}
\end{equation}
For a concrete example of \eqref{eq:mixture_general_assump}, 
suppose that the $p_{\theta_{i}}$'s are Markovian for each $1\leq i\leq k$, i.e., $p_{\theta_{i}}(z_{1:T})=p(z_{1})\prod_{t=1}^{T-1}p_{\theta_{i}}(z_{t+1} \mid z_{t})$.
The natural ergodicity assumption in this case is the following:
\begin{equation}\label{eq:mixture_general_assump_markov}
\frac{1}{T} \sum_{t=1}^{T-1}h_{ij}(z_{t},z_{t+1})\overset{T \to \infty}{\longrightarrow}  \E_{(z_{t},z_{t+1})\sim \pi_{\theta_{i}}\otimes p_{\theta_{i}}}\left[h_{ij}(z_{t},z_{t+1})\right] \quad \text{a.s.},
\end{equation}
where $h_{ij}(z_{t},z_{t+1}):=\log\left( \frac{p_{\theta_j}(z_{t+1} \mid z_{t})}{p_{\theta_i}(z_{t+1} \mid z_{t})} \right)$ and $\pi_{\theta_{i}}$ is the density of ergodic measure of the Markov process $\left\{z_{t}\right\}_{t=1}^{\infty}$ under $p_{\theta_{i}}$. Here, the $\otimes$ notation denotes the following operation between a density $\pi$ and a transition density $p$: $\pi\otimes p(z_{t},z_{t+1}):=\pi(z_{t})p(z_{t+1} \mid z_{t})$.
To see why this implies \eqref{eq:mixture_general_assump}, observe that
$\pi_{\theta_{i}}\otimes p_{\theta_{i}}$ is the density of the ergodic measure of the augmented process $\{(z_{t}, z_{t+1})\}_{t=1}^{\infty}$ under $p_{\theta_{i}}$. Hence the right hand side of \eqref{eq:mixture_general_assump_markov} reads:
\begin{align*}
\E_{(z_{t},z_{t+1})\sim \pi_{\theta_{i}}\otimes p_{\theta_{i}}}\left[h_{ij}(z_{t},z_{t+1})\right]&=\int_{\sfZ^{2}} \pi_{\theta_{i}}\otimes p_{\theta_{i}}(z_{t},z_{t+1})\log \left(\frac{p_{\theta_{j}}(z_{t+1}|z_{t})}{p_{\theta_{i}}(z_{t+1}|z_{t})}\right) \rmd z_{t}\;\rmd z_{t+1}\\
&=-\int_{\sfZ^{2}} \pi_{\theta_{i}}\otimes p_{\theta_{i}}(z_{t},z_{t+1})\log \left(\frac{\pi_{\theta_{i}}\otimes p_{\theta_{i}}(z_{t},z_{t+1})}{\pi_{\theta_{i}}\otimes p_{\theta_{j}}(z_{t},z_{t+1})}\right) \rmd z_{t}\;\rmd z_{t+1}\\
&=-\KL{\pi_{\theta_{i}}\otimes p_{\theta_{i}}}{\pi_{\theta_{i}}\otimes p_{\theta_{j}}}=:-\Delta_{ij}.
\end{align*}

Following \eqref{eq:mixture_general_posterior_collapse}, we can again argue that for large $T$, the Hessian (and therefore the Fisher information) can be controlled by a block-diagonal matrix of 
the mixture components' Fisher information matrices
\begin{equation}\label{eq:mixture_blk_diag_fisher}
\calI_{\blkdiag}(\theta):=\begin{pmatrix}
-\E^{(1)}_{\theta}\left[\nabla^{2}_{\theta_{1}}\log p_{\theta_{1}}(z_{1:T})\right] & 0 & \cdots & 0 \\
0 & -\E^{(2)}_{\theta}\left[\nabla^{2}_{\theta_{2}}\log p_{\theta_{2}}(z_{1:T})\right] & \cdots & 0 \\
\vdots & \vdots & \ddots & \vdots \\
0 & 0 & \cdots & -\E^{(k)}_{\theta}\left[\nabla^{2}_{\theta_{k}}\log p_{\theta_{k}}(z_{1:T})\right]
\end{pmatrix}
\end{equation}
under the Loewner order.  

Some remarks are in order. First, for an exact analogue of \Cref{corollary:eigenvaluebound}, we need to prove the Loewner order equivalence holds uniformly within a ball around $\theta_{\star}$. Our previous proof strategy requires characterizing non-asymptotic mixing behavior instead of the asymptotic convergence as in \eqref{eq:mixture_general_assump}. In particular, if we are working with a Markov process and consider the augmented process $\left\{(z_{t},z_{t+1})\right\}_{t=1}^{\infty}$ with an ergodic measure denoted by $\pi$, we expect the following Bernstein-type inequality to hold:\footnote{Here we only stated a schematic form. More precisely, under various mixing conditions, the right hand side might include additional $\polylog(T)$ factors.}
$$
\Pr\left(\left|\frac{1}{T}\sum_{t=1}^{T-1}h(z_{t},z_{t+1})-\E_{\pi}\left[h(z_{1},z_{2})\right]\right|\geq s\right)\lesssim \exp\left(\frac{-T s^{2}}{\Var_{\pi}(h)+C(\calH)s }\right),\quad h\in \calH,
$$
where $C(\calH)$ typically quantifies boundedness of the function class $\calH$. Such results are available for various classes of mixing processes (see e.g., \cite{Dedecker2004,MAUMEDESCHAMPS2006,Merlevde2009,Olivier2010,Hang2017,yuan2025exponentialinequalitiesmixingprocesses}). In particular, stable LDS are geometrically $\beta$-mixing~\cite{Liebscher2005}, and therefore \cite[Theorem 1 and 2]{Merlevde2009} are readily applicable. Specialized bounds for Markov chains are also available~\cite[see e.g.,][Theorem 2.4]{huang2024bernsteintypeinequalitiesmarkovchains}. This would allow us to obtain instance-optimal rates for the more general mixture dynamics, including but not limited to those considered in prior art (e.g.,~\cite{pmlr-v162-chen22t,pmlr-v202-kausik23a}).

Second, we face the technical challenge that Hellinger identifiability (analogue of \Cref{prop:mixture_identifiability}) 
does not in general hold when a subset of weights are equal.
In the scalar case, as in \Cref{prop:mixture_identifiability}, we
worked around this issue by assuming a monotone order on the corresponding parameters to guarantee unique identifiability of parameters. 
In the general case, we need to
redefine the notion of Hellinger identifiability to
be symmetry aware.
In particular, let $\mathcal{J} = (\mathcal{J}_1, \dots, \mathcal{J}_\ell)$ 
partition the indices $[k]$ into $\ell \leq k$ equivalence classes, where $f_{i_1} = f_{i_2}$ for all $i_1, i_2 \in \mathcal{J}_i$, and $f_{i_1} \neq f_{j_1}$ for all 
$i_1 \in \mathcal{J}_i$, $j_1 \in \mathcal{J}_j$ with $i \neq j$.
Next, let $\Symm_{\mathcal{J}}(\theta)$
denote the set of cardinality $\prod_{i=1}^{\ell} (\abs{\mathcal{J}_i}!)$ which given a parameter vector $\theta \in \Theta^k$ enumerates
all possible permutations within each equivalence class for $\theta$.
We then consider the following modified definition of
Hellinger identifiability (cf.~\Cref{def:identifiability}), which states that
there exists $(\gamma_1, \gamma_2)$ such that:
\begin{align*}
    \Hg{p_\theta}{p_\star} \leq \gamma_1 \Longrightarrow \min_{\theta^\pi \in \Symm_{\mathcal{J}}(\theta)} \norm{\theta^\pi - \theta_\star} \leq \gamma_2 \cdot \Hg{p_\theta}{p_\star}.
\end{align*}
We note the above symmetrized condition needs to be shown on a problem-specific basis.

We conclude by remarking that under the particular outline above, it seems necessary to require ergodicity of mixture \emph{components} to obtain non-trivial results, as otherwise the behavior of Fisher information is difficult to analyze. However, we do know that, for example, LDS with non-mixing behavior can still be identified with parametric rate from a single trajectory~\cite{simchowitz2018learningwithoutmixing}
and so we postulate that e.g., identifying mixtures of LDS may also be possible without mixing assumptions. This reflects a limitation for our current instantiation of Hellinger localization for mixture recovery, 
which we leave addressing to future work.

\paragraph{Mixture with unknown weights.} 
We now extend the above calculation to the fully general setting where the joint parameter of interest is:
$$
(\theta,f):=\begin{pmatrix}\theta_{1}^{\T} & \ldots & \theta_{k}^{\T} & f^{\T}\end{pmatrix}^{\T}=\begin{pmatrix}\theta_{1,1} & \ldots & \theta_{1,d} & \ldots & \theta_{k,1} & \ldots & \theta_{k,d} & f_{1} & \ldots & f_{k-1}\end{pmatrix}^{\T}\in \Theta^{k}\times \Delta^{k-1}_\mu.
$$
The density now takes the form
$$
p_{\theta,f}(z_{1:T}) =\sum_{i=1}^{k-1}f_{i}p_{\theta_i}(z_{1:T})+\left(1-\sum_{i=1}^{k-1}f_{i}\right)p_{\theta_{k}}(z_{1:T}).
$$
The cross-terms of the information
matrix between the mixture weights $f$ and parameters $\theta$ are:
\begin{equation}\label{eq:mixture_weight_hessian_cross}
\begin{aligned}
\partial_{f_{i}}\nabla_{\theta_{j}}\log p_{\theta,f}(z_{1:T}) 
&=\left(\frac{w_{\theta,f}^{(j)}(z_{1:T})w_{\theta,f}^{(k)}(z_{1:T})}{f_{k}}-\frac{w_{\theta,f}^{(j)}(z_{1:T})w_{\theta,f}^{(i)}(z_{1:T})}{f_{i}}\right)\nabla_{\theta_{j}}\log p_{\theta_{j}}(z_{1:T}),\\
\partial_{f_{i}}\nabla_{\theta_{i}}\log p_{\theta,f}(z_{1:T})
&=\left(\frac{1}{f_{i}}w_{\theta,f}^{(i)}(z_{1:T})\left(1-w_{\theta,f}^{(i)}(z_{1:T})\right)+\frac{w_{\theta,f}^{(i)}(z_{1:T})w_{\theta,f}^{(k)}(z_{1:T})}{f_{k}}\right)\nabla_{\theta_{i}}\log p_{\theta_{i}}(z_{1:T}),
\end{aligned}
\end{equation}
where the posterior weight
$$
w_{\theta,f}^{(i)}(z_{1:T}) := \frac{f_{i}p_{\theta_{i}}(z_{1:T})}{\sum_{i=1}^{k}f_{i}p_{\theta_{i}}(z_{1:T})}.
$$
is defined similarly as before.
Under the same ergodicity assumptions \eqref{eq:mixture_general_assump} from the previous section, both terms of \eqref{eq:mixture_weight_hessian_cross} will converge to zero. We now calculate the Hessian with respect to the mixture weights:
\begin{align*}
\partial_{f_{j}}\partial_{f_{i}}\log p_{\theta}(z_{1:T}) %
&=\frac{w_{\theta,f}^{(i)}(z_{1:T})w_{\theta,f}^{(k)}(z_{1:T})}{f_{i}f_{k}}-\frac{w_{\theta,f}^{(i)}(z_{1:T})w_{\theta,f}^{(j)}(z_{1:T})}{f_{i}f_{j}}+\frac{w_{\theta,f}^{(j)}(z_{1:T})w_{\theta,f}^{(k)}(z_{1:T})}{f_{j}f_{k}}-\left(\frac{w_{\theta,f}^{(k)}(z_{1:T})}{f_{k}}\right)^{2},\\
\partial^{2}_{f_{i}}\log p_{\theta}(z_{1:T})
&=\frac{2w_{\theta,f}^{(i)}(z_{1:T})w_{\theta,f}^{(k)}(z_{1:T})}{f_{i}f_{k}}-\left(\frac{w_{\theta,f}^{(i)}(z_{1:T})}{f_{i}}\right)^{2}-\left(\frac{w_{\theta,f}^{(k)}(z_{1:T})}{f_{k}}\right)^{2}.
\end{align*}
Now under the ergodicity assumption \eqref{eq:mixture_general_assump}, we can write the Hessian block $H_f(z_{1:T}) = \nabla^2_f \log p_{\theta,f}(z_{1:T})$
and FI matrix block $\calI(f) = - \E_{\theta,f}[ H_f(z_{1:T})]$
for $T$ large as follows:
\begin{equation}\label{eq:mixture_weight_fisher}
\begin{aligned}
H_{f}(z_{1:T})&\stackrel{T\to\infty}{\longrightarrow}-\diag\left\{\left(\frac{w_{\theta,f}^{(1)}(z_{1:T})}{f_{1}}\right)^{2},\ldots,\left(\frac{w_{\theta,f}^{(k-1)}(z_{1:T})}{f_{k-1}}\right)^{2}\right\}-\left(\frac{w_{\theta,f}^{(k)}(z_{1:T})}{f_{k}}\right)^{2}\ind\ind^{\T},\\
\calI(f)&\stackrel{T\to\infty}{\longrightarrow}\underbrace{\diag\left\{\E_{\theta,f}\left[\left(\frac{w_{\theta,f}^{(1)}(z_{1:T})}{f_{1}}\right)^{2}\right],\ldots,\E_{\theta,f}\left[\left(\frac{w_{\theta,f}^{(k-1)}(z_{1:T})}{f_{k-1}}\right)^{2}\right]\right\}}_{:=\calI_{\diag}(f)}+\E_{\theta,f}\left[\left(\frac{w_{\theta,f}^{(k)}(z_{1:T})}{f_{k}}\right)^{2}\right]\ind\ind^{\T}\\
&\preceq \calI_{\diag}(f)+\frac{1}{\left(1-\sum_{i=1}^{k-1}f_{i}\right)^{2}}\ind\ind^{\T}.
\end{aligned}
\end{equation}
Now combining \eqref{eq:mixture_blk_diag_fisher}, \eqref{eq:mixture_weight_hessian_cross} and \eqref{eq:mixture_weight_fisher} together, we obtain
\begin{equation}\label{eq:mixture_general_fisher}
\begin{aligned}
\calI(\theta,f)&\stackrel{T\to\infty}{\longrightarrow}\begin{pmatrix}\calI_{\blkdiag}(\theta) & 0  \\ 0 & \calI_{\diag}(f)+\E_{\theta,f}\left[\left(\frac{w_{\theta,f}^{(k)}(z_{1:T})}{f_{k}}\right)^{2}\right]\ind\ind^{\T}\end{pmatrix}\\
&\preceq \begin{pmatrix}\calI_{\blkdiag}(\theta) & 0  \\ 0 & \calI_{\diag}(f)+\frac{1}{\left(1-\sum_{i=1}^{k-1}f_{i}\right)^{2}}\ind\ind^{\T}\end{pmatrix}.
\end{aligned}
\end{equation}
This sheds light on how we can generalize the proof in \Cref{sec:mixture_proof} to the case where the mixture weights $f$ also need to be estimated,
in addition to the mixture parameters $\theta$.
Indeed, the proof strategy would be similar to that of \Cref{sec:mixture_proof}; however, one additional challenge is that one would need to verify the Hellinger identifiability of the mixture weights.

\subsection{Dependent Regression under General Product-Noise Distributions}
\label{sec:case_studies:l1_regression}

We next consider the following family of trajectory distributions $p_\theta(z_{1:T})$ over $\sfZ = \R^d$ parameterized by $\theta \in \Theta$ of the following form:
\begin{align}
    z_{t+1} = M(z_t) \theta + w_t, \quad z_1 \sim \rho_1. \label{eq:generalized_regression}
\end{align}
Here, the matrix-valued map $M : \R^d \mapsto \R^{d \times p}$ is allowed to be non-linear, and assumed to be known.
This setup generalizes the linear system identification problem detailed in 
\Cref{sec:general_framework:MLE} and has received considerable attention recently as a 
tractable form of
non-linear system identification, especially when a control input is added to the matrix $M$,
i.e., $z_{t+1} = M(z_t, u_t) \theta + w_t$ (see e.g.,~\cite{mania2022nonlinearactive, kakade2020nonlinearinfotheoretic, wagenmaker2023optimalexpl,musavi2024nonlinear});
a more detailed literature review is given in \Cref{sec:l1_regression:related_work}.
The noise variable $w_t$ is drawn independently across time $t$ from a distribution which has the following product density w.r.t. the Lebesgue measure on $\R^d$:
\begin{align}
    p_{\bm{\phi}}(w) = \prod_{j=1}^{d} p_{\phi}(w_j), \quad
    p_{\phi}(w_1) = \exp\left\{ - \phi(w_1) \right\}/Z(\phi), \quad Z(\phi) := \int \exp\left\{ -\phi(w_1) \right\} \rmd w_1, \label{eq:generalized_regression_noise_dist}
\end{align}
where $\phi : \R \mapsto \R$ is a known scalar function parameterizing the noise distribution.
Hence, our setup differs from more standard settings in the following way: we do not need to assume the noise is either Gaussian or sub-Gaussian, but 
the functional form of the density is needed to solve the MLE. 
In what follows, given a vector $w \in \R^d$, we let $\bm{\phi}$ denote the function mapping $\R^d \mapsto \R^d$ defined as $\bm{\phi}(w) := (\phi(w_1), \dots, \phi(w_d))$.
With this notation, we can write the MLE \eqref{eq:MLE} for \eqref{eq:generalized_regression} as:
\begin{align}
    \hat{\theta}_{m,T} \in \argmin_{\theta \in \Theta} \sum_{i=1}^{m} \sum_{t=1}^{T-1} \ip{\ind}{ \bm{\phi}( z_{t+1}^{(i)} - M(z_t^{(i)}) \theta) }. \label{eq:MLE_dependent_regression}
\end{align}
We assume the following regularity conditions on $\phi$.
\begin{mydef}[Regularity conditions on $\phi$]
\label{def:phi_regularity}
We say that $\phi : \R \mapsto \R$ is $(\beta_1,\beta_2)$-regular for constants $1 \leq \beta_1,\beta_2 < \infty$, if the following conditions hold:
\begin{enumerate}[label=(\alph*)]
    \item $\phi\in C^{2}(\R)$ and $Z(\phi) < \infty$, 
    \item Both $\lim_{\abs{x}\to\infty} \phi(x) = \infty$ and  $\lim_{\abs{x} \to \infty} \abs{\phi'(x)} \exp(-\phi(x)) = 0$, 
    \item $\sigma^2_\phi := (\E_{w \sim p_\phi}[ (\phi'(w))^2 ])^{-1} < \infty$,
    \item $\E_{w \sim p_{\phi}}[ (\phi''(w))^2 ] \leq \beta_1 /\sigma_\phi^4$, and 
    \item $\E_{w \sim p_{\phi}}[ (\phi'(w))^8 ] \leq \beta_2/\sigma_\phi^8$.
\end{enumerate}
\end{mydef}
Before we look at a few examples of distributions satisfying
\Cref{def:phi_regularity}, we briefly describe the role of each condition.
Condition (a) simply ensures that $p_\phi$ is a well-defined $C^2(\R)$ density;
having two derivatives is crucial in our framework, which relies on second order expansions.
Condition (b) controls the growth of $p_\phi$ and states that
$p_\phi$ must tend to zero in both directions which implies that
$\E_{w \sim p_\phi}[ \phi'(w) ] = 0$, and that 
the following integration by parts (IBP) identity
$\E_{w \sim p_\phi}[ (\phi'(w))^2 ] = \E_{w \sim p_\phi}[ \phi''(w) ]$ holds (see \Cref{prop:phi_properties}); both identities play a key role in our analysis.
Condition (c) ensures (via the IBP identity) that the amount of integrated curvature 
$\E_{w \sim p_\phi}[ \phi''(w) ]$ is bounded away from zero, and is necessary 
for non-degenerate Fisher Information matrices $\calI(\theta)$; 
note that in the case when $\phi$ is convex, then this condition is equivalent to
$\phi''(w)$ cannot equal zero almost everywhere.
Conditions (d) and (e) are \emph{hyper-contractivity} conditions; indeed by the IBP identity 
(d) is equivalent to 
$\E_{w \sim p_\phi}[ (\phi''(w))^2 ] \leq \beta_1 (\E_{w \sim p_\phi}[ \phi''(w) ])^2$,
and similarly (e) states that
$\E_{w \sim p_\phi}[ (\phi'(w))^8 ] \leq \beta_2 (\E_{w \sim p_\phi}[ (\phi'(w))^2 ])^4$;
by Jensen's inequality both $\beta_1, \beta_2$ must be $\geq 1$.

We next build some intuition for the generality of \Cref{def:phi_regularity},
by giving a few examples below with explicit $(\beta_1, \beta_2)$ constants;
proofs for the examples are given in \Cref{sec:l1_regression:regularity_condition_proofs}.

\begin{ex}[Multivariate normal distribution]
For any $\nu > 0$,
$$
\phi_{\nu}(x)=\frac{x^{2}}{2\nu^{2}}
$$
satisfies the conditions in \Cref{def:phi_regularity} 
with $\sigma^2_{\phi_\nu} = \nu^2$ and $(\beta_1, \beta_2) = (1, 105)$.
\end{ex}

\begin{ex}[Smoothed ``Bang-Bang'' noise]
\label{ex:smoothed_bang_bang_noise}
For $\nu > 0$, consider 
\begin{align*}
    \phi_\nu(x) = \frac{x^2 + 1}{2\nu^2} - \log\cosh(x/\nu^2).
\end{align*}
This corresponds to $p_{\phi_\nu} = \frac{1}{2} \sfN(1, \nu^2) + \frac{1}{2} \sfN(-1, \nu^2)$, a Gaussian mixture model with two $\nu^2$ variance mixtures centered as $\pm 1$.
If $\nu \in (0, 1)$, then $\phi_\nu$ satisfies the conditions in \Cref{def:phi_regularity}
for $(\beta_1, \beta_2) = (c'/\nu^6, c''/\nu^{24})$, where $c', c''$ are universal positive constants.
Note that when $\nu \to 0$, $p_{\phi_\nu}$ approaches ``bang-bang'' noise $\frac{1}{2} \delta_{1} + \frac{1}{2} \delta_{-1}$.
\end{ex}

\begin{ex}[Smoothed Laplace distribution]
\label{ex:smoothed_laplace}
Let us define $\phi$ as:
$$
    \phi_{c,\nu}(x) = \frac{1}{c} \log \cosh(cx/\nu), \quad c, \nu \in \R_{> 0}.
$$
As $c \to \infty$, we have that $\phi_{c,\nu}(x) \to \abs{x/\nu}$ pointwise, so 
$p_{\phi_{c,\nu}}$ is a smoothed Laplace distribution 
with second-order curvature. 
Define $Z(c) :=  \int \cosh(cx)^{-1/c} \rmd x$.
We have that $\phi_{c,\nu}$ satisfies the conditions of \Cref{def:phi_regularity}
for $(\beta_1, \beta_2) = \left(\frac{2c}{\tanh^2(cZ(c)/4)}, \frac{16}{\tanh^8(cZ(c)/4)}\right)$.
\end{ex}

With the data generating process $p_\theta(z_{1:T})$ in place, we now turn to the analysis of the MLE estimator in this model. We remark that the MLE estimator \eqref{eq:MLE_dependent_regression} in general for this problem is \emph{not} the solution to a least-squares regression problem (unless $p_\phi$ is Gaussian), 
nor is it generally the solution to convex optimization problem (unless $\phi$ is convex).
Furthermore, as seen in \Cref{ex:smoothed_laplace},
the noise does not necessarily have sub-Gaussian tails as well, as is the standard assumption in 
many dependent learning works~(e.g.,~\cite{simchowitz2018learningwithoutmixing,ziemann2022learning,tu2024learning,sarkar2019linear,foster2020nonlinear,musavi2024nonlinear}), although we note that some works have also considered heavier-tailed noise 
in various settings~\cite{faradonbeh2018identification,kowshik2021nonlinear,roy2021dependent,kanakeri2025}.
The following is our main result regarding parameter recovery 
for the model \eqref{eq:generalized_regression}.

\begin{mythm}
\label{thm:regression_generalized_noise}
Fix $\delta \in (0, 1)$, and
suppose the following assumptions hold:
\begin{enumerate}[label=(\alph*)]
    \item $\phi$ is $(\beta_1, \beta_2)$-regular per \Cref{def:phi_regularity},
    \item $M_{1} :=\sup_{\theta \in \Theta} \left( \E_{p_\theta}\left[\frac{1}{T-1}\sum_{t=1}^{T-1} \opnorm{M(z_t)}^4 \right]\right)^{1/4}<\infty$,
    \item $M_{2} :=\sup_{\theta \in \Theta} \left(\E_{p_\theta}\left[\frac{1}{T-1}\sum_{t=1}^{T-1} \opnorm{M(z_t)}^8 \right]\right)^{1/8}<\infty$,
    \item $\bar{\mu} := \sup_{\theta \in \Theta} \lambda_{\max}\left( \bar{\calI}(\theta) \right) < \infty$,
    \item $\underline{\mu}:= \inf_{\theta \in \Theta} \lambda_{\min}\left( \bar{\calI}(\theta) \right) > 0$, and
    \item $T \geq \beta_2^{1/2} d^2 (M_2/M_1)^4$.
\end{enumerate}
Let $\Theta = \{ \theta \in \R^p \mid \norm{\theta} \leq R \}$, and 
let $\hat{\theta}^\e_{m,T}$ denote the max FI discretized MLE estimator \eqref{eq:discrete_MLE} at resolution $\e=\delta/(2\sqrt{2m})$. 
Assume wlog that $R, M_1, \bar{\mu} \geq 1$,
and define $\kappa := \bar{\mu}/\underline{\mu}$.
Then:
\begin{enumerate}[label=\textbf{(\alph*).}]
    \item If $\calP$ is $(\gamma_1, \gamma_2)$-identifiable (cf.~\Cref{def:identifiability})
and 
the number of trajectories $m$ satisfies:
\begin{align*}
    m \gtrsim \max\bigg\{& p/\gamma_1^2 \cdot \log(c_1' p/\delta \cdot  R \bar{\mu}  T/\gamma_1 ), 
    \:\: pT\gamma_2^2 \cdot \beta_1 M_1^4 \kappa/\underline{\mu} \cdot \log( c_1'' p/\delta \cdot \beta_1 R M_1 T \kappa \max\{\gamma_2,1\} ), \\
    &\quad pT \gamma_2^2 \cdot M_1^4 \kappa / (\underline{\mu}\sigma_\phi^4) \cdot \log( c_1''' p/\delta \cdot R M_1 T \kappa \max\{\sigma_\phi^{-1}, 1\} \max\{\gamma_2,1\}) \bigg\},
\end{align*}
then with probability at least $1-\delta$,
\begin{align}
    \norm{\hat{\theta}^\e_{m,T}-\theta_\star}^2_{\bar{\calI}(\theta_\star)} \lesssim \frac{p\log(c_1 R \bar{\mu} \cdot mT/\delta)}{mT}. \label{eq:generalized_regression_final_rate}
\end{align}
Here, $c_1,c_1',c_1'',c_1'''$
are universal positive constants.
\item  On the other hand if $\phi$ is convex, then as long as the number of trajectories $m$ satisfies
\begin{align*}
    m \gtrsim \max\bigg\{& p \cdot \beta_1 M_1^4/\underline{\mu}^2 \cdot \log(c_2'p/\delta \cdot  \beta_1 R M_1 T \kappa ), \\
      &\quad p \cdot M_1^4/(\underline{\mu}^2\sigma_\phi^4) \cdot \log(c_2'' p/\delta \cdot R M_1 T \kappa \max\{\sigma_\phi^{-1}, 1\} ) \bigg\},
\end{align*}
then with probability at least $1-\delta$
the rate \eqref{eq:generalized_regression_final_rate} also holds,
with $c_1$ replaced by $c_2$. Here, $c_2, c_2', c_2''$ are universal positive constants.
\end{enumerate}
\end{mythm}

Before turning to the proof of \Cref{thm:regression_generalized_noise} (cf.~\Cref{sec:case_studies:l1_regression:proof}),
some remarks are in order. 
For the present discussion, we focus only on the characteristics of \Cref{thm:regression_generalized_noise}, deferring a detailed account and comparison to related work to \Cref{sec:l1_regression:related_work}.
Focusing only on the parameters $m, T, p$,
\Cref{thm:regression_generalized_noise} states that (a) in general, 
if $m \gtrsim \tilde{\Omega}(pT)$, then the nearly (up to logarithmic factors)
instance-optimal rate $\norm{ \hat{\theta}^\e_{m,T} - \theta_\star }^2_{\bar{\calI}(\theta_\star)} \lesssim \tilde{O}(p/(mT))$ from \eqref{eq:generalized_regression_final_rate} holds,
and (b) if the scalar function $\phi$ parameterizing the noise
distribution \eqref{eq:generalized_regression_noise_dist} is convex
then the requirement on $m$ improves to $m \gtrsim \tilde{\Omega}(p)$,
with the final rate \eqref{eq:generalized_regression_final_rate} remaining the same.
In the $\phi$ convex case (b), the requirement on $m$ is in general not improvable, as is shown by the lower bounds in \cite[Section 6]{tu2024learning} for linear regression. For case (a), the worse dependence comes from the non-concavity of the log-likelihood when $\phi$ is not convex, which requires us to use \eqref{eq:framework_B1_B2_general}, compared with the concave log-likelihood when $\phi$ is convex; see the discussion in \Cref{sec:hellinger_localization_framework}.

We next comment on the stated assumptions.
The regularity Assumption (a) was previously discussed in the remarks
following \Cref{def:phi_regularity}.
Assumptions (b) and (c) control the growth of the feature matrix $M(z_t)$ 
over the trajectory $z_{1:T}$.
By two applications of Jensen's inequality, with $T' := T-1$,
\begin{align*}
    \left(\frac{1}{T'} \sum_{t=1}^{T'} \E_{p_\theta} \opnorm{M(z_t)}^4\right)^{1/4} \leq \left(\frac{1}{T'}\sum_{t=1}^{T'} \sqrt{ \E_{p_\theta} \opnorm{M(z_t)}^8 }\right)^{1/4} \leq \left( \frac{1}{T'}\sum_{t=1}^{T'} \E_{p_\theta} \opnorm{M(z_t)}^8 \right)^{1/8} \leq M_2,
\end{align*}
and therefore $M_1 \leq M_2$, so assumption (c) actually implies (b).
The growth of both $M_1$ and $M_2$ as a function of $T$ governs the 
dependence on $T$ for the minimum number of trajectories $m$; if $M$ is almost surely bounded, then $M_1,M_2$ are trivially $O(1)$.
Assumption (d) is implied by Assumption (b), since by another application of Jensen's inequality we have
$\bar{\mu} \leq (M_1/\sigma_\phi)^2$.
Assumption (e) is states that the FI matrix $\calI(\theta)$ is not degenerate
over $\Theta$, and is necessary for parameter recovery.
Assumption (f) is made to simplify the resulting expressions for the minimum number of trajectories
$m$ required, and can be easily removed.

\subsubsection{Comparison to System Identification Literature}
\label{sec:l1_regression:related_work}

\paragraph{Linear Dynamical System Identification.}
The LDS system identification problem reviewed in \eqref{eq:LDS}
is a special case of the model \eqref{eq:generalized_regression}, with 
$p = d^2$, $\theta = \vec(A)$, and 
$M(z) = (z^\T \otimes I_d)$.
Hence \Cref{thm:regression_generalized_noise} can be thought of as a generalization
of the results from \cite{tu2024learning} for multi-trajectory learning in LDS.
However, there are some caveats/limitations to the extent that
\Cref{thm:regression_generalized_noise} truly generalizes the result. 
Focusing on Assumption (c), we have that
$\opnorm{M(z)} = \opnorm{(z^\T \otimes I_d)} = \norm{z}$, and hence Assumption (c)
posits a uniform bound on the quantity
$\chi(\theta) := \frac{1}{T'} \sum_{t=1}^{T'} \E_{p_\theta}[\norm{z_t}^8]$ as $\theta$ varies over $\Theta$.
However, the quantity $\chi(\theta)$ exhibits two phase-transitions depending 
on the operator norm of $\mathrm{mat}(\theta)$.
When $\opnorm{\mathrm{mat}(\theta)} < 1$, then $\chi(\theta) = O(1)$ (ignoring all constants other than $T$) by the ergodic theorem.
On the other hand, when $\opnorm{ \mathrm{mat}(\theta) } = 1$, then $\chi(\theta) = \mathrm{poly}(T)$.
Finally, when $\opnorm{ \mathrm{mat}(\theta) } = \rho > 1$, we have $\chi(\theta) = \rho^{O(T)}$.
In the last regime, 
the bound \eqref{eq:generalized_regression_final_rate} becomes sub-optimal compared
with \eqref{eq:LDS_asymptotic_normality}, and ends up scaling as $1/m$ instead of the optimal $1/(mT)$.
Furthermore, in the $\opnorm{ \mathrm{mat}(\theta) } = 1$ regime, the
requirement on $m$ becomes $m \gtrsim \mathrm{poly}(T)$, which is also not sharp.
Thus, for LDS system identification, \eqref{eq:generalized_regression} is only sharp
in the case when $R<1$.

It is important to clarify that
the main issue is not that the Hellinger framework requires stability/mixing
of the process $z_{1:T}$, but instead the issue is that
for LTI systems, the states $z_t$ can easily grow exponential in $T$ depending on the parameter $A$, which makes both
covering and localization arguments extremely sensitive to minor
perturbations in the parameters. 
The situation for LDS can be somewhat reconciled by utilizing a closed-form lower bound for the trajectory-level Hellinger distance~\cite[Section 4]{ziemann2024short},
which would address the sub-optimal $m \gtrsim \mathrm{poly}(T)$ requirement when $R=1$.
However, when $R > 1$, this strategy would still not yield the correct rates, as we would still need
to perform a covering argument under the hood.

In the least-squares analysis from \cite{tu2024learning}, the issue of
uniform convergence when $A$ is not stable is handled elegantly
via the special structure of the square loss.
In particular, the square loss lends itself to an \emph{offset basic inequality}~\cite{liang2015offsetcomplexity} which allows the uniform convergence to be \emph{self-normalized}, preventing unstable $A$'s from adversely affecting the resulting 
covering numbers. We leave to future work a generalized form of self-normalization that can
also be applied to log losses and the Hellinger framework.
One possible starting point for this extension is the work of \cite{vijaykumar2021localization}, which provides techniques to
define and analyze offset empirical processes for logarithmic, and more generally exp-concave losses.

\paragraph{Non-linear System Identification.}

In its general form, problem \eqref{eq:generalized_regression} is typically
studied in its controlled variant, i.e., $z_{t+1} = M(z_t, u_t) \theta + w_t$, where $z_t$ is
interpreted as the state of a discrete-time dynamical system, and $u_t$ the
control input at time $t$.
We note that \Cref{thm:regression_generalized_noise} for identifying the model \eqref{eq:generalized_regression} can be readily translated into this control setting
with some minor modifications to incorporate the expectation over the control sequence $u_t$ in the 
Fisher information matrix, and also to include the full map $M(z_t, u_t)$ in the
definitions for $M_1, M_2$ in Assumptions (b), (c);
we omit the exact result in the interest of space.
Learning in the controlled formulation of \eqref{eq:generalized_regression} is studied mostly as an active learning problem, 
with a focus on designing optimal algorithms for selecting inputs \cite{mania2022nonlinearactive, kakade2020nonlinearinfotheoretic,wagenmaker2021taskoptimal,wagenmaker2023optimalexpl}; the necessity of
active learning in the single-trajectory setting,
absent smoothness conditions on $M(z, u)$, was demonstrated
by \cite{mania2022nonlinearactive}.
The line of work from \cite{wagenmaker2023optimalexpl,wagenmaker2021taskoptimal}
considers task-guided exploration, proposing an algorithm that quantifies which system parameters are most relevant to solving the task, and actively explores to minimize uncertainty in these parameters, achieving a near instance-optimal rate for the downstream task;
this was later extended by \cite{lee2024nonlinearactive} to general parameteric dynamics models.
Extending our Hellinger localization framework for active exploration, especially for downstream
control tasks, is exciting future work.

Perhaps the most directly related work is that of \cite{musavi2024nonlinear},
which shows that the feature map $M$ being real-analytic is sufficient to allow \emph{non-active} i.i.d.\ random control signals to suffice for parameter recovery.
Their arguments proceed by showing that since the zeros of the real-analytic function have 
measure zero, this implies that the standard martingale small-ball conditions (cf.~\cite{simchowitz2018learningwithoutmixing}) used to show a lower bound on the empirical covariance matrix hold generically.
This idea is also applicable to our framework, and can be used to certify non-degeneracy of the Fisher information matrix as required by Assumption (e) in \Cref{thm:regression_generalized_noise} for real-analytic feature maps.

\subsubsection{Proof of \Cref{thm:regression_generalized_noise}}
\label{sec:case_studies:l1_regression:proof}

We first state a simple result regarding the
noise distribution $p_\phi$ which will be useful in our analysis.
\begin{myprop}
\label{prop:phi_properties}
Given (a) and (b) of \Cref{def:phi_regularity}, 
the following identities are valid:
\begin{enumerate}[label=(\alph*)]
    \item $\E_{w \sim p_\phi}[ \phi'(w) ] = 0$,
    \item $\E_{w \sim p_\phi}[ (\phi'(w))^2 ] = \E_{w \sim p_\phi}[ \phi''(w) ]$.
\end{enumerate}
\end{myprop}
\begin{proof}
We first note that:
\begin{align*}
    p'_\phi(w) = -\phi'(w) \exp(-\phi(w))/Z(w) = - \phi'(w) p_\phi(w).
\end{align*}
For (a), we see that:
\begin{align*}
    \E_{w \sim p_\phi}[ \phi'(w) ] = \int \phi'(w) p_\phi(w) \rmd w = - \int p'_\phi(w) \rmd w = - \left[ p_\phi(w) \right]^{\infty}_{-\infty} = 0,
\end{align*}
where the last equality holds from the assumption that $\phi(w) \to \infty$ as $\abs{w} \to \infty$, and hence $p_\phi(\pm \infty) = 0$.
For (b) using integration by parts,
\begin{align*}
    \E_{w \sim p_\phi}[ (\phi'(w))^2 ] &= \int (\phi'(w))^2 p_\phi(w) \rmd w = - \int \phi'(w) p'_\phi(w) \rmd w \\
    &= \int \phi''(w) p_\phi(w) \rmd w - \left[ p_\phi \phi' \right]^\infty_{-\infty} 
    = \E_{w \sim p_\phi}[ \phi''(w) ],
\end{align*}
where the last equality holds by the
limiting behavior $\lim_{\abs{x} \to \infty} \abs{\phi'(x)} \exp(-\phi(x)) = 0$.
\end{proof}

\paragraph{Covering number bound.} 

Let $\bm{\phi}'(x) := (\phi'(x_1), \dots \phi'(x_d))$ and
$\bm{\phi}''(x) := (\phi''(x_1), \dots, \phi''(x_d))$.
Using this notation, we compute the gradient and Hessian of the log probability as:
\begin{align*}
    \nabla_\theta \log p_\theta(z_{t+1} \mid z_t) &= M(z_t)^\T \bm{\phi}'(z_{t+1} - M(z_t) \theta), \\
    \nabla^2_\theta \log p_\theta(z_{t+1} \mid z_t) &= - M(z_t)^\T \diag( \bm{\phi}''(z_{t+1} - M(z_t)\theta) ) M(z_t).
\end{align*}
Consequently, we see that
\begin{align*}
    \calI(\theta) &= -\sum_{t=1}^{T-1} \E_{p_\theta}[ \nabla^2_\theta \log p_\theta(z_{t+1} \mid z_t) ] \\
    &= \sum_{t=1}^{T-1} \E_{p_\theta}[ M(z_t)^\T \diag( \bm{\phi}''(w_t)) M(z_t)  ] = \frac{1}{\sigma^2_\phi} \sum_{t=1}^{T-1} \E_{p_\theta}[ M(z_t)^\T M(z_t) ],
\end{align*}
where the last equality utilizes the IBP identity
$\E_{w \sim p_\phi}[ \phi''(w) ] = \E_{w \sim p_\phi}[ (\phi'(w))^2 ]$ from
\Cref{prop:phi_properties}.
Furthermore, we can construct a uniform bound $\calI_{\max}$ using the definition of $\bar{\mu}$:
\begin{align*}
    \calI(\theta) \preccurlyeq \calI_{\max} := T \bar{\mu} \cdot I_p, \quad \theta \in \Theta.
\end{align*}
Therefore we have an upper bound for the metric entropy under the max-FI divergence:
\begin{align*}
    \log \calN_\mathrm{\calI_{\max}}(\calP, \e) \leq \log \calN_{\norm{\cdot}}\left( \Theta, \e \sqrt{\frac{1}{T \bar{\mu}}}  \right) \leq p \log \left( \frac{3R}{\e} \sqrt{T\bar{\mu}} \right).
\end{align*}
From \Cref{thm:hellinger_bound_MLE}, with probability at least $1-\delta$,
\begin{equation}
    \HgSq{\hat{\theta}^\e_{m,T}}{\theta_\star} \lesssim \frac{ p \log(c_1 R \bar{\mu} \cdot mT/\delta ) }{m}. \label{eq:hellinger_bound_simple_generalized_regression}
\end{equation}
Hence by the arguments outlined in \eqref{eq:framework_hellinger_conv_hull_bound_general} combined with \Cref{prop:log_dominance}, as long as
\begin{align}
    m \gtrsim \gamma_1^{-2} p \log(c_1' p R \bar{\mu} \cdot T/(\gamma_1\delta) ), \label{eq:m_req_ident_generalized_regression}
\end{align}
then whenever \eqref{eq:hellinger_bound_simple_generalized_regression} holds, we have
\begin{align}
    \sup_{s \in [0, 1]} \HgSq{ (1-s)\theta_\star + s \hat{\theta}^\e_{m,T}}{\theta_\star} \lesssim \frac{ \gamma_2^2 T \bar{\mu} p \log(c_1 R \bar{\mu} \cdot mT/\delta )}{m}. \label{eq:hellinger_bound_generalized_regression}
\end{align}
Call this event $\calE_1$.
Furthermore, since the log-likelihood of $z_{1:T}$ for $\theta \in \Theta$ can be written as:
\begin{align*}
    \log p_\theta(z_{1:T}) &= \sum_{t=1}^{T-1} \log p_{\bm{\phi}}(z_{t+1} - M(z_t) \theta) + \mathrm{const} \\
    &= \sum_{t=1}^{T-1} \sum_{j=1}^{d} \log p_\phi(\ip{z_{t+1} - M(z_t) \theta}{e_j}) + \mathrm{const} \\
    &= -  \sum_{t=1}^{T-1} \sum_{j=1}^{d} \phi(\ip{z_{t+1} - M(z_t) \theta}{e_j}) + \mathrm{const},
\end{align*}
where $\mathrm{const}$ does not depend on $\theta$, when $\phi$ is convex, then $\calP$ is log-concave (cf.~\Cref{def:log_concave}).
It is not hard to see that $\mathrm{diam}(\Theta) = 2RT \bar{\mu}$.
Hence from \Cref{thm:hellinger_bound_MLE}, with probability at least $1-\delta$,
\begin{align}
    \sup_{s \in [0, 1]} \HgSq{ (1-s)\theta_\star + s \hat{\theta}^\e_{m,T}}{\theta_\star} \lesssim \frac{ p \log(c_2 R \bar{\mu} \cdot mT/\delta ) }{m}. \label{eq:hellinger_bound_generalized_regression_log_concave}
\end{align}
Call this event $\calE_{1,\mathrm{cvx}}$.

\paragraph{Estimate $B_1$ and $B_2$.}

We first focus on $B_1$.
Let us fix a test vector $v \in \R^p$, and define
$d_t := v^\T  M^\T(z_t) \bm{\phi}'(w_t)$, so that for $z_{1:T} \sim p_\theta$,
\begin{align*}
    \ip{v}{\nabla_\theta \log p_\theta(z_{1:T})} = \sum_{t=1}^{T-1} d_t.
\end{align*}
Our first observation is that, with $\calF_t := \sigma(z_{1:t+1})$,
we have $\E[ d_t \mid \calF_{t-1} ] = \E[ v^\T M^\T(z_t) \bm{\phi}'(w_t) \mid \calF_{t-1} ] = 0$
by \Cref{prop:phi_properties}, and hence $(d_t)_{t \geq 1}$ is a MDS adapted to the filtration $(\calF_t)_{t \geq 1}$.
Next, we compute:
\begin{align*}
    \E[ d_t^2 \mid \calF_{t-1} ] &= \E[ v^\T M^\T(z_t) \bm{\phi}'(w_t) \bm{\phi}'(w_t)^\T M(z_t) v \mid \calF_{t-1} ] = v^\T M^\T(z_t) \E_{w \sim p_{\bm{\phi}}}[ \bm{\phi}'(w) \bm{\phi}'(w)^\T ] M(z_t) v.
\end{align*}
Now since $\E_{w \sim p_{\bm{\phi}}}[\phi'(w_j) \phi'(w_k)] = ( \E_{w \sim p_\phi}[ \phi'(w) ]  )^2 = 0$ for $j,k \in [d]$ with $j \neq k$ by coordinate-wise independence of $p_{\bm{\phi}}$ and \Cref{prop:phi_properties},
we have that
\begin{align*}
    \E_{w \sim p_{\bm{\phi}}}[ \bm{\phi}'(w) \bm{\phi}'(w)^\T ] = \E_{w_1 \sim p_{\phi}}[ (\phi'(w_1))^2 ] \cdot I_d = \sigma^{-2}_\phi \cdot I_d.
\end{align*}
Hence,
\begin{align*}
    \E\left( \sum_{t=1}^{T-1} \E[ d_t^2 \mid \calF_{t-1} ] \right)^2 &= \sigma^{-4}_\phi \E\left( \sum_{t=1}^{T-1}  \norm{M(z_t) v}^2  \right)^2 
    \leq (T-1) \sigma^{-4}_\phi \sum_{t=1}^{T-1} \E\norm{M(z_t) v}^4 \\
    &\leq (T-1) \sigma^{-4}_\phi \norm{v}^4 \sum_{t=1}^{T-1} \E\opnorm{M(z_t)}^4 
    \leq (T-1)^2 \sigma^{-4}_\phi \norm{v}^4 M_1^4,
\end{align*}
Now let us focus on $\E[ d_t^4 ]$:
\begin{align*}
    \E[ d_t^4 ] &= \E[ (v^\T M^\T(z_t) \bm{\phi}'(w_t) \bm{\phi}'(w_t)^\T M(z_t) v)^2 ] \\
    &\leq \E[ \norm{\bm{\phi}'(w_t)}^4 (v^\T M^\T(z_t) M(z_t) v)^2 ]\\
    &\leq \sqrt{ \E_{w \sim p_{\bm{\phi}}}[ \norm{\bm{\phi}'(w)}^8 ] } \sqrt{ \E[ (v^\T M^\T(z_t) M(z_t) v)^4 ] }.
\end{align*}
Next, we control
\begin{align*}
    \E_{w \sim p_{\bm{\phi}}}[\norm{\bm{\phi}'(w)}^8] &= \E_{w \sim p_{\bm{\phi}}}\left( \sum_{j=1}^{d} (\phi'(w_j))^2 \right)^4 \leq d^3  \E_{w \sim p_{\bm{\phi}}}\left( \sum_{j=1}^{d} (\phi'(w_j))^8 \right) \\
    &= d^4 \E_{w_1 \sim p_{\phi}}[( \phi'(w_1))^8] 
    \leq d^4 \beta_2 \sigma^{-8}_\phi.
\end{align*}
where the penultimate inequality follows from H{\"o}lder's inequality, and the last inequality follows from \Cref{def:phi_regularity}. Hence,
\begin{align*}
    \sum_{t=1}^{T-1} \E[d_t^4] &\leq d^2 \beta_2^{1/2} \sigma^{-4}_\phi \sum_{t=1}^{T-1} \sqrt{ \E[ (v^\T M^\T(z_t) M(z_t) v)^4 ] } \\
    &\leq d^2 \beta_2^{1/2} \sigma^{-4}_\phi \norm{v}^4 \sum_{t=1}^{T-1} \sqrt{ \E\opnorm{M(z_t)}^8 } \\
    &\leq (T-1)  d^2 \beta_2^{1/2} \sigma^{-4}_\phi \norm{v}^4 \sqrt{ \frac{1}{T-1} \sum_{t=1}^{T-1} \E \opnorm{M(z_t)}^8 } \\
    &\leq (T-1)  d^2 \beta_2^{1/2} \sigma^{-4}_\phi \norm{v}^4 M_2^4.
\end{align*}
Now we will set $v = \calI(\theta)^{-1/2} \bar{v}$ where $\bar{v} \in \bbS^{p-1}$ is a unit test vector;
hence $\norm{v} \leq \sqrt{\sigma^2_\phi/(\underline{\mu} T)}$.
By Rosenthal's inequality for MDS (\Cref{thm:rosenthal}), we have:
\begin{align*}
    \left(\E \left(\sum_{t=1}^{T-1} d_t\right)^4 \right)^{1/4} &\lesssim \left( \E\left( \sum_{t=1}^{T-1} \E[ d_t^2 \mid \calF_{t-1} ] \right)^2 \right)^{1/4} + \left( \sum_{t=1}^{T-1} \E[d_t^4] \right)^{1/4} \\
    &\lesssim \sqrt{T} \sigma^{-1}_\phi \norm{v} M_1 + T^{1/4} d^{1/2} \beta_2^{1/8} \sigma^{-1}_\phi \norm{v} M_2 \\
    &\lesssim M_1/\sqrt{\underline{\mu}} + T^{-1/4} d^{1/2} \beta_2^{1/8} M_2/\sqrt{\underline{\mu}} \\
    &\lesssim M_1/\sqrt{\underline{\mu}},
\end{align*}
where the last inequality holds from
Assumption (f).
Hence we have
\begin{align*}
    \sup_{\theta_1,\theta_2 \in \Theta} B_1(\theta_1, \theta_2) \lesssim M_1/\sqrt{\underline{\mu}}.
\end{align*}
We next focus on $B_2$.
We first fix a vector $q \in \R^d$,
and observe that
\begin{align*}
    \E_{w \sim p_{\bm{\phi}}}[ (q^\T \diag(\bm{\phi}''(w)) q)^2 ] &= \sum_{i,j=1}^{d} q_i^2 q_j^2 \E_{w \sim p_{\bm{\phi}}}[ \phi''(w_i) \phi''(w_j) ] \\
    &= \sum_{i=1}^{d} q_i^4 \E_{w \sim p_\phi}[ (\phi''(w))^2 ] + \sum_{i \neq j}^{d} q_i^2 q_j^2 (\E_{w \sim p_\phi}[ \phi''(w) ])^2 
    \leq \beta_1 \sigma_\phi^{-4} \norm{q}^4 ,
\end{align*}
where the last inequality holds from 
\Cref{def:phi_regularity}
and the IBP identity (cf.~\Cref{prop:phi_properties})
$\E_{w \sim p_\theta}[ \phi''(w) ] = \E_{w \sim p_\theta}[ (\phi'(w))^2 ] = \sigma_\phi^{-2}$.
Hence
fixing a test vector $v \in \R^p$, we have
\begin{align*}
    \E[ (v^\T \nabla_\theta^2 \log p_\theta(z_{t+1} \mid z_t) v)^2 ] &= \E[ ( v^\T M^\T(z_t) \diag(\bm{\phi}''(w_t)) M(z_t) v )^2 ] \\
    &= \E[\E[ ( v^\T M^\T(z_t) \diag(\bm{\phi}''(w_t)) M(z_t) v )^2 \mid z_t ]] \\
    &\leq \beta_1 \sigma_\phi^{-4} \E \norm{ M(z_t) v}^4 \leq \beta_1 \sigma_\phi^{-4} \norm{v}^4 \E\opnorm{M(z_t)}^4.
\end{align*}
Hence,
\begin{align*}
    \E\left( \sum_{t=1}^{T-1} v^\T \nabla_\theta^2 \log p_\theta(z_{t+1} \mid z_t) v \right)^2 &\leq (T-1)^2 \left[ \frac{1}{T-1}\sum_{t=1}^{T-1} \E(v^\T \nabla_\theta^2 \log p_\theta(z_{t+1} \mid z_t) v)^2 \right] \\
    &\leq (T-1)^2 \beta_1 \sigma^{-4}_\phi \norm{v}^4 \left[ \frac{1}{T-1}\sum_{t=1}^{T-1} \E\opnorm{M(z_t)}^4 \right] \\
    &\leq (T-1)^2 \beta_1 \sigma^{-4}_\phi \norm{v}^4 M_1^4.
\end{align*}
Now again we choose $v = \calI(\theta)^{-1/2} \bar{v}$ for a unit norm $\bar{v} \in \R^p$.
We then have 
\begin{align*}
    \left( \E\left( \sum_{t=1}^{T-1} v^\T \nabla_\theta^2 \log p_\theta(z_{t+1} \mid z_t) v \right)^2 \right)^{1/2} \lesssim T \beta_1^{1/2} \sigma^{-2}_\phi \norm{v}^2 M_1^2 \leq \beta_1^{1/2} M_1^2/\underline{\mu}.
\end{align*}
Hence we have
\begin{align*}
    \sup_{\theta_1,\theta_2 \in \Theta} B_2(\theta_1, \theta_2) \lesssim \beta_1^{1/2} \frac{M_{1}^{2}}{\underline{\mu}}.
\end{align*}
Altogether, we can bound
\begin{align}
    \sup_{\theta_1,\theta_2 \in \Theta} \max\{ B_1^2(\theta_1, \theta_2), B_2(\theta_1, \theta_2) \} \lesssim 
    \frac{ \beta_1^{1/2} M_1^2}{\underline{\mu}}. \label{eq:B1_B2_generalized_regression}
\end{align}

\paragraph{Parameter error bound.}
We first cover the case where $\phi$ is not convex. 
Combining 
\eqref{eq:hellinger_bound_generalized_regression}, 
\eqref{eq:B1_B2_generalized_regression}, and
\Cref{prop:log_dominance}, as long as $m$ satisfies
\eqref{eq:m_req_ident_generalized_regression} and also
\begin{align}
    m \gtrsim pT\gamma_2^2 \cdot \beta_1 M_1^4 \kappa/\underline{\mu} \cdot \log( c_1'' p/\delta \cdot \beta_1 R M_1 T \max\{\gamma_2,1\} \kappa ), \label{eq:m_req_radius_generalized_regression}
\end{align}
then condition \eqref{eq:hellinger_perturbation_radius}
holds on $\calE_1$.
Hence from \Cref{prop:hellinger_perturbation}, 
combining \eqref{eq:hellinger_perturbation_avg_FI} with
\eqref{eq:hellinger_bound_generalized_regression}, we have on $\calE_1$:
\begin{align*}
    \norm{\hat{\theta}^\e_{m,T} - \theta_\star}^2 \lesssim  \frac{ p \log(c_1 R \bar{\mu} \cdot mT/\delta ) }{\underline{\mu} mT}. 
\end{align*}
We now turn to the case where $\phi$ is convex.
Combining \eqref{eq:hellinger_bound_generalized_regression_log_concave}, \eqref{eq:B1_B2_generalized_regression},
and \Cref{prop:log_dominance}, we see that if $m$ satisfies:
\begin{align}
    m \gtrsim p \cdot \beta_1 M_1^4/\underline{\mu}^2 \cdot \log(c_2'p/\delta \cdot  \beta_1 R M_1 T \kappa ), \label{eq:m_req_radius_generalized_regression_log_concave}
\end{align}
Hence from \Cref{prop:hellinger_perturbation},
combining \eqref{eq:hellinger_perturbation_avg_FI} with
\eqref{eq:hellinger_bound_generalized_regression_log_concave}, we have on $\calE_{1,\mathrm{cvx}}$,
\begin{align*}
    \norm{\hat{\theta}^\e_{m,T} - \theta_\star}^2 \lesssim  \frac{ p \log(c_2 R \bar{\mu} \cdot mT/\delta ) }{\underline{\mu} mT}.
\end{align*}

\paragraph{Verify FI radius.}
In order to verify the FI radius condition \eqref{eq:hellinger_perturbation_FI_radius}, we
will make use of \Cref{prop:hellinger_change_of_measure}.
Fix $\theta_1, \theta_2 \in \Theta$ and a unit norm $v \in \bbS^{p-1}$.
We have the following:
\begin{align*}
    \abs{ v^\T(\calI(\theta_1) - \calI(\theta_2))v } &= \frac{1}{\sigma^2_\phi} \bigabs{ \E_{p_{\theta_1}}\left[ \sum_{t=1}^{T-1} \norm{M(z_t) v}^2 \right] -\E_{p_{\theta_2}}\left[ \sum_{t=1}^{T-1} \norm{M(z_t) v}^2 \right] } \\
    &\stackrel{(a)}{\leq} \frac{\sqrt{2}}{\sigma^2_\phi} \left( \bignorm{  \sum_{t=1}^{T-1} \norm{M(z_t) v}^2 }_{\calL^2(p_{\theta_1})} + \bignorm{  \sum_{t=1}^{T-1} \norm{M(z_t) v}^2 }_{\calL^2(p_{\theta_2})}   \right) \Hg{p_{\theta_1}}{p_{\theta_2}} \\
    &\stackrel{(b)}{\leq} \frac{2\sqrt{2} (T-1) M_1^2}{\sigma^2_\phi}\Hg{p_{\theta_1}}{p_{\theta_2}},
\end{align*}
where (a) follows from \Cref{prop:hellinger_change_of_measure}
and (b) follows from Jensen's inequality.
Hence by the variational characterization of operator norm,
for any $\theta \in \Theta$:
\begin{align*}
   \frac{\opnorm{ \calI(\theta) - \calI(\theta_\star) }}{\lambda_{\min}(\calI(\theta_\star))} \lesssim \frac{M_1^2}{\underline{\mu} \sigma_\phi^2 }\Hg{\theta}{\theta_\star}.
\end{align*}
Hence from \eqref{eq:hellinger_bound_generalized_regression} 
and \Cref{prop:log_dominance},
as long as $m$ satisfies \eqref{eq:m_req_ident_generalized_regression}, 
\eqref{eq:m_req_radius_generalized_regression}, and 
\begin{align*}
    m \gtrsim pT \gamma_2^2 \cdot M_1^4 \kappa / (\underline{\mu}\sigma_\phi^4) \cdot \log( c_1''' p/\delta \cdot R M_1 T \kappa \max\{\sigma_\phi^{-1}, 1\} \max\{\gamma_2,1\} ), \label{eq:m_req_FI_radius_generalized_regression}
\end{align*}
then the FI radius condition \eqref{eq:hellinger_perturbation_FI_radius} holds on $\calE_1$.
On the other hand when $\phi$ is convex, 
from \eqref{eq:hellinger_bound_generalized_regression_log_concave}
and \Cref{prop:log_dominance},
as long as $m$ satisfies \eqref{eq:m_req_radius_generalized_regression_log_concave} and:
\begin{align}
    m \gtrsim p \cdot M_1^4/(\underline{\mu}^2\sigma_\phi^4) \cdot \log(c_2'' p/\delta \cdot R M_1 T \kappa \max\{\sigma_\phi^{-1}, 1\} ), 
\end{align}
then the FI radius condition \eqref{eq:hellinger_perturbation_FI_radius} holds on $\calE_{1,\mathrm{cvx}}$.
The result for both cases now follows from \Cref{prop:hellinger_perturbation}, specifically \eqref{eq:hellinger_perturbation}.

\subsubsection{Proof of Regularity Conditions for Example Distributions}
\label{sec:l1_regression:regularity_condition_proofs}

\begin{proof}[Proof for smoothed bang-bang noise (\Cref{ex:smoothed_bang_bang_noise}).]
We abbreviate $p_{\nu} = p_{\phi_\nu}$.
We have that $\phi'_{\nu}(x) = \frac{x}{\nu^2} - \frac{1}{\nu^2}\tanh(x/\nu^2)$
and $\E_{x \sim p_\nu}[ x^2 ] = 1 + \nu^2$.
For any $\e \in (0, 1)$, we have by Young's inequality
\begin{align*}
    (x - \tanh(x/\nu^2))^2 \geq (1-\e) x^2 + (1-1/\e) \tanh^2(x/\nu^2) \geq (1-\e) x^2 - (1/\e - 1).
\end{align*}
Hence
\begin{align*}
    \E_{x \sim p_\nu}[(x - \tanh(x/\nu^2))^2] \geq (1-\e) (1+\nu^2) - (1/\e - 1) =: \varphi_\nu(\e).
\end{align*}
Basic calculus yields
\begin{align*}
    \max_{\e \in (0, 1)} \varphi_\nu(\e) = (\sqrt{1+\nu^2} - 1)^2 \,\,\textrm{at}\,\, \e = 1/\sqrt{1+\nu^2}.
\end{align*}
Hence we have show that
\begin{align*}
    \sigma^{-2} = \E_{x \sim p_\nu}[ (\phi_\nu'(x))^2 ] = \frac{1}{\nu^2} \E_{x \sim p_\nu}[(x - \tanh(x/\nu^2))^2] \geq \frac{(\sqrt{1+\nu^2}-1)^2}{\nu^2}.
\end{align*}
Furthermore, imposing the restriction that $\nu \in (0, 1)$, a second order Taylor expansion around $\nu=0$ yields that
$\sqrt{1+\nu^2} - 1 \geq \frac{\nu^2}{2^{3/2}}$
and hence $\sigma^{-2} \geq \nu^2/8$.
Next, we compute
$\phi''_{\nu}(x) = \frac{1}{\nu^2} - \frac{1}{\nu^4} \sech^2(x/\nu^2)$,
and hence
\begin{align*}
    \abs{\phi''_\nu(x)} \leq \max\{ 1/\nu^2, 1/\nu^4 \} = 1/\nu^4 = (\sigma^2/\nu^4)/\sigma^2 \leq (8/\nu^6) / \sigma^2.
\end{align*}
Hence we can set $\beta_1 = 8/\nu^6$.
Next, we have
\begin{align*}
    \E_{x \sim p_\nu}[(\phi_\nu'(x))^8] \leq 128 \E_{x \sim p_\nu}\left[ \frac{x^8}{\nu^{16}} + \frac{1}{\nu^{16}} \right].
\end{align*}
For each mixture index $i \in \{1,2\}$, we have
\begin{align*}
    \E[ x^8 \mid i ] \leq 7^4 (\E[x^2 \mid i])^4 = 7^4 ( 1 + \nu^2 )^4 \leq 8 \cdot 7^4 (1 + \nu^8).
\end{align*}
Consequently,
\begin{align*}
    \E_{x \sim p_\nu}[(\phi_\nu'(x))^8] \lesssim \frac{1}{\nu^{16}} = \frac{\sigma^8}{\nu^{16}} \cdot \frac{1}{\sigma^8} \lesssim \frac{1}{\nu^{24}}.
\end{align*}
Hence we can set $\beta_2 \lesssim 1/\nu^{24}$.
\end{proof}

\begin{proof}[Proof for smoothed Laplace noise (\Cref{ex:smoothed_laplace}).]
The first and second derivatives of $\phi_{c,\nu}$ are:
\begin{align*}
    \phi'_{c,\nu}(x) = \frac{1}{\nu} \tanh(cx/\nu) \in [-1/\nu, 1/\nu], \quad
    \phi''_{c,\nu}(x) = \frac{c}{\nu^2} \sech^2(cx/\nu) \in [0, c/\nu^2].
\end{align*}
Define $Z(c, \nu) := \int \cosh(cx/\nu)^{-1/c} \rmd x$.
By a change of variables, $Z(c, \nu) = \nu Z(c)$.
Also for $t \in (0, 1/\nu)$,
\begin{align*}
    \Pr_{x \sim p_\phi}( \abs{\phi'_{c,\nu}(x)} \leq t \} 
    &= \Pr_{x \sim p_\phi}( x \in [ - \nu\tanh^{-1}(t\nu)/c, \nu\tanh^{-1}(t\nu)/c ]) =: \Pr_{x \sim p_\phi}(x \in I_{c,\nu}(t)).
\end{align*}
We control the RHS probability by:
\begin{align*}
    \Pr_{x \sim p_\phi}( x \in I_{c,\nu}(t) ) = \frac{1}{Z(c,\nu)} \int_{I_{c,\nu}(t)} \exp(-c^{-1}\log\cosh(cx/\nu)) \rmd x \leq \frac{\abs{I_{c,\nu}(t)}}{Z(c,\nu)} = \frac{2 \tanh^{-1}(t\nu)}{c Z (c)}.
\end{align*}
Choosing $t'$
such that $\frac{2 \tanh^{-1}(t'\nu)}{c Z (c)} = 1/2$,
i.e., $t' = \nu^{-1}\tanh( cZ(c)/4 )$:
\begin{align*}
    \sigma^{-2} = \E_{x \sim p_\phi}[ (\phi'_{c,\nu}(x))^2 ] 
    \geq (t')^2 \Pr_{x \sim p_\phi}( \abs{\phi'_{c,\nu}(x)} \geq t' ) 
    \geq \nu^{-2} \tanh^2(c Z(c)/4) / 2.
\end{align*}
Hence we have
\begin{align*}
    \sigma^2 \in \left[ \nu^2, \frac{2\nu^2}{\tanh^2(cZ(c)/4)} \right].
\end{align*}
Now let us focus on controlling $\beta_1$.
We have:
\begin{align*}
    \abs{\phi''_{c,\nu}(x)} \leq \frac{c}{\nu^2} = \frac{c \sigma^2}{\nu^2} \cdot \frac{1}{\sigma^2} \leq \frac{2c}{\tanh^2(cZ(c)/4)} \cdot \frac{1}{\sigma^2}.
\end{align*}
Hence we can take $\beta_1 = \frac{2c}{\tanh^2(cZ(c)/4)}$.
We next focus on controlling $\beta_2$.
We have:
\begin{align*}
    \E_{w \sim p_\phi}[ (\phi'(w))^8 ] \leq \frac{1}{\nu^8} = \frac{\sigma^8}{\nu^8} \cdot \frac{1}{\sigma^8} \leq \frac{16}{\tanh^8(cZ(c)/4)} \cdot \frac{1}{\sigma^8}.
\end{align*}
Hence we can take $\beta_2 = \frac{16}{\tanh^8(cZ(c)/4)}$.
\end{proof}

\subsection{Non-Monotonic Sinusoidal GLM Dynamics}
\label{sec:case_studies:glm}

For our next setup, we consider the following generalized linear model (GLM) of dynamics, given parameters $A \in \R^{d \times d}$,
\begin{align}
    z_{t+1} = \sin(A z_t) + w_t, \quad z_0 = 0, \quad w_t \sim \sfN(0, \sigma^2 I_d), \label{eq:GLM_dynamics}
\end{align}
where $\sin(\cdot)$ is overloaded to apply component-wise given a vector input, and $w_t$ is drawn independently across time.
While more general GLM dynamics $z_{t+1} = \phi(A z_t) + w_t$ have been studied in the literature in the context of system identification~\cite{foster2020nonlinear,kowshik2021nonlinear,sattar2022nonlinear,ziemann2022learning, ziemann2022singletrajnonlin}, the specific sinusoidal 
GLM we consider is more challenging as it is an instance of a \emph{non-monotonic}, \emph{non-expansive}\footnote{
An activation function $\phi(x)$ is \emph{expansive}
if there exists a $\zeta > 0$ such that $\abs{ \phi(x) - \phi(y) } \geq \zeta \abs{x - y}$ for all $x, y \in \R$.} activation function.
Furthermore, we do not impose any stability assumptions on the $A$ matrix in \eqref{eq:GLM_dynamics}, as is done in prior works. The following theorem is our main result for parameter recovery in this model.
In the following result, we let $\hat{\theta}^{\e}_{m,T} = \vec(\hat{A}^{\e}_{m,T})$ and
$\theta_\star = \vec(A_\star)$.

\begin{mythm}
\label{thm:sin_GLM}
Fix $\delta \in (0, 1)$. Consider the max FI discretized MLE at resolution $\e = \delta/(2\sqrt{2m})$
over the set $\Theta = \{ A \in \R^{d \times d} \mid \norm{A}_F \leq R \}$ for $R \geq 1$.
Put $A_{\star,\min} := \min_{j \in [d]} \norm{A_{\star}[j]}$, where $A_{\star}[j] \in \R^d$ denotes the $j$-th row of $A_\star$, and suppose $A_{\star,\min} > 0$.
Suppose also that $T \gtrsim d^2$.
There exists constants $\Phi_i$, $i \in \{1,2,3\}$, which scale as $\mathrm{poly}(\sigma, 1/\sigma, 1/A_{\star,\min}, 1/d)$, such that 
if $m$ satisfies for universal positive constants $c_0,c_1,c_2$:
\begin{align*}
    m \gtrsim \max\left\{ \Phi_1 d^2 \log\left( \frac{c_1 \Phi_1 \Xi}{\delta} \right), 
    \Phi_2 d^{10}T \log\left( \frac{c_2 \Phi_2 \Xi}{\delta} \right), 
    \Phi_3 d^{11}T \log\left( \frac{c_3 \Phi_3 \Xi}{\delta} \right)\right\}, \quad \Xi := \frac{dRT(d+\sigma^2)}{\sigma^2},
\end{align*}
then with probability at least $1-\delta$ over $\calD_{m,T}$, 
\begin{align*}
        \norm{\hat{\theta}^\e_{m,T} - \theta_\star}^2_{\bar{\calI}(\theta_\star)} \lesssim \frac{d^2}{mT} \log\left( \frac{c_0 R mT(d+\sigma^2)}{\sigma^2 \delta} \right).
\end{align*}
The precise expressions for $\Phi_i$ are given in the proof.
\end{mythm}
Note that
\begin{align*}
    \norm{\hat{\theta}^\e_{m,T} - \theta_\star}^2_{\bar{\calI}(\theta_\star)} = \frac{1}{\sigma^2(T-1)} \sum_{t=1}^{T-1} \E_{p_{\theta_\star}}[ \diag(\cos^2(A_\star z_t)) (\hat{A}^\e_{m,T} - A_\star) z_tz_t^\T   (\hat{A}^\e_{m,T} - A_\star)^\T ],
\end{align*}
where we emphasize that the expression on the RHS is over a \emph{fresh} trajectory $z_{1:T} \sim p_{\theta_\star}$ that is independent of $\calD_{m,T}$. Nevertheless, there is not a simple closed-form reduction and hence we leave it in its present form.
If we treat $\sigma$, $R$, and $A_{\star,\min}$ as constants, then
\Cref{thm:sin_GLM} states that 
whenever both $m \gtrsim \tilde{\Omega}(d^{11} T)$ and $T \gtrsim d^2$, then the nearly (up to logarithmic factors) instance-optimal rate
$\norm{\hat{\theta}^\e_{m,T} - \theta_\star}^2_{\bar{\calI}(\theta_\star)} \lesssim \tilde{O}(d^2/(mT))$ holds with high probability.
We suspect that the $m \gtrsim \tilde{\Omega}(d^{11})$ requirement is sub-optimal and can be further improved with a more refined analysis. On the other hand, the requirement on $T \gtrsim d^2$ is made
to simplify the expressions in the proof and can be removed.
Furthermore, in our proof we show that $\lambda_{\min}( \bar{\calI}(\theta_\star) ) \gtrsim 1/d^2$, which implies the parameter bound
$\norm{ \hat{\theta}^\e_{m,T} - \theta_\star }^2 \lesssim \tilde{O}( d^4/(mT) )$.
We leave to future work a sharp analysis of $\lambda_{\min}( \bar{\calI}(\theta_\star) )$ to determine the optimal \emph{un-weighted} parameter error bound.

\paragraph{Comparison to existing results.}
To the best of our knowledge, this is the first rate for parameter recovery 
in any GLM dynamics model in the multi-trajectory setting, which obtains a nearly instance-optimal rate of $\tilde{O}(d^2/(mT))$. %
The previous sharpest rate for this problem utilizes the fact that
the MLE $\hat{\theta}_{m,T}$ for this problem involves solving a realizable, parametric least-squares ERM problem, and hence the reduction described in \Cref{sec:general_framework:MLE}
to the result of \cite{ziemann2024sharp} would provide a similar bound on the \emph{excess risk},
but not the weighted parameter error directly; as described in \Cref{sec:general_framework:MLE}, however, without verification of the weakly sub-Gaussian condition specifically for the function class $\{ \sin(A_1 x) - \sin(A_2 x) \mid A_1, A_2 \in \Theta \}$ (which to the best of our knowledge has not been shown in the literature), the best burn-in requirement on $m, T$ that this reduction can provide depends exponentially on the process dimension $d$. On the other hand, \Cref{thm:sin_GLM} requires that both $m \gtrsim \mathrm{poly}(d) \cdot T$ and $T \gtrsim \mathrm{poly}(d)$ requirement; while it is likely that our exact polynomial dependence is not optimal, we are able to break the exponential in $d$ barrier of existing results.

In the single-trajectory setting, several recent works have explicitly studied 
parameter recovery for GLM dynamics~\cite{foster2020nonlinear, ziemann2022learning, ziemann2022singletrajnonlin, kowshik2021nonlinear, sattar2022nonlinear}. However, the specific setting \eqref{eq:GLM_dynamics} we consider does not satisfy the requisite assumptions
for any of these works, and hence these works cannot be used as a basis for reduction.
To start, the works \cite{foster2020nonlinear, ziemann2022learning, kowshik2021nonlinear} all assume a model class with a 1-Lipschitz monotonic activation function $\phi$, %
with fast rates further requiring $\phi$ to be expansive.
The $\sin$ function only satisfies the $1$-Lipschitz requirement; the bounded and oscillatory nature of $\sin$ violates the other assumptions.  
The work \cite{sattar2022nonlinear} requires a one-point convexity assumption 
on the population loss, which is challenging to verify; they are only able to 
verify their condition assuming uniformly monotonic activations (i.e., $\phi' \geq \zeta > 0$).
Regarding stability of $A_\star$,
\cite{foster2020nonlinear, ziemann2022learning} additionally assume Lyapunov stability conditions, specifically that there exists a diagonal positive definite $K$ and scalar $\rho < 1$ such that $A_\star^\top K A_\star \preceq \rho \cdot K$; however, it is immediately obvious that this does not hold in our setting as we permit $A_\star = c \cdot I_d$ for $c > 1$, which would require $\rho \geq c^2  > 1$.
This also violates the explicit assumption made in some works that $\opnorm{A_\star} < 1$~\cite{ziemann2022singletrajnonlin}.
Other works such as \cite{sattar2022nonlinear, kowshik2021nonlinear} made explicit exponential regularity assumptions on the trajectories that given a noise sequence $\{ w_t \}_{t=1}^{T-1}$, the difference in states from two initial states will expand at most by a factor of $\rho = 1 + O(1/T)$ every timestep; however, this is also violated in our setting.\footnote{Concretely, our
setup does not satisfy \cite[Assumption 4]{kowshik2021nonlinear} for any $\rho \leq 1 + O(1/T)$, as we now show.
Let $\Phi_t(z)$ denote the value of $z_t$ following the dynamics $z_{t+1} = \sin(2 z_t)$ starting at $z_1 = z$.
Suppose there exists positive $c_1, \rho = 1 + c_2/T$ such
$\abs{ \Phi_t(z) - \Phi_t(z') } \leq c_1 \rho^t \abs{z - z'}$ for all $z,z' \in \R$ and $t \in \N$.
Clearly we have $\Phi_t(0) = 0$ for all $t$. 
Furthermore, one can show that $\lim_{t \to \infty} \Phi_t(z) = r^\star$, where $r^\star \approx 0.94775$ is the unique solution to $r = \sin(2 r)$ in $(0, 1]$,
for all $z \in (0, 1]$.
Now, let $T_0$ be such that 
$\abs{ \Phi_t(\bar{z}) - r^\star } \leq r^\star/2$ for all $t \geq T_0$,
where $\bar{z} := r^\star/(4 c_1 e^{c_2})$ (we can always take $c_1, c_2$ large enough so that $\bar{z} \in (0, 1)$).
Hence, for any $T \geq T_0$, we have 
$r^\star/2 \leq \abs{ \Phi_T(\bar{z}) } \leq c_1 \rho^T \abs{\bar{z}} \leq c_1 e^{c_2} \abs{\bar{z}} = r^\star/4$, a contradiction.
}

\paragraph{Hellinger identifiability for sinusoidal GLMs.}
Before we turn to applying the Hellinger localization framework to this problem, we 
discuss the main technical challenge: absent the strict monotonically increasing activation function assumption $\phi'(x) \geq \gamma > 0$, 
establishing both that (a) $\calI(\theta) \succcurlyeq \Omega(T) \cdot I_{d^2}$
and that (b) $\HgSq{\hat{p}^{\e}_{m,T}}{p_\star} \leq \gamma^2$ implies $\norm{ \hat{\theta}^{\e}_{m,T} - \theta_\star}^2 \lesssim \gamma^2$ (i.e., Hellinger identifiability (\Cref{def:identifiability})) becomes substantially more challenging.
A key step towards establishing identifiability is to show the bound
$\gamma^2 := \E_{z \sim \sfN(0, \sigma^2 I_d)}[ (\sin(\ip{u_1}{z}) - \sin(\ip{u_2}{z}))^2 ] \gtrsim \sigma^2 \norm{u_1-u_2}^2$ when $\gamma^2$ is sufficiently small. 
We believe this result, which is detailed in \Cref{sec:appendix:glm}, to be of independent interest, and may be helpful in e.g.,
analyzing neural networks with sinusoidal activation functions~\cite{sitzmann2020implicit}.
We remark that a similar bound is shown for ReLU activations in \cite[Lemma 11]{foster2020nonlinear}, in particular
for $z \sim \sfN(\mu, \sigma^2 I_d)$, $\E_z [ (\mathrm{ReLU}( \ip{u_1}{z} ) - \mathrm{ReLU}(\ip{u_2}{z}))^2] \geq \frac{\sigma^2}{4} e^{- \| \mu \|^2 / \sigma^2} \| u_1 - u_2 \|^2$.
A key difference is in how this style of result is used in our analysis versus
in \cite{foster2020nonlinear}.
In our analysis, the Hellinger identifiability is only used 
for the first two timestep as noted in \Cref{rmk:single_step_identifiability},
and hence we only need to consider taking expectation over $z \sim \sfN(0, \sigma^2 I_d)$.
On the other hand, \cite{foster2020nonlinear} proves identifiability at every timestep in order to relate parameter recovery error to average prediction error;
hence their analysis is required to consider non-zero means $\mu$'s, leading to 
parameter error rates that have an $e^d$ dependence on the dimension in general 
(cf.~\cite[Theorem 3]{foster2020nonlinear}).

\subsubsection{Proof of \Cref{thm:sin_GLM}}

We will let $\theta = \vec(A) \in \R^{d^2}$,
$\Theta = \{ \theta \in \R^{d^2} \mid \norm{\theta} \leq R \}$, 
and define the map $M(z) := (z^\T \otimes I_d)$ so that
$A z = M(z) \vec(A) = M(z) \theta$.
For what follows, we will often use $A$ and $\theta$ interchangeably, and similarly for $A_\star$ and $\theta_\star$, choosing whichever notation is more convenient.

\paragraph{Step 1: Covering number bound.}
Define $h_\theta(z) := \sin(M(z) \theta)$
and its Jacobian w.r.t.\ $\theta$ 
$D_\theta h_\theta(z) = \diag(\cos(M(z) \theta)) M(z)$ (similar to $\sin(\cdot)$, $\cos(\cdot)$ is also overloaded to apply component-wise given a vector input).
We observe that
\begin{align*}
    \nabla_\theta \log p_\theta(z_{t+1} \mid z_t) = - \frac{1}{\sigma^2} (D_\theta h_\theta(z_t))^\T (h_\theta(z_t) - z_{t+1}),
\end{align*}
and hence
\begin{align*}
    \calI(\theta) = \frac{1}{\sigma^2} \sum_{t=1}^{T-1} \E_{z_t \sim p_\theta(\cdot | z_{t-1})}[ (D_\theta h_\theta(z_t))^\T (D_\theta h_\theta(z_t)) ].
\end{align*}
We evaluate:
\begin{align*}
    \E[ (D_\theta h_\theta(z_t))^\T(D_\theta h_\theta(z_t)) ] &= \E[ M(z_t)^\T \diag(\cos^2(M(z_t) \theta)) M(z_t) ].
\end{align*}
Let us start with an upper bound.
Since $\cos(x)^2 \in [0, 1]$, we can see that $\diag(\cos^2(M(z_t) \theta)) \preccurlyeq I_d$.  This allows us to simplify the expression:
\begin{align*}
    \calI(\theta) &\preccurlyeq \frac{1}{\sigma^2} \sum_{t=1}^{T-1} \E_{z_t \sim p_\theta(\cdot \mid z_{t-1})}[ M(z_t)^\T M(z_t) ] \\
    &= \frac{1}{\sigma^2} \sum_{t=1}^{T-1} \E_{z_t \sim p_\theta(\cdot \mid z_{t-1})}[ (z_t^\T \otimes I_d)^\T (z_t^\T \otimes I_d) ]  \\
    &= \frac{1}{\sigma^2} \sum_{t=1}^{T-1} \E_{z_t \sim p_\theta(\cdot \mid z_{t-1})}[ z_t z_t^\T ] \otimes I_d.
\end{align*}
We now turn to analyzing $\E_{z_t \sim p_\theta(\cdot \mid z_{t-1})}[ z_t z_t^\T ]$.  Expanding $z_t = \mu_{t-1} + w_{t-1}$ for $\mu_{t-1} = h_\theta(z_{t-1})$ and observing that $\E[w_{t-1}] = 0$, we can see:
\begin{align*}
    \E_{z_t \sim p_\theta(\cdot \mid z_{t-1})}[ z_t z_t^\T ] &= \E_{w_{t-1}}[ (\mu_{t-1} + w_{t-1}) (\mu_{t-1} + w_{t-1})^\T ] \\
    &= \E_{w_{t-1}}[ \mu_{t-1} \mu_{t-1}^\T + \mu_{t-1} w_{t-1}^\T + w_{t-1} \mu_{t-1}^\T + w_{t-1} w_{t-1}^\T ] \\
    &= \mu_{t-1} \mu_{t-1}^\T + \sigma^2 I_d.
\end{align*}
We fix a $v \in \mathbb{R}^d$ with unit norm, observe $\sin(x)^2 \in [0, 1]$, and bound the outer product of $\mu_{t-1}$:
\begin{align*}
    v^\T \mu_{t-1} \mu_{t-1}^\T v &= (v ^\T \mu_{t-1})^2 \leq \norm{v} \norm{ \mu_{t-1} } \leq d.
\end{align*}
Putting this all together gives the maximum eigenvalue of the Fisher information:
\begin{align*}
    \lambda_{\max}(\calI(\theta)) &\leq \frac{T}{\sigma^2} (d + \sigma^2).
\end{align*}
We note that this is a parameter agnostic upper bound, and as such
we can set $\calI_{\max} = \frac{T}{\sigma^2} (d + \sigma^2) \cdot I_p$.
By \Cref{prop:local_hellinger_identifiability_sin_GLM} and the data processing inequality, 
for any $\theta \in \Theta$,
\begin{align*}
    \HgSq{\theta}{\theta_\star} \lesssim \min\left\{1, \frac{1}{\sigma^2} \right\} \Longrightarrow \norm{\theta - \theta_\star}^2 \lesssim d \max \left\{1, \frac{1}{\sigma^2} \right\} \HgSq{\theta}{\theta_\star}.
\end{align*}
This shows that $\calP$ is $(\gamma_1, \gamma_2)$-identifiable (cf.~\Cref{def:identifiability})
for constants:
\begin{align*}
    \gamma_1 \asymp \min\{1, 1/\sigma\}, \quad \gamma_2 \asymp \sqrt{d} \max\{1,1/\sigma\}.
\end{align*}
Hence from \eqref{eq:hellinger_error_via_identifiability},
\begin{align}
    \HgSq{\hat{\theta}^\e_{m,T}}{\theta_\star} \lesssim \min\left\{1, \frac{1}{\sigma^2} \right\} \Longrightarrow \sup_{\theta \in \mathrm{conv}\{ \hat{\theta}^\e_{m,T}, \theta_\star \}} \HgSq{\theta}{\theta_\star} \lesssim \frac{dT(d+\sigma^2)}{\sigma^2} \max\left\{1, \frac{1}{\sigma^2} \right\} \HgSq{\hat{\theta}^\e_{m,T}}{\theta_\star}. \label{eq:sin_GLM_cond_conv}
\end{align}
Our next step is to apply \Cref{thm:hellinger_bound_MLE} to ensure that the LHS condition in \eqref{eq:sin_GLM_cond_conv} holds.
To do this, we shall estimate the covering number of $\calP$ in the max FI divergence through the $\ell_2$ covering number.
If we fix some $\theta \in \Theta$ and let $\theta'$ denote its closest element in an $\e$-covering of $\Theta$ in $\ell_2$, %
\begin{align*}
    \FIMax{\theta}{\theta'} &= \| \theta - \theta' \|_{\calI_{\max}} \leq \sqrt{\lambda_{\max}(\calI_{\max})} \norm{ \theta - \theta' } \leq \frac{\sqrt{T}}{\sigma}\sqrt{d + \sigma^2} \e.
\end{align*}
This allows the relation 
\begin{align*}
    \calN_{\mathrm{\calI_{\max}}}(\calP, \e) \leq \calN_{\norm{\cdot}}\left(\Theta, \frac{\e \sigma}{\sqrt{T(d + \sigma^2)}} \right) 
    \leq \left( 3R \frac{\sqrt{T(d + \sigma^2)}}{\e \sigma} \right)^{d^2}. 
\end{align*}
We now apply \Cref{thm:hellinger_bound_MLE} with $\e = \delta / (2 \sqrt{2m})$ 
to conclude that with probability at least $1-\delta$:
\begin{align}
    \HgSq{ \hat{\theta}^\e_{m,T} }{ \theta_\star } \lesssim \frac{d^2}{m} \log\left( \frac{c_0 R mT (d+\sigma^2)}{\sigma^2 \delta} \right), \label{eq:hellinger_error_sin_GLM_simple}
\end{align}
where $c_0$ is a universal positive constant.
Call this event $\calE_1$.
If we define $1 / \Phi_0 := \min \{1, 1 / \sigma^2\}$,
plugging this bound into \eqref{eq:sin_GLM_cond_conv} and applying \Cref{prop:log_dominance} yields that
if $m$ satisfies %
\begin{align}
    m \gtrsim \Phi_0 d^2 \log\left(  \frac{c_0' \Phi_0 d RT (d+\sigma^2)}{\sigma^2 \delta}  \right), \label{eq:phi0_bound}
\end{align}
where $c_0'$ is a universal constant,
then the following also holds on $\calE_1$:
\begin{align}
    \sup_{\theta \in \mathrm{conv}\{ \hat{\theta}^\e_{m,T}, \theta_\star \}} \HgSq{\theta}{\theta_\star} 
    &\lesssim \frac{d^4 T}{m} \max\left\{ \frac{1}{d}, \frac{1}{\sigma^2}, \frac{1}{\sigma^4} \right\} \log\left( \frac{c_0 R mT(d + \sigma^2)}{\sigma^2 \delta} \right) . \label{eq:hellinger_error_sin_GLM}
\end{align}

For what follows, we define the set
\begin{align*}
    \Theta' := \left\{ \theta \in \Theta \mid \forall j \in [d], \,\, \norm{ \mat(\theta)[j] - \mat(\theta_\star)[j] } \leq \norm{\mat(\theta_\star)[j]}/2 \right\}.
\end{align*}
A key property of $\Theta'$ that we will utilize is that
$\theta \in \Theta'$ implies $\norm{\mat(\theta)[j]} \geq \norm{\mat(\theta_\star)[j]}/2$ for all $j \in [d]$
by the triangle inequality.

\paragraph{Ensuring that $\hat{\theta}^\e_{m,T} \in \Theta'$ on $\calE_1$.}

Using \Cref{prop:local_hellinger_identifiability_sin_GLM} and \eqref{eq:hellinger_error_sin_GLM_simple}, we have that if $m$ satisfies,
\begin{align*}
    \frac{d^2}{m} \log\left( \frac{c_0 R mT (d+\sigma^2)}{\sigma^2 \delta} \right)
    &\lesssim \min \left\{ (A_{\star,\min}^2 \min\{1, \sigma^2\}), \min\{1, 1/\sigma^2\} \right\} \triangleq 1 / \Phi_1.
\end{align*}
then we have that $\hat{\theta}^\e_{m,T} \in \Theta'$ on $\calE_1$.
Using \Cref{prop:log_dominance}, this holds whenever $m$ satisfies:
\begin{align}
    m &\gtrsim \Phi_1 d^2 \log\left(  \frac{c_1 \Phi_1 d RT (d+\sigma^2)}{\sigma^2 \delta}  \right), \label{eq:m_req_1_sin_GLM}
\end{align}
where $c_1$ is another univeral constant.
We can note that since $\Phi_1 \geq \Phi_0$, this constraint automatically satisfies \eqref{eq:phi0_bound} (possibly after adjusting the value of $c_1$).

\paragraph{Lower bound FI matrix.} 
Our first task is to estimate a lower bound on the FI matrix for $\theta \in \Theta'$.
Recall the Fisher information from earlier:
\begin{align*}
    \calI(\theta) = \frac{1}{\sigma^2} \sum_{t=1}^{T-1} \E_{z_t \sim p_\theta}[ (D_\theta h_\theta(z_t))^\T (D_\theta h_\theta(z_t)) ]
\end{align*}
We can expand $z_t = \mu_{t-1} + w_{t-1}$ for $\mu_{t-1} = h_{\theta}(x_{t-1})$ to expand the dependency of the expectation on the previous timestep.
\begin{align*}
    \E[ (D_\theta h_\theta(z_t))^\T &(D_\theta h_\theta(z_t)) ] \\
    &= \E[ M(z_t)^\T \diag(\cos^2(M(z_t) \theta)) M(z_t) ] \\
    &= \E[\E[ M(\mu_{t-1} + w_{t-1})^\T \diag(\cos^2(M(\mu_{t-1} + w_{t-1})\theta))M(\mu_{t-1} + w_{t-1}) \mid \mu_{t-1} ]],
\end{align*}
We choose a $\nu_0$ to be specified later, and observe that by 
\Cref{prop:cos_anticoncentration} and a union bound,
with $A = \mat(\theta)$ and $A[j] \in \R^d$ denoting the $j$-th row of $A$, %
\begin{align*}
    \Pr\left( \bigcup_{j \in [d]} \left\{ \cos^2( \ip{A[j]}{\mu_{t-1}} + \ip{A[j]}{w_{t-1}} ) \leq \frac{\nu_0}{c_0 d} \min\{1,\sigma \norm{A[j]} \} \right\} \,\Big|\, \mu_{t-1} \right) \leq \nu_0.
\end{align*}
Call this event, which depends on $\mu_{t-1}$, 
$\calE_{\mu_{t-1}}$. Letting $A_{\min} := \min_{j \in [d]} \norm{A[j]}$,
we now write:
\begin{align*}
    &\E[ (D_\theta h_\theta(z_t))^\T(D_\theta h_\theta(z_t)) ] \\
    &= \E[\E[ M(\mu_{t-1} + w_{t-1})^\T \diag(\cos^2(M(\mu_{t-1} + w_{t-1})\theta))M(\mu_{t-1} + w_{t-1}) \mid \mu_{t-1} ]] \\
    &\succcurlyeq \E[\E[ M(\mu_{t-1} + w_{t-1})^\T \diag(\cos^2(M(\mu_{t-1} + w_{t-1})\theta))M(\mu_{t-1} + w_{t-1}) \ind_{\calE_{\mu_{t-1}}} \mid \mu_{t-1} ]] \\
    &\succcurlyeq \frac{\nu_0^2}{c_0^2 d^2} \min\{1, \sigma^2 A_{\min}^2 \} \cdot \E[ \E[    
 ((\mu_{t-1} + w_{t-1})^{\otimes 2} \otimes I_d ) \ind_{\calE_{\mu_{t-1}}} \mid \mu_{t-1}] ] \\
    &= \frac{\nu_0^2}{c_0^2 d^2} \min\{1, \sigma^2 A_{\min}^2 \} \cdot ( \E[ \E[    
 (\mu_{t-1} + w_{t-1})^{\otimes 2} \ind_{\calE_{\mu_{t-1}}} \mid \mu_{t-1}]] ) \otimes I_d,
\end{align*}
where the last equality follows from the bi-linearity of the Kronecker product.
We now fix a $v \in \R^{d}$ with unit-norm, 
and observe, dropping subscripts on $t$ for clarity,
\begin{align*}
    \E_w[ \ip{v}{\mu + w}^2 \ind_{\calE_\mu} ] &= \E_w[ \ip{v}{\mu + w}^2 ] - \E_w[ \ip{v}{\mu + w}^2 \ind_{\calE_\mu^c} ] \\
    &\geq \E_w[ \ip{v}{\mu + w}^2 ] - \sqrt{\E_w[ \ip{v}{\mu + w}^4]} \sqrt{\Pr(\ind_{\calE_\mu^c})} \\
    &\geq \E_w[ \ip{v}{\mu + w}^2 ] - \sqrt{\E_w[ \ip{v}{\mu + w}^4]} \sqrt{\nu_0} \\ 
    &\geq (1 - 9 \sqrt{\nu_0}) \E_w[ \ip{v}{\mu + w}^2 ] \\
    &= (1 - 9 \sqrt{\nu_0}) v^\T( \mu\mu^\T + \sigma^2 I_d ) v \\
    &\geq (1- 9 \sqrt{\nu_0}) \sigma^2, 
\end{align*}
where the last inequality holds by hyper-contractivity of Gaussian polynomials \cite[see e.g.,][Ch.~5]{janson1997gaussian}.
Hence we set $\nu_0 = 1/324$,
from which we conclude
\begin{align*}
    \E[ (\mu_{t-1}+w_{t-1})^{\otimes 2} \ind_{\calE_{\mu_{t-1}}} ] \succcurlyeq \sigma^2/2 \cdot I_d.
\end{align*}
Consequently, we have that
\begin{align*}
    \E[ (D_\theta h_\theta(z_t))^\T (D_\theta h_\theta(z_t)) ] \succcurlyeq \frac{\nu_0^2}{c_0^2 d^2} \min\{1, \sigma^2\} \sigma^2/2 \cdot I_{d^2} = \frac{c_1^2}{d^2} \min\{1, \sigma^2 A_{\min}^2 \} (T-1) \sigma^2 \cdot I_{d^2}.
\end{align*}
Therefore, we conclude that:
\begin{align*}
    \calI(\theta) = \frac{1}{\sigma^2} \sum_{t=1}^{T-1} \E_{z_t \sim p_\theta}[ (D_\theta h_\theta(z_t))^\T (D_\theta h_\theta(z_t)) ] \succcurlyeq \frac{c_1^2}{d^2} \min\{1, \sigma^2 A_{\min}^2 \} (T-1) \cdot I_{d^2}.
\end{align*}
Up until this point, we have not used the assumption that
$\theta \in \Theta'$. 
Observe that $A_{\min} \geq A_{\star,\min}/2$ whenever $\theta \in \Theta'$, we finally conclude that for $\theta \in \Theta'$,
\begin{align}
    \calI(\theta) = \frac{1}{\sigma^2} \sum_{t=1}^{T-1} \E_{z_t \sim p_\theta}[ (D_\theta h_\theta(z_t))^\T (D_\theta h_\theta(z_t)) ] \succcurlyeq \frac{c_1^2}{4d^2} \min\{1, \sigma^2 A_{\star,\min}^2 \} (T-1) \cdot I_{d^2}. \label{eq:FI_matrix_lower_bound_sin_GLM}
\end{align}

\paragraph{Step 2: Estimating $B_1$ and $B_2$.}

We will estimate $B_1$ and $B_2$ over $\Theta'$.
We start with $B_1$.
First, given a trajectory $z_{1:T} \sim p_\theta$, we have that
\begin{align*}
    \nabla_\theta \log p_\theta(z_{1:T}) = \sum_{t=1}^{T-1} \nabla_\theta \log p_\theta(z_{t+1} \mid z_t) = \frac{1}{\sigma^2} \sum_{t=1}^{T-1} (D_\theta h_\theta(z_t))^\T w_t.
\end{align*}
Now fix a test vector $v \in \R^{d^2}$, and consider:
\begin{align*}
    v^\T \nabla_\theta \log p_\theta(z_{1:T}) = \frac{1}{\sigma^2} \sum_{t=1}^{T-1} \ip{D_\theta h_\theta(z_t) v}{w_t} =: \sum_{t=1}^{T-1} d_t.
\end{align*}
A useful inequality is the following.
\begin{align*}
    \opnorm{D_\theta h_\theta(z)} = \opnorm{\diag(\cos(M(z)\theta))M(z)} \leq \opnorm{M(z)} = \norm{z}.
\end{align*}
We first compute:
\begin{align*}
    \E[ d_t^2 \mid \calF_{t-1} ] &= \E[ \ip{D_\theta h_\theta(z_t) v}{w_t}^2  \mid \calF_{t-1} ] 
    = \sigma^2 \norm{ D_\theta h_\theta(z_t) v }^2 \leq  \sigma^2 \norm{z_t}^2 \norm{v}^2.
\end{align*}
We next bound $\E[ \norm{z_t}^4 ]$ as:
\begin{align*}
    \E[ \norm{z_t}^4 ] &= \E[ \norm{ h_\theta(x_{t-1}) + w_{t-1}  }^4 ] \\
    &\leq 8\E[ \norm{h_\theta(x_{t-1})}^4 + \norm{w_{t-1}}^4 ] 
    \leq 8(1 + \E[\norm{w_{t-1}}^4]) 
    \leq 8(1 + 3d^2\sigma^4).
\end{align*}
Using this bound,
\begin{align*}
    \E\left( \sum_{t=1}^{T-1} \E[ d_t^2 \mid \calF_{t-1} ] \right)^2 &\leq (T-1) \sum_{t=1}^{T-1} (\E[ d_t^2 \mid \calF_{t-1}])^2 \leq (T-1) \sigma^4 \norm{v}^4 \sum_{t=1}^{T-1} \E[\norm{z_t}^4] \\
    &\leq (T-1)^2 8(1+3d^2\sigma^4) \sigma^4 \norm{v}^4.
\end{align*}
Hence,
\begin{align*}
    \left( \E \left(\sum_{t=1}^{T-1} \E[ d_t^2 \mid \calF_{t-1} ] \right)^2 \right)^{1/4} \lesssim \sigma(1+\sigma\sqrt{d}) \norm{v} \sqrt{T}.
\end{align*}
We next bound,
\begin{align*}
    \E[ d_t^4 ] \leq \E[ \norm{D_\theta h_\theta(z_t) v}^4 \norm{w_t}^4 ] \leq \norm{v}^4 \E[ \norm{z_t}^4 \norm{w_t}^4 ] \leq \norm{v}^4 \sqrt{ \E[\norm{z_t}^8] } \sqrt{ \E[\norm{w_t}^8] }.
\end{align*}
Since $\norm{w_t}$ is a $\sigma$-sub-Gaussian random variable, we know that $\E[\norm{w_t}^8] \lesssim \sigma^8 d^4$.
On the other hand,
\begin{align*}
    \E[\norm{z_t}^8] = \E[ \norm{\mu_{t-1} + w_{t-1}}^8 ] \leq 128 (1 + \E[\norm{w_{t-1}}^8] ) \lesssim 1 + \sigma^8 d^4.
\end{align*}
Hence we have
\begin{align*}
    \sum_{t=1}^{T-1} \E[d_t^4] \lesssim \norm{v}^4 T \sigma^4 d^2 ( 1 + \sigma^4 d^2).
\end{align*}
Therefore,
\begin{align*}
    \left( \E\left(\sum_{t=1}^{T-1} d_t\right)^4 \right)^{1/4} \lesssim \norm{v} \sigma(1+\sigma\sqrt{d}) T^{1/2}(1 + T^{-1/4} \sqrt{d}) \leq \norm{v} \sigma(1 + \sigma\sqrt{d}) T^{1/2},
\end{align*}
where the last inequality holds since we assume $T \gtrsim d^2$.
Now we set $v = \calI(\theta)^{-1/2} \bar{v}$ for a unit norm $\bar{v}$, 
we have that
\begin{align*}
    \sup_{\theta_0, \theta_1 \in \Theta'} B_1(\theta_0, \theta_1) &\lesssim \frac{\sigma(1+\sigma\sqrt{d}) T^{1/2}}{\sqrt{\inf_{\theta \in \Theta'} \lambda_{\min}(\calI(\theta))}} 
    \lesssim \sigma d(1 + \sigma\sqrt{d}) \max\{ 1, 1/(\sigma A_{\star,\min}) \} .
\end{align*}
We now move to $B_2$.
First we define the vector-valued function for a fixed $q \in \R^{d}$:
\begin{align*}
    g(\theta; z, q) = D_\theta h_\theta(z)^\T q = M^\T(z) \diag(\cos(M(z)\theta)) q.
\end{align*}
The Jacobian of $g(\theta; z, q)$ is given by:
\begin{align*}
    D_\theta g(\theta; z, q) = - M^\T(z) \diag( h_\theta(z) \odot q ) M(z),
\end{align*}
where $\odot$ denotes the Hadamard (entry-wise) product.
Now we have
\begin{align*}
    -\sigma^2 \nabla_\theta^2 \log p_\theta(z' \mid z) &= (D_\theta h_\theta(z))^\T (D_\theta h_\theta(z)) + (D_\theta (D_\theta h_\theta(z))^\T) (h_\theta(z) - z') \\ 
    &= (D_\theta h_\theta(z))^\T (D_\theta h_\theta(z)) + D_\theta g(\theta; z, h_\theta(z)-z') \\ 
    &= M^\T(z) [\diag(\cos^2(M(z)\theta)) - \diag(h_\theta(z) \odot (h_\theta(z)-z')) ] M(z).
\end{align*}
Hence, given a test vector $v \in \R^{d^2}$, given $z_{1:T} \sim p_\theta$, for all $t \in [T-1]$ we have:
\begin{align*}
    -\sigma^2 v^\T \nabla_\theta^2 \log p_\theta(z_{t+1} \mid z_t) v &= v^\T M^\T(z_t) \diag( \cos^2(M(z_t)\theta) ) M(z_t) v + v^\T M^\T(z_t) \diag (h_\theta(z_t) \odot w_t ) M(z_t) v.
\end{align*}
We next compute:
\begin{align*}
    \E[ (v^\T M^\T(z_t) \diag( \cos^2(M(z_t)\theta) ) M(z_t) v )^2 ] &\leq \E[ \norm{M(z_t) v}^4 ] \leq \norm{v}^4 \E[ \norm{z_t}^4 ] \lesssim \norm{v}^4 (1 + \sigma^4 d^2).
\end{align*}
Next we compute
\begin{align*}
    \E[ (v^\T M^\T(z_t) \diag( h_\theta(z_t) \odot w_t ) M(z_t) v )^2 ] &= \sigma^2 \E[ \norm{ M(z_t) v \odot \sqrt{h_\theta(z_t)} }_4^4 ] \\
    &\leq \sigma^2 \E[ \norm{ M(z_t) v}_4^4 ] 
    \leq \sigma^2 \E[ \norm{M(z_t) v}^4 ] \\
    &\leq \sigma^2 \norm{v}^4 \E[ \norm{z_t}^4 ] 
    \lesssim \sigma^2 \norm{v}^4 (1 + \sigma^4 d^2).
\end{align*}
Hence,
\begin{align*}
     (\E[ (v^\T \nabla^2_\theta \log p_\theta(z_{1:T}) v)^2 ])^{1/2} \leq \sum_{t=1}^{T-1} (\E[ v^\T \nabla^2_\theta \log p_\theta(z_{t+1} \mid z_t) v ])^{1/2} \lesssim \frac{\norm{v}^2}{\sigma^2} T (1+\sigma) (1+\sigma^2 d).
\end{align*}
Now we set $v = \calI(\theta)^{-1/2} \bar{v}$ for a unit norm $\bar{v}$, 
we have that
\begin{align*}
    \sup_{\theta_0, \theta_1 \in \Theta'} B_2(\theta_0, \theta_1) &\lesssim \frac{T (1+\sigma)(1+\sigma^2 d)/\sigma^2}{\inf_{\theta \in \Theta'} \lambda_{\min}(\calI(\theta))} \\
    &\lesssim d^2 \max\left\{ 1, \frac{1}{\sigma^2 A^2_{\star,\min}} \right\} \frac{(1+\sigma)(1+\sigma^2 d)}{\sigma^2} \\
    &\lesssim d^3 \max\left\{ \sigma, 1 + \frac{1}{\sigma^2 d} \right\} \max\left\{ 1, \frac{1}{\sigma^2 A^2_{\star,\min}} \right\}.
\end{align*}

Finally, we conclude that
\begin{align}
    \sup_{\theta_0, \theta_1 \in \Theta'} \max\{ B_1^2(\theta_0, \theta_1), B_2(\theta_0, \theta_1) \} \lesssim d^3 \max\left\{ 1, \sigma^4, \frac{1}{\sigma^2 d} \right\} \max\left\{ 1, \frac{1}{\sigma^2 A^2_{\star,\min}} \right\}. \label{eq:B1_B2_sin_GLM}
\end{align}

\paragraph{Step 3: Parameter error bound.}
Utilizing \eqref{eq:hellinger_error_sin_GLM} and \eqref{eq:B1_B2_sin_GLM}, we have that to verify the condition \eqref{eq:hellinger_perturbation_radius} 
for $\theta_0 = \theta_\star$ and $\theta_1 = \hat{\theta}^\e_{m,T}$
on $\calE_1$, 
we need $m$ to satisfy:
\begin{align*}
    m &\gtrsim d^{10} T \cdot \max\left\{ \frac{1}{d}, \frac{1}{\sigma^2}, \frac{1}{\sigma^4} \right\} \max\left\{ 1, \sigma^8, \frac{1}{\sigma^4 d^2} \right\} \max\left\{1 , \frac{1}{\sigma^4 A^4_{\star,\min}} \right\} \cdot \log\left( \frac{c_0 RmT(d+\sigma^2)}{\sigma^2\delta} \right).
\end{align*}
Defining 
$$
    \Phi_2 := \max\left\{ \frac{1}{d}, \frac{1}{\sigma^2}, \frac{1}{\sigma^4} \right\} \max\left\{ 1, \sigma^8, \frac{1}{\sigma^4 d^2} \right\} \max\left\{1 , \frac{1}{\sigma^4 A^4_{\star,\min}} \right\},
$$
by \Cref{prop:log_dominance} it suffices for $m$ to satisfy: 
\begin{align}
    m &\gtrsim \Phi_2 d^{10}T \log\left( \frac{c_2 \Phi_2 d R T (d+\sigma^2)}{\sigma^2 \delta} \right). \label{eq:m_req_2_sin_GLM}
\end{align}
From \eqref{eq:FI_matrix_lower_bound_sin_GLM}, we have that on $\Theta'$, the following lower bound 
$\inf_{\theta \in \Theta'} \lambda_{\min}(\calI(\theta)) \gtrsim \frac{T}{d^2} \cdot \min\{1, \sigma^2 A_{\star,\min}^2 \}$
holds.
Since 
(a) \eqref{eq:hellinger_perturbation_radius} holds for $\theta_0=\theta_\star$ and
$\theta_1=\hat{\theta}^\e_{m,T}$ implies \eqref{eq:hellinger_perturbation_radius}
also holds for $\theta_0=\theta_\star$ and any $\theta_1 \in \mathrm{conv}\{ \hat{\theta}^\e_{m,T}, \theta_\star \}$
and (b) $\hat{\theta}^\e_{m,T} \in \Theta'$ implies
$\mathrm{conv}\{ \hat{\theta}^\e_{m,T}, \theta_\star \} \subset \Theta'$,
by \eqref{eq:hellinger_perturbation_avg_FI} from \Cref{prop:hellinger_perturbation}, we have on $\calE_1$,
for every $\theta \in \mathrm{conv}\{ \hat{\theta}^\e_{m,T}, \theta_\star \}$,
\begin{align}
    \norm{ \theta - \theta_\star }^2 &\lesssim \frac{d^2 \max\{1, 1/(\sigma^2 A^2_{\star,\min})\}}{T} \HgSq{\theta}{\theta_\star}, \nonumber \\
    &\lesssim \max\{1, 1/(\sigma^2 A^2_{\star,\min})\} \HgSq{\theta}{\theta_\star}, \label{eq:param_error_v2_sin_GLM}
\end{align}
where the last inequality holds since we assume that $T \gtrsim d^2$.

\paragraph{Step 4: Verify FI radius.}
We will utilize \Cref{prop:FI_lipschitz} to verify the FI radius condition \eqref{eq:hellinger_perturbation_FI_radius}.
Since $\nabla_\theta \log p_\theta(z_{t+1} \mid z_t) = - \frac{1}{\sigma^2} (D_\theta h_\theta(z_t))^\T (h_\theta(z_t) - z_{t+1})$ for $t \geq 1$, we have for any $\theta \in \Theta$,
\begin{align*}
    \calI_{t+1}(\theta \mid z_{1:t}) &= \frac{1}{\sigma^2} (D_\theta h_\theta(z_t))^\T (D_\theta h_\theta(z_t)), \quad t \in \{1, \dots, T-1\}, \\
    &= \frac{1}{\sigma^2} M(z_t)^\T \diag(\cos^2(M(z_t)\theta)) M(z_t).
\end{align*}
We also have the base case $\calI_1(\theta) = 0$.
Hence for $t \geq 1$, $\theta_1,\theta_2 \in \Theta$,
and any unit-norm $v \in \R^{d^2}$,
\begin{align*}
    \abs{v^\T(\calI_{t+1}(\theta_1 \mid z_{1:t}) - \calI_{t+1}(\theta_2 \mid z_{1:t}))v} &= \frac{1}{\sigma^2} \abs{ v^\T M(z_t)^\T \diag( \cos^2(M(z_t) \theta_1) - \cos^2(M(z_t)\theta_2) ) M(z_t) v } \\
    &\leq \frac{2}{\sigma^2} \norm{M(z_t) v}^2 \norm{M(z_t)(\theta_1-\theta_2)} \\
    &\leq \frac{2}{\sigma^2} \opnorm{M(z_t)}^3 \norm{\theta_1-\theta_2} \\
    &\leq \frac{2}{\sigma^2} \norm{z_t}^3 \norm{\theta_1-\theta_2}.
\end{align*}
Therefore by the variational form of the operator norm, 
we have for all $t \geq 1$ and $\theta_1,\theta_2 \in \Theta$:
\begin{align*}
    \opnorm{ \calI_{t+1}(\theta_1 \mid z_{1:t}) - \calI_{t+1}(\theta_2 \mid z_{1:t}) } \leq \Lip_{t+1}(z_{1:t}) \norm{\theta_1-\theta_2}, \quad \Lip_{t+1}(z_{1:t}) = \frac{2}{\sigma^2} \norm{z_t}^3.
\end{align*}
Note that for any $\theta \in \Theta$,
$\E_{p_\theta}[ \Lip_{t+1}(z_{1:t}) ] \lesssim \frac{1}{\sigma^2}(1 + \sigma^3 d^{3/2}) \asymp \max\{ 1/\sigma^2, \sigma d^{3/2} \}$,
and consequently $\Lip \lesssim \max\{1/\sigma^2, \sigma d^{3/2} \}$.
On the other hand, for a unit-norm $v \in \R^{d^2}$ and $\theta_1,\theta_2 \in \Theta$,
\begin{align*}
    \E_{p_{\theta_1}}[ (v^\T \calI_{t+1}(\theta_2 \mid z_{1:t}) v)^2 ] &= \frac{1}{\sigma^4} \E_{p_{\theta_1}}[ (v^\T M(z_t)^\T \diag(\cos^2(M(z_t) \theta_2)) M(z_t) v)^2 ] \\
    &\leq \frac{1}{\sigma^4} \E_{p_{\theta_1}}[ \norm{M(z_t) v}^4 ] 
    \leq \frac{1}{\sigma^4} \E_{p_{\theta_1}}[ \norm{z_t}^4 ] \lesssim \frac{1}{\sigma^4}(1 + \sigma^4 d^2) \asymp \max\{1/\sigma^4, d^2\}.
\end{align*}
Hence, we have that
$B_{\calI} \lesssim \max\{ 1/\sigma^2, d \}$.
By \Cref{prop:FI_lipschitz} and
\eqref{eq:param_error_v2_sin_GLM}
we have on $\calE_1$ 
for any $\theta \in \mathrm{conv}\{ \hat{\theta}^\e_{m,T}, \theta_\star \}$,
\begin{align}
    \opnorm{\calI(\theta) - \calI(\theta_\star)} &\lesssim T \left[ \max\{1/\sigma^2, \sigma d^{3/2} \} \norm{\theta-\theta_\star} +  \max\{ 1/\sigma^2, d \} \Hg{\theta}{\theta_\star} \right] \nonumber \\
    &\lesssim T \left[ \max\{1/\sigma^2, \sigma d^{3/2} \} \max\{ 1, 1/(\sigma A_{\star,\min}) \} + \max\{ 1/\sigma^2, d \}  \right] \Hg{\theta}{\theta_\star} \nonumber \\
    &\lesssim T \max\{ 1/\sigma^2, \sigma d^{3/2}, d \} \max\{1, 1/(\sigma A_{\star,\min})\} \Hg{\theta}{\theta_\star}. \label{eq:sin_GLM_FI_lipschitz_intermediate}
\end{align}
Let us define 
\begin{align*}
    \Phi_3 := \max \left\{ \frac{1}{\sigma^4 d^3}, \sigma^2, \frac{1}{d} \right\} \max \left\{1, \frac{1}{\sigma^6 A^6_{\star,\min}} \right\} \max \left\{\frac{1}{d}, \frac{1}{\sigma^2}, \frac{1}{\sigma^4} \right\}.
\end{align*}
Combining \eqref{eq:hellinger_error_sin_GLM},
\eqref{eq:FI_matrix_lower_bound_sin_GLM}, 
and \eqref{eq:sin_GLM_FI_lipschitz_intermediate}, we have that on $\calE_1$,
\begin{align*}
    \sup_{\theta \in \mathrm{conv} \{\hat{\theta}^\e_{m,T}, \theta_\star\}} &\frac{\opnorm{\calI(\theta) - \calI(\theta_\star)}}{\lambda_{\min}(\calI(\theta_\star))} \\
    &\lesssim \sup_{\theta \in \mathrm{conv} \{\hat{\theta}^\e_{m,T}, \theta_\star\}} d^2 \max\{ 1/\sigma^2, \sigma d^{3/2}, d \} \max\{1, 1/(\sigma^3 A^3_{\star,\min})\} \Hg{\theta}{\theta_\star} \\
    &= \sup_{\theta \in \mathrm{conv} \{\hat{\theta}^\e_{m,T}, \theta_\star\}} d^{7/2} \max\{1/(\sigma^2 d^{3/2}), \sigma, 1/d^{1/2} \} \max\{1, 1/(\sigma^3 A^3_{\star,\min})\} \Hg{\theta}{\theta_\star}  \\
    &\lesssim \sqrt{ \frac{\Phi_3 d^{11} T}{m} \log\left( \frac{c_0 RmT(d+\sigma^2)}{\sigma^2 \delta} \right) }.
\end{align*}
Hence by \Cref{prop:log_dominance}, we have that \eqref{eq:hellinger_perturbation_FI_radius} 
holds on $\calE_1$ as long as $m$ satisfies:
\begin{align}
    m \gtrsim \Phi_3 d^{11} T \log\left( \frac{c_3 \Phi_3 dRT(d+\sigma^2)}{\sigma^2 \delta} \right). \label{eq:m_req_3_sin_GLM}
\end{align}

\paragraph{Step 5: Final result.} 
If $m$ satisfies conditions
\eqref{eq:m_req_1_sin_GLM},
\eqref{eq:m_req_2_sin_GLM}, and 
\eqref{eq:m_req_3_sin_GLM}, then on $\calE_1$ we have
from \Cref{prop:hellinger_perturbation} and \eqref{eq:hellinger_error_sin_GLM_simple}:
\begin{align*}
        \norm{\hat{\theta}^\e_{m,T} - \theta_\star}^2_{\bar{\calI}(\theta_\star)} \lesssim \frac{d^2}{mT} \log\left( \frac{c_0 R mT(d+\sigma^2)}{\sigma^2 \delta} \right).
\end{align*}

\subsection{Sequence Modeling with Linear Attention}
\label{sec:case_studies:sequence_modeling}

Since their introduction, transformer models and architectures have found popularity in modern sequence modeling tasks, finding use in fields such as language modeling, computer vision, and reinforcement learning \cite{ouyang2022traininglanguagemodelsfollow, li2023surveytransformersreinforcementlearning, dosovitskiy2021imageworth16x16words}.
Despite their widespread application however, full theoretical analysis of multi-layer transformer models is currently out of reach.  
As a result, stylized and simplified attention modules that isolate core mechanisms are commonly used in the literature as analytically tractable proxies for analyzing the full models.
Single-layer linear self-attention models have been used to explore the dynamics of in-context learning \cite{zhang2023trainedtransformerslearnlinear, vonoswald2023transformerslearnincontextgradient} and emergent inductive biases in transformers \cite{pmlr-v162-edelman22a, tian2023scansnapunderstandingtraining}.
Recent works show that attention can operate as a max-margin token selection mechanism, even establishing an equivalence with hard-margin SVM \cite{tarzanagh2023maxmargintokenselectionattention, tarzanagh2024transformerssupportvectormachines}.
Furthermore, \cite{li2024mechanics} established the global convergence of gradient descent for this framework, and a finite sample bounds were established by \cite{ildiz2024transformers} with parameter estimation upper bounds;
we take particular inspiration from this line of work, especially \cite{ildiz2024transformers},
and analyze a simple linear transformer~\cite{katharopoulos2020linearattention}
with a single-layer cross-attention and linear activation.

Let us consider a vocabulary of $K$ tokens denoting the states $z \in \sfZ$, each assigned a $d$-dimensional embedding by an embedding matrix $E = [e_1 \ \cdots \ e_K]^\T \in \mathbb{R}^{K \times d}$, such that $\sfZ = \{e_i \mid i \in [K]\}$. 
We further emphasize that these embeddings $\{e_k\}_{k=1}^K$ are \emph{not} the standard basis vectors, which we denote instead $\mathbf{1}(k) \in \mathbb{R}^K$
in this section.
We assume that the first \emph{two} tokens $z_0, z_1$ are drawn from a given initial distribution $\rho_1$ over $\sfZ \times \sfZ$ to create an initial state
for cross-attention.
To sample a new token $z_{t+1} \in \sfZ$ in a sequence $z_{0:t}$ where $z_{0:t} \in \mathbb{R}^{(t+1) \times d}$, we sample \emph{auto-regressively} by taking the last token $z_t$ to be the query token in a cross-attention layer given as:
\begin{align*}
    p_\theta(z_{t+1} \mid z_{0:t}) &= \mathbb{S}( \Phi(C (z_{0:t-1} V)^\T  \mathbb{A}( z_{0:t-1} \mathcal{K} Q^\T z_t))), \quad t \in [T-1],
\end{align*}
where $\mathcal{K}, Q, V \in \mathbb{R}^{d \times d}$ denote the key, query, and value matrices, $C \in \mathbb{R}^{(K-1) \times d}$ denotes the classifier head, $\mathbb{A}$ denotes an activation function, $\mathbb{S}$ denotes the softmax function, and  $\Phi:\mathbb{R}^{K-1} \mapsto \mathbb{R}^K$ denotes
the function that embeds $\Phi(x) := (x,0)$.\footnote{The function $\Phi$ is
introduced to allow for parameter recovery; otherwise 
parameterizing a distribution over $[K]$ using $K$ logits is not identifiable, as softmax is invariant under affine transforms (i.e., $\mathbb{S}(x + c \ind) = \mathbb{S}(x)$ for any $c \in \R$).}
We shall denote $\theta := \vec(\mathcal{K} Q^\T)$ to be the parametrization such that $\Theta = \{ \theta \in \mathbb{R}^{d^2} \mid \norm{\theta} \leq R \}$, 
and we constrain $V = I_d$ and $C$ to be fixed matrices.  Finally, we use a linear activation~\cite{katharopoulos2020linearattention}, where we normalize by $1/t$ for key trajectory length $t$.\footnote{This normalization ensures that the sum of key-query attention scores does not scale with trajectory length, which would increase the magnitude inside the outer softmax over time and cause exponential decay in the minimum probability of a token.
This scaling is implicit in softmax activation settings, which would divide by the sum of the exponentiated weights.
We note that this scaling is also used in practice in~\cite{zhuoran2021efficientattention}.
}
This all simplifies to the following: 
\begin{align}
    p_\theta(z_{t+1} \mid z_{0:t}) &= \mathbb{S} \left(\Phi \left( \frac{1}{t}C z_{0:t-1}^\T   z_{0:t-1} \mat(\theta) z_t \right) \right), \quad t \in [T-1], \label{eq:linear_attention_model}
\end{align}
where we recall that $\mat$ is the matricization function such that $\mat(\theta) \in \mathbb{R}^{d \times d}$.

\begin{mythm}
\label{thm:sequence_model}
    Fix $\delta \in (0,1)$, and suppose that 
    the embedding matrix $E$ and classifier head $C$ are both full column rank, with normalized embeddings such that $\max_{k \in [K]} \norm{e_k} = 1$ and classifier head such that $\opnorm{C} \geq 1$.
    Let $\Theta = \{\theta \in \mathbb{R}^{d^2} \mid \norm{\theta} \leq R\}$ for $R \geq 1$ and $d > 1$, and let $\hat{\theta}_{m,T}^\e$ denote the max FI discretized MLE estimator at resolution $\e= \frac{\delta}{2 \sqrt{2m}}$. 
    If the number of trajectories $m$ and the trajectory length $T$ satisfies 
    \begin{align}
        m &\gtrsim d^2 T_0^2 \log \left( \frac{c_0 d^2 T_0^2 R^2 \opnorm{C}^2 T}{\delta^2} \right), \quad
        T \gtrsim T_0, \quad
        T_0 := \frac{K^4 \kappa(C)^2}{\sigma_{\min}(E)^4} \exp(6 R \opnorm{C}), \label{eq:sequence_model_reqs}
    \end{align}
    where $\kappa(C)$ denotes the condition number of $C$,
    then with probability at least $1 - \delta$ for $\delta \in (0, 1)$ and for a universal positive constant $c_0$, we have
    \begin{align}
        \| \hat{\theta}_{m,T}^\e - \theta_\star \|_{\bar{\calI}(\theta_\star)}^2 \lesssim \frac{d^2 \log(c_0 R^2 \opnorm{C}^2 mT / \delta^2)}{mT}. \label{eq:sequence_model_rate}
    \end{align}
\end{mythm}

To the best of our knowledge, \Cref{thm:sequence_model} is the first result for learning the parameters of an auto-regression linear transformer model
in the multiple trajectory setting which achieves a nearly instance-optimal
rate (cf.~\eqref{eq:sequence_model_rate}), 
which also includes a rate of convergence that decreases with all the data $mT$ instead of just the number of trajectories $m$.
The most related result to \Cref{thm:sequence_model}
comes from \cite[Corollary 4.3]{ildiz2024transformers}, which
we will compare with in detail in a moment.
Before discussion this related result, we make a few remarks on \Cref{thm:sequence_model}. First, we know that the assumption 
that both $E$ and $C$ are full column rank implies the constraint
$d + 1 \leq K$, which states that the embedding size $d$ is 
\emph{less} than the vocabulary size $K$ minus one.
This is a realistic assumption in practice, as typical vocabulary sizes
for modern LLMs are often in the $100$k range,
whereas typical embedding sizes are typically no more than $10$k.\footnote{For example, Meta's Llama3 8B model has a vocabulary size of $128$k with an embedding dimension of $4096$~\cite{llama3}.}
We next focus on the trajectory requirement on $m$ in \eqref{eq:sequence_model_reqs}. We first note that 
the constraint on trajectory length $T \geq T_0$ in \eqref{eq:sequence_model_reqs} 
can be elided at the expense of a more complex expression for the required number of trajectories, but the final rate would remain the same.
Next, the requirement on $m$, ignoring the contributions of $C, E$,
is $m \gtrsim \tilde{\Omega}(d^2 K^8 \exp(R))$.
While the dependence on $d$ is correct, we anticipate that the dependence on $K, R$ is not sharp, and can be improved with further analysis.
Similarly, in our analysis we show a bound of the form
$\lambda_{\min}(\bar{\calI}(\theta_\star)) \gtrsim K^{-4} \exp(-R)$
(also ignoring contributions of $C, E$), which implies from \eqref{eq:sequence_model_rate} a parameter recovery 
bound of $\norm{ \hat{\theta}^\e_{m,T} - \theta_\star }^2 \lesssim \tilde{O}( d^2 K^4 \exp(R) / (mT))$; as with the requirement on $m$, we anticipate this
parameter recovery bound is also not optimal in its dependence on $K, R$
(but is optimal in $d, m, T$).

\paragraph{Comparison with \cite[Corollary 4.3]{ildiz2024transformers}.}

As mentioned previously, the most comparable result to \Cref{thm:sequence_model}
is \cite[Corollary 4.3]{ildiz2024transformers}. Here, the authors also study a multi-trajectory
data model, but one key difference is that in \cite{ildiz2024transformers}, the trajectory
$z_{0:T}$ is not auto-regressively generated. Instead, there is a distribution $\calD_X$ over \emph{prompts} $z_{0:T-1}$, followed by a last token $z_T$ generated from a self-attention model conditioned on 
the prompt $z_{0:T-1}$, resembling a standard supervised learning setup.
Consequently, their final parameter recovery rate only decays with the number of trajectory $m$,
in comparison to our rate \eqref{eq:sequence_model_rate} which decreases with the total data budget $mT$.
Another difference between our two settings is a structural one. 
We choose to analyze a setting with linear activation instead of softmax activation $\mathbb{A}$, and we constrain the outputs of the classifier head $C$ to $\Delta^{K-1}$ (the probability simplex in $\R^K$)
by means of an outer softmax activation (i.e., we treat the outputs of the classifier head as logits, as is typically done in practice), as detailed in \eqref{eq:linear_attention_model}.
On the other hand in \cite{ildiz2024transformers} they consider a softmax activation $\mathbb{A}$,
but omit the softmax activation after the classifier head. Hence, they 
require additional assumptions on the classifier head and the embeddings matrix
to ensure that the output of the classifier head is a valid probability distribution.
This may seem like a minor difference, but their setup requires that vocabulary embeddings
$E$ are linearly independent~\cite[Assumption 2.3]{ildiz2024transformers}, 
which requires that $d \geq K$ (i.e., the embedding dimension exceeds the vocabulary size). As we discussed previously, in practice we typically have the opposite trend (i.e., embedding dimension is much smaller than vocabulary size), which our model allows for.

With these remarks in place, we can now directly compare our bounds to \cite{ildiz2024transformers},
keeping in mind the differences in problem setup and assumptions described previously.
To keep the comparison simple, we will suppress dependency on $C, R, E$, and only focus on $m, T, K$ in the bounds. The main parameter recovery result in \cite{ildiz2024transformers} states that with high probability:
\begin{align}
    m \gtrsim \tilde{\Omega}\left(\frac{K^2}{\alpha^2}\right) \Longrightarrow \norm{\hat{\theta}_{m,T} - \theta_\star}^2 \lesssim \tilde{O}\left( \frac{K^2}{\alpha^4 m} \right), \label{eq:ildiz_seq_model_rate}
\end{align}
where $\alpha > 0$ is the strong convexity constant of the population loss over a ball around $\theta_\star$. This quantity $\alpha$ is left unspecified in their argument; they only argue
that $\alpha > 0$, but do not provide an explicit lower bound for it. 
Since for negative log likelihood, both Fisher information and Hessian of the population 
loss coincide, in the notation of our work,
$\alpha = \inf_{\theta \in B(\theta_\star, r_0)} \lambda_{\min}(\E_{z_{0:T-1} \sim \calD_X}[ \calI_T(\theta \mid z_{0:T-1}) ])$ (note that the conditional Fisher Information notation is defined in \eqref{eq:conditional_FI_matrix}) where $r_0$ is a localization parameter
which we consider as a constant.
With this in mind, \Cref{thm:sequence_model} implies that with high probability:
\begin{align}
    m \gtrsim \tilde{\Omega}\left(\frac{d^2}{\bar{\alpha}^2}\right), \,\, T \gtrsim \frac{1}{\bar{\alpha}} \Longrightarrow \norm{\hat{\theta}^\e_{m,T} - \theta_\star}^2 \lesssim \tilde{O}\left( \frac{d^2}{\overline{\alpha} mT} \right), \label{eq:our_seq_model_rate}
\end{align}
where $\bar{\alpha} := \lambda_{\min}(\bar{\calI}(\theta_\star))$.
As mentioned previously, we show a lower bound on 
$\bar{\alpha} \gtrsim K^{-4}$, which again is most likely
not optimal.
Comparing \eqref{eq:ildiz_seq_model_rate}
with \eqref{eq:our_seq_model_rate}, we see that
the dependence on the parameter dimension ($K$ for the former as they consider a subspace of 
$d \times d$ matrices with dimension $\leq K^2$, $d$ for our case) is equivalent up to log factors.
On the other hand, our bound yields an improvement on the dependence of the $\bar{\alpha}$ (vs. $\alpha$) parameter in the final rate. 
Finally and most importantly, as our setting studies auto-regressive generation, our rate is able to capture the dependence on all the data points $mT$, rather than just the number of trajectories $m$.

\subsubsection{Proof of \Cref{thm:sequence_model}}

\paragraph{Step 1: Covering number bound.}
As earlier, we first estimate the covering number of $\mathcal{P}$ in the FI norm through the $\ell_2$ covering number. 
Let $J := \begin{bmatrix} I_{K-1} \\ 0 \end{bmatrix} \in \R^{K \times (K-1)}$.
If we denote $M_{0:t} := \frac{1}{t} z_t^\T \otimes (J C z_{0:t-1}^\T z_{0:t-1}) \in \mathbb{R}^{K \times d^2}$ and $\bar{M}_{0:t} := \frac{1}{t} z_t^\T \otimes (C z_{0:t-1}^\T z_{0:t-1}) \in \mathbb{R}^{(K-1) \times d}$,
we can see that the following holds by vectorization:
\begin{align*}
    \Phi \left( \frac{1}{t} C z_{0:t-1}^\T z_{0:t-1} \mat(\theta) z_t \right) &= \Phi(\bar{M}_{0:t} \theta) = J \bar{M}_{0:t} \theta = M_{0:t} \theta
\end{align*}
As such we can say $p_\theta(z_{t+1} \mid z_{0:t}) = \mathbb{S}(M_{0:t} \theta)_{k}$ for $z_{t+1} = e_k$.
Next, since we have
\begin{align*}
    \FIMax{p_{\theta_0}}{p_{\theta_1}} &=\| \theta_0 - \theta_1 \|_{\calI_{\max}} \leq \sqrt{\lambda_{\max}(\calI_{\max})} \norm{ \theta_0 - \theta_1},
\end{align*}
this motivates finding an expression for the Fisher information matrix and its maximum eigenvalue.
If we let $(z_t) \in [K]$ denote the index of the token associated with entry $z_t$ and $[M_{0:t}]_i$ denote the $i$-th row of $M_{0:t}$, we calculate the Hessian of the log likelihood:
\begin{align*}
    \nabla_\theta \log p_\theta(z_{t+1} \mid z_{0:t}) &= [M_{0:t}]_{(z_{t+1})} - M_{0:t}^\T \mathbb{S}(M_{0:t} \theta), \\
    \nabla_\theta^2 \log p_\theta(z_{t+1} \mid z_{0:t}) &= M_{0:t}^\T \left( \mathbb{S}(M_{0:t} \theta) \mathbb{S}(M_{0:t} \theta)^\T - \diag(\mathbb{S}(M_{0:t} \theta)) \right) M_{0:t}.
\end{align*}
If we denote the conditional expectation $\E_t^\theta[\cdot] \triangleq \E_{p_\theta}[ \cdot \mid z_{0:t}]$,
\begin{align*}
    \mathbb{E}^\theta_{t}[\nabla_\theta \log p_\theta(z_{t+1} \mid z_{0:t})] &= 0, \\
    \mathbb{E}^\theta_{t}[\nabla_\theta^2 \log p_\theta(z_{t+1} \mid z_{0:t})] &= M_{0:t}^\T \left( \mathbb{S}(M_{0:t} \theta) \mathbb{S}(M_{0:t} \theta)^\T - \diag(\mathbb{S}(M_{0:t} \theta)) \right) M_{0:t}.
\end{align*}
Hence, the FI matrix can be represented as: 
\begin{align}
    \calI_{t+1}(\theta \mid z_{0:t}) &= - \mathbb{E}^\theta_{t} [ \nabla_\theta^2 \log p_\theta(z_{t+1} \mid z_{0:t})] = M_{0:t}^\T \left( \diag(\mathbb{S}(M_{0:t} \theta)) - \mathbb{S}(M_{0:t} \theta) \mathbb{S}(M_{0:t} \theta)^\T \right) M_{0:t}, \label{eq:seq_mod_cond_FI} \\
    \calI(\theta) &= \sum_{t=1}^{T-1} \mathbb{E}_{z_{0:t} \sim p_{\theta}} \left[ M_{0:t}^\T \left( \diag(\mathbb{S}(M_{0:t} \theta)) - \mathbb{S}(M_{0:t} \theta) \mathbb{S}(M_{0:t} \theta)^\T \right) M_{0:t} \right]. \label{eq:seq_mod_FI}
\end{align}
We may see that for all $t \in [T-1]$, $\opnorm{M_{0:t}} \leq \sup_{k \in [K]} \norm{e_k}^3 \opnorm{C} = \opnorm{C}$.
Expanding out the multiplication and upper bounding gives our result.
\begin{align*}
    \calI(\theta) &= \sum_{t=1}^{T-1} \mathbb{E}_{z_{0:t} \sim p_{\theta}} \left[ M_{0:t}^\T \left( \diag(\mathbb{S}(M_{0:t} \theta)) - \mathbb{S}(M_{0:t} \theta) \mathbb{S}(M_{0:t} \theta)^\T \right) M_{0:t} \right] \\
    &= \sum_{t=1}^{T-1} \mathbb{E}_{z_{0:t} \sim p_{\theta}} \left[ M_{0:t}^\T  \diag(\mathbb{S}(M_{0:t} \theta))  M_{0:t} - M_{0:t}^\T \mathbb{S}(M_{0:t} \theta) \mathbb{S}(M_{0:t} \theta)^\T M_{0:t} \right] \\
    &\preceq \sum_{t=1}^{T-1} \mathbb{E}_{z_{0:t} \sim p_{\theta}} \left[ M_{0:t}^\T  \diag(\mathbb{S}(M_{0:t} \theta))  M_{0:t} \right] \\
    &\preceq T \opnorm{C}^2 I_{d^2}.
\end{align*}
This gives that for all $\theta \in \Theta$, $\lambda_{\max}(\calI(\theta)) \leq T \opnorm{C}^2$: notably, this expression is agnostic of the exact parametrization.  
This allows us to set $\calI_{\max} = T \opnorm{C}^2 I_{d^2}$, from which we have $\lambda_{\max}(\calI_{\max}) = T \opnorm{C}^2$.
If we fix some $\theta \in \Theta$ and let $\hat{\theta}$ denote its closest element in an $\e$-covering of $\Theta$ in $\ell_2$, substituting
$\calI_{\max}$ into the previous bound on $\FIMax{p_\theta}{p_{\hat{\theta}}}$
gives $\FIMax{p_\theta}{p_{\hat{\theta}}} \leq \sqrt{T} \opnorm{C} \e$.
This implies the relation for $\e \in (0, 1)$:
$$
\mathcal{N}_{\mathrm{\calI_{\max}}}(\mathcal{P}, \e) \leq \mathcal{N}_{\norm{\cdot}}\left(\Theta, \frac{\e}{\sqrt{T} \opnorm{C}}\right) \leq \left( 3 R \frac{\sqrt{T} \opnorm{C}}{\e} \right)^{d^2}.$$
Applying \Cref{thm:hellinger_bound_MLE} with $\e = \frac{\delta}{2 \sqrt{2m}}$ and $\eta = \frac{1}{2 R \opnorm{C} \sqrt{mT}}$, and upper bounding $ R \opnorm{C} \sqrt{mT} / \delta \leq ( R \opnorm{C} \sqrt{mT} / \delta)^{d^2}$ in the logarithm, for some universal constant $c_0 > 0$ we obtain with probability at least $1 - \delta$,
\begin{align*}
    \sup_{\theta \in \mathrm{conv}\{ \hat{\theta}^\e_{m,T}, \theta_\star \}} \HgSq{\theta}{\theta_\star} &\leq \inf_{\eta>0}\left\{\frac{6}{m} \log\left(\frac{2 \calN_{\calI_{\max}}(\calP, \e) }{\delta} \bigceil{\frac{1}{2\eta}} \right) + \frac{3\eta^2}{4} \mathrm{diam}^2(\Theta) + 3\e^2\right\} \\ 
    &\lesssim \frac{1}{m} \log\left(\frac{ c_0 \calN_{\calI_{\max}}(\calP, \e) \bigceil{ R \opnorm{C} \sqrt{mT}} }{\delta} \right) + \frac{\mathrm{diam}^2(\Theta)}{ R^2 \opnorm{C}^2 mT} + \e^2 \\ 
    &\lesssim \frac{d^2}{m} \log \left( \frac{ c_0 R^2  \opnorm{C}^2 m T}{\delta^2} \right) + \frac{1 + \delta^2}{m}. 
\end{align*}
For satisfactory $c_0$, the final $\frac{1 + \delta^2}{m}$ term can additionally be collapsed into the first term given that $d > 1$.
Let us denote $\mathbf{1}(z) \in \{0, 1\}^K$ to be the one-hot standard basis vector such that for $i \in [K]$, $[\mathbf{1}(z)]_i = 1$ if $z = e_i$ and $[\mathbf{1}(z)]_i = 0$ otherwise.
When invoking \Cref{thm:hellinger_bound_MLE}, 
we can use the log-concave conclusion \eqref{eq:fisher_bound_MLE_discrete},
since $\log p_\theta(z_{0:T})$ is given by:
\begin{align*}
    \log p_\theta(z_{0:T}) &= \log \rho_1(z_0, z_1) + \sum_{t=1}^{T-1} \log p_\theta(z_{t+1} \mid z_{0:t}) \\
    &=  \log \rho_1(z_0, z_1) + \sum_{t=1}^{T-1} \log( \ip{\mathbb{S}(M_{0:t} \theta)}{\mathbf{1}(z_{t+1})} ) \\
    &= \log \rho_1(z_0, z_1) + \sum_{t=1}^{T-1} \ip{ M_{0:t}\theta }{\mathbf{1}(z_{t+1})} - \mathrm{LSE}( M_{0:t}\theta ),
\end{align*}
which is the sum of affine terms in $\theta$
minus the sum of terms which are given by taking the log-sum-exp (LSE) of a linear function of $\theta$, which is convex~\cite[cf.][Section 3.1.5]{boyd2004convex}.
Hence, $\log p_\theta(z_{0:T})$ is a concave function by basic composition rules.

\paragraph{Step 2: Estimating $B_1$ and $B_2$.}
Crucial in bounding $B_1$ and $B_2$ is bounding the smallest eigenvalue of $\calI(\theta_\star)$, so let us note the  expansion of the Fisher information from \cref{eq:seq_mod_FI}:
\begin{align*}
    \calI(\theta) 
    &= \sum_{t=1}^{T-1} \mathbb{E}_{z_{0:t} \sim p_{\theta}} \left[ \bar{M}_{0:t}^\T J^\T \left( \diag(\mathbb{S}(M_{0:t} \theta)) - \mathbb{S}(M_{0:t} \theta) \mathbb{S}(M_{0:t} \theta)^\T \right) J \bar{M}_{0:t} \right].
\end{align*}
We note that if one simply looks at $\diag(\mathbb{S}(M_{0:t} \theta)) - \mathbb{S}(M_{0:t} \theta) \mathbb{S}(M_{0:t} \theta)^\T$, the minimum eigenvalue would be zero since there is a constrained direction due to $\mathbb{S}(M_{0:t}\theta)$ lying on the $K-1$ dimensional simplex. Hence, we look at the reduced matrix by ignoring the last redundant coordinate, which restores a non-zero minimum eigenvalue.
As such, we begin by examining the inner expression $J^\T(\diag(\mathbb{S}(M_{0:t} \theta)) - \mathbb{S}(M_{0:t} \theta) \mathbb{S}(M_{0:t} \theta)^\T)J$.   
\Cref{prop:min_eig_cov} states that if we can show there exists $\mu > 0$ such that $\forall k \in [K]$, $\mathbb{S}(M_{0:t} \theta)_k \geq \mu$, then $\lambda_{\min} \left( J^\T(\diag(\mathbb{S}(M_{0:t} \theta)) - \mathbb{S}(M_{0:t} \theta) \mathbb{S}(M_{0:t} \theta)^\T)J \right) \geq \frac{\mu}{4(K-1)}$. 
Let us now find such a $\mu$ by lower bounding $\log p_\theta(e_k \mid z_{0:t})$ and exponentiating it to find a bound for $p_\theta(e_k \mid z_{0:t})$ for arbitrary $k \in [K]$:
\begin{align*}
    \log p_\theta(e_k \mid z_{0:t}) &= [M_{0:t}]_{k}^\T \theta - \log \sum_{i=1}^K \exp\left( [M_{0:t}]_i^\T \theta \right) \\
    &\geq \min_{i,j \in [K]} [M_{0:t}]_i^\T \theta - \log \left( K \exp \left( [M_{0:t}]_j^\T \theta \right) \right) \\
    &\geq \min_{i,j \in [K]} \left( [M_{0:t}]_i - [M_{0:t}]_j \right)^\T \theta - \log K.
\end{align*}
We now exponentiate to recover a bound for the conditional likelihood:
\begin{align*}
    p_\theta(e_k \mid z_{0:t}) &\geq \min_{i,j \in [K]} \exp\left( \left( [M_{0:t}]_i - [M_{0:t}]_j \right)^\T \theta - \log K \right) \\
    &= \min_{i,j \in [K]} \frac{1}{K} \exp \left( -\left( [M_{0:t}]_i - [M_{0:t}]_j \right)^\T \theta \right).
\end{align*}
Through Cauchy-Schwartz we may bound the quantity inside the exponent, giving our final bound.
\begin{align*}
    \max_{i,j \in [K]} \left([M_{0:t}]_i - [M_{0:t}]_j \right)^\T \theta &\leq \max_{i,j \in [K]}  \left\| [M_{0:t}]_i - [M_{0:t}]_j\right\| \norm{\theta} \\
    &\leq \max_{i \in [K]} 2 \norm{ [M_{0:t}]_i } \norm{\theta} \\
    &\leq 2  R \opnorm{C}, \\
    \Longrightarrow p_\theta(e_k \mid z_{0:t}) &\geq \frac{1}{K} \exp\left(-2  R \opnorm{C} \right).
\end{align*}
Therefore we can set $\mu = \frac{1}{K} \exp\left(-2  R \opnorm{C} \right)$ in order to satisfy the condition.
This gives that $\lambda_{\min} \left(J^\T (\diag(\mathbb{S}(M_{0:t} \theta)) - \mathbb{S}(M_{0:t} \theta) \mathbb{S}(M_{0:t} \theta)^\T) J \right) \geq \frac{\exp\left(-2 \opnorm{C} R \right)}{4K(K-1)}$.
Once we have this, we can extract the inner matrix from $\calI(\theta_\star)$ by lower bounding the Rayleigh quotient, giving:
\begin{align*}
    \mathbb{E}_{z_{0:t} \sim p_{\theta}}[\calI_{t+1}(\theta \mid z_{0:t})] &= \mathbb{E}_{z_{0:t} \sim p_{\theta}} \left[ \bar{M}_{0:t}^\T J^\T \left( \diag(\mathbb{S}(M_{0:t} \theta)) - \mathbb{S}(M_{0:t} \theta) \mathbb{S}(M_{0:t} \theta)^\T \right) J \bar{M}_{0:t} \right] \\
    &\succeq \frac{\exp\left(-2 \opnorm{C} R \right)}{4K(K-1)} \mathbb{E}_{z_{0:t} \sim p_{\theta}} \left[ \bar{M}_{0:t}^\T \bar{M}_{0:t} \right] \\
    &= \frac{\exp\left(-2 \opnorm{C} R \right)}{4K(K-1)} \mathbb{E}_{z_{0:t} \sim p_{\theta}} \left[ (z_t z_t^\T) \otimes \left( \frac{1}{t^2} z_{0:t-1}^\T z_{0:t-1} C^\T C z_{0:t-1}^\T z_{0:t-1} \right) \right] \\
    &= \frac{\exp\left(-2 \opnorm{C} R \right)}{4K(K-1)}  \mathbb{E}_{z_{t} \sim p_{\theta}} [z_t z_t^\T] \otimes \mathbb{E}_{z_{0:t-1} \sim p_{\theta}} \left[ \frac{1}{t^2} z_{0:t-1}^\T z_{0:t-1} C^\T C z_{0:t-1}^\T z_{0:t-1} \right].
\end{align*}
The smallest eigenvalue of this quantity can be lower bounded by the minimum eigenvalue
of both sides of the Kronecker product.
The first quantity can be handled by lower bounding the probability of seeing a particular token,
\begin{align*}
    \lambda_{\min}(\mathbb{E}_{z_{t} \sim p_{\theta}} [z_t z_t^\T]) &= \lambda_{\min} \left( \sum_{k \in [K]} (e_k e_k^\T) \ p_{\theta}(e_k \mid z_{0:t-1}) \right) \\
    &\geq \frac{1}{K} \sigma_{\min}^2 \left( E \right) \exp(-2 \opnorm{C} R).
\end{align*}
For the second quantity, we note that the minimum eigenvalue function is concave so we may lower bound the expression by moving it into the expectation:
\begin{align*}
    \lambda_{\min} &\left( \mathbb{E}_{z_{0:t-1} \sim p_{\theta}} \left[ \frac{1}{t^2} z_{0:t-1}^\T z_{0:t-1} C^\T C z_{0:t-1}^\T z_{0:t-1} \right] \right) \\
    &\geq \sigma^2_{\min}(C) \cdot \lambda_{\min}^2( \mathbb{E}_{z_{0:t-1} \sim p_\theta} [t^{-1} z_{0:t-1}^\T z_{0:t-1}]) \\
    &\geq \sigma^2_{\min}(C) \cdot \frac{1}{t} \sum_{s=0}^{t-1} \lambda_{\min}^2(\mathbb{E}_{z_s \sim p_\theta \mid z_{0:s-1}} [z_s z_s^\T]) \\ 
    &\geq \frac{1}{K} \sigma^2_{\min}(C) \sigma^2_{\min}(E) \exp(-2\opnorm{C} R).
\end{align*}
Finally, we may put this all together for an expression for the minimum eigenvalues of the conditional and unconditional Fisher information matrices:
\begin{align}
    \lambda_{\min} \left( \mathbb{E}_{z_{0:t} \sim p_\theta}[ \calI_{t+1}(\theta \mid z_{0:t})] \right) &\geq \frac{\sigma^2_{\min}(C) \sigma^4_{\min}(E)}{4K^3 (K-1)} \exp(-6 \opnorm{C} R), \label{eq:seq_mod_cond_FI_lmin} \\
    \lambda_{\min} \left( \mathbb{E}_{z_{0:T} \sim p_\theta}[ \calI(\theta)] \right) &\geq (T-1) \frac{\sigma^2_{\min}(C) \sigma^4_{\min}(E)}{4K^3 (K-1)} \exp(-6 \opnorm{C} R). \label{eq:seq_mod_FI_lmin}
\end{align}
We can now begin working on finding bounds for the constants $B_1$ and $B_2$.  
Let us take a test vector $v \in \mathbb{R}^d$ with magnitude $\norm{v} = \opnorm{\calI(\theta_\star)^{1/2}}$ and analyze $\psi_{t+1}$ defined as follows:
\begin{align*}
    \psi_{t+1} &:= v^\T (\nabla_\theta \log p_\theta(z_{t+1} \mid z_{0:t})) \\
    &= v^\T \left( [M_{0:t}]_{(z_{t+1})} - M_{0:t}^\T \mathbb{S}(M_{0:t} \theta) \right).
\end{align*}
We can immediately see that $\mathbb{E}_{p_\theta}[\psi_{t+1} \mid z_{0:t}] = 0$.   
We may observe that $\mathbf{1}(z_{t+1}) - \mathbb{S}(M_{0:t} \theta)$ is a bounded random variable vector:
\begin{align*}
    \norm{\mathbf{1}(z_{t+1}) - \mathbb{S}(M_{0:t} \theta)}^2 &= (1 - \mathbb{S}(M_{0:t} \theta)_{(z_{t+1})})^2 + \sum_{i:e_i \not= z_{t+1}} \mathbb{S}(M_{0:t} \theta)_i^2 \\
    &\leq (1 - \mathbb{S}(M_{0:t} \theta)_{(z_{t+1})})^2 + \left( \sum_{i:e_i \not= z_{t+1}} \mathbb{S}(M_{0:t} \theta)_i \right)^2 \\
    &= 2 (1 - \mathbb{S}(M_{0:t} \theta)_{(z_{t+1})})^2 \leq 2.
\end{align*}
From this we can observe the following:
\begin{align*}
    \psi_{t+1} &= v^\T \left( [M_{0:t}]_{(z_{t+1})} - M_{0:t}^\T \mathbb{S}(M_{0:t} \theta) \right) \\
    &= v^\T M_{0:t}^\T \left( \mathbf{1}(z_{t+1}) - \mathbb{S}(M_{0:t} \theta) \right) \\
    &\leq \norm{v} \opnorm{M_{0:t}} \norm{ \mathbf{1}(z_{t+1}) - \mathbb{S}(M_{0:t} \theta) } \\
    &\leq \sqrt{2} \norm{v} \opnorm{C}.
\end{align*}
This all gives that $\psi_{t+1}$ is a zero-mean bounded random variable given by $\sigma^2 = 2 \norm{v}^2 \opnorm{C}^2$, and we can see that $\langle v, \nabla_\theta \log p_\theta(z_{0:T}) \rangle = \sum_{t=2}^{T} \psi_{t}$ is a martingale sum.  
This allows us to apply Azuma-Hoeffding (\Cref{thm:azuma_hoeffding}), giving us that 
\begin{align*}
    \mathbb{P}\left[ \ip{v}{\nabla_\theta \log p_\theta(z_{0:t})}^4 \geq t \right] &= \mathbb{P}\left[ \sum_{s=2}^t |\psi_s| \geq u^{1/4} \right] \leq 2 \exp \left( -\frac{\sqrt{u}}{4(t-1) \norm{v}^2 \opnorm{C}^2} \right),
\end{align*}
and as such $\mathbb{E}[ \ip{v}{\nabla_\theta \log p_\theta(z_{0:t})}^4]^{1/4} \leq 4 \sqrt{2T} \norm{v} \opnorm{C}$. 
We note that applying Rosenthal's inequality for MDS (cf.~\Cref{thm:rosenthal}) retrieves a similar result, but requires a longer proof in order to express a bound for both terms.
Taking $v = \calI(\theta_\star)^{1/2} \bar{v}$ for unit vector $\bar{v}$ results in $B_1 \leq 4\sqrt{\frac{2T}{\lambda_{\min}(\calI(\theta_\star))}} \opnorm{C}$.

Since the Hessian of the conditional log-likelihood has no dependence on the new token, bounding $B_2$ reduces to $\sup_{\theta \in \Theta} \opnorm{ \calI(\theta)^{-1/2} M_{0:t}^\T ( \diag(\mathbb{S}(M_{0:t} \theta)) - \mathbb{S}(M_{0:t} \theta)\mathbb{S}(M_{0:t} \theta)^\T ) M_{0:t} \calI(\theta)^{-1/2} }$ which may be bounded by $\frac{2}{\lambda_{\min}(\calI(\theta))} \opnorm{C}^2$.

\paragraph{Step 3: Parameter error bound.}
From here, we can unlock the first set of bounds by verifying \eqref{eq:hellinger_perturbation_radius}:
\begin{align*}
    \sup_{\theta \in \mathrm{conv}\{ \hat{\theta}^\e_{m,T}, \theta_\star \}} \HgSq{\theta}{\theta_\star} &\lesssim \frac{d^2}{m} \log \left( \frac{ c_0 R^2  \opnorm{C}^2 m T}{\delta^2} \right) + \frac{1 + \delta^2}{m} \\
    &\lesssim \frac{\lambda_{\min}(\calI(\theta_\star))^2}{T^2 \opnorm{C}^4}.
\end{align*}
In order to satisfy this condition, we can take $m \geq \left( d^2 \log\left(\frac{T R^2 \opnorm{C}^2 m}{\delta^2}\right) + 1 + \delta^2 \right) \cdot \frac{T^2 \opnorm{C}^4}{\lambda_{\min}(\calI(\theta_\star))^2}$. 
If we apply \Cref{prop:log_dominance} and expand out the minimum eigenvalue, for $T > 1$ this gives us: 
\begin{align}
    \label{eq:seqmod_param_error_m}
    m \gtrsim \frac{K^8 \opnorm{C}^4 \exp(12R\opnorm{C})}{\sigma_{\min}(C)^4 \sigma_{\min}(E)^8} \left( 1 + \delta^2 + d^2 R \opnorm{C} + d^2 \log \left( \frac{ c_0 TK^8 d^2 R^2 \opnorm{C}^6}{\delta^2 \sigma_{\min}(C)^4 \sigma_{\min}(E)^8} \right) \right).
\end{align}

\paragraph{Step 4: Verify FI radius.}
We now need to show \eqref{eq:hellinger_perturbation_FI_radius} in order to unlock the second bound.
To this end, we show a Lipschitz condition on the conditional Fisher information from \cref{eq:seq_mod_cond_FI}:
\begin{align*}
    &\opnorm{ \calI_{t+1}(\theta_1 \mid z_{0:t}) - \calI_{t+1}(\theta_2 \mid z_{0:t}) } \\
    &= \opnorm{ M_{0:t}^\T ( \diag(\mathbb{S}(M_{0:t} \theta_1)) - \mathbb{S}(M_{0:t} \theta_1)^{\otimes 2} - \diag(\mathbb{S}(M_{0:t} \theta_2)) + \mathbb{S}(M_{0:t} \theta_2)^{\otimes 2} ) M_{0:t} } \\
    &\leq \opnorm{ M_{0:t} }^2 \left( \| \mathbb{S}(M_{0:t} \theta_1) - \mathbb{S}(M_{0:t} \theta_2) \|_\infty + \opnorm{ \mathbb{S}(M_{0:t} \theta_1)^{\otimes 2} - \mathbb{S}(M_{0:t} \theta_2)^{\otimes 2} } \right) \\
    &\leq  \opnorm{C}^2 \left( \opnorm{M_{0:t}} \norm{\theta_1 - \theta_2} + \left( \norm{\mathbb{S}(M_{0:t} \theta_1)} + \norm{\mathbb{S}(M_{0:t} \theta_2)} \right) \norm{\mathbb{S}(M_{0:t} \theta_1) - \mathbb{S}(M_{0:t} \theta_2)} \right) \\
    &\leq \opnorm{C}^2 \left( \opnorm{C} \norm{\theta_1 - \theta_2} + 2 \norm{\mathbb{S}(M_{0:t} \theta_1) - \mathbb{S}(M_{0:t} \theta_2)} \right) \\
    &\leq 3 \opnorm{C}^3 \norm{\theta_1 - \theta_2} .
\end{align*}
This gives us that $\Lip \leq 3 \opnorm{C}^3$.
Let us now find an expression for the second moment bound:
\begin{align*}
    B_{\calI} &= \sup_{\theta_1, \theta_2 \in \Theta_s} \sup_{v \in \bbS^{p-1}} \max_{t \in [T-1]} \norm{ v^\T \calI_{t+1}(\theta_1 \mid z_{0:{t}}) v}_{\calL^2(p_{\theta_2})} \\
    &\leq \sup_{\theta_1, \theta_2 \in \Theta_s} \sup_{v \in \bbS^{p-1}} \max_{t \in [T-1]} \left( \mathbb{E}_{z_{0:t} \sim p_{\theta_2}} \mathbb{E}_{z_{t+1} \sim p_{\theta_1}} \left[ \left( v^\T M_{0:t}^\top \left( \diag \left( \mathbb{S}(M_{0:t} \theta_1) \right) - \mathbb{S}(M_{0:t} \theta_1)^{\otimes 2} \right) M_{0:t} v \right)^2 \right] \right)^{1/2} \\
    &\leq \opnorm{C}^2.
\end{align*}
With both a globally bounded Lipschitz constant and a finite $B_{\calI}$, we can apply \Cref{prop:FI_lipschitz} to bound the difference in Fisher informations:
\begin{align*}
    \opnorm{ \calI(\theta_1) - \calI(\theta_2) } &\leq 3 T \opnorm{C}^3 \norm{\theta_1 - \theta_2} + 2 \sqrt{2} T \opnorm{C}^2 \Hg{p_{\theta_1}}{p_{\theta_2}}.
\end{align*}
Note that the lower bound on $\lambda_{\min}(\calI(\theta))$ in \eqref{eq:seq_mod_FI_lmin} is agnostic to the value of $\theta \in \Theta$. Therefore,
\begin{align*}
    \lambda_{\min}(\calI(\theta_\star, \hat{\theta})) &\geq (T-1) \frac{\sigma^2_{\min}(C) \sigma^4_{\min}(E)}{4 K^3 (K-1)} \exp(-6 \opnorm{C} R).
\end{align*}
If we apply this to \eqref{eq:hellinger_perturbation_avg_FI}, we can upper bound the parameter distance:
\begin{align*}
    \frac{1}{8} \| \theta_0 - \theta_1 \|^2_{\calI(\theta_0, \theta_1)} &\leq \HgSq{\theta_0}{\theta_1} \lesssim \frac{d^2}{m} \log \left( \frac{c_0 R^2 \opnorm{C}^2 mT}{\delta^2} \right) + \frac{1}{m} + \frac{\delta^2}{m}, \\
    \Longrightarrow \norm{\theta_0 - \theta_1} &\leq \frac{2 \sqrt{2}}{\sqrt{\lambda_{\min}(\calI(\theta_0, \theta_1))}} \Hg{p_{\theta_0}}{p_{\theta_1}} .
\end{align*}
If we put these together, we have the following:
\begin{align*}
    \sup_{\theta \in \mathrm{conv}\{\theta_\star, \hat{\theta}_{m,T}^\e\}} & \opnorm{ \calI(\theta_\star)^{-1/2} \calI(\theta) \calI(\theta_\star)^{-1/2} - I_{d^2} } \\
    &\leq \frac{1}{\lambda_{\min}\left(\calI(\theta_\star)\right)}  \sup_{\theta \in \mathrm{conv}\{\theta_\star, \hat{\theta}_{m,T}^\e\}} \opnorm{ \calI(\theta)  - \calI(\theta_\star) } \\
    &\leq \frac{2 \sqrt{2} T \opnorm{C}^2}{\lambda_{\min}\left(\calI(\theta_\star)\right)} \left( \frac{3 \opnorm{C}}{\sqrt{\lambda_{\min}\left(\calI(\theta_\star)\right)}} + 1 \right)  \sup_{\theta \in \mathrm{conv}\{\theta_\star, \hat{\theta}_{m,T}^\e\}} \Hg{p_{\theta}}{p_{\theta_\star}} .
\end{align*}
If we take $T \gtrsim \frac{K^4 \opnorm{C}^2}{\sigma_{\min}(C)^2 \sigma_{\min}(E)^4} \exp(6 R \opnorm{C})$ such that $\opnorm{C} \lambda_{\min}(\calI(\theta_\star))^{-1/2} \lesssim 1$,
this is bounded above by $1/2$ for:
\begin{align*}
    m &\gtrsim \frac{T^2 \opnorm{C}^4}{\lambda_{\min}(\calI(\theta_\star))^2} \left( d^2 \log \left( \frac{ c_0 R^2 \opnorm{C}^2 mT}{\delta^2} \right) + 1 + \delta^2 \right).
\end{align*}

\paragraph{Step 5: Final result.}
Finally, we can plug in our expression for the minimum eigenvalue in \cref{eq:seq_mod_FI_lmin} to take the following bound 
\begin{align*}
    m &\gtrsim \frac{K^{8} \opnorm{C}^4 \exp(12R\opnorm{C})}{\sigma_{\min}(C)^4 \sigma_{\min}(E)^{8}} \left( d^2 \log \left( \frac{c_0 R^2 \opnorm{C}^2 mT}{\delta^2} \right) + 1 + \delta^2 \right),
\end{align*}
and applying \Cref{prop:log_dominance} extracts $m$ to satisfy the Fisher radius:
\begin{align*}
    m \gtrsim \frac{K^{8} \opnorm{C}^4 \exp(12R\opnorm{C})}{\sigma_{\min}(C)^4 \sigma_{\min}(E)^{8}} \left( 1 + \delta^2 + d^2 R \opnorm{C} + d^2 \log \left( \frac{ c_0 T K^{8} d^2 R^2 \opnorm{C}^6}{\delta^2 \sigma_{\min}(C)^4 \sigma_{\min}(E)^{8}} \right) \right).
\end{align*}
Notably, this additionally satisfies \eqref{eq:seqmod_param_error_m}.
In this case, we unlock the following bound for $\delta \in (0, 1)$:
\begin{align*}
    \| \hat{\theta}_{m,T}^\e - \theta_\star \|_{\bar{\calI}(\theta_\star)} &\leq \frac{32}{3T} \HgSq{\hat{\theta}_{m,T}^\e}{\theta_\star} \\
    &\leq \sup_{\theta \in \mathrm{conv}\{\hat{\theta}^\e_{m,T}, \theta_\star\}} \frac{32}{3T} \HgSq{\theta}{\theta_\star} \\
    &\lesssim \frac{d^2}{mT} \log \left( \frac{ c_0 R^2 \opnorm{C}^2 mT}{\delta^2} \right) + \frac{1 + \delta^2}{mT} \\
    &\lesssim \frac{d^2}{mT} \log \left( \frac{c_0 R^2 \opnorm{C}^2 mT}{\delta^2} \right) .
\end{align*}

\section{Conclusion}
\label{sec:conclusion}

We introduced the Hellinger localization framework for deriving nearly instance-optimal
parameter recovery rates for multi-trajectory learning setups. 
We applied our framework to a diverse set of case studies, including
a mixture of Markov chains example, a dependent linear regression problem with general noise distributions, a non-monotonic sinusoidal GLM example, and a linear attention sequence modeling
setup. In each case, we showed that our Hellinger localization framework was able to 
provide nearly instance-optimal rates that significantly improve upon the prior art. 

Our work further opens up several avenues for future investigation. We list out a few ideas, 
starting with technical improvements, and ending with more broader, high-level directions.
\begin{enumerate}[label=(\alph*)]
    \item \textit{Extensions of self-normalization:} As discussed in
    \Cref{sec:l1_regression:related_work}, one particular drawback of our current
    framework is that it places un-necessary requirements on the regularity of the trajectory process $z_{1:T}$.
    Concretely, in the context of dependent linear regression, the process $z_{1:T}$ can not grow more than 
    $\mathrm{poly}(T)$, which rules out e.g., recovering linear dynamical systems with spectral radius $> 1$.
    We believe this restriction is purely a technical limitation of our argument, which currently
    does not have a method to self-normalize as is done in analysis specialized for least-squares linear regression. The work of \cite{vijaykumar2021localization} which generalizes offset complexity
    to exp-concave losses is a natural starting point for such an inquiry.

    \item \textit{Improved minimum trajectory requirements for non-log-concave families:}
    As we discussed in \Cref{sec:hellinger_localization_framework}, another limitation of our current analysis
    is that whenever the family of distribution $\calP$ is not log-concave, our requirements on
    the number of trajectory $m$ grows from $m \gtrsim \mathrm{polylog}(T)$ in the log-concave setting,
    to $m \gtrsim T \cdot \mathrm{polylog}(T)$. We believe this scaling should generally be improvable. 
    One possible pathway is to utilize the local geodesic convexity of the squared Hellinger distance
    in the Fisher-Rao metric~\cite{khesin2011geometry}, and conduct our second-order Taylor analysis (cf.~\Cref{prop:hellinger_perturbation}) over geodesics.
    
    \item \textit{Non-realizable settings:} Our work is carried out in the realizable setting, i.e., where the data generating distribution $p_\star \in \calP$. A natural and useful extension would be to 
    allow for $p_\star \not\in \calP$, and study convergence to the
    best distribution in $\calP$, i.e., $\theta_\star^\calP := \argmin_{p \in \calP} \KL{p_\star}{p}$.
    One key technical challenge for the non-realizable setting is extending \Cref{thm:hellinger_bound_MLE} to measure squared Hellinger distance $\HgSq{\hat{p}^\e_{m,T}}{p_\star^\calP}$ without relying on e.g., max divergence coverings (cf.~\Cref{thm:hellinger_bound_mle_og}), but instead allowing for some less stringent tail behavior for the log-likelihoods which is still practical to verify. This could also be useful in allowing \Cref{thm:hellinger_bound_MLE} to apply directly to the MLE estimator
    and not its discretized counterpart (cf.~\Cref{rmk:hellinger_new_vs_OG}).

    \item \textit{Applications to non-sequentially dependent data:} 
    While our work focuses on sequentially-ordered stochastic processes,
    our main tools in \Cref{sec:general_framework} (i.e., \Cref{thm:hellinger_bound_MLE}
    and \Cref{prop:hellinger_perturbation}) are actually agnostic
    to this sequential structure.
    It is only when we analyze the score function 
    and observed information matrix moments (i.e., \eqref{eq:hellinger_MDS} and
    \eqref{eq:hellinger_hessian} from \Cref{prop:hellinger_perturbation})
    that we impose a temporal dependence in the data.
    Hence, an interesting future direction
    is to apply our main tools to other problem settings with
    different correlation structures, such as for Ising models (cf.~related work from \Cref{sec:related_work})
    and other graph/network structures~\cite{usunier2005,ralaivola2010noniid,shalizi2013graphmodels}.

    \item \textit{Applications for filtering and control problems:}
    Finally, through the case studies in \Cref{sec:case_studies},
    we have looked at parameter recovery
    in various types of dynamical systems. A natural next step is to consider
    the downstream control task where the recovered model parameters
    would be applied, by extending our results to enable task-specific
    optimal exploration for a broader family of parametric models and
    loss functions (cf.~discussion in \Cref{sec:l1_regression:related_work}).
    Another direction is to apply our framework
    for filtering problems in state estimation, which can be cast as a latent maximum likelihood 
    estimation problems. Here, an important sub-direction would be to 
    study the application of our techniques to analyzing not just the
    exact MLE estimate, but also practical algorithms
    such as expectation-maximization and variational inference, which are
    necessary in situations where directly computing the MLE is computationally
    intractable.
    
\end{enumerate}

\subsubsection*{Acknowledgments}
NM and ES were supported in part by AFOSR Award FA9550-24-1-0102.

{\footnotesize
\bibliographystyle{unsrtnat}
\bibliography{paper}
}

\newpage
\appendix

\section{Additional Results for Hellinger Localization}
\label{sec:appendix:framework}

\begin{myprop}[{cf.~\cite[Lemma F.2]{du2021bilinearRL}}]
\label{prop:log_dominance}
Suppose that $\nu, a, b \geq 0$. Then, we have:
\begin{align*}
    m \geq (1+\nu)^\nu a \log^\nu((1+\nu)^\nu ab) \Longrightarrow m \geq a \log^\nu(bm).
\end{align*}
\end{myprop}

\begin{myprop}
\label{prop:hellinger_tensorization}
Let $p, q$ be two measures. Fix $\delta \in (0, 1)$ and $m \in \N_+$.
We have that:
\begin{align*}
    \Hg{p}{q} \leq \frac{\delta}{\sqrt{2m}} \Longrightarrow \Hg{p^{\otimes m}}{q^{\otimes m}} \leq \delta.
\end{align*}
\end{myprop}
\begin{proof}
We have that
$\HgSq{p^{\otimes m}}{q^{\otimes m}} = 2(1-\rho(p, q)^m)$ where $\rho(p, q) := \int \sqrt{p q} \rmd \mu$ denotes the Bhattacharyya coefficient between $p$ and $q$ (note that $\HgSq{p}{q} = 2(1-\rho(p, q))$).
Let $\rho = \rho(p, q)$.
Note that if $\rho \geq (1-\delta^2/2)^{1/m}$ then we have
$\sqrt{2(1-\rho^m)} \leq \delta$.
On the other hand
since $(1-\delta^2/2)^{1/m} \leq \exp(-\delta^2/(2m))$,
it suffices to take $\rho \geq \exp(-\delta^2/(2m))$.
The latter condition is equal to $\HgSq{p}{q}/2 \leq 1 - \exp(-\delta^2/(2m))$.
Using the inequality $\exp(-x) \leq 1 - x + x^2/2$ valid for $x \in [0, 1]$, we have that:
\begin{align*}
    \exp(-\delta^2/(2m)) \leq 1 - \frac{\delta^2}{2m}\left(1 - \frac{\delta^2}{4m}\right) \leq 1 - \frac{\delta^2}{4m} \Longrightarrow 1-\exp(-\delta^2/(2m)) \geq \frac{\delta^2}{4m}.
\end{align*}
Hence, we have shown that:
\begin{align*}
    \Hg{p}{q} \leq \frac{\delta}{\sqrt{2m}} \Longrightarrow \Hg{p^{\otimes m}}{q^{\otimes m}} \leq \delta.
\end{align*}
\end{proof}

\begin{myprop}
\label{prop:hellinger_change_of_measure}
Let $\mu, \nu$ be two probability measures on the same measure space $\sfX$, and let $f : \sfX \mapsto \R$ be a real-valued function. We have:
\begin{align*}
    \abs{\E_\mu[f] - \E_\nu[f]} \leq \sqrt{2} ( \norm{f}_{L^2(\mu)} + \norm{f}_{L^2(\nu)} ) \Hg{\mu}{\nu}.
\end{align*}
\end{myprop}
\begin{proof}
Let $\lambda$ be a common measure and let $p_\mu, p_\nu$ denote the resulting Radon-Nikodym derivatives. We have:
\begin{align*}
    \E_\mu[f] - \E_\nu[f] &= \int f(x) (p_\mu(x) - p_\nu(x)) \rmd \lambda \\
    &= \int f(x) (\sqrt{p_\mu(x)} + \sqrt{p_\nu(x)})(\sqrt{p_\mu(x)} - \sqrt{p_\nu(x)}) \rmd \lambda \\
    &\leq \sqrt{ \int f(x)^2 (\sqrt{p_\mu(x)} + \sqrt{p_\nu(x)})^2 \rmd \lambda  } \sqrt{ \int (\sqrt{p_\mu(x)} - \sqrt{p_\nu(x)})^2 \rmd \lambda } \\
    &\leq \sqrt{ 2\int f(x)^2 p_\mu(x) \rmd \lambda + 2 \int f(x)^2 p_\nu(x) \rmd \lambda } \cdot \Hg{\mu}{\nu} \\
    &= \sqrt{2}( \norm{f}_{L^2(\mu)} + \norm{f}_{L^2(\nu)} ) \Hg{\mu}{\nu}.
\end{align*}
Above, the first inequality is Cauchy-Schwarz, and the second is
$(a+b)^2 \leq 2(a^2 + b^2)$.
The claim now follows by reversing the role of $\mu, \nu$.
\end{proof}

\begin{myprop}
\label{prop:FI_lipschitz}
For $t \in [T]$, define the conditional Fisher information matrices as:
\begin{align}
    \calI_t(\theta \mid z_{1:t-1}) := -\E_{p_\theta}\left[ \nabla^2 \log p_\theta(z_t \mid z_{1:t-1}) \mid z_{1:t-1} \right]. \label{eq:conditional_FI_matrix}
\end{align}
(We interpret $z_{1:0}$ to condition on no information.)
Let $\Theta' \subseteq \Theta$,
and suppose that the following conditions hold:
\begin{enumerate}[label=(\alph*)]
    \item For all $t \in [T]$, we have for a.e.\ $z_{1:t-1} \in \sfZ^{t-1}$
    and $\theta_1, \theta_2 \in \Theta'$,
    \begin{align*}
        \opnorm{ \calI_t(\theta_1 \mid z_{1:t-1}) - \calI_t(\theta_2 \mid z_{1:t-1}) } \leq \Lip_t(z_{1:t-1}) \norm{\theta_1 - \theta_2}.
    \end{align*}
    \item The following bound on the Lipschitz conditions holds:
    \begin{align*}
        \Lip := \sup_{\theta \in \Theta_s} \max_{t \in [T]} \E_{p_\theta}[ \Lip_t(z_{1:{t-1}}) ] < \infty.
    \end{align*}
    \item The following second moment bound holds:
    \begin{align*}
        B_{\calI} := \sup_{\theta_1, \theta_2 \in \Theta_s} \sup_{v \in \bbS^{p-1}} \max_{t \in [T]} \norm{ v^\T \calI_t(\theta_1 \mid z_{1:{t-1}}) v}_{\calL^2(p_{\theta_2})} < \infty.
    \end{align*}
\end{enumerate}
Then we have:
\begin{align*}
    \opnorm{\calI(\theta_1) - \calI(\theta_2)} \leq T \left[\Lip \cdot \norm{\theta_1-\theta_2} + 2\sqrt{2} B_{\calI} \cdot \Hg{p_{\theta_1}}{p_{\theta_2}} \right].
\end{align*}
\end{myprop}
\begin{proof}
We first use the tower property to decompose $\calI(\theta)$ as:
\begin{align*}
    \calI(\theta) = \sum_{t=1}^{T} \E_{p_\theta}[ \calI_t(\theta \mid z_{1:t-1}) ].
\end{align*}
Hence we have for a fixed unit-norm $v \in \R^p$,
\begin{align*}
    \abs{v^\T(\calI(\theta_1) - \calI(\theta_2))v} &= \bigabs{\sum_{t=1}^{T} \E_{p_{\theta_1}}[ v^\T \calI_t(\theta_1 \mid z_{1:t-1}) v] - \E_{p_{\theta_2}}[ v^\T \calI_t(\theta_2 \mid z_{1:t-1}) v]} \\
    &= \bigabs{\sum_{t=1}^{T} \E_{p_{\theta_1}}[ v^\T (\calI_t(\theta_1 \mid z_{1:t-1}) - \calI_t(\theta_2 \mid z_{1:t-1})) v] + (\E_{p_{\theta_1}} - \E_{p_{\theta_2}})[ v^\T I_t(\theta_2 \mid z_{1:t-1}) v ]} \\
    &\leq \sum_{t=1}^{T}\E_{p_{\theta_1}}[ \Lip_t(z_{1:t-1})] \norm{\theta_1-\theta_2} + \sum_{t=1}^{T} \abs{ (\E_{p_{\theta_1}} - \E_{p_{\theta_2}})[ v^\T I_t(\theta_2 \mid z_{1:t-1}) v ] } \\
    &\leq T \Lip \cdot \norm{\theta_1-\theta_2} + 2\sqrt{2}T B_{\calI} \cdot \Hg{p_{\theta_1}}{p_{\theta_2}},
\end{align*}
where the last inequality holds from
\Cref{prop:hellinger_change_of_measure}.
The claim now follows by the variational form of the operator norm for symmetric matrices.
\end{proof}

\begin{myprop}[{cf.~\cite[Lemma A.4]{foster2021statistical}}]
\label{prop:logMGF_bound}
Let $(X_t)_{t \in \N_+}$ be a sequence of real-valued random variables adapted to a filtration $(\calF_t)_{t \in \N_+}$.
With probability at least $1-\delta$ for all $\tau \in \N_+$:
\begin{align*}
    \sum_{t=1}^{\tau} - \log(\E[ e^{-X_t} \mid \calF_{t-1} ]) \leq \sum_{t=1}^{\tau} X_t + \log(1/\delta).
\end{align*}
By negating $X_t$, the following inequality
also holds with probability at least $1-\delta$ for all $\tau \in \N_+$:
\begin{align*}
    \sum_{t=1}^{\tau} X_t \leq \sum_{t=1}^{\tau} \log( \E[e^{X_t} \mid \calF_{t-1}] ) + \log(1/\delta).
\end{align*}
\end{myprop}

\begin{mythm}[Rosenthal's inequality for MDS, \cite{Burkholder1973,Hitczenko1990}]\label{thm:rosenthal} Let $(d_n)_{n \geq 1}$ be a martingale difference sequence (MDS)
adapted to a filtration $(\calF_n)_{n \geq 1}$. For any $2 \leq p < \infty$,
$$
    \left( \E\bigabs{\sum_{k=1}^{n} d_k}^p \right)^{1/p} \leq C_p\left\{  \left(\E\left(  \sum_{k=1}^{n} \E[ d_k^2 \mid \calF_{k-1} ] \right)^{p/2}\right)^{1/p} + \left( \sum_{k=1}^{n}\E \abs{d_k}^p \right)^{1/p}  \right\},
$$
where the constant $C_p$ only depends on $p$.
\end{mythm}

\begin{mythm}[Azuma-Hoeffding, \cite{wainwright2019high}]
    \label{thm:azuma_hoeffding}
    Let $(\{(D_k, \mathcal{F}_k)\}_{k=1}^\infty)$ be a martingale difference sequence for which there are constants $\{(a_k, b_k)\}_{k=1}^n$ such that $D_k \in [a_k, b_k]$ almost surely for all $k \in [n]$.  Then, for all $t \geq 0$,
    $$
        \mathbb{P}\left[ \left| \sum_{k=1}^n D_k \right| \geq t \right] \leq 2 \exp \left(- \frac{2t^2}{\sum_{k=1}^n (b_k - a_k)^2} \right).
    $$
\end{mythm}

We next restate and prove \Cref{prop:hellinger_FI_upper_bound}.
\HellingerFIUpperBound*
\begin{proof}
For $\theta \in \Theta$ and $z \in \sfZ^T$, define $h(\theta; z) := \sqrt{p_\theta(z)}$.
We take the first derivative of
$\theta \mapsto h(\theta; z)$:
\begin{align*}
    \nabla_\theta h(\theta; z) &= \frac{1}{2} \sqrt{p_\theta(z)} \nabla_\theta \log p_\theta(z).
\end{align*}
The integral form of Taylor's theorem yields for $\mu^{\otimes T}$ a.e.\ $z \in \sfZ^T$:
\begin{align*}
    h(\theta_1; z) = h(\theta_0; z) + \int_0^1 \ip{\nabla_\theta h(\theta(s); z)}{\theta_1 - \theta_0} \rmd s, \quad \theta(s) := (1-s)\theta_0 + s \theta_1.
\end{align*}
Hence, we have, overloading $\mu = \mu^{\otimes T}$,
\begin{align*}
    \HgSq{p_{\theta_0}}{p_{\theta_1}} &= \int (h(\theta_1; z) - h(\theta_0; z))^2 \rmd \mu \\
    &= \int \left( \int_0^1 \ip{\nabla_\theta h(\theta(s); z)}{\theta_1 - \theta_0} \rmd s \right)^2 \rmd \mu \\
    &\leq \int \int_0^1 (\ip{\nabla_\theta h(\theta(s); z)}{\theta_1 - \theta_0})^2 \rmd s \rmd \mu &&\text{[Jensen's inequality]} \\
    &= \int_0^1 \int (\ip{\nabla_\theta h(\theta(s); z)}{\theta_1 - \theta_0})^2 \rmd \mu \rmd s &&\text{[Fubini's lemma]} \\
    &= \frac{1}{4} \int_0^1 \int (\ip{\nabla_\theta \log p_{\theta(s)}(z)}{\theta_1-\theta_0})^2 p_{\theta(s)}(z) \rmd \mu \rmd s \\
    &= \frac{1}{4} \FISq{p_{\theta_0}}{p_{\theta_1}}.
\end{align*}
\end{proof}

\begin{myprop}[Hellinger identifiability under general conditions]
\label{prop:hellinger_identifiability_generic}
In addition to the assumptions on $\calP = \{ p_\theta \mid \theta \in \Theta\}$ stated in \Cref{sec:general_framework:MLE}, assume that $\Theta$ is compact, $\calI(\theta_\star)$ has full rank, and that the map
$\theta \mapsto \HgSq{\theta}{\theta_\star}$ has Lipschitz Hessians.
Then, $\calP$ is $(\gamma_1, \gamma_2)$-identifiable (cf.~\Cref{def:identifiability}) for some positive $\gamma_1,\gamma_2$.
\end{myprop}
\begin{proof}
For every $\delta > 0$, we define the set
$\Theta_\delta := \Theta \cap \{ \theta \in \R^p \mid \norm{\theta - \theta_\star} \geq \delta \}$.
As $\Theta_\delta$ is the intersection of two closed sets, $\Theta_\delta$ is closed. Furthermore, since $\Theta_\delta \subset \Theta$, it is also bounded, and hence compact.
Define $r(\theta) := \HgSq{\theta}{\theta_\star}$,
and $r_\delta := \inf\{ r(\theta) \mid \theta \in \Theta_\delta \}$.
Since $r_\delta$ is the infimum of a continuous function over a compact set, the infimum is achieved by some $\theta_\delta \in \Theta_\delta$. Furthermore, since $\theta_\delta \neq \theta_\star$ by definition, we must have that $r_\delta = r(\theta_\delta) > 0$ 
(since $r(\theta) = 0$ iff $p_\theta = p_{\theta_\star}$, and we assumed that $p_{\theta_\star}$ is uniquely represented in $\calP$, i.e., $p_\theta = p_{\theta_\star}$ iff $\theta=\theta_\star$).
Hence, this yields the conclusion:
\begin{align*}
    \forall \theta \in \Theta,\,\, r(\theta) \leq r_\delta/2 \Longrightarrow \norm{\theta-\theta_\star} \leq \delta.
\end{align*}
By the star-convexity assumption of $\Theta$ around $\theta_\star$, the function $r(\theta)$ is $C^2$ on the entire ray between $\theta$ and $\theta_\star$, and hence by the integral form of Taylor's theorem we obtain with $\Delta := \theta - \theta_\star$:
\begin{align*}
    r(\theta) &= r(\theta_\star) + \ip{\nabla r(\theta_\star)}{\Delta} + \int_0^1 (1-t) \Delta^\T \nabla^2 r((1-t)\theta_\star + t\theta) \Delta \rmd t \\
    &\stackrel{(a)}{=} \frac{1}{2} \Delta^\T \nabla^2 r(\theta_\star) \Delta + \int_0^1 (1-t) \Delta^\T [ \nabla^2 r((1-t)\theta_\star + t\theta) - \nabla^2 r(\theta_\star) ] \Delta \rmd t \\
    &\stackrel{(b)}{\geq} \frac{1}{2} \Delta^\T \nabla^2 r(\theta_\star) \Delta - L\int_0^1 t(1-t) \norm{\Delta}^3 \rmd t = \frac{1}{2} \Delta^\T \nabla^2 r(\theta_\star) \Delta - \frac{L}{6} \norm{\Delta}^3.
\end{align*}
where (a) holds since $r(\theta_\star)=0$ and $\nabla r(\theta_\star) = 0$ and (b) holds by the assumption that $r(\theta)$ has Lipschitz Hessians.
Suppose now that that $\nabla^2 r(\theta_\star)$ is non-degenerate; we will check this momentarily.
Hence, we have:
\begin{align*}
    \norm{\Delta} \leq \delta_0 := \frac{3\lambda_{\min}(\nabla^2 r(\theta_\star))}{2} \Longrightarrow r(\theta) \geq \frac{\lambda_{\min}(\nabla^2 r(\theta_\star))}{4} \norm{\Delta}^2.
\end{align*}
We therefore have:
\begin{align*}
    r(\theta) \leq r_{\delta_0/2} \,\,\Longrightarrow\,\, \norm{\Delta} \leq \delta_0 \,\,\Longrightarrow\,\, \norm{\Delta}^2 \leq \frac{4}{\lambda_{\min}(\nabla^2 r(\theta_\star))} r(\theta),
\end{align*}
which shows the desired Hellinger identifiability.
To finish the proof, we confirm that $\nabla^2 r(\theta_\star)$ is
non-degenerate.
A standard computation (as done in the 
proof of \Cref{prop:hellinger_perturbation})
shows that $\nabla^2 r(\theta_\star) = \frac{1}{2} \calI(\theta_\star)$, from which the proof concludes.
\end{proof}

\begin{myprop}
\label{prop:log_mgf_sub_exp}
Suppose that $Y$ is a zero-mean random variable
satisfying for some $\theta \in (0, 1)$ and $\alpha > 0$,
\begin{align*}
    \log \E[\exp(\lambda Y)] \leq h(\lambda) := \alpha \log\left( e^\lambda \theta + 1-\theta \right) - \alpha \theta \lambda , \quad \lambda \in \R.
\end{align*}
Then we have with probability at least $1-\delta$,
\begin{align*}
    \abs{Y} \leq 2 \sqrt{2e\theta(1-\theta) \alpha \log(2/\delta)} + 2 \log(2/\delta).
\end{align*}
\end{myprop}
\begin{proof}
We first differentiate $h(\lambda)$ twice:
\begin{align*}
    h'(\lambda) = \frac{\alpha \theta e^\lambda}{\theta e^\lambda + (1-\theta)} - \alpha\theta, \quad h''(\lambda) = \frac{\alpha \theta(1-\theta) e^\lambda}{(\theta e^\lambda + (1-\theta))^2}.
\end{align*}
We next observe that:
\begin{align*}
    \abs{\lambda} \leq 1 \Longrightarrow \frac{e^\lambda}{(\theta e^\lambda + (1-\theta))^2} \leq \min\left\{ \frac{1}{\theta^2 e^\lambda}, \frac{e^\lambda}{(1-\theta)^2}  \right\} \leq e \min\left\{ \frac{1}{\theta^2}, \frac{1}{(1-\theta)^2} \right\} \leq 4e.
\end{align*}
Hence by a second order Taylor expansion of $h(\lambda)$:
\begin{align*}
    \abs{\lambda} \leq 1 \Longrightarrow h(\lambda) &\leq h(0) + h'(0) \lambda + \frac{1}{2} \sup_{\abs{c} \leq 1} h''(c) \lambda^2 
    \leq 2e \alpha \theta(1-\theta) \lambda^2.
\end{align*}
Therefore $Y$ is $(2\sqrt{e\alpha\theta(1-\theta)}, 1)$-sub-Exponential~\cite[see e.g.,][Chapter 2]{wainwright2019high}, and hence from using a sub-Exponential tail bound \cite[Proposition 2.9]{wainwright2019high} we have with probability at least $1-\delta$,
\begin{align*}
    \abs{Y} \leq \sqrt{8 e \alpha \theta(1-\theta) \log(2/\delta)} + 2 \log(2/\delta).
\end{align*}
\end{proof}

\section{Additional Derivations for Two-State Markov Chain Example}
\label{sec:appendix:two_state}

\paragraph{Log-likelihood and FI matrix computations.}
We start with the expression for the log-likelihood.
The conditional log-likelihood is:
\begin{align*}
    \log p_\theta(z' \mid z) = \log{\theta} \cdot \ind\{ z' = z \} + \log(1-\theta) \cdot \ind \{ z' \neq z \}.
\end{align*}
Hence, we have:
\begin{align*}
    \log p_\theta(z_{1:T}) = \log{\theta} \cdot N_{\mathrm{stay}}(z_{1:T}) + \log(1-\theta) \cdot N_{\mathrm{switch}}(z_{1:T}) + \log\rho_1(z_1),
\end{align*}
where $N_{\mathrm{stay}}(z_{1:T}) := \sum_{t=1}^{T-1} \ind\{ z_{t+1} = z_t \}$ and $N_{\mathrm{switch}}(z_{1:T}) := T - 1 - N_{\mathrm{stay}}(z_{1:T})$.
We immediately see that $\calP$ is log-concave (cf.~\Cref{def:log_concave}).
We next compute the first and second derivatives of the
conditional log-likelihood:
\begin{align*}
    \partial_\theta \log p_\theta(z' \mid z) &= \frac{1}{\theta} \ind\{ z' = z \} - \frac{1}{1-\theta} \ind\{ z' \neq z \}, \\
    \partial^2_\theta \log p_\theta(z' \mid z) &= - \left[ \frac{1}{\theta^2} \ind\{z' = z\} + \frac{1}{(1-\theta)^2} \ind\{z' \neq z\} \right].
\end{align*}
Taking conditional expectations,
\begin{align*}
    \E_{p_\theta}[ \partial_\theta \log p_\theta(z_{t+1} \mid z_t) \mid z_t] = 0, \quad \E_{p_\theta}[ \partial^2_\theta \log p_\theta(z_{t+1} \mid z_t) ] = - \frac{1}{\theta(1-\theta)}.
\end{align*}
Hence, the FI matrix is:
\begin{align*}
    \calI(\theta) = - \E_{p_\theta}[ \partial^2_\theta \log p_\theta(z_{1:T}) ] = - \sum_{t=1}^{T-1} \E_{p_\theta}[ \partial^2_\theta \log p_\theta(z_{t+1} \mid z_t) ] = \frac{T-1}{\theta(1-\theta)}.
\end{align*}
Hence, for every $\theta \in \Theta$, we have:
\begin{align*}
    \calI(\theta) = \frac{T-1}{\theta(1-\theta)} \leq \frac{T-1}{\mu(1-\mu)} =: \calI_{\max}.
\end{align*}
We also have the bound
$\mathrm{diam}(\Theta) \leq \frac{T-1}{\mu(1-\mu)}$.

\paragraph{Moment computations.}
We now turn to the moment computations of
$\E_{p_\theta}[ (\partial_\theta \log p_\theta(z_{1:T}))^4 ]$
and $\E_{p_\theta}[ (\partial_\theta^2 \log p_\theta(z_{1:T}))^2 ]$.
Using the expressions for the conditional log-likelihoods, we have that:
\begin{align*}
    \partial_\theta \log p_\theta(z_{1:T}) &= \frac{1}{\theta} N_{\mathrm{stay}}(z_{1:T}) - \frac{1}{1-\theta} (T-1-N_{\mathrm{stay}}(z_{1:T})), \\
    \partial_\theta^2 \log p_\theta(z_{1:T}) &= -\left[ \frac{1}{\theta^2} N_{\mathrm{stay}}(z_{1:T}) + \frac{1}{(1-\theta)^2} (T-1-N_{\mathrm{stay}}(z_{1:T})) \right].
\end{align*}
We also observe that $N_{\mathrm{stay}}(z_{1:T}) \sim \mathrm{Bin}(T-1, \theta)$ when $z_{1:T} \sim p_\theta$.
Hence, utilizing standard moment expressions for binomial distributions,
it is straightforward to compute:
\begin{align*}
    \E_{p_\theta}[ (\partial_\theta \log p_\theta(z_{1:T}))^4 ] &= (T-1)\left(\frac{1}{\theta^3} + \frac{1}{(1-\theta)^3}\right) + 3(T-1)(T-2)\frac{1}{\theta^2(1-\theta)^2}, \\
    \E_{p_\theta}[ (\partial_\theta^2 \log p_\theta(z_{1:T}))^2 ] &= (T-1) \left( \frac{1}{\theta^3} + \frac{1}{(1-\theta)^3}\right) + (T-1)(T-2) \frac{1}{\theta^2(1-\theta)^2}.
\end{align*}

\paragraph{Direct analysis.}
Let $\psi^{(i)} := N_{\mathrm{stay}}(z_{1:T}^{(i)})$ for $i \in [m]$ and $T' := T-1$.
We know that $\psi^{(i)} \sim \mathrm{Bin}(T', \theta_\star)$.
Hence, its MGF is given as $\E[\exp(\lambda \psi^{(i)})] = (\theta_\star e^\lambda + (1-\theta_\star))^{T'}$ for $\lambda \in \R$.
Since $\psi^{(i)}$ are iid across $i \in [m]$, we have
$\E[ \exp(\lambda \sum_{i=1}^{m} \psi^{(i)}) ] = (\theta_\star e^\lambda + (1-\theta_\star))^{mT'}$.
Hence, defining $Y := \sum_{i=1}^{m} \psi^{(i)} - mT' \theta_\star$,
we have that $\E[Y] = 0$ and 
$$
    \log \E[ \exp(\lambda Y) ] = mT' \log(\theta_\star \e^\lambda + (1-\theta_\star)) - mT' \theta_\star \lambda, \quad \lambda \in \R.
$$
From \Cref{prop:log_mgf_sub_exp}, with probability at least $1-\delta$ (over $\calD_{m,T}$),
we have \begin{align*}
    \abs{Y} \lesssim \sqrt{mT \sigma_\star^2 \log(2/\delta)} + \log(2/\delta).
\end{align*}
Dividing both sides by $mT'$, we have with probability at least $1-\delta$,
\begin{align*}
    \abs{\hat{\theta}_{m,T} - \theta_\star} \lesssim \sqrt{ \frac{\sigma_\star^2 \log(1/\delta)}{mT}} + \frac{\log(1/\delta)}{mT}.
\end{align*}
Call this event $\calE_2$.
From this, we finally obtain:
\begin{align*}
    mT \gtrsim \sigma^{-2}_\star \log(1/\delta) \Longrightarrow \abs{\hat{\theta}_{m,T} - \theta_\star}^2 \lesssim \frac{\sigma_\star^2 \log(1/\delta)}{mT} \textrm{ on $\calE_2$},
\end{align*}
which is the claimed rate in \eqref{eq:two_state_optimal_rate}.

\section{Additional Results for Mixture of Two-State Markov Chains}
\label{sec:appendix:mixture}

\begin{myprop}\label{prop:blockmajorization}Given real matrix $A\in \R^{kd\times kd}$ with
$$
A =
\begin{pmatrix}
A_{11} & \cdots & A_{1k}  \\
\vdots & \ddots & \vdots \\
A_{k1}  & \cdots & A_{kk}
\end{pmatrix},\quad \forall\;1\leq i,j\leq k,\;A_{ij}\in \R^{d\times d},
$$
and a real symmetric matrix $A'\in\R^{kd\times kd}$ with
$$
A' =
\begin{pmatrix}
A'_{11} & \cdots & A'_{1k}  \\
\vdots & \ddots & \vdots \\
A_{1k}^{\T}  & \cdots & A_{kk}
\end{pmatrix},\quad \forall\;1\leq i,j\leq k,\;A'_{ij}\in \R^{d\times d},\;A'_{ii}\text{ symmetric},
$$
such that for any $1\leq i,j\leq d$, $\opnorm{A_{ij}}\leq \opnorm{A'_{ij}}$, we have
$$
A\preceq \blkdiag\left\{\left(\sum_{j=1}^{k}\opnorm{A'_{1j}}\right)I_{d},\;\ldots,\;\left(\sum_{j=1}^{k}\opnorm{A'_{kj}}\right)I_{d}\right\}.
$$
\end{myprop}
\begin{proof}
Given a test vector $v\in \R^{kd}$, we decompose $v$ into corresponding blocks:
$$
v=\begin{pmatrix} v_{1}^{\T} & \ldots & v_{k}^{\T}\end{pmatrix}^{\T},\quad \forall\;1\leq i\leq k,\;v_{i}\in \R^{d}.
$$
Then we have:
\begin{align*}
v^{\T}Av&=\sum_{i=1}^{k}v_{i}^{\T}A_{ii}v_{i} +2\sum_{1\leq i<j\leq k}v_{i}^{\T}A_{ij}v_{j}\\
&\stackrel{(a)}{\leq} \sum_{i=1}^{k}\opnorm{A'_{ii}}\norm{v_{i}}^{2}+\sum_{1\leq i<j\leq k}\opnorm{A'_{ij}}\norm{v_{i}}\norm{v_{j}}\\
&\stackrel{(b)}{\leq} \sum_{i=1}^{k}\opnorm{A'_{ii}}\norm{v_{i}}^{2}+\sum_{1\leq i<j\leq k}\opnorm{A'_{ij}}\left(\norm{v_{i}}^{2}+\norm{v_{j}}^{2}\right)\\
&\stackrel{(c)}{\leq }\sum_{i=1}^{k}\opnorm{A'_{ii}}\norm{v_{i}}^{2}+\sum_{1\leq i\neq j\leq k}\opnorm{A'_{ij}}\norm{v_{i}}^{2}\\
&=\sum_{i=1}^{k}\left(\sum_{j=1}^{k}\opnorm{A'_{ij}}\right)\norm{v_{i}}^{2}\\
&=v^{\T}\blkdiag\left\{\left(\sum_{j=1}^{k}\opnorm{A'_{1j}}\right)I_{d},\;\ldots,\;\left(\sum_{j=1}^{k}\opnorm{A'_{kj}}\right)I_{d}\right\}v,
\end{align*}
where for (a) we applied the Cauchy-Schwarz inequality, for (b) we applied the AM-GM inequality, and for (c) we used the fact that for any square matrix $M$, $\opnorm{M}=\opnorm{M^{\T}}$.
\end{proof}

\begin{mycorollary}\label{prop:diagmajorization}Given real matrix $A\in \R^{d\times d}$ and real symmetric matrix $A'\in\R^{d\times d}$ such that for any $1\leq i,j\leq d$, $\left|A_{ij}\right|\leq A'_{ij}$, we have
$$
A\preceq \diag\left\{\sum_{j=1}^{d}A'_{1j},\;\ldots,\;\sum_{j=1}^{d}A'_{dj}\right\}.
$$
\end{mycorollary}
\begin{proof}
This is a special case of \Cref{prop:blockmajorization} with $k=1$.
\end{proof}

\begin{myprop}\label{prop:loewnerprecond} Given positive definite matrices $A_{1}$, $A_{2}\in\R^{d\times d}$, a vector-valued function $M_{1}:\;\mathbb{X}\to\R^{d}$, and a matrix valued function $M_{2}:\;\mathbb{X}\to\R^{d\times d}$. Assume that
$
A_{1}\succeq A_{2}.
$ Let $\psi:\;\R^{\mathbb{X}}\to\R_{\geq 0}$ be a non-negative linear functional. Then we have
\begin{align*}
&\sup_{\norm{v}=1}\psi\left( v^{\T}A^{-1/2}_{1}M_{1}(x)\right)\leq \sup_{\norm{v}=1}\psi\left( v^{\T}A^{-1/2}_{2}M_{1}(x)\right),\\
&\sup_{\norm{v}=1}\psi\left( v^{\T}A^{-1/2}_{1}M_{2}(x)A_{1}^{-1/2}v\right)\leq \sup_{\norm{v}=1}\psi\left( v^{\T}A^{-1/2}_{2}M_{2}(x)A_{2}^{-1/2}v\right).
\end{align*}
In particular, we have for any $p,q>0$, suppose 
\begin{align*}
\sup_{\norm{v}=1}\left\Vert v^{\T}A^{-1/2}_{2}M_{1}(x)\right\Vert_{\calL^{p}(\mu)}<\infty,\quad\sup_{\norm{v}=1}\left\Vert v^{\T}A^{-1/2}_{2}M_{2}(x)A_{2}^{-1/2}v\right\Vert_{\calL^{q}(\mu)}<\infty,
\end{align*}
then,
\begin{align*}
&\sup_{\norm{v}=1}\left\Vert v^{\T}A^{-1/2}_{1}M_{1}(x)\right\Vert_{\calL^{p}(\mu)}\leq \sup_{\norm{v}=1}\left\Vert v^{\T}A^{-1/2}_{2}M_{1}(x)\right\Vert_{\calL^{p}(\mu)},\\
&\sup_{\norm{v}=1}\left\Vert v^{\T}A^{-1/2}_{1}M_{2}(x)A_{1}^{-1/2}v\right\Vert_{\calL^{q}(\mu)}\leq \sup_{\norm{v}=1}\left\Vert v^{\T}A^{-1/2}_{2}M_{2}(x)A_{2}^{-1/2}v\right\Vert_{\calL^{q}(\mu)}.
\end{align*}
\end{myprop}
\begin{proof}
Denote $u:=A_{1}^{-1/2}v$, by symmetricity, we have
\begin{align*}
\sup_{\norm{v}=1}\psi\left( v^{\T}A^{-1/2}_{1}M_{1}(x)\right)&=\sup_{\norm{v}=1}\psi \left( \left(A^{-1/2}_{1}v\right)^{\T}M_{1}(x)\right)\\
&=\sup_{\norm{A_{1}^{1/2}u}=1}\psi\left( u^{\T}M_{1}(x)\right)\\
&=\sup_{\norm{u}_{A_{1}}=1}\psi\left( u^{\T}M_{1}(x)\right)\\
&=\sup_{u}\psi\left( \left(\frac{u}{\norm{u}_{A_{1}}}\right)^{\T}M_{1}(x)\right)\\
&=\sup_{u}\frac{\psi\left( u^{\T}M_{1}(x)\right)}{\sqrt{u^{\T}A_{1}u}}\\
&\leq \sup_{u}\frac{\psi\left( u^{\T}M_{1}(x)\right)}{\sqrt{u^{\T}A_{2}u}}=\sup_{\norm{v}=1}\psi\left( v^{\T}A^{-1/2}_{2}M_{1}(x)\right),
\end{align*}
and similarly
\begin{align*}
\sup_{\norm{v}=1}\psi\left( v^{\T}A^{-1/2}_{1}M_{2}(x)A_{1}^{-1/2}v\right)&=\sup_{\norm{v}=1}\psi \left( \left(A^{-1/2}_{1}v\right)^{\T}M_{2}(x)\left(A_{1}^{-1/2}v\right)\right)\\
&=\sup_{\norm{A_{1}^{1/2}u}=1}\psi\left( u^{\T}M_{2}(x)u\right)\\
&=\sup_{\norm{u}_{A_{1}}=1}\psi\left( u^{\T}M_{2}(x)u\right)\\
&=\sup_{u}\psi\left( \left(\frac{u}{\norm{u}_{A_{1}}}\right)^{\T}M_{2}(x)\left(\frac{u}{\norm{u}_{A_{1}}}\right)\right)\\
&=\sup_{u}\frac{\psi\left( u^{\T}M_{2}(x)u\right)}{u^{\T}A_{1}u}\\
&\leq \sup_{u}\frac{\psi\left( u^{\T}M_{2}(x)u\right)}{u^{\T}A_{2}u}=\sup_{\norm{v}=1}\psi\left( v^{\T}A^{-1/2}_{2}M_{2}(x)A_{1}^{-1/2}v\right).
\end{align*}
\end{proof}

\begin{myprop}\label{prop:loewnerprecond2}
Let $\Delta$ be a $d \times d$ matrix, and $M, N$ be two $d \times d$ positive definite matrices satisfying $M \succcurlyeq N$.
We have:
\begin{align*}
    \opnorm{ M^{-1/2} \Delta M^{-1/2} } \leq \opnorm{ N^{-1/2} \Delta N^{-1/2} }.
\end{align*}
\end{myprop}
\begin{proof}
Observe that:
\begin{align*}
    \opnorm{M^{-1/2} \Delta M^{-1/2}}^2 &= \lambda_{\max}( M^{-1/2} \Delta^\T M^{-1} \Delta M^{-1/2} ) \\
    &\stackrel{(a)}{\leq} \lambda_{\max}( M^{-1/2} \Delta^\T N^{-1} \Delta M^{-1/2} ) \\
    &= \lambda_{\max}( M^{-1/2} \Delta^\T N^{-1/2} \cdot N^{-1/2} \Delta M^{-1/2} ) \\
    &\stackrel{(b)}{=} \lambda_{\max}( N^{-1/2} \Delta M^{-1/2} \cdot M^{-1/2} \Delta^\T N^{-1/2}  ) \\
    &= \lambda_{\max}( N^{-1/2} \Delta M^{-1} \Delta^\T N^{-1/2} ) \\
    &\stackrel{(c)}{\leq} \lambda_{\max}( N^{-1/2} \Delta N^{-1} \Delta^\T N^{-1/2} ) \\
    &= \opnorm{ N^{-1/2} \Delta N^{-1/2} }^2,
\end{align*}
where (a) follows since $M \succcurlyeq N \succ 0$ implies
$N^{-1} \succcurlyeq M^{-1}$~\cite[see e.g.,][Prop.~V.1.6]{bhatia2013matrix}
and so conjugating both sides of the latter inequality by $\Delta M^{-1/2}$ yields
$M^{-1/2} \Delta^\T M^{-1} \Delta M^{-1/2} \preccurlyeq M^{-1/2} \Delta^\T N^{-1} \Delta M^{-1/2}$,
(b) uses $\lambda(A B) = \lambda(B A)$ for two square matrices $A, B$, where $\lambda(\cdot)$ refers to spectrum of its argument 
and (c) uses the same argument as (a).

\end{proof}

\section{Hellinger Identifiability for Sinusoidal GLMs}
\label{sec:appendix:glm}
Here we cover the necessary Gaussian anti-concentration results and local identifibility needed in the proof of \Cref{thm:sin_GLM}.
We believe these set of results to be of independent interest.

\begin{myprop}
\label{prop:cos_anticoncentration}
Let $a \in \R$, $t \in (0, 1)$, and $\sigma > 0$. We have for $g \sim \sfN(0, 1)$,
\begin{align*}
    \Pr_g( |\cos(\sigma g + a)| \leq t ) \leq \left(1 + \frac{3\sqrt{\pi/2}}{\sigma} \right) \left(1 - \frac{2\cos^{-1}(t)}{\pi} \right).
\end{align*}
Hence there exists a universal constant $c_0$ such that:
\begin{align*}
    \forall t > 0, \,\, \Pr_g( |\cos(\sigma g + a)| \leq t ) \leq c_0 \max\{1, 1/\sigma\} t.
\end{align*}
Note that since $\cos(x) = \sin(x + \pi/2)$, the same result also holds for $\sin$ in place of $\cos$.
\end{myprop}
\begin{proof}
Fix a $t \in (0, 1)$.
For $k \in \bbZ$, define the interval:
\begin{align*}
    I_k := \left[k\pi + \cos^{-1}(t), (k+1)\pi - \cos^{-1}(t)\right].
\end{align*}
Since these intervals are disjoint and their union
covers the $t$-sub-level set of $\abs{\cos(x)}$, i.e.,
\begin{align*}
    \bigsqcup_{k \in \bbZ} I_k = \{ x \in \R \mid \abs{\cos(x)} \leq t \},
\end{align*}
we have that, letting $X \sim \sfN(a, \sigma^2)$,
\begin{align*}
    \Pr_{g}( \abs{\cos(\sigma g + a)} \leq t ) = \Pr\left( X \in \bigsqcup_{k \in \bbZ} I_k \right) = \sum_{k \in \bbZ} \Pr(X \in I_k) = \sum_{k \in \mathbb{Z}} \int_{I_k} \frac{1}{\sqrt{2\pi\sigma^2}} e^{-\frac{(x - a)^2}{2\sigma^2}} \rmd x.
\end{align*}
Define $k_0 := \inf\{ k \in \mathbb{Z} \mid k\pi + \cos^{-1}(t) \geq a \}$,
and $\phi_k(x) := x - (k - k_0) \pi$ for $k \geq k_0$.
Now for any $k \geq k_0$ and $x \in I_k$, we have that
$\phi_k(x) \in I_{k_0}$, and furthermore:
\begin{align*}
    \frac{\exp( -(x - a)^2 / (2\sigma^2) )}{\exp( -(\phi_k(x) - a)^2/(2\sigma^2))} &= \exp\left( -\frac{1}{2\sigma^2} \left[  (x-a)^2 - (\phi_k(x)-a)^2  \right]  \right) \\
    &= \exp\left( -\frac{1}{2\sigma^2} \left[ ((x-a) + (\phi_k(x) - a))( x - \phi_k(x)   )    \right]   \right) \\
    &= \exp\left( - \frac{(k-k_0)\pi}{2\sigma^2}((x-a) + (\phi_k(x) - a))  \right) \\
    &\leq \exp\left( - \frac{(k-k_0)^2\pi^2}{2\sigma^2} \right).
\end{align*}
The last inequality holds since: (a) we know that $\phi_k(x) - a \geq 0$ since $\phi_k(x) \in I_{k_0}$ and every $x \in I_{k_0}$ satisfies $x \geq a$ by definition,
and (b) since $x \in I_k$ and $k \geq k_0$, we know that $x-a = (k-k_0)\pi + \phi_k(x) - a \geq (k-k_0)\pi$.
Hence, for each $k \geq k_0$,
\begin{align*}
    \Pr(X \in I_k) &= \int_{I_k} \frac{1}{\sqrt{2\pi\sigma^2}} e^{-\frac{(x-a)^2}{2\sigma^2}} \rmd x \leq e^{-(k-k_0)^2\pi^2/(2\sigma^2)} \int_{I_k} \frac{1}{\sqrt{2\pi\sigma^2}}  e^{-\frac{(\phi_k(x)-a)^2}{2\sigma^2}} \rmd x \\
    &= e^{-(k-k_0)^2\pi^2/(2\sigma^2)} \Pr( X \in I_{k_0}) \leq e^{-(k-k_0)^2\pi^2/(2\sigma^2)} \sup_{k \in \bbZ} \Pr( X \in I_k ).
\end{align*}
We now consider the case when $k \leq \bar{k}_0 := k_0 - 2$.
We next define $\bar{\phi}_k(x) := x + (\bar{k}_0 - k) \pi$
for $k \leq \bar{k}_0$.
Similar to before, we have that for any $k \leq \bar{k}_0$ and
$x \in I_k$, $\bar{\phi}_k(x) \in I_{\bar{k}_0}$.
Also similar to before, we can show that:
\begin{align*}
    \frac{\exp( -(x - a)^2 / (2\sigma^2) )}{\exp( -(\bar{\phi}_k(x) - a)^2/(2\sigma^2))} \leq \exp\left( -\frac{(\bar{k}_0-k)^2 \pi^2}{2\sigma^2} \right),
\end{align*}
and hence for $k \leq \bar{k}_0$,
\begin{align*}
    \Pr(X \in I_k) \leq e^{-(\bar{k}_0-k)^2\pi^2/(2\sigma^2)} \Pr( X \in I_{\bar{k}_0}) \leq e^{-(\bar{k}_0-k)^2\pi^2/(2\sigma^2)} \sup_{k \in \bbZ} \Pr( X \in I_k ).
\end{align*}
Consequently,
\begin{align*}
    \sum_{k \in \bbZ} \Pr(X \in I_k) &= \sum_{k \leq \bar{k}_0}\Pr(X \in I_k) + \sum_{k \geq k_0} \Pr(X \in I_k) + \Pr(X \in I_{k_0 - 1}) \\
    &\leq \sup_{k \in \bbZ} \Pr(X \in I_k) \left[ 1 + \sum_{k \leq \bar{k}_0}e^{-(\bar{k}_0-k)^2\pi^2/(2\sigma^2)} + \sum_{k \geq k_0}e^{-(k-k_0)^2\pi^2/(2\sigma^2)}  \right] \\
    &=  \sup_{k \in \bbZ} \Pr(X \in I_k) \left[ 1 + 2 \sum_{k \in \N} e^{-k^2\pi^2/(2\sigma^2)} \right].
\end{align*}
Next since $x \mapsto e^{-x^2 \pi^2/(2\sigma^2)}$ is decreasing on $\R_{\geq 0}$ we have that
\begin{align*}
    \sum_{k \in \N}  e^{-k^2\pi^2/(2\sigma^2)} = 1 + \sum_{k \geq 1}  e^{-k^2\pi^2/(2\sigma^2)} \leq 1 + \int_0^\infty  e^{-x^2\pi^2/(2\sigma^2)} \rmd x = 1 + \frac{\sigma}{\sqrt{2\pi}}.
\end{align*}
Since $\abs{I_k} = \pi - 2 \cos^{-1}(t)$, we also have
\begin{align*}
    \sup_{k \in \bbZ} \Pr(X \in I_k) \leq \frac{ \pi - 2 \cos^{-1}(t)}{ \sqrt{2\pi} \sigma }.
\end{align*}
Therefore,
\begin{align*}
    \sum_{k \in \bbZ} \Pr(X \in I_k) \leq \left( 3 + \frac{2\sigma}{\sqrt{2\pi}} \right) \frac{ \pi - 2 \cos^{-1}(t)}{ \sqrt{2\pi} \sigma } = \left(1 + \frac{3\sqrt{\pi/2}}{\sigma} \right) \left(1 - \frac{2\cos^{-1}(t)}{\pi} \right).
\end{align*}
\end{proof}

The following result is similar to \cite[Lemma 10]{foster2020nonlinear}, except for $\sin$
instead of ReLU activations.

\begin{myprop}
\label{prop:param_error_from_square_error_sin_GLM}
Let $u_1, u_2 \in \R^d$. There exists a universal positive constants $\gamma_0, c_0$ such that for all $\gamma \in [0, \gamma_0]$,
\begin{align*}
     \E_{z \sim \sfN(0, \sigma^2 I_d)}[ (\sin(\ip{u_1}{z}) - \sin(\ip{u_2}{z}))^2 ] \leq \gamma^2 \Longrightarrow \norm{u_1-u_2}^2 \leq \frac{c_0 \gamma^2}{\sigma^2}.
\end{align*}
\end{myprop}
\begin{proof}
We first use the identity
\begin{align*}
    \sin(\ip{u_1}{z}) - \sin(\ip{u_2}{z}) &= 2 \cos\left( \frac{\ip{u_1+u_2}{z}}{2}\right) \sin\left( \frac{\ip{u_1-u_2}{z}}{2} \right),
\end{align*}
so that
\begin{align*}
    (\sin(\ip{u_1}{z}) - \sin(\ip{u_2}{z}))^2 &= 4 \cos^2\left( \frac{\ip{u_1+u_2}{z}}{2}\right) \sin^2\left( \frac{\ip{u_1-u_2}{z}}{2} \right).
\end{align*}
We fix $\delta = 1/4$ and consider two cases.

\paragraph{Case $\sigma \norm{u_1+u_2} \leq 1/ \sqrt{2\log(2/\delta)}$.}
We define two events:
\begin{align*}
    \calE_1 &:= \{ \abs{\ip{u_1+u_2}{z}} \leq \sigma \norm{u_1+u_2}\sqrt{2\log(2/\delta)} \}, \\
    \calE_2 &:= \{  \abs{ \sin( \ip{u_1-u_2}{z}/2 )} \geq \delta/c_0 \cdot \min\{1, \sigma/2 \cdot\norm{u_1-u_2} \} \},
\end{align*}
where $c_0$ is from \Cref{prop:cos_anticoncentration}.
By standard Gaussian concentration results, we know that $\Pr(\calE_1^c) \leq \delta$.
From \Cref{prop:cos_anticoncentration} we also know that $\Pr(\calE_2^c) \leq \delta$.
By a union bound, we have
$\Pr(\calE_1 \cap \calE_2) \geq 1 - 2\delta = 1/2$.
Putting these together,
\begin{align*}
    &\E_{z \sim \sfN(0, \sigma^2 I_d)}[ (\sin(\ip{u_1}{z}) - \sin(\ip{u_2}{z}))^2  ] \\
    &= 4 \E_{z \sim \sfN(0, \sigma^2 I_d)}\left[ \cos^2\left( \frac{\ip{u_1+u_2}{z}}{2}\right) \sin^2\left( \frac{\ip{u_1-u_2}{z}}{2} \right) \right] \\
    &\geq 4 \E_{z \sim \sfN(0, \sigma^2 I_d)}\left[ \cos^2\left( \frac{\ip{u_1+u_2}{z}}{2}\right) \sin^2\left( \frac{\ip{u_1-u_2}{z}}{2} \right) \ind\{\calE_1 \cap \calE_2\} \right] \\
    &\geq 4 \E_{z \sim \sfN(0, \sigma^2 I_d)}\left[ \cos^2(1/2) (\delta/c_0)^2 \min\{ 1, \sigma^2/4 \cdot \norm{u_1-u_2}^2 \}  \ind\{\calE_1 \cap \calE_2\} \right] \\
    &\geq 2 \cos^2(1/2) (\delta/c_0)^2 \min\{ 1, \sigma^2/4 \cdot \norm{u_1-u_2}^2 \}.
\end{align*}

\paragraph{Case $\sigma \norm{u_1+u_2} > 1/ \sqrt{2\log(2/\delta)}$.}

In this case, we define two events:
\begin{align*}
    \calE_1 &:= \{ \abs{ \cos( \ip{u_1+u_2}{z}/2 )} \geq \delta/c_0 \cdot \min\{1, \sigma/2 \cdot\norm{u_1+u_2} \} \}, \\
    \calE_2 &:= \{ \abs{ \sin( \ip{u_1-u_2}{z}/2 )} \geq \delta/c_0 \cdot \min\{1, \sigma/2 \cdot\norm{u_1-u_2} \} \},
\end{align*}
where again $c_0$ is from \Cref{prop:cos_anticoncentration}.
Using \Cref{prop:cos_anticoncentration} and a union bound, we have $\Pr(\calE_1 \cap \calE_1) \geq 1/2$.
Furthermore,
\begin{align*}
    &\E_{z \sim \sfN(0, \sigma^2 I_d)}[ (\sin(\ip{u_1}{z}) - \sin(\ip{u_2}{z}))^2  ] \\
    &\geq 4 \E_{z \sim \sfN(0, \sigma^2 I_d)}\left[ \cos^2\left( \frac{\ip{u_1+u_2}{z}}{2}\right) \sin^2\left( \frac{\ip{u_1-u_2}{z}}{2} \right) \ind\{\calE_1 \cap \calE_2\} \right] \\
    &\geq 4 (\delta/c_0)^4 \min\{ 1, \sigma^2/4 \cdot \norm{u_1+u_2}^2 \} \min\{ 1, \sigma^2/4 \cdot \norm{u_1-u_2}^2 \} \Pr(\calE_1 \cap \calE_2) \\
    &\geq 2 (\delta/c_0)^4 \min\{1, 1/(8 \log(2/\delta)) \} \min\{ 1, \sigma^2/4 \cdot \norm{u_1-u_2}^2 \}.
\end{align*}

\paragraph{Combining both cases.}
Combining both cases, we have that:
\begin{align*}
    \E_{z \sim \sfN(0, \sigma^2 I_d)}[ (\sin(\ip{u_1}{z}) - \sin(\ip{u_2}{z}))^2 ] \geq c_1 \min\{ 1, \sigma^2/4 \cdot \norm{u_1-u_2}^2 \},
\end{align*}
where $c_1 > 0$ is a universal constant (recall that $\delta=1/4$ is fixed).
Hence, for any $\gamma^2 \leq c_1/2$, we must have that
\begin{align*}
    \E_{z \sim \sfN(0, \sigma^2 I_d)}[ (\sin(\ip{u_1}{z}) - \sin(\ip{u_2}{z}))^2 ] \leq \gamma^2 \Longrightarrow 
    \min\{1, \sigma^2/4 \cdot \norm{u_1-u_2}^2 \} = \sigma^2/4 \cdot \norm{u_1-u_2}^2,
\end{align*}
otherwise we would have the contradiction
$c_1/2 \geq \gamma^2 \geq c_1$.
Hence, we conclude
\begin{align*}
     \E_{z \sim \sfN(0, \sigma^2 I_d)}[ (\sin(\ip{u_1}{z}) - \sin(\ip{u_2}{z}))^2 ] \leq \gamma^2 \Longrightarrow \norm{u_1-u_2}^2 \leq \frac{4\gamma^2}{c_1 \sigma^2}.
\end{align*}
\end{proof}

\begin{myfact}[{Hellinger distance for multivariate Gaussians~\cite[cf.][]{devroye2018total}}]
\label{fact:hellinger_gaussians}
Let $\sfN(\mu_i, \Sigma_i)$ for $i \in \{1,2\}$ be two multivariate Gaussians in $\R^d$. The squared Hellinger distance has the following closed-form expression:
\begin{align*}
    \frac{1}{2}\HgSq{\sfN(\mu_1, \Sigma_1)}{\sfN(\mu_2, \Sigma_2)} = 1 - \frac{\det(\Sigma_1)^{1/4} \det(\Sigma_2)^{1/4}}{\det((\Sigma_1+\Sigma_2)/2)^{1/2}} \exp\left\{ -\frac{1}{8}(\mu_1-\mu_2)^\T \left(\frac{\Sigma_1+\Sigma_2}{2}\right)^{-1} (\mu_1-\mu_2) \right\}.
\end{align*}
Hence, a special case when $\Sigma_1=\Sigma_2=\sigma^2 I_d$ is:
\begin{align*}
    \frac{1}{2} \HgSq{\sfN(\mu_1, \sigma^2 I_d)}{ \sfN(\mu_2, \sigma^2 I_d)} = 1 - \exp\left(-\frac{1}{8\sigma^2} \norm{\mu_1-\mu_2}^2 \right).
\end{align*}
\end{myfact}

\begin{myprop}
\label{prop:hellinger_to_square_error_sin_GLM}
Fix parameters $A_1, A_2 \in \R^{d \times d}$ and
let $\theta_i = \vec(A_i) \in \R^{d^2}$ for $i \in \{1,2\}$.
For any $\gamma \geq 0$, we have that
$\HgSq{p_{\theta_1}(z_{1:2})}{p_{\theta_2}(z_{1:2})} \leq \gamma^2$ implies the following bound holds:
\begin{align*}
    \max_{j \in [d]} \E_{z_1 \sim \sfN(0, \sigma^2 I_d)} [ ( \sin(\ip{A_1[j]}{z_1}) - \sin(\ip{A_2[j]}{z_1}) )^2  ] \leq 4\max\{ 2\sigma^2, 1\} \gamma^2.
\end{align*}
Here, $A_i[j] \in \R^d$ denotes the $j$-th row of $A_i$.
\end{myprop}
\begin{proof}
Since $z_1 \sim \sfN(0, \sigma^2 I_d)$ regardless of $\theta$, we have using \Cref{fact:hellinger_gaussians},
\begin{align*}
    \frac{1}{2}\HgSq{p_{\theta_1}(z_{1:2})}{p_{\theta_2}(z_{1:2})} &= \frac{1}{2}\E_{z_1}[ \HgSq{\sfN(\sin(A_1 z_1), \sigma^2 I_d)}{\sfN(\sin(A_2 z_1), \sigma^2 I_d)} ] \\
    &= 1 - \E_{z_1}\left[ \exp\left(-\frac{1}{8\sigma^2} \norm{ \sin(A_1 z_1) - \sin(A_2 z_1) }^2 \right) \right].
\end{align*}
Hence,
\begin{align*}
    \HgSq{p_{\theta_1}(z_{1:2})}{p_{\theta_2}(z_{1:2})} \leq \gamma^2 \Longleftrightarrow 1-\frac{\gamma^2}{2} \leq \E_{z_1}\left[ \exp\left(-\frac{1}{8\sigma^2} \norm{ \sin(A_1 z_1) - \sin(A_2 z_1) }^2 \right) \right].
\end{align*}
Let $c = \max\{8, 4/\sigma^2\}$, $x \in [0, 2]$, and observe that by the inequality
$\exp(-x) \leq 1 - x + x^2/2$ which is valid for all $x \geq 0$, 
\begin{align*}
    \exp\left(-\frac{x^2}{8\sigma^2}\right) &\leq \exp\left(-\frac{x^2}{c \sigma^2} \right) \leq 1 - \frac{x^2}{c\sigma^2} + \frac{x^4}{2 c^2 \sigma^4} \leq 1 - \frac{x^2}{c\sigma^2} + \frac{2 x^2}{c^2 \sigma^4} \\
    &= 1 - \frac{x^2}{c \sigma^2} \left( 1 - \frac{2}{c\sigma^2} \right) \leq 1 - \frac{x^2}{2 c \sigma^2}.
\end{align*}
Fixing any index $j_0 \in [d]$, we now observe that:
\begin{align*}
    &\E_{z_1}\left[ \exp\left(-\frac{1}{8\sigma^2} \norm{ \sin(A_1 z_1) - \sin(A_2 z_1) }^2 \right) \right]\\
    &\leq \E_{z_1}\left[ \exp\left(-\frac{1}{8\sigma^2}( \sin(\ip{A_1[j_0]}{z_1}) - \sin(\ip{A_2[j_0]}{z_1}) )^2 \right) \right]. \\
    &\leq 1 - \frac{1}{2c\sigma^2} \E_{z_1} [ ( \sin(\ip{A_1[j_0]}{z_1}) - \sin(\ip{A_2[j_0]}{z_1}) )^2  ].
\end{align*}
From this, we conclude that
\begin{align*}
     \HgSq{p_{\theta_1}(z_{1:2})}{p_{\theta_2}(z_{1:2})} \leq \gamma^2 \Longrightarrow \E_{z_1} [ ( \sin(\ip{A_1[j_0]}{z_1}) - \sin(\ip{A_2[j_0]}{z_1}) )^2  ] \leq c \sigma^2 \gamma^2.
\end{align*}
Since $j_0 \in [d]$ is arbitrary, the claim follows.
\end{proof}

\begin{myprop}
\label{prop:local_hellinger_identifiability_sin_GLM}
Fix parameters $A_1, A_2 \in \R^{d \times d}$ and
let $\theta_i = \vec(A_i) \in \R^{d^2}$ for $i \in \{1,2\}$.
There exists universal positive constants $\gamma_0, c_0$ such that for all $\gamma \in [0, \gamma_0/ \max\{1, \sigma\}]$,
\begin{align*}
    \HgSq{p_{\theta_1}(z_{1:2})}{p_{\theta_2}(z_{1:2})} \leq \gamma^2 \Longrightarrow \max_{j \in [d]} \norm{A_1[j] -A_2[j]}_F^2 \leq c_0 \max\{1, 1/\sigma^2\} \gamma^2.
\end{align*}
\end{myprop}
\begin{proof}
Given the condition $\HgSq{p_{\theta_1}(z_{1:2})}{p_{\theta_2}(z_{1:2})} \leq \gamma^2$, from \Cref{prop:hellinger_to_square_error_sin_GLM} for every $j \in [d]$,
\begin{align*}
    \E_{z_1 \sim \sfN(0, \sigma^2 I_d)} [ ( \sin(\ip{A_1[j]}{z_1}) - \sin(\ip{A_2[j]}{z_1}) )^2  ] \lesssim \max\{ 1, \sigma^2 \} \gamma^2.
\end{align*}
Next, from \Cref{prop:param_error_from_square_error_sin_GLM},
this implies that for every $j \in [d]$,
\begin{align*}
    \norm{ A_1[j] - A_2[j] }^2 \lesssim \max\{1, 1/\sigma^2\} \gamma^2.
\end{align*}
\end{proof}

\section{Additional Results for Sequence Modeling}
\label{sec:appendix:seq}
\begin{myprop}
    \label{prop:min_eig_cov}
    Suppose that for $\Psi: \R^d \mapsto \R^{d-1}$, $d > 1$ defined as
    $$
        \Psi(v) := J^\top v \quad \text{for} \quad  
        J := \begin{bmatrix}
            I_{d-1} \\ 0
        \end{bmatrix}.
    $$
    For $p \in \R^d_{> 0}$ such that $\| p \|_1 = 1$ with $\mu := \min_{i \in [d]} p_i > 0$, it holds that
    $$
        \lambda_{\min}\left(\diag(\Psi(p)) - \Psi(p) \Psi(p)^\T\right) \geq \frac{\mu}{4(d-1)}.
    $$
\end{myprop}
\begin{proof}
    For convenience, we define $q = \Psi(p)$ such that $q \in \R^{d-1}$.
    Let us begin by briefly establishing an upperbound for the minimum eigenvalue.
    First, 
    noting that $\ip{\ind_{d-1}}{q} = 1 - p_d$ for all ones vector $\ind_{d-1} \in \R^{d-1}$ and setting $v = \frac{1}{\sqrt{d-1}} \ind_{d-1}$ such that $\|v\| = 1$, we can see
    \begin{align*}
        \lambda_{\min}(\diag(q) - qq^\T) &\leq v^\T (\diag(q) - qq^\T) v \\
        &= \frac{1}{d-1}\left[ \ip{\ind_{d-1}}{q} - \ip{\ind_{d-1}}{q}^2 \right] \\
        &= \frac{p_d(1-p_d)}{d-1}.
    \end{align*}
    Therefore, for $d > 1$ we can conclude that $\lambda_{\min}(\diag(q) - qq^\T) < p_d$. 
    Now, for $j := \argmin_{i \in [d-1]} q_i$ let us set $v$ to be the basis vector in $\R^{d-1}$ such that $v_j = 1$ and $v_i = 0$ for all $i \in [d-1] \setminus \{j\}$.
    \begin{align*}
        \lambda_{\min}(\diag(q) - qq^\T) &\leq v^\T (\diag(q) - qq^\T) v \\
        &= q_j - q_j^2 < q_j.
    \end{align*}
    Putting these together, we can see that for $d > 1$, $\lambda_{\min}(\diag(q) - qq^\T) < \min\{q_j, p_d\} = \mu$.
    Now let us move on to the lowerbound.
    Let $\lambda$ be the smallest eigenvalue of $\diag(q) - qq^\top$: this means that 
    $$
        0 = \det( \lambda I - ( \diag(q) - qq^\T ) ) = \det(\lambda I - \diag(q) + qq^\T ).
    $$
    Since we have established that $\lambda \not\in \{p_1, \dots, p_{d-1}\}$, we can see that the matrix $\lambda I - \diag(q)$ must be invertible. Hence, by the matrix determinant lemma,
    \begin{align*}
        \det(\lambda I - \diag(q)) (1 + q^\T (\lambda I - \diag(q))^{-1} q) = 0,
    \end{align*}
    and since $\det(\lambda I - \diag(q)) \not= 0$, we can see that
    \begin{align*}
        1 + q^\top(\lambda I - \diag(q))^{-1} q &= 0, \\
        \sum_{i=1}^{d-1} \frac{p_i^2}{p_i - \lambda} &= 1.
    \end{align*}
    Let us define $f_a(\lambda) := a/(a-\lambda)$ for some $a > 0$ and $\lambda \not= a$, such that $f'_a(\lambda) = a/(a-\lambda)^2$.
    By the mean value theorem, for some $c \in [0, \lambda]$,
    $$
        f_a(\lambda) = f_a(0) + \lambda f'_a(c) = 1 + \frac{a \lambda}{(a-c)^2} \leq 1 + \frac{a\lambda}{(a-\lambda)^2}.
    $$
    Using this in the original equation and noting $\sum_{i=1}^{d-1} p_i \leq 1 - \mu$ we have
    \begin{align*}
        \sum_{i=1}^{d-1} \frac{p_i^2}{p_i-\lambda} 
        = \sum_{i=1}^{d-1} p_i \cdot f_{p_i}(\lambda) 
        \leq \sum_{i=1}^{d-1} p_i \left(1 + \frac{p_i\lambda}{(p_i-\lambda)^2} \right) 
        \leq 1-\mu + \sum_{i=1}^{d-1} \frac{p_i^2 \lambda}{(p_i-\lambda)^2}.
    \end{align*}
    Next, it is quick to check that $x \mapsto x^2/(x-\lambda)^2$ is \emph{decreasing} for $x > \lambda$, since
    $\frac{\rmd}{\rmd x} x^2/(x-\lambda)^2 = - 2\lambda x/(x-\lambda)^3$.
    Since we have established that the minimum eigenvalue is upperbounded by $\mu$ (for $d > 1$), we have that $p_i \geq \mu > \lambda$ for all $i \in [d]$ so we can upperbound the expression by lowerbounding the elements of $p$ for 
    $$
        \sum_{i=1}^{d-1} \frac{p_i^2\lambda}{(p_i-\lambda)^2} \leq (d-1) \frac{\mu^2 \lambda}{(\mu - \lambda)^2}.
    $$
    Now we write $\lambda(c) = c \mu$ for $c \in (0, 1)$, and
    compute a $c_0$ such that for all $c \leq c_0$, $(d-1) \frac{ \mu^2 \lambda(c) }{ (\mu - \lambda(c))^2 } < \mu$:
    $$
        \frac{c}{(1-c)^2} \mu = \frac{\mu^2 c \mu}{(1-c)^2 \mu^2} < \frac{\mu}{d-1} \Longleftrightarrow \frac{c}{(1-c)^2} < \frac{1}{d-1}.
    $$
    For the RHS, it suffices to take $c < 1/(4(d-1))$.
    Thus, we conclude that:
    \begin{align*}
        \lambda_{\min}(\diag(q) - qq^\T) \geq \frac{\mu}{4(d-1)}.
    \end{align*}
\end{proof}

\end{document}